%% file: main.tex
\documentclass[10pt,twocolumn,letterpaper]{article}

\usepackage[pagenumbers]{static/cvpr} % To force page numbers, e.g. for an arXiv version

\input{sec_preprint/subsec/preamble}

\definecolor{cvprblue}{rgb}{0.21,0.49,0.74}
\usepackage[pagebackref,breaklinks,colorlinks,allcolors=cvprblue]{hyperref}
\usepackage{tablefootnote}
\usepackage{fancyhdr}
\usepackage{tocloft}
\newcommand{\wsmcurrentsectionnumber}{}
\newcommand{\wsmcurrentsectiontitle}{}

\fancypagestyle{wsmrunning}{%
  \fancyhf{}%
  \fancyhead[R]{\parbox[b]{0.9\headwidth}{\raggedleft\normalsize\bfseries
    \ifx\wsmcurrentsectionnumber\empty\else
      Section~\wsmcurrentsectionnumber\ ---\ \nouppercase{\wsmcurrentsectiontitle}%
    \fi
    \ifx\rightmark\empty\else\\[-0.1em]\nouppercase{\rightmark}\fi}}%
  \fancyfoot[C]{\thepage}%
}
\fancypagestyle{wsmnoheader}{%
  \fancyhf{}%
  \fancyfoot[C]{\thepage}%
}
\renewcommand{\sectionmark}[1]{}
\renewcommand{\cftsecfont}{\normalsize\bfseries\color{cvprblue}}
\renewcommand{\cftsecpagefont}{\normalsize\bfseries\color{cvprblue}}
\renewcommand{\cftsubsecfont}{\small\color{black}}
\renewcommand{\cftsubsecpagefont}{\small\color{black}}
\makeatletter
\def\cvprsect#1{\cvprsection[#1]{\texorpdfstring{\hskip -1em.~}{}#1}}
\def\cvprsubsect#1{\cvprsubsection[#1]{\texorpdfstring{\hskip -1em.~}{}#1}}
\def\cvprsubsubsect#1{\cvprsubsubsection[#1]{\texorpdfstring{\hskip -1em.~}{}#1}}
\makeatother

\usepackage{graphicx}
\usepackage{simpleicons}
\usepackage{tgadventor}
\definecolor{ghcolor}{rgb}{0.15,0.15,0.18}
\definecolor{webcolor}{rgb}{0.31,0.45,0.62}
\definecolor{gradiocolor}{rgb}{1.00,0.49,0.00}
\definecolor{docpagecolor}{rgb}{0.15,0.15,0.18}
\newcommand{\iconbox}[1]{\makebox[5.1em][c]{#1}}
\newcommand{\labelbox}[1]{\makebox[5.1em][c]{\fontfamily{qag}\selectfont\footnotesize\strut #1}}
\newcommand{\ghlogo}{\includegraphics[height=2.06em]{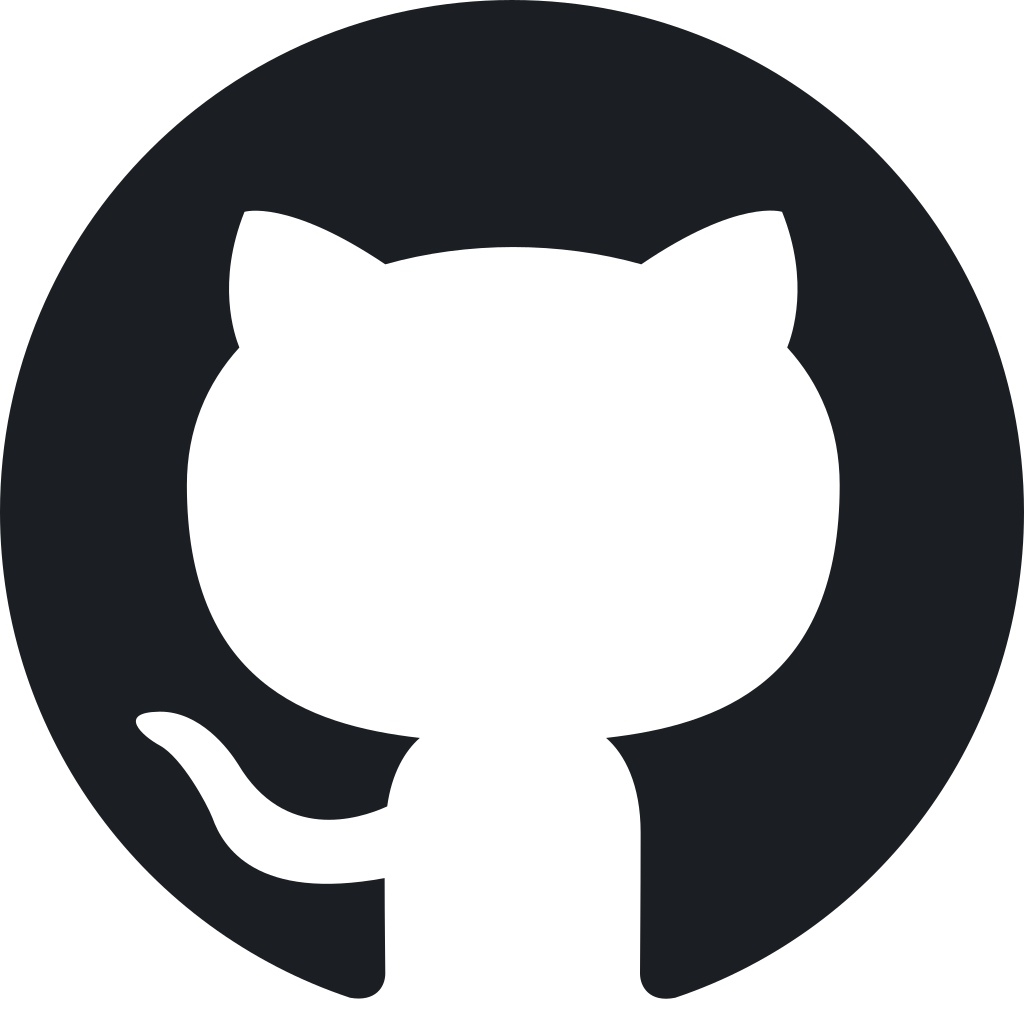}}
\newcommand{\hflogo}{\includegraphics[height=2.06em]{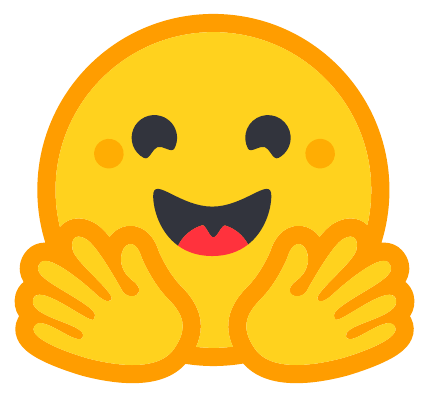}}
\newcommand{\webuilogo}{\includegraphics[height=2.06em]{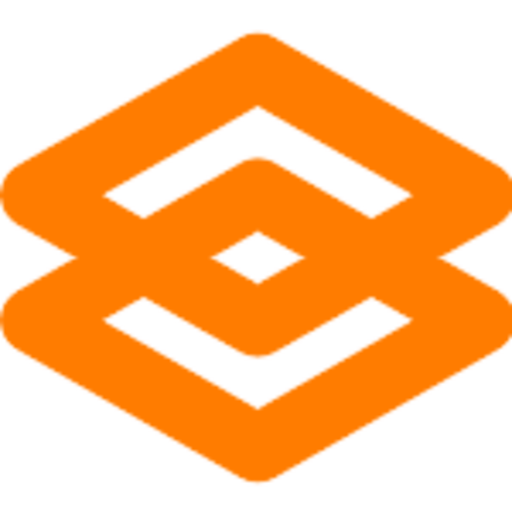}}
\newcommand{\doclogo}{\textcolor{docpagecolor}{\fontsize{22}{22}\selectfont\simpleicon{googledocs}}}
\newcommand{\iconentry}[3]{%
  \begin{tabular}{@{}c@{}}
    \iconbox{\href{#1}{#2}}\\[-0.1em]
    \labelbox{#3}
  \end{tabular}%
}

\usepackage{cuted}
\usepackage{etoolbox}
\makeatletter
\patchcmd{\@maketitle}{\begin{center}}{\begin{flushleft}}{}{}
\patchcmd{\@maketitle}{\end{center}}{\end{flushleft}}{}{}
\patchcmd{\@maketitle}{\begin{tabular}[t]{c}}{\begin{tabular}[t]{@{}l@{}}}{}{}
\makeatother

\def\confName{CVPR}
\def\confYear{2026}

\newcommand{\blueboxnum}[1]{%
    \begingroup
    \setlength{\fboxsep}{1.8pt}%
    \raisebox{0pt}[1.08\ht\strutbox][0.96\dp\strutbox]{%
        \colorbox{blue}{\textcolor{white}{\bfseries\footnotesize #1}}%
    }%
    \endgroup
}
\newcommand{\numtag}[1]{\blueboxnum{#1}}

\newcommand{\iclrtitle}{%
\vspace{-1.75em}
\parbox{\textwidth}{%
\raggedright
\fontsize{21}{22}\selectfont
\scshape
Unified Law for Visual Tasks (The Four Laws of Computer Vision)%
}%
}

\title{\iclrtitle}

\input{sec_preprint/subsec/author_preprint}

\input{tables/pwc_task_scope/pwc_cv_task_scope_macros}
\input{tables/pwc_task_scope/wsm_historical_overlap_macros}
\begin{document}
\maketitle
\ifdefined\anonymousprojectlinks
  \anonymousprojectlinks
\fi
\pagestyle{wsmrunning}
\renewcommand{\sectionmark}[1]{}
\renewcommand{\subsectionmark}[1]{}
\let\wsmoriginalsection\section
\RenewDocumentCommand{\section}{s o m}{%
  \IfBooleanTF{#1}{%
    \wsmoriginalsection*{#3}%
  }{%
    \IfNoValueTF{#2}{\wsmoriginalsection{#3}}{\wsmoriginalsection[#2]{#3}}%
    \xdef\wsmcurrentsectionnumber{\thesection}%
    \gdef\wsmcurrentsectiontitle{#3}%
    \markboth{Section~\thesection\ ---\ #3}{}%
  }%
}
\let\wsmoriginalsubsection\subsection
\RenewDocumentCommand{\subsection}{s o m}{%
  \IfBooleanTF{#1}{%
    \wsmoriginalsubsection*{#3}%
  }{%
    \IfNoValueTF{#2}{\wsmoriginalsubsection{#3}}{\wsmoriginalsubsection[#2]{#3}}%
    \markright{\thesubsection\ ---\ #3}%
  }%
}
\thispagestyle{wsmnoheader}

\vspace*{-4pt}
\input{fig/teaser_with_comment}

\input{sec_preprint/0_abstract}
\input{sec_preprint/1_intro}

% The table of contents occupies a full-width single-column page between Introduction and Related Work.
\onecolumn
\markboth{Contents}{}
\thispagestyle{wsmnoheader}
\phantomsection
\pdfbookmark[1]{Contents}{wsm.contents}
\begingroup
\input{static/full_row_toc_links}
% \enlargethispage{0.8\baselineskip} % one-page TOC compression
% \vspace*{-1.5em}                   % one-page TOC compression
\begin{center}
  {\LARGE\bfseries\color{cvprblue} Contents}
\end{center}
% \vspace{-0.15em}                   % one-page TOC compression
% \setlength{\cftbeforesecskip}{0.12em}    % one-page TOC compression
% \setlength{\cftbeforesubsecskip}{0em}    % one-page TOC compression
  \makeatletter
  \@starttoc{toc}
  \makeatother
\endgroup
\twocolumn
\input{sec_preprint/2_related}
\input{sec_preprint/3_method}
\input{sec_preprint/4_ablation}
\input{sec_preprint/5_quantitative}
\input{sec_preprint/6_qualitative}
\input{sec_preprint/7_discussion}
\onecolumn
\input{sec_preprint/8_conclusion}
\twocolumn
\FloatBarrier
{
    \small
    \bibliographystyle{unsrtnat}
    \bibliography{ref/main}
}

% Keep the appendix in the document's continuous Arabic page sequence so the
% PDF footer and table-of-contents page references use the same numbering.
\renewcommand{\thepage}{\arabic{page}}
\input{sec_preprint/X_suppl}

\end{document}

%% file: sec_preprint/subsec/preamble.tex
\usepackage{cuted} %< for strip environment
\usepackage{multicol} %< for local supplementary two-column layouts
\makeatletter
\providecommand{\@setmarks}{}
\makeatother

\usepackage{currfile} %< for \currfilebase macro

\usepackage{caption} %< for "\captionof" commands (teaser)
\usepackage{array} %< for vertically centered table columns
\usepackage{booktabs} %< reproducible evaluation tables
\usepackage{colortbl} %< subtle header and result-row backgrounds
\usepackage{longtable} %< multi-page historical task inventory
\usepackage{tabularx} %< flexible-width analysis summary tables
\usepackage{placeins} %< keep evaluation floats with their text
\usepackage{algorithm} %< modality-task construction pseudocode
\usepackage{algpseudocode}
\DeclareRobustCommand{\circlednote}[1]{%
    \raisebox{0.12ex}{%
        \textcircled{\raisebox{-0.18ex}{\scriptsize\bfseries #1}}%
    }%
}

%% file: sec_preprint/subsec/author_preprint.tex
\author{
\textbf{Sen Fang}\textsuperscript{1} \hspace{0.5em}
\textbf{Yalin Feng}\textsuperscript{2} \hspace{0.5em}
\textbf{Yancheng Zhang}\textsuperscript{3} \hspace{0.5em}
\textbf{Yanxin Zhang}\textsuperscript{4} \hspace{0.5em}
\textbf{Zonglin Lyu}\textsuperscript{3} \hspace{0.5em}
\textbf{Hongbin Zhong}\textsuperscript{5} \\
\textbf{Liwei Che}\textsuperscript{1} \quad
\textbf{Rui Wu}\textsuperscript{1} \quad
\textbf{Chen Chen}\textsuperscript{3} \quad
\textbf{Dimitris N. Metaxas}\textsuperscript{1}
\\[0.5em]
\textsuperscript{1}Rutgers, USA \quad
\textsuperscript{2}NTU, Singapore \quad
\textsuperscript{3}UCF, USA \quad
\textsuperscript{4}UW--Madison, USA \quad
\textsuperscript{5}Georgia Tech, USA
\\[0.5em]
\textbf{Project Page:} {\hypersetup{urlcolor=magenta}\textbf{\url{https://world-snapshot.github.io/WSMs}}}
\\[0.3em]
\begin{tabular}{@{}cccc@{}}
    \iconentry{https://github.com/World-Snapshot/WSMs}{\ghlogo}{Code} &
    \iconentry{https://huggingface.co/World-Snapshot/datasets}{\hflogo}{Dataset} &
    \iconentry{https://github.com/World-Snapshot/ControlWorld}{\webuilogo}{WebUI} &
    \iconentry{https://world-snapshot.github.io/doc}{\doclogo}{Document}
  \end{tabular}
\vspace{-24pt}
}

%% file: fig/teaser_with_comment.tex
\begin{strip}
\centering
\begin{minipage}[t]{.49\linewidth}
    \centering
    \includegraphics[width=\linewidth]{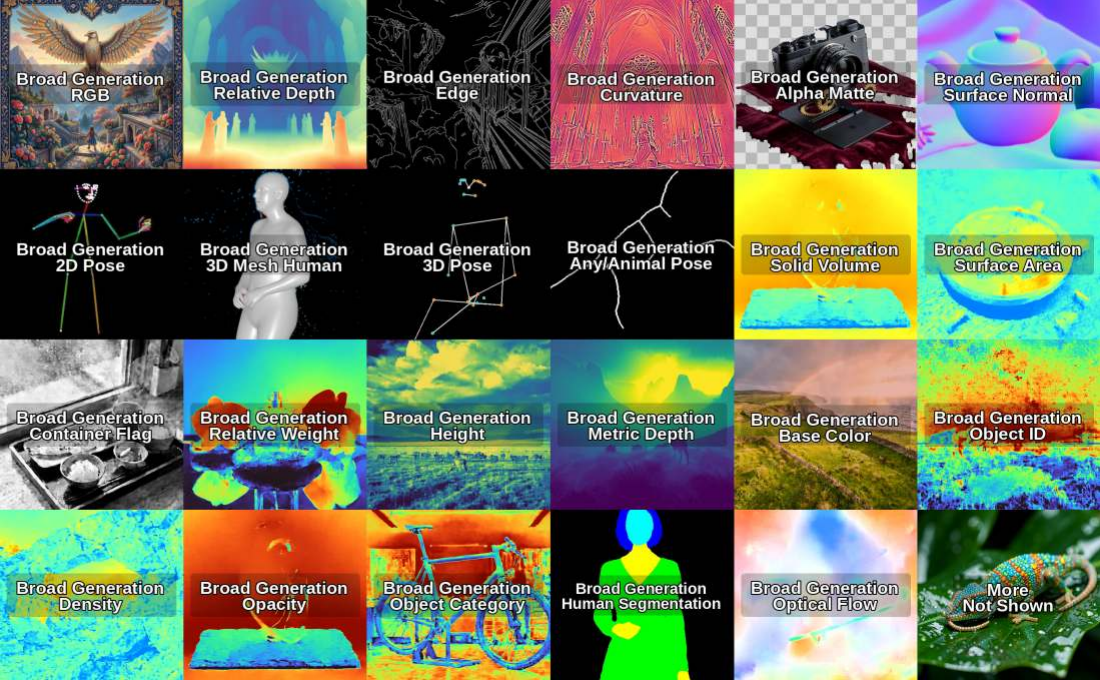}

    \small Lex I. The previous generation was mostly generating RGB as a special case of visual generation.
\end{minipage}\hfill
\begin{minipage}[t]{.49\linewidth}
    \centering
    \includegraphics[width=\linewidth]{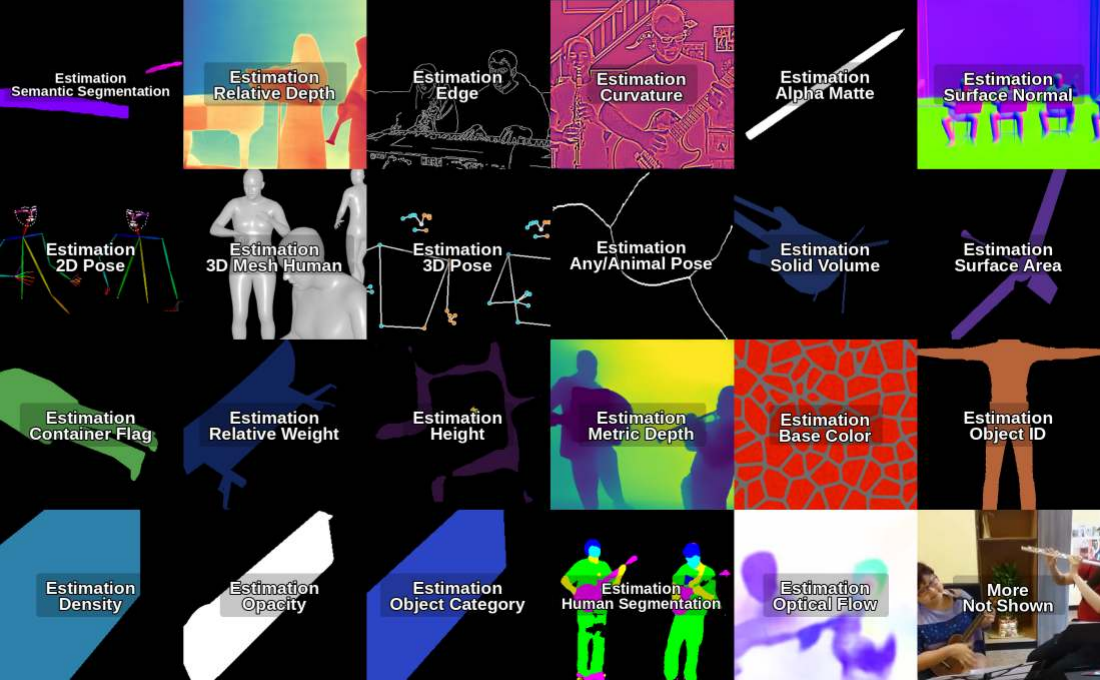}

    \small Lex II. The previous estimations (understand, seg, detect) were obtaining world information (additional channels) from RGB.
\end{minipage}

\medskip

\begin{minipage}[t]{.49\linewidth}
    \centering
    \includegraphics[width=\linewidth]{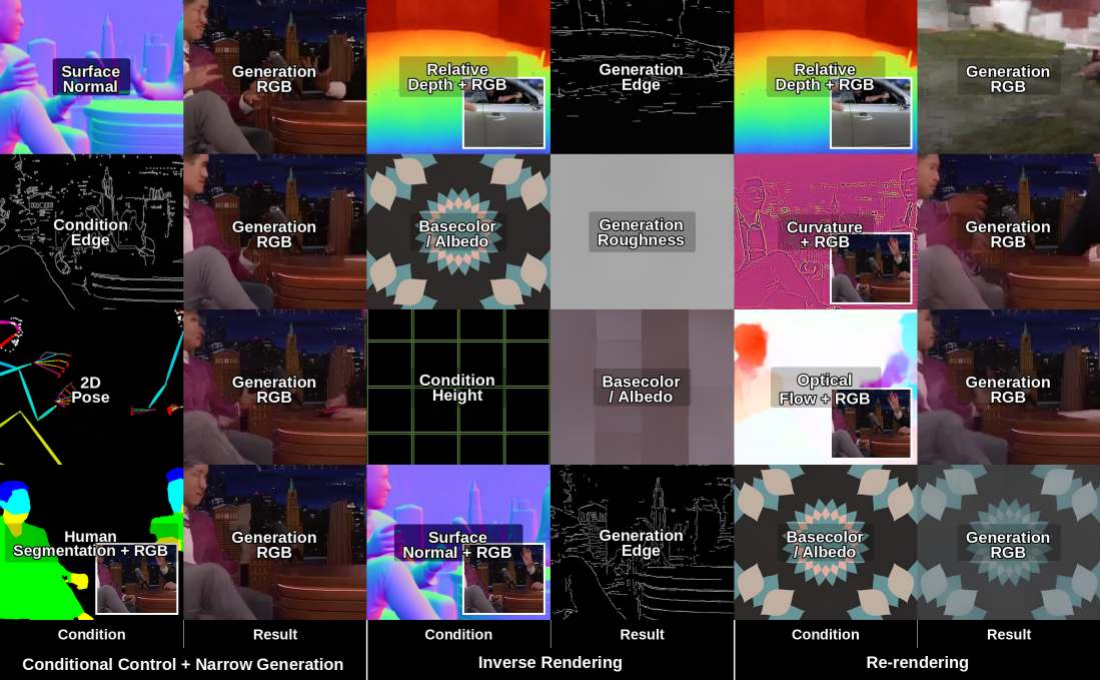}

    \small Lex III. The previous conditional control was generating RGB based on world information (additional channels).
\end{minipage}\hfill
\begin{minipage}[t]{.49\linewidth}
    \centering
    \includegraphics[width=\linewidth]{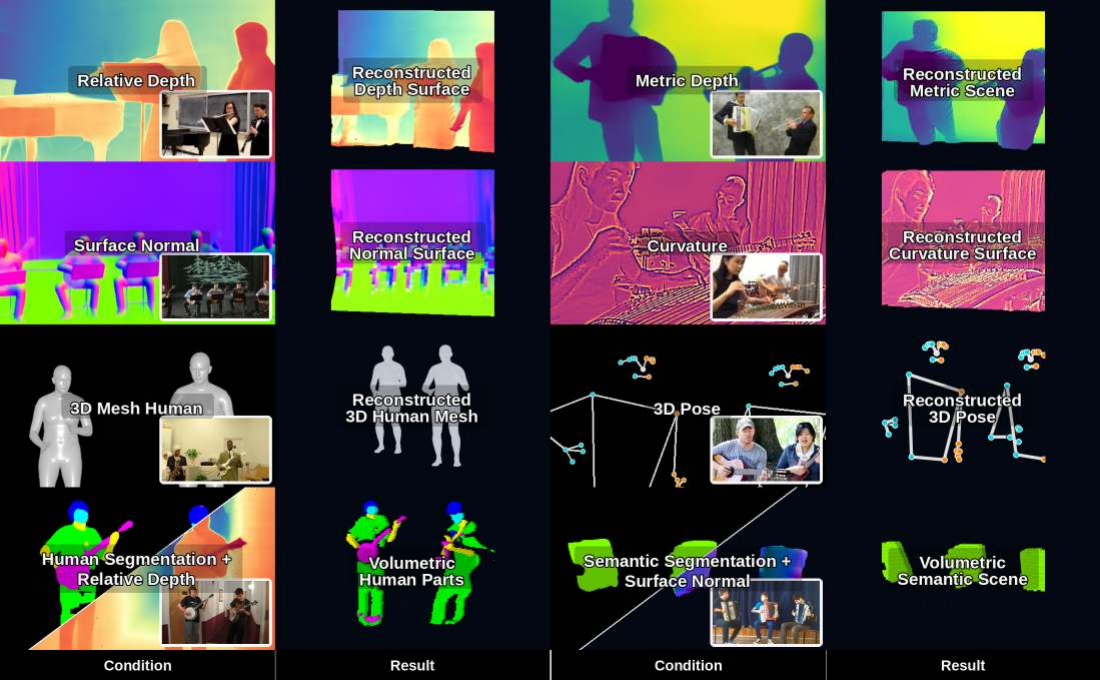}

    \small Lex IV. Reconstruction estimates world information from RGB, with all of this information encoding shape.
\end{minipage}
\captionsetup{hypcap=false}
\captionof{figure}{\textbf{The Unified Law for Visual Tasks}---flattens the world into a two-dimensional space. In this representation, image processing is world processing. If we summarize all visual tasks into four broad categories and reformulate them all as generation tasks (\textbf{The Four Laws of Computer Vision}), then a generative method can perform thousands of visual tasks. Representative tasks are shown above.}
\captionsetup{hypcap=true}
\label{fig:\currfilebase}%
\par
\vspace{6pt}
\raggedright
\small
\noindent\textbf{Comment:} This work began in December 2024. Project page at \href{https://world-snapshot.github.io/WSMs}{link}; GitHub org at \href{https://github.com/World-Snapshot}{link}; Hugging Face org at \href{https://huggingface.co/World-Snapshot}{link}; This is an empirical article that attempts to unify \textcolor{red}{\textbf{thousands of visual tasks}} through a single methodology. Patent pending (inventors: SF and DM).
\end{strip}

%arXiv Comment: Project homepage at xx; Contributions and timeline at xx; Model at xx; This is an empirical article that attempts to unify thousands of visual tasks through a single methodology.

%Title: The official name of this article in Camera Ready and OpenReview format is "The Four Laws of Computer Vision". This is purely for the aesthetic consideration of the CV template. It will be used in arXiv, repositories, informal PDFs, and project homepages as "Unified Law for Visual Task (The Four Laws of Computer Vision)" or its variations, as we believe the latter fully conveys the meaning and feeling we intend to express.

%% file: sec_preprint/0_abstract.tex
\begin{abstract}
%在这篇文章中，我们提出了视觉任务统一定律，细分为计算机视觉的四大定律，作用是使得单个框架/方法可以统一训练和执行几乎所有可视化任务。
%首先我们得到一个灵感并假设：将世界编码到图像内（称为世界快照），图像处理就是世界处理，图像生成就是世界（快照）生成。
%进而，如果将所有视觉、可视化任务都重新定义成生成任务，那么我们只需要开发一种模型（暂时称为世界快照模型）就能执行绝大多数视觉任务，就像LLM几乎统一NLP一样，
%本文旨在总结并于1000+项视觉任务上验证这些规律，并提出一个统一的训练框架，以指导未来的视觉任务研究和应用。

In this paper, we propose a Unified Law for Visual Tasks, which could be further divided into Four Laws of Computer Vision, with the goal of enabling a single framework or method to uniformly train on and perform nearly all visual tasks.
Our starting point is an inspiration and hypothesis: if the world information is encoded into image, which we call World Snapshots, then image processing is world processing, and image generation is World Snapshot generation.
Furthermore, if all visual and visualization tasks can be reformulated as generation tasks, then a unified method, World Snapshot Models (WSMs), can perform the vast majority of visual tasks, much as large language models have largely unified natural language processing.
This paper validates these laws across more than 1,000 trained and evaluated visual tasks (defined by distinct input--output mappings) and presents a unified training framework for future visual-task research and applications.

\end{abstract}

%% file: sec_preprint/1_intro.tex
\section{Introduction}
\label{sec:intro}

%目前,很多研究人员期待计算机视觉出现LLMs一样强大的统一模型,通常称之为世界模型. 希望它能尝试将“预测下一个token/状态”这样的方法使用到视觉领域。 但实际上这种想法常常面临两大困境：1. 视觉数据不是“Prompt-Answer”就能概括的。2. 计算机视觉的任务有多种形式。因此,在这篇文章,我们获得尝试提出一种理想世界模型,使它可以走出相应困境,尽量增强视觉方法的统一、通用性。

%首先,我们认为视觉领域真正需要的是“将所有信息编码到图像本身”。更详细的说，将所有信息编码到图像本身, 二维图像生成就是无损世界生成、flootnote{这句话就是我们所提到的统一定律。另外，上个世纪末的物理学家们将之称为物理全息论,同时有系统的论证，他们的结论是在特殊条件下，n+1维空间可以编码到n维空间，这也为我们的富信息图像等效3D世界的想法提供了支持}。

%为了验证这个统一定律的存在，我们可以创建一种名为世界快照图形(.wsg)的256通道图像数据格式,将更多信息（空间、姿势、物理、音频、语义等）存在额外的通道当中,尝试训练一个模型生成这种图像格式进而验证它。

%在这个过程中，我们必须将生成、估计(understand, detect, segment)、条件控制(re-rendering,re-rendering)、重建全部重新定义为生成任务、footnote{这就是细分的4个定律，在实验中我们发现准备四类接口可以兼容几乎所有视觉任务。}

Recently, many researchers now expect computer vision to develop a unified model as powerful as large language models~\cite{brown2020languagemodelsfewshotlearners,openai2024gpt4technicalreport}, often referring to such a model as a world model~\cite{ha2018world,lecun2022path}.
This expectation is usually motivated by the hope that ideas such as ``predicting the next token'' or ``predicting the next state'' can be transferred to the visual domain~\cite{ren2025nexttoken,wang2024emu3,wang2025bridging}.
However, this view faces two fundamental difficulties: visual data cannot be fully summarized by a simple prompt-answer format, and computer vision tasks appear in many different forms.
In this paper, we therefore attempt to formulate an ideal World Model that can move beyond these limitations and improve the unity and generality of visual methods.

\noindent\textbf{Insight and Hypothesis.} First, we argue that what the visual domain truly needs is to encode all information into the image itself.
More specifically, \textbf{\textcolor{purple}{if all information can be encoded into the image itself, then two-dimensional image generation becomes lossless world generation.}}\footnote{This statement is the \textbf{\textcolor{purple}{Unified Law}} proposed in this paper. Physicists at the end of the last century referred to a similar idea as the \href{https://en.wikipedia.org/wiki/Holographic_principle}{holographic principle} and provided systematic arguments showing that~\cite{maldacena1999large,Fields_2022,HosseinEybposh:20}, under special conditions, an $(n+1)$-dimensional space can be encoded into an $n$-dimensional space. This also supports our view that an information-rich image can approximate a world frame.}

To verify the existence of this Unified Law, we introduce an image data format with up to 256 channels named World Snapshot Graphics (.wsg).
This format stores additional information, including spatial structure, pose, physics, audio, and semantics, in extra channels.
We then attempt to train a model to generate images in this format, using the generation process itself as a test of the proposed law.

\noindent\textbf{Derived Implications and Research Reframing.} Furthermore, a task is a visual task as long as either its input or its output is visual information. Then, visual tasks can be abstracted at a macro level into generation, estimation, conditional control, and reconstruction.\footnote{These correspond to the four sub-laws. In our experiments, we find that preparing four types of interfaces can support almost all visual tasks.}
\textcolor{purple}{These four broad categories are not arbitrary: they correspond to the four fundamental directional relations formed when RGB, text, and non-RGB world attributes serve as conditions or targets.} Prior work covers each category in a narrow form, such as generating only RGB or estimating a few attributes from RGB; we call this narrow vision, and its extension to every world attribute broad vision. More directly, we do not want to train four separate classes of world-processing models for generation, estimation, conditional control, and reconstruction; we treat them all as forms of generation.
% Here, estimation includes understanding, detection, and segmentation, while conditional control includes re-rendering and related controllable visual transformations.
Figure~\ref{fig:teaser_with_comment} presents these four task families as Lex
I--IV. Together, they refine and extend the Unified Law by treating all four as forms of generation, thereby avoiding having to train four broad classes of world snapshot/frame processing models.

%对于具体的方法，我们应该考虑如何训练一个统一视觉生成模型，让无数条件利用它的主干完成一次或者多次推理。
%但是这时出现了一个长久的问题，即训练效率。我们不可以为了让一个模型掌握1000+视觉任务而大量增加参数，反而应该尽量达到当模型能力扩大千倍，而参数仅仅有轻微增长。
%ControlNet、LoRA是目前常用的两种解决方法，但是即使是10小时-数天就能掌握一种新能力的LoRA，面临上千任务也显得力不从心。

%因此我们提出了一种新方法，名为隧穿网络，用于训练WSM\footnote{如果说ResNet可以让模型训练特别深甚至是上千层，我们的TunnelingNet就可以使得模型训练的模态数量特别多}。
%它的核心在于完全摒弃传统，我们完全不要求模型理解两种模态，而是让他们各自学习。他们就像正常人和红绿色盲，他们可以描述相同的事物模态，但是不要求互相理解。
%在三个结构化非RGB任务上，相比相同参数量的LoRA式原生注意力基线，MTA达到展示质量所需的更新数减少19.0--38.5倍，首次可识别结构约早77--100倍出现。

\noindent\textbf{The Elephant in the Room.} For a concrete method, we should consider how to train a unified visual generation model whose backbone can be reused by countless conditions to perform one or multiple rounds of inference.
However, this immediately raises a long-standing challenge: training efficiency.
We should not substantially increase the number of parameters merely to make one model master more than 1,000 visual tasks; instead, we should aim for a regime in which the model's capabilities expand by orders of magnitude while its parameter count grows only slightly.
ControlNet \cite{zhang2023adding}, LoRA \cite{hu2022lora}, MoE \cite{shazeer2017outrageouslylargeneuralnetworks,zhang2025mixtureexpertslargelanguage,ye2026modus,Ravishankar_2025_CVPR}, IP-Adapter \cite{ye2023ipadapter}, and IC-Control Style \cite{huang2024incontextlora,zhang2025icedit, dalva2025lorashop,guo2025longcontext} are five commonly used approaches for expanding the capabilities of existing models, but even LoRA, which can acquire a new capability within several hours to several days, becomes insufficient when facing thousands of tasks.

% \noindent\textbf{Method.} We propose a new method named TunnelingNet for training WSMs.
% Its core idea is that one modality need not understand another modality in the
% latter's representational language. Instead, each modality retains an
% independent learning stream connected through a shared generative backbone.
% Consider two observers, one with typical color vision and one with red--green
% color blindness. They can describe the same underlying object through their
% respective visual experiences without either observer needing to see as the
% other sees. TunnelingNet gives modalities the same relationship: each evidence
% stream preserves its own representation, while a target-specific tunnel learns
% how that evidence should alter the shared generative trajectory.

\noindent\textbf{Method.} We propose a new method named TunnelingNet for training WSMs.
Its core idea is to assign each modality to a separate learning stream connected
through a shared generative backbone. The streams are analogous to two
observers, one with typical color vision and the other with red--green color
blindness: each describes the same underlying object through its own
representation without adopting the other's visual perception.
With this method, a model can master thousands of visual tasks within a very short time.
On structured non-RGB mappings, our proposed Modality Tunneling Attention (MTA) yields $19.0$--$38.5\times$ update-based
speedups over a parameter-matched LoRA native-attention baseline and produces
the first visually identifiable Edge structure after roughly 50 updates,
versus 3,850--5,000 updates for the matched NJA baseline
($77$--$100\times$ fewer updates; Fig.~\ref{fig:ablation_validation_curves}).

\onecolumn
\noindent\textbf{Contributions.}
The main contributions of this paper are summarized as follows:
\begin{itemize}
    \item We propose \href{https://github.com/World-Snapshot/Unified-Law-for-Visual-Tasks}{\textbf{Unified Law for Visual Tasks}}, which organizes all visual tasks into four broad categories---generation, estimation, conditional control, and reconstruction---and reformulates all four under a common generative perspective, providing a theoretical path toward unifying large-scale visual tasks.
    \item To support large-scale, complex visual reasoning through arbitrary, cyclic, and chained transformations among many modalities, we introduce \href{https://github.com/World-Snapshot/wsg}{\textbf{World Snapshot Graphics (.wsg)}}, a conceptual world-frame representation and flexible image format. Its world-frame abstraction provides a macro-level logic for understanding and managing modalities, while up to 256 channel IDs provide a flexible operational address space for spatial, pose, physical, audio, semantic, and other world information.
    \item We introduce \href{https://github.com/World-Snapshot/WSMs}{\textbf{World Snapshot Models (WSMs)}}, a unified framework connecting the Four Laws to four generative interfaces through a many-to-many correspondence. The Laws provide a macro-level conceptual organization of visual tasks; the interfaces are the minimal operational basis determined by the properties of task conditions and targets and provide concrete training schemes.
    \item We propose \href{https://github.com/World-Snapshot/WSMs}{\textbf{TunnelingNet}}, an efficient capability-expansion method that independently trains and registers lightweight task networks inside a shared frozen visual generator, enabling rapid scaling to thousands of tasks. Each generated world state can directly condition the next TunnelingNet, forming a self-contained visual-processing loop across all four broad task categories and revealing an initial visual chain of thought in a purely visual model. This composition opens a path to large-scale, complex visual reasoning: the task space expands as modalities are added, while executable visual chains grow exponentially with chain depth.
\end{itemize}

%% file: sec_preprint/2_related.tex
\section{Related Work}
\label{sec:related}

\begin{table}[t]
\centering
\caption{Comparison between World Snapshot Models (WSMs) and Large Language Models (LLMs).}
\label{tab:wsm_llm_comparison}
\footnotesize
\setlength{\tabcolsep}{3pt}
\renewcommand{\arraystretch}{1.18}
\begin{tabular}{@{}>{\raggedright\arraybackslash}m{0.31\columnwidth}
                >{\raggedright\arraybackslash}m{0.32\columnwidth}
                >{\raggedright\arraybackslash}m{0.32\columnwidth}@{}}
\hline
\textbf{Comparison Aspect} & \textbf{WSMs} & \textbf{LLMs} \\
\hline
\textbf{Domain} &
Vision &
NLP \\
\hline
\textbf{Methodology} &
Encode Information into Image &
Predict Next Token \\
\hline
\textbf{Main Laws/Principles Realized} &
\href{https://github.com/World-Snapshot/Unified-Law-for-Visual-Tasks}{\textbf{Unified Laws}}:\newline all vision tasks are generative forms &
\href{https://arxiv.org/abs/2001.08361}{\textbf{Scaling Laws}}: emergent intelligence, prompt-answer pairs \\
\hline
\textbf{Can Unify All Corresponding Tasks} &
Yes &
Yes \\
\hline
\textbf{Can Subsume Other Domains} &
Yes &
Yes \\
\hline
\textbf{Can Scale Domains and Tasks Multiplicatively} &
Yes &
Yes \\
\hline
\textbf{Zero-Shot/Training-Free/Plug-and-Play/Scalable} &
Yes &
Yes \\
\hline
\textbf{Transcends Original Domain as Almost Entirely New Research} &
Yes &
Yes \\
\hline
\textbf{Chain of Thought/Visual}\tablefootnote{The Chain of Visual refers to the process of transferring within the visual network we have constructed, such as ordinary transfers; complex transfers involving the addition or reduction of input types; and the process of reasoning based on visual evidence (For example, the color brown can be associated with tree bark, roughness, and hardness.), etc.}\textbf{ (CoT/CoV)} &
Yes &
Yes \\
\hline
\end{tabular}
\par\vspace{0.35em}
\parbox{\columnwidth}{\footnotesize \textit{Note.}
According to the \href{https://github.com/World-Snapshot/papers-with-code}{Papers with Code data}, there are 1477 commonly recorded Vision tasks and 858 commonly recorded NLP tasks.}

\end{table}

\noindent\textbf{Physical Holographic Principle}~\cite{maldacena1999large,Fields_2022,HosseinEybposh:20}. Proposed in theoretical physics, it suggests that all the information contained within a volume of space can be encoded on its boundary surface. This principle originated from black hole thermodynamics, where the entropy $S$ of a black hole is proportional to its surface area $A$ rather than its volume:
$S = \frac{k_B A}{4 l_p^2}$
where $k_B$ is the Boltzmann constant and $l_p$ is the Planck length. This implies that a three-dimensional universe might be described by information living on its two-dimensional boundary. 
%The information content of any spatial region is bounded by:
%$I \leq \frac{A}{4l_p^2}$
%where $A$ is the surface area of the region. 
This principle suggests that the complete universe can be fully represented by a lower-dimensional hologram \cite{maldacena1999large}. Put differently, a special image can be no different from a world in one dimension higher, and they are inherently the same.
%our seemingly three-dimensional reality could be a projection from a lower-dimensional surface, similar to how a 2D hologram encodes 3D information. 
%The holographic principle has profound implications for quantum gravity and the fundamental nature of spacetime. This is one of the foundations of our theory. 
Our approach originated from our own inspiration, but it coincides with this theory. To a large extent, it supports our methodology.

\noindent\textbf{Rectified Flow Models}~\cite{yan2024perflowpiecewiserectifiedflow,liu2022rectified,esser2024scalingrectifiedflowtransformers,guo2025variationalrectifiedflowmatching,labs2025flux1kontextflowmatching}. Rectified Flow presents a simple and efficient approach to learning ordinary differential equation (ODE) models for transporting distributions between two empirically observed distributions $\pi_0$ and $\pi_1$~\cite{liu2022flowstraightfastlearning,liu2024instaflowstephighqualitydiffusionbased,lipman2023flowmatchinggenerativemodeling}. The core idea is to learn ODEs that follow straight paths connecting points sampled from the source and target distributions as much as possible. Given samples $X_0 \sim \pi_0$ and $X_1 \sim \pi_1$, the method constructs a linear interpolation:
$$X_t = (1-t)X_0 + tX_1, \quad t \in [0,1]$$
and trains a neural network to predict the velocity field $v(x,t)$ by minimizing the straightness loss:
$$\mathcal{L} = \mathbb{E}_{t, X_0, X_1}\left[\|v(X_t, t) - (X_1 - X_0)\|^2\right]$$
The learned velocity field defines an ODE that transports the distribution:
$\frac{\mathrm{d}Z_t}{\mathrm{d}t} = v(Z_t, t), \quad Z_0 \sim \pi_0$
Straight paths are preferred because they minimize discretization error and enable fast inference with very few Euler steps. The method supports a ``reflow'' procedure that iteratively straightens trajectories, ultimately achieving high-quality generation with a single discretization step. The rectified flow is the main reason why generative models have become more powerful in the past two years. Unlike the classical diffusion \cite{rombach2021highresolution}, this model 
%is learning the operation rather than directly predicting the outcome. It 
predicts the velocity field along the transport trajectory.
%Rectified flow has been successfully applied to large-scale image generation tasks and serves as the foundation for models like Stable Diffusion 3.

\noindent
%多模态联合生成
%我们的做法基于上一段介绍的生成概念进行，因为基于RF/FM的DiT在近些年常常用于更加强大的视觉生成主干。
%首先是Flux、Wan、LTX等开源的强大图像或视频生成模型，他们推动视觉生成的发展。
%然后最近有Google Veo、OmniControl、Vision Banana、UniVidX等模型尝试大规模进行多模态联合生成。
%这些生成方法是利用旧方法进行我们相同目标探索的体现，不过他们受到资源的限制，探索的不够彻底和深入。\footnote{根据第三方系统记录时间线，我们比相关工作早2年开始探索，并且我们远比他们深入，使得我们公开的论文多覆盖50-150倍左右的任务数量}
\noindent\textbf{Multimodal Joint Generation}. Our method builds on the generative formulation discussed above, 
since RF/FM-based DiTs have recently become a common backbone for more powerful visual generation systems:
Open-source image and video generation models such as Flux \cite{labs2025flux1kontextflowmatching,flux-2-2025}, Wan-Video \cite{wan2025}, and LTX \cite{hacohen2024ltxvideorealtimevideolatent,hacohen2025ltx2} have first advanced the capability of visual generation.
More recently, models such as Google Veo \cite{wiedemer2025videomodelszeroshotlearners}, OminiControl \cite{tan2025ominicontrol,tan2026ominicontrol2}, Vision Banana \cite{gabeur2026imagegeneratorsgeneralistvision}, and UniVidX \cite{chen2026unividx} have attempted large-scale multimodal joint generation.
These generative methods can be viewed as attempts to pursue a goal similar to ours; however, due to resource constraints,
their exploration is still not sufficiently thorough or deep.\footnote{According to timelines recorded by third-party systems,
we began exploring this direction two years earlier than related work, and our exploration is substantially deeper, allowing our public papers to cover roughly 50--150 times more tasks.}

%% file: sec_preprint/3_method.tex
\section{Method}
\label{sec:method}

\begin{figure*}[t]
    \centering
    \includegraphics[width=0.98\linewidth]{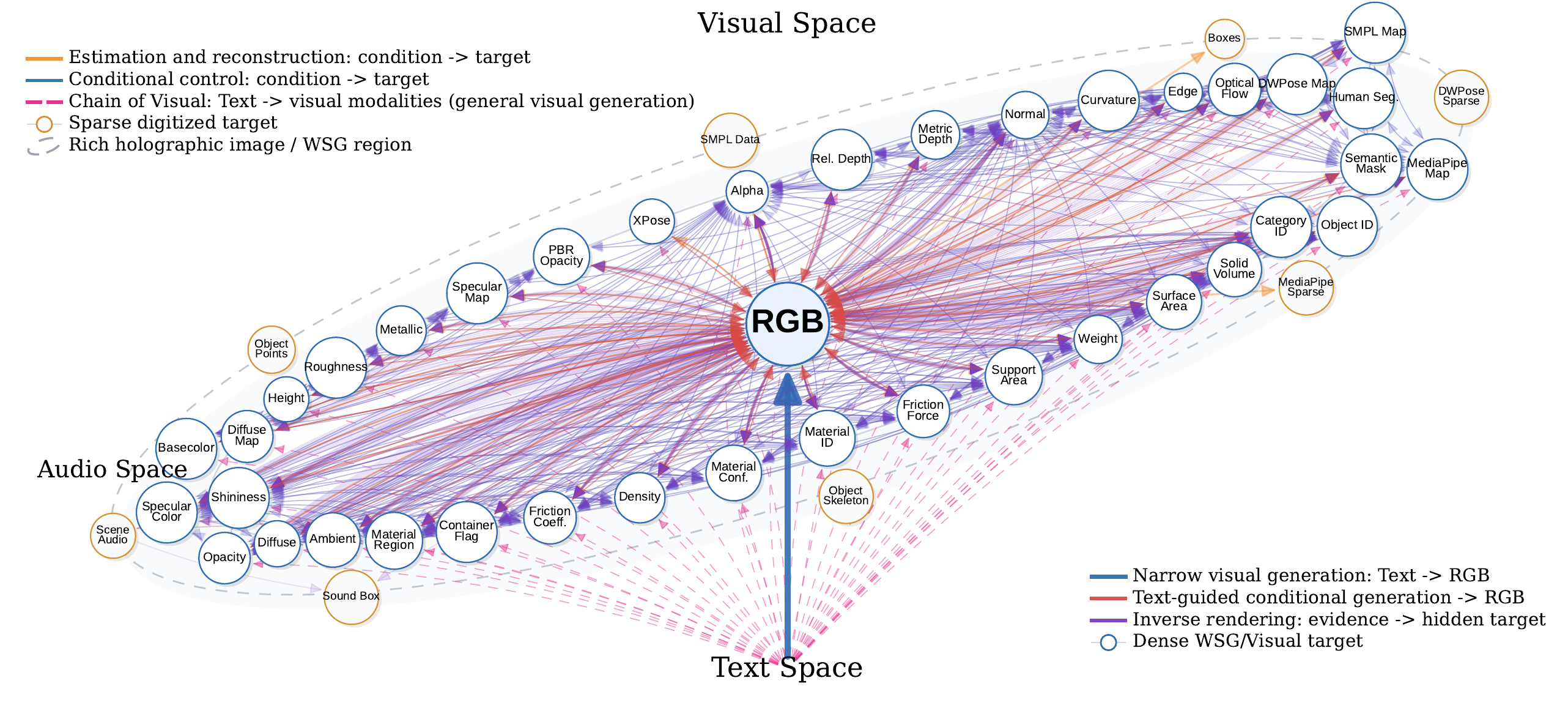}
    \caption{\textbf{WSG task graph and the four generative interfaces.}
    RGB is the center of the visual task space. Text enters from the bottom and
    generates RGB; WSG modalities surround RGB as target or condition nodes.
    Each directed arrow is a visual task realized by a lightweight TunnelingNet
    distributed through the frozen shared generator to steer its trajectory. Arrow color indicates the task type: visual generation,
    estimation/reconstruction, conditional control, text-guided conditional
    generation, or inverse rendering. Dashed arrows denote Chain of Visual,
    where text can reach non-RGB World Snapshot modalities through generated
    World Snapshots. Blue-outline nodes are dense WSG/visual targets, and orange-outline
    nodes are sparse digitized targets such as boxes, body parameters,
    skeletons, keypoints, or audio tokens. Each line represents one task.}
    \label{fig:wsg_task_graph}
\end{figure*}

\subsection{World Snapshot Graphics}
\label{subsec:world_snapshot_graphics}

We introduce World Snapshot Graphics (\href{https://github.com/World-Snapshot/wsg}{.wsg}) as a unified image format for representing World Snapshots.
A .wsg sample is an image-like representation addressed by up to 256 channel IDs, where standard visual appearance occupies part of the representation and auxiliary channels encode richer world attributes. The 256 channels are a management convention: channel IDs form an address space currently capped at 256, and string names would serve equally well. Each task reads and generates only its own channels, never a full 256-channel tensor; all configured tasks currently use 115 channel IDs, so new types of world information can be added without changing the format.
We used datasets including AVIS~\cite{guo2023audio}, Articulation-XL2.0~\cite{song2025magicarticulatemake3dmodels}, ShapeNetSem~\cite{chang2015shapenetinformationrich3dmodel}, and MatSynth~\cite{Vecchio_2024}, %and MMFi \cite{yang2023mm},
and processed this data using our developed WSG dependency library, extracting information from RGB estimation and then injecting the RGB and additional information into WSG files.

These components contain at least the following information respectively:
AVIS includes Depth and Spatial Geometry Domain~\cite{guo2023audio} (relative depth~\cite{Zavadski_2024_ACCV} + absolute depth~\cite{Bochkovskii2024:arxiv} + normal~\cite{Bin_2025_ICCV} + curvature + edge~\cite{4767851} + optical flow~\cite{ilg2016flownet20evolutionoptical}), Pose and Human Information Domain~\cite{yang2023effective,lugaresi2019mediapipeframeworkbuildingperception,arbol2024bodyshapegptsmplbodyshape} (dwpose + mediapipe + smpl + seg), and Sound And Language Properties Domain (Original Audio + Description Audio + Semantic Segmentation).
Articulation-XL2.0 has Object/Animal Information Domain~\cite{Song_2025_CVPR} (category + vertex + mesh + rig + uv + Joint).
ShapeNetSem, in addition to similar basic obj information as the previous one, has Physics Properties Domain~\cite{shapenet2015,savva2015semgeo} (relative solidVolume + relative surfaceVolume + relative weight + supportSurfaceArea + relative staticFrictionForce + material\_id + material\_confidence + density + staticFrictionCoeff + isContainerLike + Object category id) and Tradition and Control Optics of Material \& Light Properties Domain (Material Name + Ambient light color + Diffuse reflection color + Opacity + Specular reflection color + hyperspectral index).
MatSynth has PBR Material of Material \& Light Properties Domain~\cite{Vecchio_2024} (Basecolor + Diffuse + Normal + Height + Roughness + Metallic + Specular + Opacity).
%MMFi has Wireless Sensing of Abstract Information Domain~\cite{yang2023mmfimultimodalnonintrusive4d,bhat2025mmhsensemultimodaldistributedmmwave,yang2023lidarllmexploringpotentiallarge} (LiDAR + mmWave + WiFi).
The processed datasets are available at \href{https://huggingface.co/World-Snapshot/datasets}{WSM Datasets}.
%We will use this information for training capabilities including generation, estimation, conditional control, and reconstruction.

\subsection{Four Generative Interfaces of Training}

We organize WSG training as a directed graph centered at RGB. Text first enters
visual space through $\mathrm{text}\to\mathrm{RGB}$. Around RGB, every WSG
modality is a node, and every directed edge is one visual task realized by a
TunnelingNet. Each TunnelingNet is a lightweight capability network distributed
through multiple layers inside the frozen generative backbone; training it
redirects the backbone's generation trajectory toward one task without modifying
the shared model.
As shown in Figure~\ref{fig:wsg_task_graph}, the released dense WSM graph
configures 3,042 visual task mappings around one shared generative backbone.
The benchmark in this paper provides trained models and end-to-end inference results on the evaluation set for 1,148 dense TunnelingNets; accordingly,
1,148 is the task count reported for WSM in the main paper. Sparse digitized
tasks are evaluated separately in Table~\ref{tab:native_sparse_heads} and are
not included in this count. Extending evaluation to the remaining dense mappings requires no change to the unified interfaces or
model architecture.\footnote{If ResNet enables models to be trained especially deep, even with thousands of layers, our TunnelingNet enables models to be trained on thousands of visual tasks.}

We use a pretrained text-and-image-to-image (TI2I) model as the starting point.
Such models are commonly referred to as image-editing models and have already
acquired the ability to take text and RGB as inputs and produce an image as
output. The backbone is a DiT, so text, images, and conditions are all converted
into tokens and concatenated into one input sequence. Based on our exploration, almost all visual tasks can be realized
through four interfaces. An important question then arises: should RGB be sent
to the native reference-image input for the model to modify it, or should it be
used only as evidence/condition? We finally define the four interfaces as
follows:

\numtag{A}\hspace{0.55em}\textbf{RGB-to-World-Information estimation and reconstruction.} This interface
takes the currently observed RGB image as input; its target is World Information stored in channels
additional to RGB, i.e., the non-RGB channels of WSG.
This interface covers depth, normal, curvature, edge, optical flow, pose maps,
semantic masks, material maps, physical properties, human structure, object
structure, and visual-to-audio reconstruction. In the default dense graph, this
interface uses only the current RGB image as condition:
$\mathcal{F}_{\mathrm{estimation}}: \mathbf{x}^{\mathrm{current}}_{\mathrm{RGB}} \rightarrow \mathbf{x}^{\mathrm{target}}_{\mathrm{world}}$.

\numtag{B}\hspace{0.55em}\textbf{Text-and-World-Information narrow visual generation.} Text supplies
the semantic intent; the World Information condition supplies the structural
control; and a reference RGB frame, placed in the input sequence,
supplies the appearance control; during training, it is the first frame of the
same video as the target frame. The corresponding non-RGB WSG channel enters a WSM condition lane, while the first
RGB frame enters the native reference slot of the backbone. The target is the
current RGB image, as in text-and-pose-to-video. In
short, $\mathcal{F}_{\mathrm{narrow\ generation}}: (\mathbf{y}^{\mathrm{text}}, \mathbf{x}^{\mathrm{condition}}_{\mathrm{world}}, \mathbf{x}^{\mathrm{first}}_{\mathrm{RGB}}) \rightarrow \mathbf{x}^{\mathrm{current}}_{\mathrm{RGB}}$.
And broad generation can be achieved through visual chains: ($\mathcal{F}_{\mathrm{broad\ generation}}: \mathbf{y}^{\mathrm{text}} \rightarrow \mathbf{x}^{\mathrm{current}}_{\mathrm{RGB}} \rightarrow \mathbf{x}^{\mathrm{target}}_{\mathrm{world}}$).

\numtag{C}\hspace{0.55em}\textbf{World-Information-to-World-Information conditional control.} Generated or
observed World Information becomes the condition, and other World Information
becomes the target (e.g., Edge to Normals):
$\mathcal{F}_{\mathrm{conditional\ control}}: \mathbf{x}^{\mathrm{condition}}_{\mathrm{world}} \rightarrow \mathbf{x}^{\mathrm{target}}_{\mathrm{world}}$.

\numtag{D}\hspace{0.55em}\textbf{First-RGB-frame-plus-World-Information inverse rendering.} A fixed first RGB
frame gives the scene-level visual evidence, while observed World Information
from the target time helps infer hidden World Information: shape,
material, pose, audio source, or another world property. In the default dense
graph, the RGB token $0$--$2$ is routed through the first-frame condition branch,
and \texttt{Reference\_IMG} is not used  (e.g., First RGB Frame as Condition and Normals to Edge). In short,
$\mathcal{F}_{\mathrm{inverse\ rendering}}: (\mathbf{x}^{\mathrm{first}}_{\mathrm{RGB}}, \mathbf{x}^{\mathrm{observation}}_{\mathrm{world}}) \rightarrow \mathbf{x}^{\mathrm{hidden}}_{\mathrm{world}}$.

Sparse-output TunnelingNets use the same input--output task definition as dense TunnelingNets, but need specialized output modules when the target is a list, box set, skeleton, body parameter, or audio
token sequence rather than a dense image-like map. In this way, dense visual
channels and sparse digitized structures are both treated as endpoints in the
same generative WSG graph.

The Four Laws and the four interfaces describe WSM at complementary levels.
The Four Laws organize the conceptual relationships between RGB and World
Information into four broad categories. The four interfaces form the minimal
operational basis shared by all visual tasks and provide their training
schemes. Their many-to-many correspondence is explicit: \numtag{B} and
\numtag{B}$\rightarrow$\numtag{A} support
narrow and broad Lex I; \numtag{A} supports Lex II; \numtag{B} and \numtag{C}
support the RGB- and World-Information-target forms of Lex III; and \numtag{A}
and \numtag{D} support reconstruction and inverse rendering under Lex IV.
Image classification, localization, detection,
segmentation, estimation, reconstruction, restoration, image and shape completion, image and point-cloud registration,
pixel- and feature-level correspondence, tracking, retrieval, visual-database
indexing, prediction, generation, editing,
rendering, simulation, planning, and control all fall within these four
categories.

\begin{algorithm}[t]
    \caption{Constructing and learning a dense modality-task edge}
    \label{alg:modality_task_construction}
    \footnotesize
    \begin{algorithmic}[1]
        \Require edge $\mathcal{H}_k$, paired observations $\mathcal{D}_k$, shared generator $f_\theta$
        \Ensure task residual $\phi_k^{\star}$
        \State $\rho_k\gets\Gamma(\mathcal{H}_k)$ \Comment{assign semantic roles}
        \State Associate boundary maps $\{\mathcal{E}_m:m\in\mathcal{C}_k\cup\mathcal{T}_k\}$
        \State Couple task residual $\phi_k$ to the shared field $f_\theta$
        \For{$(\mathbf{x}_{\mathcal C_k},\mathbf{x}_{\mathcal T_k},\mathbf{p})\sim\mathcal{D}_k$}
            \State $\mathbf{z}_0\gets\mathcal{E}_{\mathcal T_k}(\mathbf{x}_{\mathcal T_k})$
            \State $(\mathbf{z}_{\mathcal C_k},\mathbf{r},\mathbf{p})\gets\Gamma_{\rho_k}(\mathbf{x}_{\mathcal C_k},\mathbf{p})$
            \State Sample $t$ and $\epsilon\sim\mathcal{N}(0,I)$
            \State $\mathbf{z}_t\gets(1-t)\mathbf{z}_0+t\epsilon$
            \State $\mathbf{v}^{\star}\gets\epsilon-\mathbf{z}_0$
            \State $\widehat{\mathbf{v}}\gets f_{\theta,\phi_k}(\mathbf{z}_t,t,\mathbf{z}_{\mathcal C_k},\mathbf{r},\mathbf{p};\rho_k)$
            \State $\mathcal{L}_k\gets\lVert\widehat{\mathbf{v}}-\mathbf{v}^{\star}\rVert_2^2$
            \State $\phi_k\gets\phi_k-\eta\nabla_{\phi_k}\mathcal{L}_k$
        \EndFor
        \State \Return $\phi_k^{\star}$
    \end{algorithmic}
\end{algorithm}

\subsection{Modality Architectures}
\label{subsec:modality_architectures}

A WSG mapping begins with a typed, directed relation that assigns every
participating modality a semantic role:
\begin{equation}
    \mathcal{H}_k=(\mathcal{C}_k,\mathcal{T}_k,\tau_k),
\end{equation}
where $\mathcal{C}_k$ and $\mathcal{T}_k$ denote the condition and target
modality sets and $\tau_k$ denotes one of the four generative interfaces. This
relation induces a role assignment
\begin{equation}
    \rho_k=\Gamma(\mathcal{H}_k)
    =(R_p,R_t,R_c,R_r,a_k),
\end{equation}
which partitions information into prompt tokens $R_p$, a noised target stream
$R_t$, clean modal evidence $R_c$, and an optional RGB appearance prior $R_r$;
$a_k$ selects the interaction rule between these roles. The construction
therefore determines both the direction of prediction and the information
available to it.

Each modality $m$ has a boundary map $\mathcal{E}_m$ into the model's token
space. Dense image-like fields and spatial surrogates share the normalized VAE
boundary, while native sparse structures and audio use the compact adapters
described below. After this boundary mapping, the learning principle is shared:
the target follows a stochastic rectified-flow trajectory, whereas modal
evidence remains clean. This clean-evidence/noised-target asymmetry turns a
heterogeneous set of modalities into a directed generative problem.
Algorithm~\ref{alg:modality_task_construction} summarizes the construction for
a dense task edge.

\subsection{TunnelingNet}

Around an RGB-based generative backbone, the WSG graph represents visual tasks
as directed condition--target edges. A scalable WSM must add edges without
replicating the full generator. TunnelingNet
therefore factorizes learning into a shared World Snapshot generator and
lightweight capability networks distributed throughout the frozen backbone. Given the typed edge $\mathcal{H}_k$
and its role assignment $\rho_k$, TunnelingNet models the transformation as
\begin{equation}
    p_{\theta,\phi_k}\left(\mathbf{x}_{\mathcal{T}_k}
    \mid \mathbf{x}_{\mathcal{C}_k}, \mathbf{p}\right),
\end{equation}
where $\theta$ is a frozen rectified-flow visual generator, $\phi_k$ contains the
trainable parameters of TunnelingNet $k$, and $\mathbf{p}$ is the text prompt or
modality prompt. The same backbone therefore supports many directed edges, and
the marginal parameter cost of a new edge is the cost of its tunnel. WSM is one
modular model comprising the shared backbone $\theta$, a resident task-indexed
set of TunnelingNets $\{\phi_k\}$, and an explicit task router. TunnelingNets can be
optimized independently and physically sharded for parallel training and
distribution; these are training and storage choices, not model boundaries.
At inference, a task key activates one TunnelingNet, while a visual chain
activates a selected TunnelingNet sequence. This yields one WSM with many generative
paths.

A useful intuition is that TunnelingNet internalizes ControlNet: instead of
attaching a large copied control branch outside the generator, it places
low-rank residual steering inside every transformer block. MTA gives condition evidence its own independently normalized path
rather than mixing it with target tokens in native attention. These layer-wise
pushes redirect the full generation trajectory while leaving the backbone and
evidence streams frozen, limiting disturbance to the native prior. Under
matched trainable parameters, this routing reaches the selected structured
outputs with $19.0$--$38.5\times$ fewer updates and recognizable structure
$77$--$100\times$ earlier than LoRA NJA.

\begin{figure}[t]
    \centering
    \includegraphics[width=0.98\linewidth]{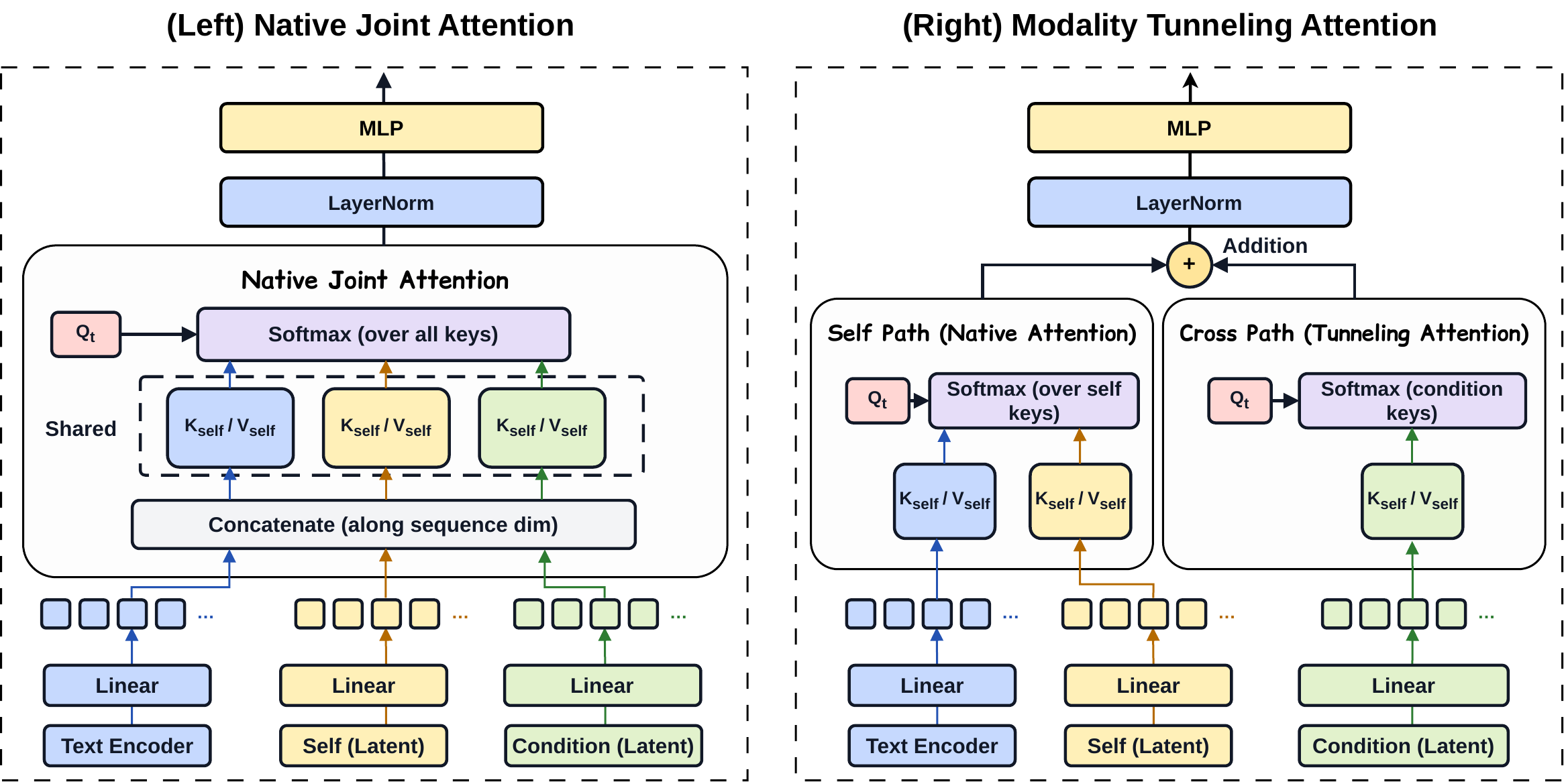}
    \caption{\textbf{Two attention routes in TunnelingNet.}
    (Left) \numtag{B} retains NJA, where text, target, condition, and
    RGB evidence share one joint attention field.
    (Right) Interfaces \numtag{A}, \numtag{C}, and \numtag{D} use MTA, which
    preserves target self-attention and adds an independently normalized route
    from conditions to the target.}
    \label{fig:tunnelingnet_two_paths}
\end{figure}

As shown in Figure~\ref{fig:tunnelingnet_two_paths}, a TunnelingNet separates what is learned from how evidence is routed. Its
trainable component $\phi_k$ is a lightweight target-side tunnel, whereas its
attention topology determines how clean evidence enters the noised target
lane. For estimation/reconstruction, conditional control, and inverse
rendering (Interfaces \numtag{A}, \numtag{C}, and \numtag{D}), we introduce
\emph{Modality Tunneling Attention} (MTA). MTA decomposes the target update into
self-attention over prompt and target tokens and an independently normalized
target-to-condition path:
\begin{equation}
    \begin{aligned}
    \mathrm{Attn}_{\mathrm{tun}}(Q_t)
    &= \mathrm{Attn}(Q_t,K_{\{p,t\}},V_{\{p,t\}}) \\
    &\quad + \mathrm{Attn}(Q_t,K_{\mathcal{C}_k},V_{\mathcal{C}_k}).
    \end{aligned}
\end{equation}
Here, $\mathcal{C}_k$ includes every active WSG condition and any RGB or
reference observation routed as visual evidence. The evidence lanes themselves
remain on the frozen native transformer path. MTA introduces no additional
parameters; its separately normalized condition-to-target route allows a small
target-side tunnel to learn heterogeneous non-RGB mappings efficiently.

We do not replace the native attention topology universally. For text-guided
narrow visual generation (Interface \numtag{B}), whose target is RGB, we retain
\emph{Native Joint Attention} (NJA). NJA places text, target, condition, and the
RGB evidence in one joint attention field, allowing RGB synthesis to reuse the
backbone's image-generation prior.

%TunnelingNet uses lane-wise routing. Each batch is packed into separate streams: text, target, optional WSG condition, optional current-frame RGB condition, optional first-frame RGB condition, optional native RGB reference, and optional audio condition.
%The target stream carries the noised target latent. Condition and reference streams carry clean evidence. A task recipe decides which streams activate the Channel-Head tunnel and which streams remain on the frozen backbone.
%In the default recipes, the target lane is trainable for all dense WSG heads, while condition lanes are frozen so that shared conditions can be reused across many heads.

In our implementation, $\theta$ is initialized from FLUX.2-klein-4B~\cite{flux-2-2025}. The text
encoder, VAE, and base transformer weights are frozen. Each TunnelingNet is
implemented as a small target-side tunnel
placed on the transformer input embedding, modulation layers, attention
projections, and feed-forward projections. Its trainable parameters are
therefore
\begin{equation}
    \phi_k = \{\Delta W^{(k)}_{\mathrm{embed}},
    \Delta W^{(k)}_{\mathrm{mod}},
    \Delta W^{(k)}_{\mathrm{attn}},
    \Delta W^{(k)}_{\mathrm{ffn}}\},
\end{equation}
and optimization updates only $\phi_k$:
\begin{equation}
    \theta \leftarrow \theta, \qquad
    \phi_k \leftarrow \phi_k - \eta \nabla_{\phi_k}\mathcal{L}_k .
\end{equation}
When the condition is an audio feature, the TunnelingNet also contains a small
audio projection module that maps audio embeddings into the visual token space.

Training follows rectified flow in the latent space. Given a target latent
$\mathbf{z}_0$ and noise $\epsilon$, we sample $t$ and construct
\begin{equation}
    \mathbf{z}_t=(1-t)\mathbf{z}_0+t\epsilon, \qquad
    \mathbf{v}^{\star}=\epsilon-\mathbf{z}_0 .
\end{equation}
TunnelingNet predicts the velocity from the noised target lane and clean
condition lanes:
\begin{equation}
    \mathcal{L}_k=
    \mathbb{E}\left[\left\|
    f_{\theta,\phi_k}(\mathbf{z}_t,t,
    \mathbf{z}_{\mathcal{C}_k},\mathbf{r},\mathbf{p})
    -\mathbf{v}^{\star}\right\|_2^2\right].
\end{equation}
For deterministic dense targets, the implementation supports low-noise timestep
shifting, Min-SNR weighting, Huber loss, high-noise anchoring, multi-resolution
noise, and single-step clean-latent prediction. The default recipe uses
high-noise anchoring for the four dense task types and keeps the other objective
shaping switches off unless enabled. The task interface remains the
same condition--target edge.

\begin{figure*}[t]
    \centering
    \includegraphics[width=0.98\textwidth]{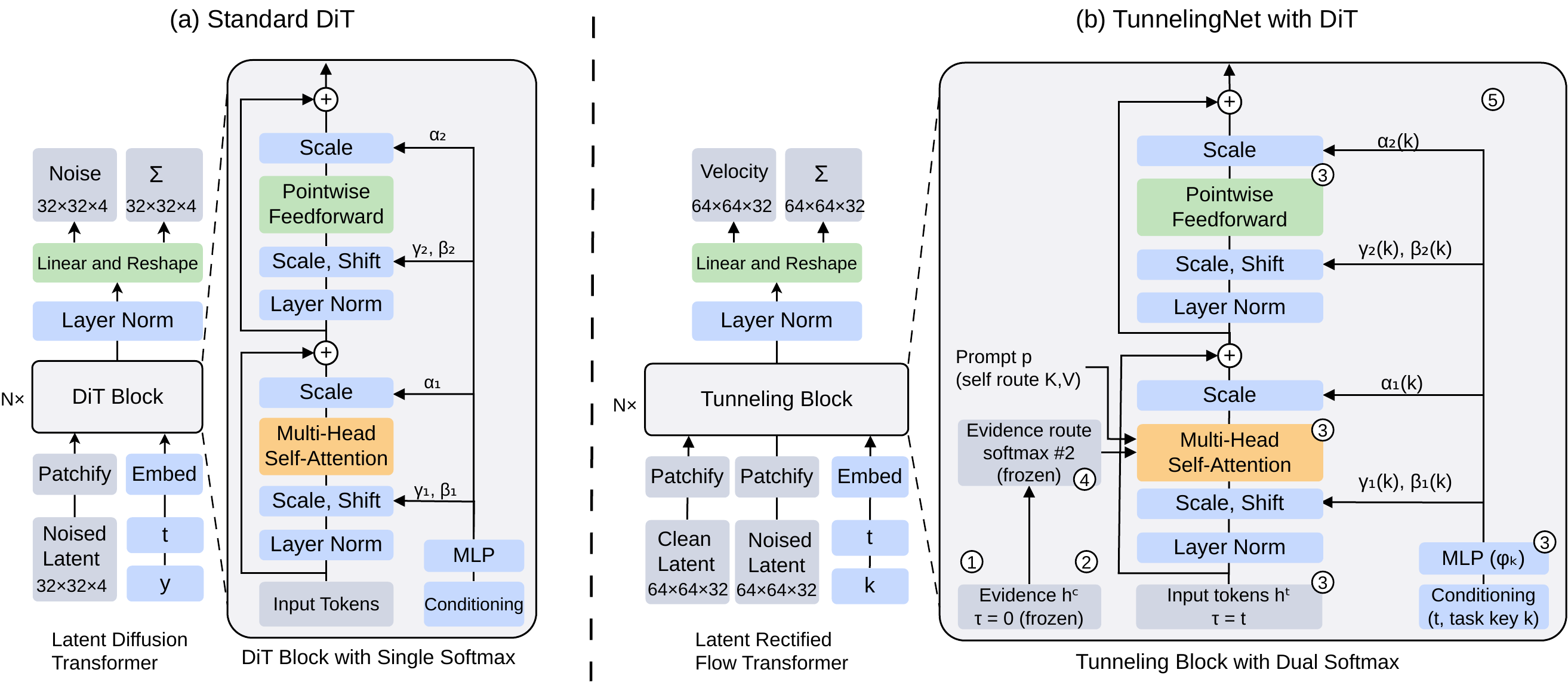}
    \caption{\textbf{From a standard DiT block to the Tunneling Block.}
    Panel \textbf{(a)} shows the standard DiT block, and panel \textbf{(b)} its
    task-indexed conversion. \circlednote{1} Target, evidence, and prompt tokens
    receive distinct computational roles. \circlednote{2} The target follows
    flow time $\tau=t$, whereas clean evidence remains fixed at $\tau=0$.
    \circlednote{3} The task key $k$ selects TunnelingNet $\phi_k$, whose
    task-specific updates cover the input embedding, conditioning MLP,
    attention projections, and pointwise feed-forward projections.
    \circlednote{4} In MTA, target tokens provide queries $Q_t$; prompt and
    target tokens provide keys and values
    $K_{\{p,t\}},V_{\{p,t\}}$ for the self route, while evidence provides keys
    and values $K_{\mathcal C_k},V_{\mathcal C_k}$ for a second,
    independently normalized Softmax route. \circlednote{5} The same pattern
    repeats through all $N$ Tunneling Blocks. Together, these changes reuse and
    extend a generative backbone into heterogeneous modality spaces with almost
    zero inter-task interference.}
    \label{fig:tunneling_block}
\end{figure*}

\subsection{Deep Tunneling Learning}
\label{subsec:deep_tunneling_learning}

\paragraph{Why Deep Tunneling Learning Works.}
Figure~\ref{fig:tunneling_block} summarizes how Deep Tunneling Learning extends
a standard DiT block while preserving its shared residual backbone. In
Figure~\ref{fig:tunneling_block}(a), the conditioning MLP produces the
adaptive-LayerNorm scale and shift $(\gamma_i,\beta_i)$ and residual gates
$\alpha_i$ for the attention and pointwise feed-forward branches; the plus
nodes denote ordinary residual addition. The two panels also retain the native
objectives of their respective backbones: the standard latent-diffusion example
predicts noise, while the FLUX.2 rectified-flow backbone predicts velocity. The
$64\!\times\!64\!\times\!32$ tensors in panel (b) are FLUX.2 VAE latents for a
$512\!\times\!512$ image. Latent shapes vary with the VAE rather than with the
Tunneling Block; the standard-DiT example in panel (a) uses the smaller
$32\!\times\!32\!\times\!4$ representation.

The method rests on the premise that heterogeneous visual
modalities do not describe independent worlds. They expose different
observations of the same underlying world state. A new modality mapping
therefore need not reconstruct the visual generator; it needs a task-specific
path that redirects the shared generative dynamics toward the target
representation.

\paragraph{Why modalities differ even when they describe the same world.}
Consider two observers viewing the same object, one with typical color vision
and the other with red--green color blindness. Their sensory representations
differ, yet both remain grounded in the same object geometry, composition, and
world state. Neither observer must reproduce the other's perception before
using evidence supplied by the other. We express this shared-world hypothesis
as
\begin{equation}
    \mathbf{x}_m=g_m(\mathbf{s}), \qquad m\in\mathcal{M},
\end{equation}
where $\mathbf{s}$ is the latent world state and $g_m$ is the observation map
of modality $m$. WSG makes these heterogeneous observations addressable within
one typed task graph, while the shared generator supplies their common visual
prior.

\paragraph{How lightweight tunnels redirect generation.}
TunnelingNet preserves the representation learned by the shared generator and
specializes its dynamics through a task-indexed residual field inside every
transformer stage. For layer $\ell$ and module type
$m\in\{\mathrm{embed},\mathrm{mod},\mathrm{attn},\mathrm{ffn}\}$, the effective
linear map of task $k$ is
\begin{equation}
    W_{\ell,m}^{(k)}\mathbf{h}
    =W_{\ell,m}\mathbf{h}
    +\frac{\alpha_k}{r_k}
    B_{\ell,m}^{(k)}A_{\ell,m}^{(k)}\mathbf{h}.
\end{equation}
The shared term $W_{\ell,m}$ preserves the visual prior and a common coordinate
system across modalities. The low-rank term changes only the vector field
followed by the selected target modality. Under the shared-world hypothesis, a
new task can therefore learn how to move through the common representation
instead of learning that representation again.

\paragraph{How residual corrections accumulate across layers.}
For MTA, let $\mathbf{h}_{\ell}^{t}$ denote target tokens and
$\mathbf{h}_{\ell}^{\mathcal C}$ denote the clean evidence tokens at block
$\ell$. Their interaction can be written as two residual components:
\begin{equation}
    \mathbf{h}_{\ell+1}^{t}
    =\mathbf{h}_{\ell}^{t}
    +R_{\ell,k}^{\mathrm{self}}
      (\mathbf{h}_{\ell}^{t},\mathbf{p})
    +R_{\ell,k}^{\mathrm{evidence}}
      (\mathbf{h}_{\ell}^{t},\mathbf{h}_{\ell}^{\mathcal C}).
\end{equation}
The self residual advances the target along its generative trajectory, while
the evidence residual corrects that trajectory using independently normalized
observations. Repeating this correction through the transformer depth turns
local updates into a complete modality tunnel: boundaries can constrain
geometry, geometry can constrain surface orientation, and RGB evidence can
constrain otherwise ambiguous world properties. Interface \numtag{B} retains
the same task residuals but absorbs both interactions into NJA, preserving the
native RGB-generation prior.

\paragraph{Shared backbone knowledge and task-specific learning.}
This construction separates two responsibilities. Frozen evidence pathways
retain reusable observations in the shared feature space, while the selected
target tunnel learns how those observations should alter one particular output
modality. Deep Tunneling Learning consequently acquires new modality relations
as sequences of small, task-indexed corrections while keeping the shared
generator and previously learned tunnels unchanged.

\subsection{Categorical Route Output}
\label{subsec:categorical_refinement}

For semantic categorical targets, a TunnelingNet includes an integrated
categorical output path. The rectified-flow branch predicts spatial support,
while a lightweight classification branch assigns the exact semantic palette
value to each generated region. These branches define one deployable TunnelingNet; the
intermediate pseudo-color field is an internal representation rather than a
separate model output. Continuous fields and edge maps use only the flow path.

Let $\hat{\mathbf{x}}_k$ be the spatial field produced inside categorical
TunnelingNet $k$, and let $\{R_j\}$ denote its foreground connected components.
For each region, we mask the encoded condition used by the task:
\begin{equation}
    \mathbf{h}_j =
    \operatorname{Concat}(\mathbf{z}_{\mathcal C_k})
    \odot \operatorname{Resize}(R_j).
\end{equation}
A compact spatial encoder pools $\mathbf{h}_j$ into a region embedding, which is
compared with learned category prototypes using cosine similarity:
\begin{equation}
    \ell_{j,c} =
    \exp(s)\left\langle
    \overline{e_\psi(\mathbf{h}_j)},\overline{\mathbf{p}_c}
    \right\rangle .
\end{equation}
During training, regions and labels are obtained directly from the ground-truth
categorical map. We use class-balanced cross-entropy and select the classifier
state with the highest validation balanced accuracy. At inference, regions are
extracted from the generated spatial field, classified independently, and
written with the exact palette value of the predicted category. This naturally
supports multiple objects and categories in one image.

The complete categorical prediction therefore preserves the generative
localization while resolving category identity from the task condition. Its
classification parameters are stored with the corresponding TunnelingNet run
and loaded automatically by evaluation and inference. For a categorical target,
both parts are required; a checkpoint without its classification parameters is
not treated as a deployable categorical TunnelingNet.

\subsection{Sparse-Structure Modeling}
\label{subsec:sparse_structure_modeling}

We find that dense and sparse prediction need not form two independent task
systems. Their difference can instead be confined to the representation
boundary of a TunnelingNet. When a sparse structure admits a spatial surrogate,
we encode it as an image-like field and retain the same latent-flow learning
path used by dense WSG modalities. For example, keypoint tuples are converted
to semantic Gaussian heatmaps, while instance boxes are represented as filled
pseudo-color regions. After generation, peak extraction, non-maximum
suppression, or connected-component analysis recovers the original sparse
structure.

Some outputs, such as body parameters, object skeleton trees, and audio tokens,
do not have a natural dense representation. These tasks use compact
structure-specific adapters instead. Dense visual evidence is converted only at
the adapter boundary---for example into full-image features, person-crop
features, or masked modality features---and the predicted structure is written
back into its designated WSG channels. Consequently, a dense WSG prediction can
condition a sparse task, and a reconstructed sparse result can re-enter the same
world representation without introducing a separate task ontology.

The shared generative backbone is initialized from FLUX.2-klein-4B, while its
transformer, text encoder, and VAE remain frozen. Dense tasks update only their
target-side tunnels using AdamW in bfloat16. Training and visualization share a
portable set of normalization rules, which are inverted before values are written
back to WSG. Categorical output branches are trained for 3,000 updates,
validated every 50 updates, and retain only the checkpoint with the best
balanced accuracy.

%% file: sec_preprint/4_ablation.tex
\makeatletter
\newcommand{\wsmnewphysicalpage}{%
  \if@firstcolumn
    \newpage\null\newpage
  \else
    \newpage
  \fi}
\makeatother

\begin{table*}[t]
\centering
\caption{\textbf{Final optimal configuration selected by our ablation evaluation.} MTA and NJA denote Modality Tunneling Attention and Native Joint Attention, respectively; RGB$_c$ and RGB$_1$ denote current- and first-frame RGB. All four TunnelingNets train only the target tunnel, freeze the evidence lanes, and use high-noise anchoring without additional dense-objective shaping.}
\label{tab:selected_ablation_recipe}
\footnotesize
\setlength{\tabcolsep}{7pt}
\renewcommand{\arraystretch}{1.12}
\begin{tabular}{@{}c l l c l@{}}
\hline
\textbf{TunnelingNet} & \textbf{Interface} & \textbf{Condition $\rightarrow$ target} & \textbf{Attention} & \textbf{RGB evidence route} \\
\hline
003 & Estimation/reconstruction & RGB$_c$ $\rightarrow$ Normal$_c$ & MTA & RGB$_c$ condition \\
270 & Narrow generation & Text + Normal$_c$ + RGB$_1$ $\rightarrow$ RGB$_c$ & NJA & Native reference \\
275 & Conditional control & Normal$_c$ $\rightarrow$ Edge$_c$ & MTA & None \\
313 & Inverse rendering & RGB$_1$ + Normal$_c$ $\rightarrow$ Edge$_c$ & MTA & RGB$_1$ condition \\
\hline
\end{tabular}
\end{table*}

\wsmnewphysicalpage

\definecolor{wsmUnavailable}{HTML}{C5CBD3}
\DeclareRobustCommand{\wsmunavailablekey}{%
  \raisebox{-0.08em}{\textcolor{wsmUnavailable}{\rule{0.78em}{0.78em}}}\,\textup{N/A}}

\section{Ablation Evaluation}
\label{sec:ablation}

We organize the experimental study into three main sections. This section
establishes the final TunnelingNet design through controlled ablations;
Section~\ref{sec:quantitative} measures large-scale task coverage and
efficiency; and Section~\ref{sec:qualitative} examines model outputs across
the four generative interfaces.

Our ablation evaluation has three primary objectives: \textbf{(I) Attention
Topology}, determining how target and condition tokens should interact;
\textbf{(II) Evidence Routing and Tunnel Placement}, establishing how
condition/reference evidence should enter the model and which tunnels require
adaptation; and \textbf{(III) Rectified-Flow Objective Shaping}, identifying an
objective formulation that remains effective across heterogeneous output
modalities. Together, these objectives determine the architecture, evidence
pathways, and optimization recipe of the final TunnelingNet.

To ensure that the methods selected for these three objectives are broadly
effective across task types rather than specialized to a single mapping, we
choose four representative TunnelingNets from our broader task suite, with one
task from each interface: RGB-to-Normal estimation (TunnelingNet 003),
text-and-Normal-to-RGB narrow generation (TunnelingNet 270), Normal-to-Edge conditional
control (TunnelingNet 275), and first-RGB-plus-Normal-to-Edge inverse rendering
(TunnelingNet 313). Table~\ref{tab:selected_ablation_recipe} summarizes these four
representative TunnelingNets and the final optimal configuration identified by our
three-part ablation evaluation. Together, these tasks span dense geometric and photorealistic RGB
outputs, sparse structural outputs, and distinct uses of current observations
and first-frame visual evidence. We apply each ablation to the representative
TunnelingNets for which the corresponding design choice is applicable. Agreement across
interfaces supports a generally effective choice, whereas consistent differences
identify where interface-specific treatment is necessary. Unless stated
otherwise, the FLUX.2-Klein-4B backbone, text encoder, and VAE are frozen; each TunnelingNet
activates only 12.5M target-side tunnel parameters, or approximately 0.32\% of
the 3.9B-parameter visual backbone. We use the same data split and fixed
validation set within each controlled sweep. Loss values across different
target modalities are reported to establish convergence, not to rank the
intrinsic difficulty of the modalities.

\subsection{Attention Topology}
\begin{figure*}[t]
    \centering
    \includegraphics[width=0.98\textwidth]{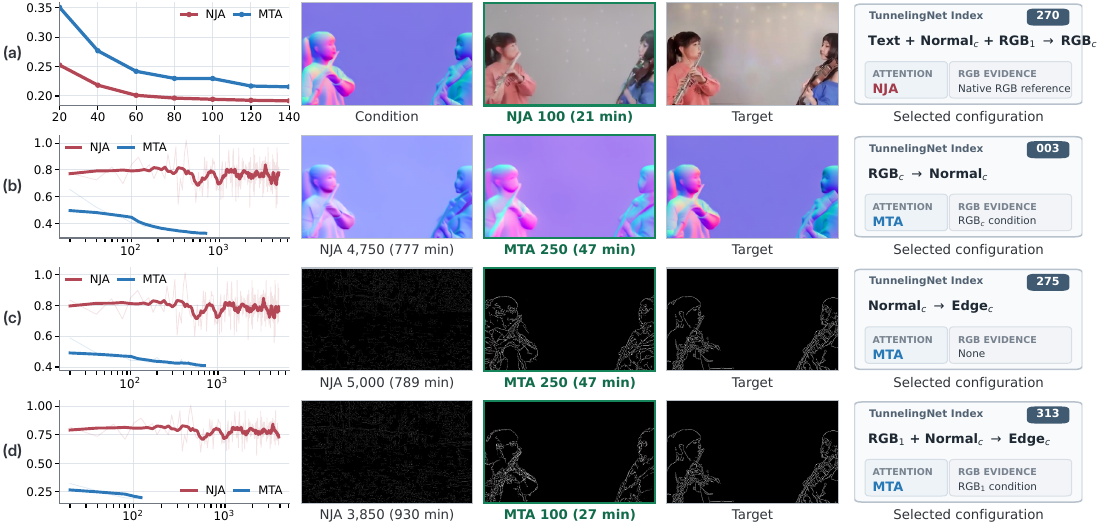}
    \caption{\textbf{Attention ablation and selected TunnelingNet configurations.}
    Rows show (a) narrow generation, (b) estimation/reconstruction,
    (c) conditional control, and (d) inverse rendering. Curves plot validation
    MSE (lower is better) against optimizer updates; thin/thick traces denote
    measurements/nine-point moving averages, and (b--d) use logarithmic
    horizontal axes. Under matched parameters, data, optimizer, batch size, and
    schedule, NJA is preferred for the RGB target. For the three structured
    non-RGB targets, MTA produces clean, recognizable outputs in 27--47 minutes
    with $19.0$--$38.5\times$ fewer updates, while NJA remains pale or fragmented
    after 777--930 minutes; recognizable structure appears $77$--$100\times$
    earlier. These endpoint time gaps of $16.6$--$34.1\times$ are conservative
    because NJA has not reached comparable quality. The legacy NJA previews in
    (c,d) use different evaluation examples; all Edge maps use the same
    grayscale reader. The rightmost cards summarize the deployed mapping,
    attention topology, and RGB-evidence route.}
    \label{fig:ablation_validation_curves}
\end{figure*}

In this ablation, NJA uses the same target-side low-rank parameterization as MTA
and serves as our LoRA native-attention baseline. MTA changes only the
attention routing; the trainable parameter count, data, optimizer, batch size,
and training schedule remain fixed.
Native Joint Attention performs one softmax over the pooled text, target,
condition, and reference keys. In a controlled TunnelingNet 270 comparison at 140
updates, NJA obtained a validation MSE of 0.1917, compared with 0.2155 for MTA,
an 11.0\% reduction. We therefore retain NJA for narrow visual generation.
Figure~\ref{fig:ablation_validation_curves}(a) pairs the directly recorded
validation curves with the condition, selected NJA output, and target at
update 100. This result reflects the different requirements for conditioning
information in RGB generation and other modality estimation tasks. Modality
Tunneling Attention aims to limit the influence of the input RGB on the target
generation path, allowing conditioning information to act on the target through
an independent one-way channel. This mechanism is suitable for estimation,
simple conditional control, and inverse-rendering tasks because it can preserve
the target modality's own generation process and prevent the input evidence
from dominating. However, conditional RGB generation typically requires edges,
normals, or other conditions to make larger changes to the structure and
content of the reference RGB, and therefore requires fuller interaction among
the condition, text, reference image, and target sequences. Native Joint
Attention places these sequences in the same attention space, allowing every
token to influence the entire generation sequence and thereby enabling the
conditioning information to globally reorganize the RGB generation path.
Therefore, Native Joint Attention is better suited to conditional RGB
generation, whereas Modality Tunneling Attention is better suited to non-RGB
estimation and reconstruction tasks that require the scope of the input
influence to be controlled.

The other three interfaces have heterogeneous, often deterministic non-RGB
targets. Here, joint normalization forces target self-attention and condition
attention to compete for the same probability mass. We identify this competition
as the central optimization bottleneck and introduce MTA to remove it. Rather
than requiring heterogeneous modalities to be jointly understood within one
shared attention distribution, our method retains each modality's own
representation path and learns a separately normalized tunnel between different
modal descriptions of the same underlying visual content. Concretely, MTA
preserves the text--target self path and adds an independently normalized
target-to-condition path.

The effect is most visible for sparse, high-frequency Edge prediction. With
batch size one and no gradient accumulation, MTA recovers recognizable Edge
structure after roughly 50 updates, corresponding to only 50 sampled training
examples, and produces clean boundaries by update 250. NJA also learns the
mapping, but its structural acquisition remains at a rudimentary stage even
after thousands of updates. RGB-to-Normal exhibits the same
contrast: NJA converges much more slowly and produces pale, fog-like normals
with weak boundaries. Figure~\ref{fig:ablation_validation_curves}(b)
shows that MTA obtains a structured Normal prediction by update 250, while NJA
requires 4,750 updates. Figures~\ref{fig:ablation_validation_curves}(c,d) show
the same effect on conditional control and inverse rendering: MTA recovers
clean Edge structure by updates 250 and 100, whereas NJA has only begun to
produce a few fragmented short contours after 5,000 and 3,850 updates,
respectively, remaining at an early stage of learning.
Thus, the LoRA NJA baseline has sufficient capacity to learn these
mappings; its limitation is optimization efficiency. Parameter efficiency
alone does not remove the competition between target and condition evidence
inside native joint attention.

The modest, non-monotonic NJA validation MSE is expected: this global velocity
metric can favor mean or background solutions and therefore need not track
decoded structural quality. The curves and samples must be read together.
Measured by optimizer updates, MTA reaches the displayed structured outputs at
least $19.0\times$, $20.0\times$, and $38.5\times$ sooner than the
same-parameter LoRA NJA baseline on TunnelingNets 003, 275, and 313,
respectively. The first recognizable structure emerges approximately
$77$--$100\times$ earlier. Because the NJA Edge outputs remain lower quality at
their cited endpoints, the corresponding update ratios are conservative lower
bounds. MTA introduces no additional parameters. These results also
show why one attention topology should not be imposed on all four interfaces:
MTA is preferable for transferring structured non-RGB evidence, while NJA is
preferable when synthesizing RGB with the native image prior.
% Figure~\ref{fig:ablation_design_choices}(a,c) enlarges the corresponding
% qualitative comparisons for TunnelingNets 003 and 275.

\begin{table}[t]
\centering
\caption{RGB-routing ablations for estimation and narrow generation.
(a) TunnelingNet 003 follows the final 200-update recipe except for the current-RGB
route. (b) TunnelingNet 270 uses the same legacy protocol for all variants and is
evaluated at update 2,220. Comparisons are controlled within each block;
absolute MSE should not be compared across blocks. \emph{Target} indicates a
trainable target-side tunnel. Bold marks the lowest MSE, while $\dagger$ marks
the retained route.}
\label{tab:routing_ablation}
\scriptsize
\renewcommand{\arraystretch}{1.10}
\textbf{(a) TunnelingNet 003: RGB$_c\rightarrow$Normal$_c$}\\[2pt]
\setlength{\tabcolsep}{2.5pt}
\begin{tabular}{@{}l l c c@{}}
\hline
\textbf{Attention} & \textbf{RGB$_c$ route} & \textbf{Target} & \textbf{Val. MSE $\downarrow$} \\
\hline
MTA & Condition branch$^\dagger$ & Yes & \textbf{0.1844} \\
MTA & Native reference & Yes & 0.1853 \\
\hline
\end{tabular}
\\[6pt]
\textbf{(b) TunnelingNet 270: Text + Normal$_c$ + RGB$_1\rightarrow$RGB$_c$}\\[2pt]
\begin{tabular}{@{}l l c c@{}}
\hline
\textbf{Attention} & \textbf{RGB$_1$ route} & \textbf{Target} & \textbf{Val. MSE $\downarrow$} \\
\hline
NJA & Native reference$^\dagger$ & Yes & \textbf{0.6277} \\
NJA & Condition branch & Yes & 0.6279 \\
NJA & Native reference & No & 0.6311 \\
MTA & Condition branch & Yes & 0.6481 \\
\hline
\end{tabular}
\end{table}

\subsection{Condition Routing and Tunnel Placement}
Routing is nontrivial because the pretrained backbone introduced in
Section~\ref{sec:method} is an image-editing model. Its native interface accepts
a text prompt and an optional RGB reference---the image to be edited---and uses
the resulting preservation prior while generating the target. TunnelingNet
additionally exposes clean condition lanes for WSG modalities and RGB
observations. The same RGB evidence can therefore enter either as a native
editable reference or as a condition: the former encourages the backbone to
preserve the supplied image, whereas the latter treats it as evidence rather
than content to copy. A second choice is tunnel placement: we can adapt both
the condition and target lanes, or train only the target-side tunnel while
leaving all evidence lanes on the frozen backbone.

These choices yield a different candidate set for each representative task.
\textbf{TunnelingNet 003 (estimation)} maps RGB$_c$ to Normal$_c$; we test whether the
current RGB should occupy the native editable-reference slot or a current-RGB
condition lane. \textbf{TunnelingNet 270 (narrow generation)} synthesizes RGB$_c$ from
text, Normal$_c$, and RGB$_1$; we test whether the first frame should be a
native appearance reference or a condition, and whether a target-side tunnel
is needed. \textbf{TunnelingNet 275 (conditional control)} maps Normal$_c$ to Edge$_c$
without any RGB input, so its evidence route is fixed and it isolates whether
the Normal condition lane itself needs a trainable tunnel. \textbf{TunnelingNet 313
(inverse rendering)} predicts Edge$_c$ from current Normal$_c$ and RGB$_1$; we
test whether RGB$_1$ should use the native reference slot or a separate
first-frame condition lane, and whether adapting the condition side improves
over a target-only tunnel.

\begin{figure*}[t]
    \centering
    \includegraphics[width=0.98\textwidth]{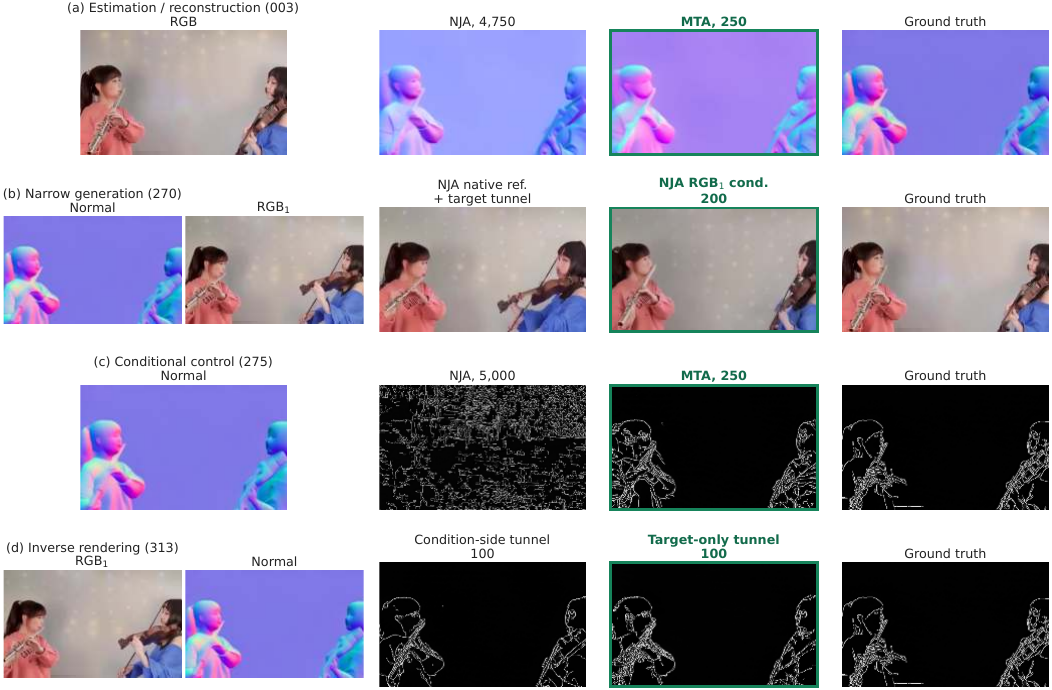}
    \caption{\textbf{Qualitative effects of task-dependent structural choices.}
    Each row uses an evaluation example from one of the four interfaces.
    (a,c) compare NJA with MTA for estimation/reconstruction and conditional
    control. (b) compares native-reference and first-RGB-condition routing
    under NJA for narrow generation. (d) compares condition-side and
    target-only tunnel placement for inverse rendering. Green borders mark the visually preferred outputs in the displayed
    comparisons. The highlighted outputs in rows (b) and (d) use 200 and 100
    optimizer updates, reached in 45 min 32 s and 27 min 18 s, respectively.
    In (b), the condition-branch result follows the Normal/Pose condition while
    retaining the appearance supplied by RGB$_1$; the deployed TunnelingNet follows the
    slightly lower validation MSE and uses the native reference route reported
    in Table~\ref{tab:selected_ablation_recipe}. Checkpoint indices are shown
    above outputs.}
    \label{fig:ablation_design_choices}
\end{figure*}

Together, these alternatives define our routing-and-placement sweep.
Table~\ref{tab:selected_ablation_recipe} consolidates the final configuration
for all four TunnelingNets, Table~\ref{tab:routing_ablation} reports the controlled
RGB-routing comparisons for TunnelingNets 003 and 270, and
Figure~\ref{fig:ablation_design_choices} summarizes the corresponding
qualitative choices, including tunnel placement for TunnelingNet 313.

\begin{figure}[t]
    \centering
    \includegraphics[width=\columnwidth]{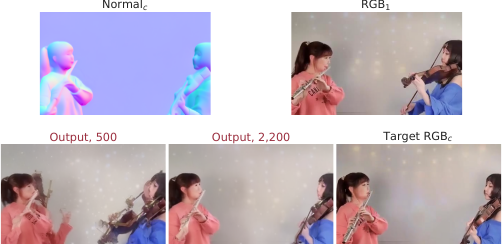}
    \caption{\textbf{Native-reference routing can override current geometry.}
    True 10-step inference from the TunnelingNet 270 native-reference run. At update
    500, conflicting poses produce duplicate limbs; at update 2,200, the output
    still retains the RGB$_1$ pose instead of following the current Normal.}
    \label{fig:routing_native_failure}
\end{figure}

The TunnelingNet 003 block of Table~\ref{tab:routing_ablation} compares the two current-RGB routes for
TunnelingNet 003 under the same 200-update MTA recipe. The condition branch obtains
0.1844 MSE, compared with 0.1853 for the native reference. The small 0.43\%
gap shows that both routes can transmit RGB evidence, but the condition route
is marginally better and better matches the estimation task: the current RGB
provides the visual evidence from which the Normal map is inferred, rather than
an appearance reference that the image-editing backbone is encouraged to
preserve.

The TunnelingNet 270 block summarizes the controlled routing sweep. The
two NJA variants with trainable target tunnels are numerically tied: routing
RGB$_1$ through the native reference slot improves MSE by only 0.0002
(0.03\%). Figure~\ref{fig:routing_native_failure} reveals the more important
distinction: the native slot strongly preserves the reference appearance and
can resist changes specified by the current Normal, as if the task were direct
image editing.
TunnelingNet 270 generates the current RGB from text, current Normal, and first-frame
appearance evidence. We retain NJA for its RGB prior and select the native
reference route, which obtains the slightly lower validation MSE and matches the
final configuration of the selected TunnelingNet. Removing the target tunnel weakens
the direct adaptation path to RGB, while replacing NJA with MTA gives up part
of the native RGB synthesis prior. Figure~\ref{fig:ablation_design_choices}(b) fixes Normal, RGB$_1$, and
the target across both routes. The condition-branch result is true 10-step
inference at update 200 (45 min 32 s) and illustrates that the alternative route
can follow a deliberately conflicting pose while retaining first-frame
appearance.

\begin{figure*}[t]
    \centering
    \includegraphics[width=0.98\textwidth]{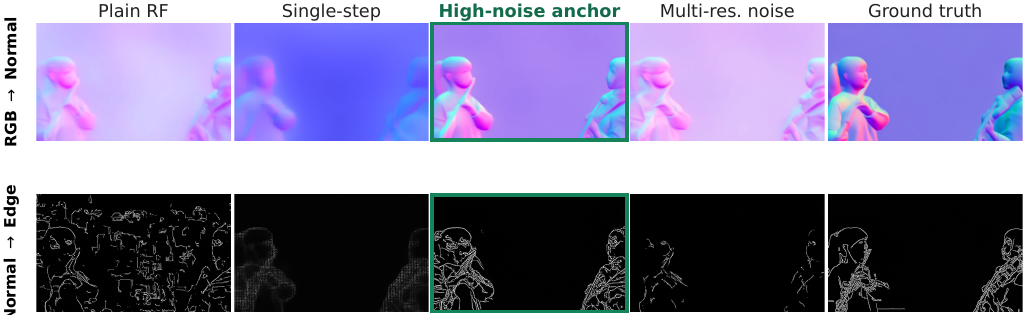}
    \caption{\textbf{Qualitative effects of rectified-flow objective shaping.}
    At update 250, each row compares plain rectified flow, single-step
    clean-latent prediction, high-noise anchoring, multi-resolution noise, and
    the ground truth. With batch size one and no gradient accumulation, each
    model has completed only 250 optimizer steps and trained on exactly
    \emph{250 sampled examples}. High-noise anchoring is the only shared choice that
    preserves both Normal saturation and sparse Edge structure. Green borders
    mark the selected objective; all Edge maps use the same grayscale reader.}
    \label{fig:ablation_objective_qualitative}
\end{figure*}

Inverse rendering makes this separation stricter. In TunnelingNet 313, current Normal
defines the target state, whereas RGB$_1$ is only a scene-level clue. Routing
RGB$_1$ as a native reference encourages a shortcut that copies stale
appearance or geometry. We instead pass its tokens through the first-frame
condition branch and disable the native reference slot, so the current
Normal remains the state-defining evidence. Accordingly, TunnelingNet 270 is the only selected configuration that uses the native
RGB-reference slot. TunnelingNets 003 and 313 route RGB as condition evidence, while
TunnelingNet 275 has no RGB input.

\begin{table}[t]
\centering
\caption{Task-agnostic objective ablation. Values are fixed-$t=1$ validation velocity MSE, so all rows are directly comparable.}
\label{tab:objective_ablation}
\footnotesize
\setlength{\tabcolsep}{3.5pt}
\renewcommand{\arraystretch}{1.10}
\begin{tabular}{@{}l c c c@{}}
\hline
\textbf{Objective} & \textbf{003 Normal} & \textbf{275 Edge} & \textbf{Mean $\downarrow$} \\
\hline
Plain rectified flow & 0.6805 & 0.5961 & 0.6383 \\
Single-step clean latent & 0.4308 & 0.4186 & 0.4247 \\
High-noise anchor & \textbf{0.3263} & \textbf{0.4062} & \textbf{0.3662} \\
Multi-resolution noise & 0.6505 & 0.5917 & 0.6211 \\
\hline
\end{tabular}
\end{table}
Figure~\ref{fig:ablation_design_choices}(d) isolates tunnel placement under the
same 100-update budget. Both variants recover comparable major Edge structure,
and adapting the condition side adds no visible structural gain. This does not
mean that condition evidence is ignored: clean Normal and RGB$_1$ tokens still
enter the separately normalized MTA path. Rather, the result locates the
task-specific learning burden at the target readout, which must translate
already informative frozen evidence into the requested output modality. It is
the practical counterpart of the color-vision analogy in
Section~\ref{sec:intro}: modality-specific streams need not acquire a common
internal interpretation, or learn to ``understand one another,'' to describe
the same underlying state. The target tunnel learns only the translation needed
by its output. Removing condition-side reinterpretation therefore reduces the
optimization burden and keeps condition states adapter-independent, allowing
one condition forward pass to be shared by many TunnelingNets. We consequently
train only the target tunnel, the smallest effective design. The panel also
reveals the complementary role of visual
evidence: the Normal estimate alone omits fine structures smoothed away during
estimation, whereas providing RGB$_1$ as additional evidence enables the
inverse-rendering TunnelingNet to recover them. This cross-modal completion provides
an empirical basis for Chain of Visual (CoV) reasoning, in which evidence
retained in one visual state resolves information absent from another. With
routing and tunnel placement fixed, the
remaining objective-shaping choice is quantified in
Table~\ref{tab:objective_ablation} and analyzed next.
% Figure~\ref{fig:ablation_objective_qualitative} provides the corresponding
% qualitative comparison.

\begin{figure*}[t]
    \centering
    \includegraphics[width=0.98\textwidth]{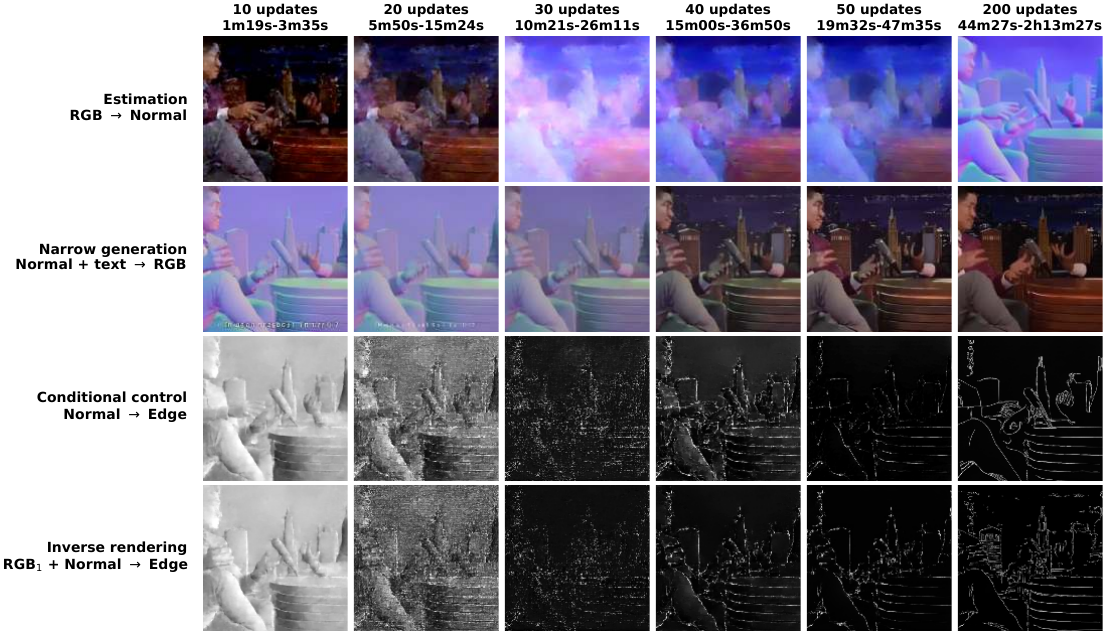}
    \caption{\textbf{Rapid acquisition across the first 200 updates.}
    Rows correspond to estimation, narrow generation, conditional control, and
    inverse rendering. Columns show true 10-step Euler inference from pure noise
    after 10, 20, 30, 40, 50, and 200 updates. Times below the column headings
    report the observed wall-clock range across the four tasks, including
    scheduled diagnostics. Each row reuses one fixed evaluation example and the
    same inference seed. With batch size one and no gradient accumulation,
    \emph{each update uses exactly one sampled training example}; thus even the
    final column has seen only 200 examples. No condition or ground-truth image
    is displayed. Edge outputs use the same per-image contrast stretch for
    visibility.}
    \label{fig:minute_scale_acquisition}
\end{figure*}

\subsection{Rectified-Flow Objective}

The initial dense-task experiments exposed two failure modes. Normal
predictions retained the correct coarse shape but drifted toward low-saturation
mean colors; sparse Edge predictions were progressively overwhelmed by their
black background. We therefore tested task-agnostic objective changes rather
than modality-specific losses. Table~\ref{tab:objective_ablation} compares
plain rectified flow, single-step clean-latent prediction, high-noise anchoring,
and annealed multi-resolution noise. Every model is evaluated by the same
plain velocity MSE at $t=1$. High-noise anchoring, which assigns half of the
training samples to $t=0.999$, is best on both representative target types and
reduces their mean validation error by 42.6\% relative to the baseline.
Figure~\ref{fig:ablation_objective_qualitative} shows that this advantage is not
merely numerical. For RGB-to-Normal, plain rectified flow and multi-resolution
noise recover the coarse layout but weaken the chromatic separation that
encodes surface orientation; the anchored model retains a more saturated,
spatially coherent field. For Normal-to-Edge, the same alternatives lose much
of the sparse contour support to the dominant black background, whereas
anchoring preserves connected structure. Single-step prediction is competitive
in Table~\ref{tab:objective_ablation}, but its decoded outputs are less
consistent across these dense and sparse targets. High-noise anchoring is thus
the only tested objective that improves the shared validation metric while also
preserving the modality-specific information visible after decoding.

We also ablated low-noise timestep shifting, Min-SNR weighting, and Huber loss.
Huber produced the cleanest early visual results, and Min-SNR better resisted
late degradation, but neither was a consistent default. In particular,
enabling all three gradually drove outputs toward the target mean: Normal kept
its shape but faded, while Edge disappeared into the black background. This
qualitative behavior is not captured reliably by a weighted training loss,
which is why we retain fixed plain-MSE validation and inspect the decoded output
throughout training. The final concise recipe is therefore high-noise
anchoring alone, with timestep shifting, Min-SNR, Huber, single-step prediction,
and multi-resolution noise disabled. With this recipe, the four representative
TunnelingNets reach stable solutions within a 500-update ablation budget, compared with
the 5,000-update budget used by the earlier native setup.
\subsection{Minute-Scale Rapid Learning}
Figure~\ref{fig:minute_scale_acquisition} tracks end-to-end inference for all four
interfaces at 10-update intervals through update 50 and again at update 200.
Because batch size is one, update count is also the number of sampled training
examples. The first predictions are obtained after only 1\,min\,19\,s to
3\,min\,35\,s, the 50-update milestone after 19\,min\,32\,s to
47\,min\,35\,s, and the 200-update endpoint after 44\,min\,27\,s to
2\,h\,13\,min\,27\,s. This wall-clock spread reflects the different
conditioning contracts and their diagnostic overhead; update count provides
the cleaner measure of data exposure, which is identical across rows.

Using a fixed evaluation example and inference seed within each row, the sequence
reveals a consistent coarse-to-structured transition. The first tens of updates
establish the target modality and broad spatial support; later updates sharpen
task-specific geometry, improve structural continuity, and suppress residual
artifacts. This trend appears across dense estimation, RGB synthesis, sparse
conditional control, and inverse rendering despite their different output
statistics. Most importantly, the update-200 predictions have been trained on
only 200 sampled examples. They are therefore evidence that the proposed
routing and objective acquire a new visual mapping at minute scale, rather than
requiring conventional long-horizon fine-tuning. These early outputs are not treated as final convergence; Appendix
Figure~\ref{fig:minute_scale_detailed} provides condition--prediction--target
comparisons at the later selected checkpoints.

%% file: sec_preprint/5_quantitative.tex
\clearpage

\section{Large-Scale Quantitative Evaluation}
\label{sec:quantitative}

\paragraph{Protocol.}
We freeze one deployable checkpoint for each trained dense TunnelingNet and
evaluate it end to end from pure Gaussian noise through the complete generation
path. Every run uses 10 Euler steps, a fixed
task-specific seed, and deterministically selected examples from the evaluation
set. Across 1,148 TunnelingNets, the resulting snapshot contains 18,191
end-to-end inference evaluations.
Table~\ref{tab:evaluation_coverage} reports coverage across the four interfaces.

\begin{table}[t]
    \centering
    \caption{\textbf{Coverage of the frozen dense evaluation snapshot.}
    Input signatures count distinct conditioning contracts, not individual
    images. All outputs are obtained by true 10-step Euler inference.}
    \label{tab:evaluation_coverage}
    \footnotesize
    \resizebox{\columnwidth}{!}{%
        \input{tables/evaluation_interface_coverage_normv2}
    }
\end{table}

\begin{figure}[t]
    \centering
    \includegraphics[width=\columnwidth]{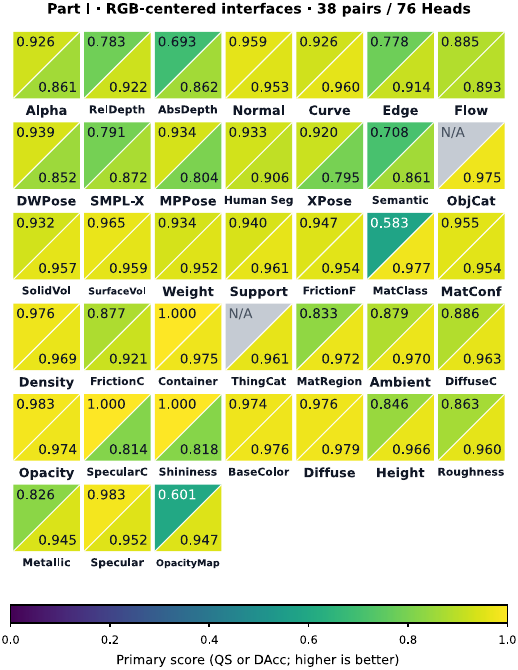}
    \caption{\textbf{Part I: RGB-centered interfaces (38 modality pairs; 76 TunnelingNets).}
    Each cell pairs upper-left RGB-to-WSG estimation/reconstruction with
    lower-right text+WSG+RGB$_1$-to-RGB narrow generation. Continuous
    entries use QS and categorical estimation entries display DAcc.
    Their triangle color uses DAcc; continuous triangle
    color uses QS on the same
    zero-to-one scale used throughout Figures~\ref{fig:rgb_centered_matrix}--
    \ref{fig:packed_paired_bridges}. Cells marked \wsmunavailablekey{} are
    unavailable in the current release.}
    \label{fig:rgb_centered_matrix}
\end{figure}

\begin{figure*}[t]
    \centering
    \includegraphics[width=0.96\textwidth]{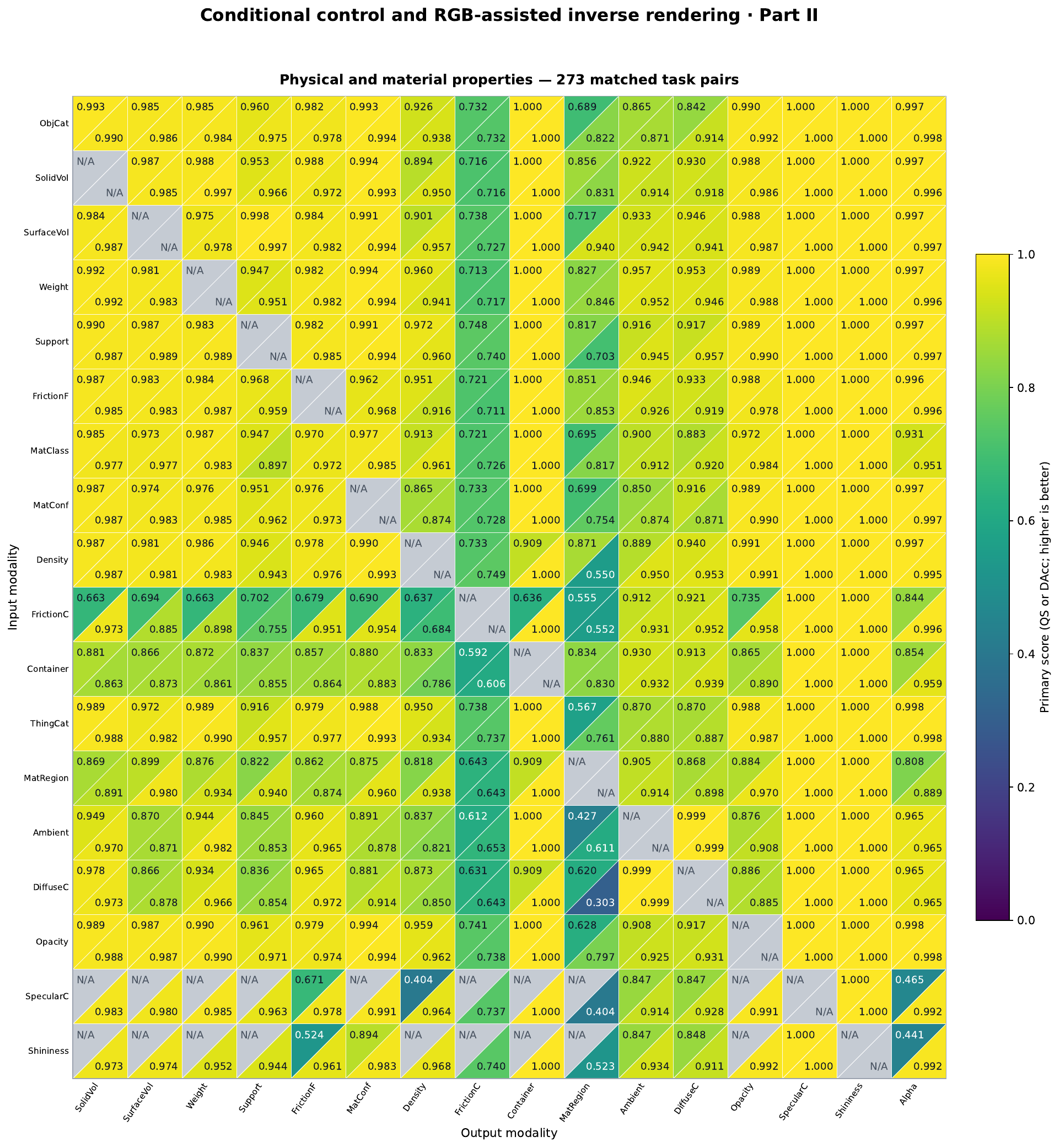}
    \caption{\textbf{Paired cross-modal matrix, Part II: physical and material
    properties (273 pairs; 546 TunnelingNets).} Each cell places conditional control in
    the upper-left triangle and RGB-assisted inverse rendering in the
    lower-right. Continuous entries use
    $\mathrm{QS}=1-\text{decoded MSE}$; categorical entries display
    DAcc. Triangle colors encode QS for continuous outputs and DAcc for
    categorical outputs on a
    common zero-to-one scale. Cells marked \wsmunavailablekey{} are unavailable
    in the current release.}
    \label{fig:packed_paired_physical}
\end{figure*}

\paragraph{Metrics and task matrices.}
For continuous or image-like channels we report decoded pseudo-RGB MSE and
SSIM. Because modality-specific encodings have different numeric scales, MSE
is compared only among tasks sharing the same output channel. For semantic categorical outputs, the main matrices report dominant-class
accuracy (DAcc) from the final deployed TunnelingNet and categorical refiner. DAcc measures
whether the principal generated region receives the correct identity. The
released tables additionally retain mIoU as a stricter spatial and multi-region
metric; strict pixel accuracy remains secondary because background can dominate it.
The complete input--output matrices retain every trained edge numerically.
Figure~\ref{fig:rgb_centered_matrix} first presents the 76 RGB-centered TunnelingNets
by pairing the two directions for each of 38 WSG modalities: RGB-to-WSG
estimation in the upper-left and text+WSG-to-RGB narrow generation in the
lower-right. The remaining 1,072 TunnelingNets form 536 shared input--output pairs
across conditional control and RGB-assisted inverse rendering.
Figures~\ref{fig:packed_paired_physical}--
\ref{fig:packed_paired_bridges} place both results in each diagonally divided
cell rather than repeating the axes: conditional control occupies the
upper-left and RGB-assisted inverse rendering the lower-right.

Continuous results use the higher-is-better quality score
$\mathrm{QS}=1-\text{decoded MSE}$, where decoded MSE is computed between the
prediction and ground truth in the normalized VAE pseudo-RGB decoder space;
categorical cells show dominant-class accuracy (DAcc). Triangles are colored by
QS for continuous outputs and DAcc for categorical outputs. Released tables
retain exact metrics for all
mappings, including those below the display threshold. To avoid a mostly empty
global rectangle, we partition the trained support into five compact blocks:
RGB-centered interfaces, physical and material properties, visual structure
and pose, PBR texture space, and cross-family bridges. These Parts define the
layout---not a task taxonomy or method components---and summarize 1,034 TunnelingNets.
Mappings unavailable in the current release are marked N/A.
% The compact paired matrices withhold the three categorical target families
% ObjCat, MatClass, and ThingCat. Their complete measurements remain in the
% released evaluation inventory.
% The cross-modal panels retain square matrix cells, white grid lines, rotated modality axes,
% and vector metric annotations; the compact RGB-centered panel uses seven labeled
% cells per row. The source decoded MSE values and exact
% rectangular matrices remain available as CSVs.

\subsection{Cross-Modal Matrix Analysis}
\label{subsec:cross_modal_matrix_analysis}

\paragraph{Performance by task direction.}
Figure~\ref{fig:rgb_centered_matrix} exposes a consistent target-side
asymmetry. RGB estimation reaches a median primary score of 0.886, whereas
narrow RGB generation reaches 0.953; 68.4\% and 97.4\% of the respective TunnelingNets
exceed 0.8. The representation explains this gap: RGB synthesis can reuse the pretrained appearance
prior, while RGB-to-WSG translation must reproduce the exact geometry or
pseudo-color convention of each target channel. The same tendency recurs in
the paired matrices, where the RGB-assisted direction is usually stronger than
condition-only control. It therefore measures the value of a visual reference,
not an intrinsic ranking of the two interfaces.

\begin{figure*}[p]
    \centering
    \includegraphics[width=0.94\textwidth]{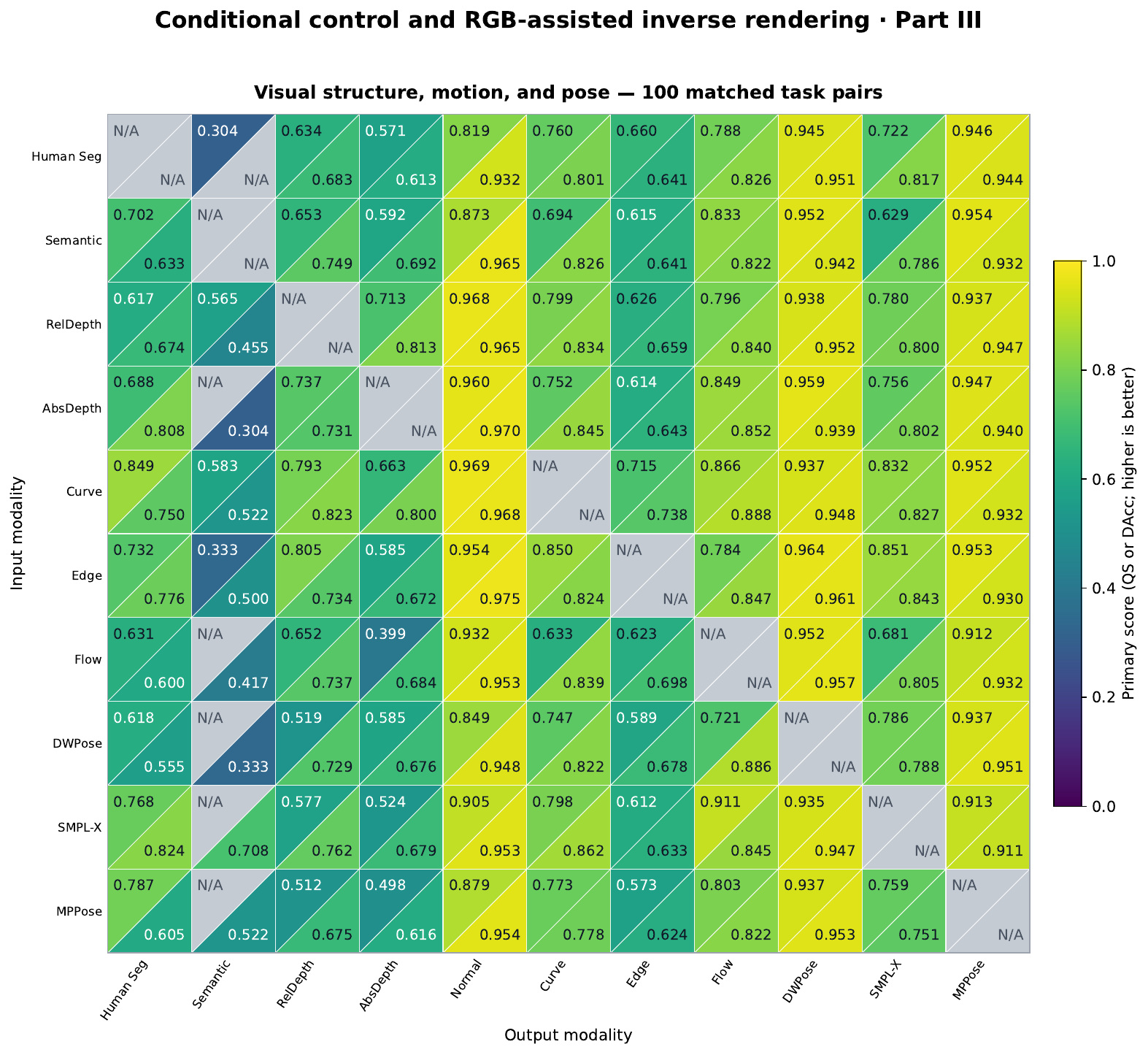}
    \caption{\textbf{Paired cross-modal matrix, Part III: visual structure,
    motion, and pose (100 pairs; 200 TunnelingNets).} Diagonal cell ordering and metrics
    follow Figure~\ref{fig:packed_paired_physical}. Separating this family from
    the global matrix enlarges pose, geometry, and semantic annotations while
    preserving their original input--output coordinates. Cells marked
    \wsmunavailablekey{} are unavailable in the current release.}
    \label{fig:packed_paired_visual}
\end{figure*}

\FloatBarrier

\begin{strip}
    \centering
    \begin{minipage}[t]{0.48\textwidth}
        \vspace{0pt}
        \centering
        \includegraphics[width=\linewidth]{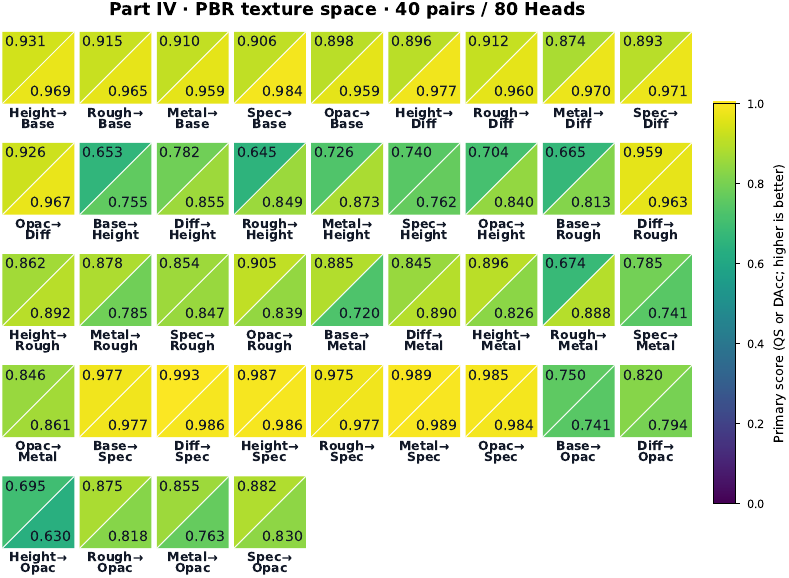}
        \vspace{-12pt}
        \captionof{figure}{\textbf{Part IV: PBR texture space (40 pairs;
        80 TunnelingNets).} The compact card layout retains only trained
        input--output pairs; upper-left and lower-right triangles report
        conditional control and RGB-assisted inverse rendering. Cells marked
        \wsmunavailablekey{} are unavailable in the current release.}
        \label{fig:packed_paired_pbr}
    \end{minipage}
    \hfill
    \begin{minipage}[t]{0.48\textwidth}
        \vspace{0pt}
        \centering
        \includegraphics[width=\linewidth]{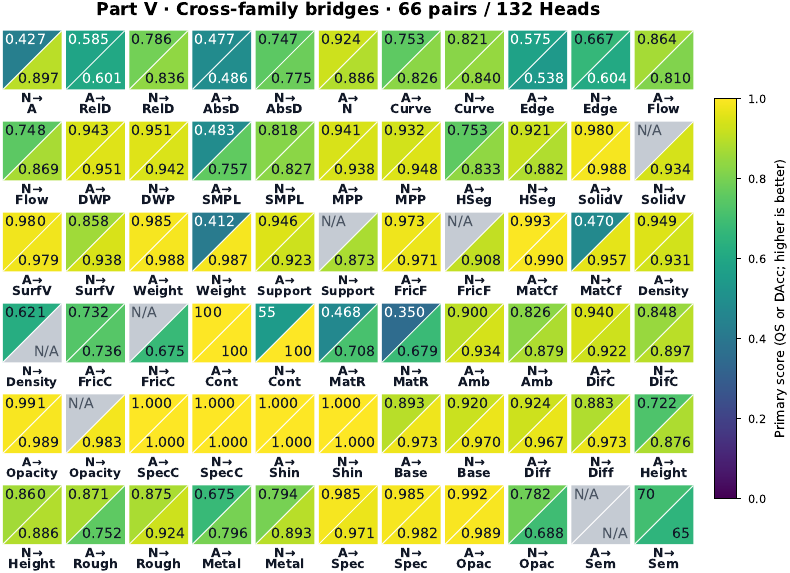}
        \vspace{-12pt}
        \captionof{figure}{\textbf{Part V: cross-family bridges (66 pairs;
        132 TunnelingNets).} The companion compact layout wraps the trained Alpha-
        and Normal-input pairs. Categorical cells show percentage
        DAcc and use DAcc for color; continuous cells
        show and color by QS. Cells marked
        \wsmunavailablekey{} are unavailable in the current release.}
        \label{fig:packed_paired_bridges}
    \end{minipage}
\end{strip}

% Continue beneath the paired Part IV--V strip so both columns fill naturally.
\paragraph{Physical and visual structure.}
The physical/material block in Figure~\ref{fig:packed_paired_physical} is the
strongest large family: its median score is 0.947 for conditional control and
0.973 for RGB-assisted inverse rendering, with 79.5\% and 89.0\% of mappings
above 0.8. Spatially smooth scalar fields and appearance channels dominate this
high-score region. Friction coefficient, material region, and the categorical
container mask are harder, which separates approximate spatial support from
exact semantic calibration. Figure~\ref{fig:packed_paired_visual} shows a more
selective pattern. Its medians are 0.798 and 0.839, but Normal and articulated
pose outputs remain consistently strong: Normal reaches median scores of 0.919
and 0.960, while DWPose reaches 0.945 and 0.951. This pattern is consistent with
human geometric reasoning: Depth and Edge cannot be recovered reliably when
the conditioning modality does not preserve the corresponding spatial cues.

\paragraph{PBR attributes and cross-modal conversion.}
Figure~\ref{fig:packed_paired_pbr} is nearly directionally symmetric. The
median scores are 0.880 for control and 0.881 for inverse rendering, and
72.5\%/77.5\% of its mappings exceed 0.8. Base color, diffuse, and specular
channels transfer particularly reliably, while height and opacity expose the
remaining direction-dependent cases. The bridge matrix in
Figure~\ref{fig:packed_paired_bridges} clarifies when a reference image matters.
Alpha-conditioned mappings remain strong in both directions (median
0.932/0.933), whereas Normal-conditioned mappings rise from 0.790 under
condition-only control to 0.897 with RGB assistance. The RGB reference thus
supplies appearance and scene-specific evidence that a structural channel
alone does not contain, while the structural condition still determines the
requested transformation.

\paragraph{Interpretation.}
Taken together, the matrices support a structured claim: a single tunneling
mechanism covers hundreds of heterogeneous transfers, and performance follows
the information shared by the input and target modalities.
Large coherent regions indicate reusable cross-modal transport; directional
gaps identify mappings that benefit from first-frame RGB evidence; and N/A
cells keep unavailable mappings from being mistaken for valid low-confidence
predictions. We retain their exact records in the released tables; an extended
analysis of categorical limitations and the corresponding upgrade inventory is
provided in Appendix~\ref{subsec:supp_categorical_limitations}.

\subsection{Prediction--Target Correspondence Calibration}
\label{subsec:prediction_target_calibration}

The matrix scores above compare each generated output with its corresponding
held-out target. We additionally measure how strongly those scores depend on
the correct sample correspondence. For every TunnelingNet, we keep its predictions
fixed and cyclically shift the targets by a deterministic nonzero offset chosen
away from adjacent samples. This produces a mismatched control with the same
prediction set, target set, modality, and decoder statistics. We then compute
MSE for the matched and shifted pairings in the normalized decoder space used
by QS. The calibration requires no retraining or additional inference.

\begin{table}[t]
    \centering
    \caption{\textbf{Prediction--target correspondence across all four
    interfaces.} Error reduction compares the mean matched and shifted MSE.}
    \label{tab:prediction_target_calibration}
    \footnotesize
    \setlength{\tabcolsep}{6pt}
    \begin{tabular}{@{}lr@{}}
        \toprule
        Interface &
        \shortstack{Error\\reduction} \\
        \midrule
        RGB estimation / reconstruction & 57.2\% \\
        Text + WSG narrow generation & 78.3\% \\
        WSG conditional control & 33.5\% \\
        RGB-assisted inverse rendering & 46.3\% \\
        \midrule
        \textbf{All interfaces} & \textbf{41.4\%} \\
        \bottomrule
    \end{tabular}
\end{table}
% \begin{table}[t]
%     \centering
%     \caption{\textbf{Prediction--target correspondence across all four
%     interfaces.} ``Varying/all'' gives the number of TunnelingNets whose held-out
%     targets change under the deterministic shift and the total number of TunnelingNets.
%     Win rate is computed over the target-varying TunnelingNets; equal-target cases
%     provide no correspondence comparison. Error reduction compares the mean
%     matched and shifted MSE over all TunnelingNets.}
%     \label{tab:prediction_target_calibration}
%     \footnotesize
%     \setlength{\tabcolsep}{3.1pt}
%     \resizebox{\columnwidth}{!}{%
%     \begin{tabular}{@{}lrrrrr@{}}
%         \toprule
%         Interface & \shortstack{TunnelingNets\\varying/all} &
%         \shortstack{Matched\\MSE}$\downarrow$ &
%         \shortstack{Shifted\\MSE}$\downarrow$ &
%         \shortstack{Error\\reduction} &
%         \shortstack{Matched\\wins} \\
%         \midrule
%         RGB estimation / reconstruction & 36/38 & 0.153 & 0.357 & 57.2\% & 86.1\% \\
%         Text + WSG narrow generation & 38/38 & 0.074 & 0.339 & 78.3\% & 97.4\% \\
%         WSG conditional control & 498/536 & 0.243 & 0.365 & 33.5\% & 87.6\% \\
%         RGB-assisted inverse rendering & 498/536 & 0.176 & 0.327 & 46.3\% & 89.8\% \\
%         \midrule
%         \textbf{All interfaces} & \textbf{1,070/1,148} & \textbf{0.203} &
%         \textbf{0.346} & \textbf{41.4\%} & \textbf{88.9\%} \\
%         \bottomrule
%     \end{tabular}
%     }
% \end{table}

Correct target correspondence lowers mean error in every interface. Across the
full matrix, it reduces error by 41.4\%, and the median per-task reduction is
51.9\%. A fixed output template is unchanged when targets are permuted and
would therefore produce zero correspondence reduction. The strictly positive
reduction in every interface, including a greater-than-half median reduction
across TunnelingNets, directly rejects that behavior. The
per-task and per-target records and deterministic calibration script accompany
the evaluation tables. These consistent gaps show that the matrix scores
capture sample-conditioned target structure beyond modality-level averages or
fixed output templates. The calibration therefore establishes nontrivial
prediction--target correspondence across all four interfaces of the large task
graph.

\FloatBarrier
\clearpage
\begin{table*}[th]
    \centering
    \caption{\textbf{Three complementary levels of evidence for the Unified Law.}
    The evidence asks whether the law works broadly, whether the trained routes
    learn the requested relation for each example, and why one target
    representation is easier for a frozen generator to learn than another.
    Together, they establish broad applicability, genuine task learning, and
    the mechanism that makes aligned representations easier to acquire.}
    \label{tab:three_level_law_validation}
    \footnotesize
    \setlength{\tabcolsep}{5pt}
    \begin{tabularx}{\textwidth}{@{}c p{0.22\textwidth} p{0.36\textwidth} X@{}}
        \toprule
        Level & The question & What we did & What it shows \\
        \midrule
        I: broad use & Was the law fitted only to the four tasks that first
        revealed it? & We established the four interfaces with four initial
        Tunnels, then kept the shared backbone frozen and trained the other
        1,144 task routes. & The same rule realizes 1,148 trained and evaluated
        tasks, covering 39 output groups and all four interfaces. \\
        II: real input--output learning & Did each route learn the requested
        relation for each example, or only a typical output for its modality? &
        We keep every prediction fixed and deliberately pair it with the wrong
        held-out target. A generic output would be equally close before and
        after this shift. & Correct pairs have 41.4\% lower error overall, with
        a 51.9\% median reduction per TunnelingNet. The routes therefore follow the
        individual input and target. \\
        III: why representation matters & Does a frozen generator learn the
        same information more easily when it is written in a representation
        the generator already knows? & We write the same surface-normal target
        in five lossless coordinate systems: its native VAE coordinates and
        four reversible channel rearrangements. Every version reconstructs
        exactly the same normal image; only its familiarity to the backbone
        changes. & Native coordinates lower checkpoint-averaged error by
        20.1\%--26.6\%, produce the lowest final error, and win 59 of 64 final
        sample--transformation comparisons. \\
        \bottomrule
    \end{tabularx}
\end{table*}

\begin{figure*}[th]
    \centering
    \includegraphics[width=0.94\textwidth]{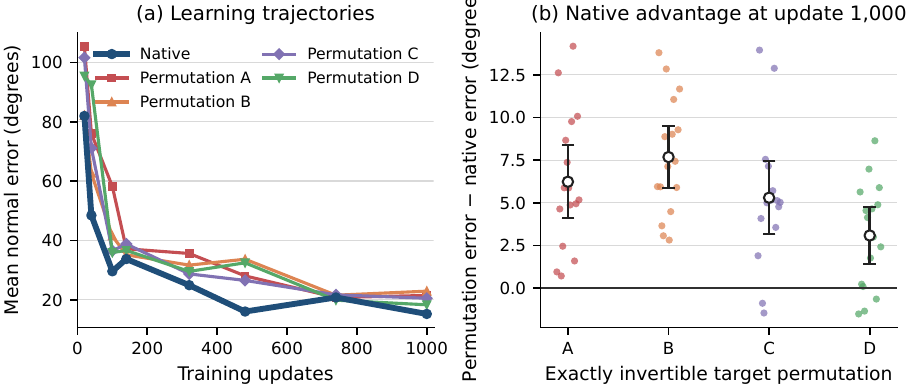}
    \caption{\textbf{The same target is easier to learn in coordinates already
    familiar to the frozen generator.} We encode every surface-normal target
    in its native VAE coordinates and in four reversible rearrangements of the
    same channels. All five versions retain exactly the same information and
    reconstruct the same normal image. The rearrangements simulate alternative
    representations that the backbone did not learn during pretraining.
    \textbf{(a)} Native coordinates give the lowest checkpoint-averaged error
    across the training trajectory and the lowest final error. \textbf{(b)}
    Positive values mean that native coordinates have lower final angular error
    for one held-out sample under one rearrangement. Native coordinates improve
    all four mean comparisons and 59 of the 64 individual comparisons. This
    directly shows that backbone familiarity changes learning difficulty even
    when target information is identical.}
    \label{fig:representation_alignment_learning_curves}
\end{figure*}

\clearpage
\subsection{Three Complementary Levels of Evidence: Breadth, Genuine Task
Learning, and Mechanism}
\label{subsec:three_level_law_validation}

\noindent\textbf{What this section establishes.}
The same rule extends from four initial tasks to 1,148 trained tasks; the
resulting routes learn the requested relation for individual examples; and a
target becomes substantially easier to learn when it uses a representation
already understood by the frozen generator. Together, these results support
the Unified Law as both a way of organizing visual tasks and a learning
principle. The law explains the large-scale success of WSM, predicts which
representation of a new target will be easier to acquire, and provides a
practical rule for building future task routes.

This subsection asks three progressively harder questions. First, did the rule
continue to work after the four tasks that revealed it? Second, did the
resulting routes learn the requested relation for each example? Third, can we
directly test the proposed reason that some representations are easier for a
frozen generator to acquire? The three levels move from breadth, to genuine
input--output correspondence, to a controlled test of the learning mechanism.
Table~\ref{tab:three_level_law_validation} gives the complete logic in one
place.

\paragraph{Level I: the rule extends far beyond the tasks that revealed it.}
We first developed TunnelingNets 003, 270, 275, and 313, one representative task for
each of the four interfaces. These four experiments established how a task
should be expressed and where a TunnelingNet should connect to the shared
generator. We then applied that rule to the other 1,144 routes. Each route was
trained and evaluated while the shared generator remained frozen. The result
is 1,148 trained tasks across 39 output groups. This order matters: thousands
of results were obtained after the common rule had already been formed, so the
rule is not a summary invented separately for each finished task. The complete
chronological record, including sparse, categorical, and pose variants, appears
in Sec.~\ref{pre:subsec:memoir_prospective_confirmation}.

\paragraph{Level II: the routes learn each requested relation, not only the
appearance of a modality.}
A model could produce a plausible depth map, normal map, or mask without
following the particular input image. We test this possibility without changing
any prediction. For every TunnelingNet, we first compare each prediction with its true
held-out target. We then shift the targets between examples and compare the
same predictions with targets belonging to other inputs. An output that only
repeats a generic modality pattern receives essentially the same error after
the shift. A prediction that follows its own input loses agreement.
Section~\ref{subsec:prediction_target_calibration} shows the latter: correct
pairing reduces error in every interface, by 41.4\% over the whole task graph,
and the median TunnelingNet loses 51.9\% of its agreement after targets are shifted.
The thousand-task result therefore reflects sample-specific visual relations.

\paragraph{Level III: the same information is easier to learn when its
representation matches the backbone.}
The first two levels show that the law works broadly and that the routes learn
real relations. The third level tests why representation choice changes the
difficulty of learning. We use RGB-to-surface-normal estimation and write every
normal target in five different coordinate systems. The first is the native VAE
representation seen by the generator during pretraining. The other four simply
rearrange the 32 VAE channels. Each rearrangement is perfectly reversible: it
removes no information and reconstructs exactly the same normal image after the
inverse rearrangement. It therefore simulates a modality encoded in a complete
but unfamiliar organization.

This comparison separates two ideas that are often confused. A representation
can encode and decode a target perfectly, while a pretrained backbone may still
be unfamiliar with how that representation organizes the information. Here all
five versions are equally complete; only their match to the frozen generator
changes. We keep the task, data, backbone, TunnelingNet, optimization, update
budget, and normal evaluator identical. Thus, any consistent learning advantage
comes from representation--backbone alignment.

Before training the five Tunnels, we ask the frozen backbone to predict each
coordinate system. Its residual is lowest for the native coordinates
($1.249$, versus $1.653$--$1.835$), confirming that these are the coordinates
most familiar to the pretrained generator. We then train all five versions and
measure the same native normal-angle error at eight shared checkpoints on 16
held-out examples. The result follows the prediction. Native coordinates have
the lowest checkpoint-averaged error ($23.39^\circ$, versus
$29.28^\circ$--$31.85^\circ$), a 20.1\%--26.6\% reduction. They also achieve
the lowest final error ($15.20^\circ$, versus
$18.29^\circ$--$22.89^\circ$) and win 59 of 64 final
sample--transformation comparisons.
Figure~\ref{fig:representation_alignment_learning_curves} shows both the full
trajectories and the per-example results.

\paragraph{Takeaway: the law predicts both what can be unified and how to make
it learn efficiently.}
The three levels establish one connected result. A rule discovered from four
Tunnels extends to 1,148 trained visual tasks. Those routes learn the requested
relation for individual examples across all four interfaces. When the target
information is held exactly constant, the representation already understood by
the frozen generator is learned substantially faster and more accurately. The
last result provides controlled evidence for the comparative prediction in
Eq.~\ref{pre:eq:comparative_learning_prediction}.

This gives the Unified Law practical force. Encoding and decoding a modality is
only the first requirement. Efficient capability expansion also places the
target in a representation whose organization is familiar to the shared
backbone. Dense visual targets can often use the backbone's native visual
coordinates directly. Sparse or highly abstract targets benefit from a visual
reorganization or a task-appropriate output boundary. This principle explains
the successful choices observed across the development record and provides a
concrete way to design new Tunnels before paying for full training. The formal
statement of this result, its information-preservation and backbone-coverage
conditions, and its extension to other generative backbones appear in
Sec.~\ref{pre:sec:formal_unified_law}.

\begin{table*}[p]
    \centering
    \caption{\textbf{External comparison on representative tasks across three WSM interfaces.}
    Evaluation uses the shared AVIS evaluation set. External methods run released
    official checkpoints, and each WSM request activates its registered task route. Each quality
    metric occupies its own column. Parameters report the active route; WSM activates the 4B base and one
    12.5M-parameter TunnelingNet.
    Training Sample Budget is the number of optimizer updates multiplied by the
    global batch size, including all devices and gradient accumulation. Relative
    Sample Budget is the ratio of processed examples to the corresponding WSM
    TunnelingNet. Frame-based models retain their native
    training units. Training Time/Hardware gives the training cost reported in each paper or
    official repository. WSM uses the elapsed time to the selected checkpoint on
    one A100, including periodic validation. Hardware-specific values remain in
    their original units. For MODUS, the reported token budgets, per-device packed
    sequence lengths, and 64-GPU launch configuration recover approximately
    68.8K optimizer updates and 4.40M packed sequences; because a packed
    sequence may contain multiple examples, its reported ratio is a lower
    bound. Scaling Diffusion Perception reports 4K updates after MoE upcycling,
    but does not disclose the preceding mixed-task update count, global batch
    size, or wall time. N/R denotes a factor not reported by the paper or recoverable from
    the official code. Bold marks the best reported value in each
    arrow-directed column within a task.
    Inference Steps counts denoising/sampling updates (GeoWizard reports $50{\times}10$
    for 50 steps and an ensemble of 10; feed-forward VGGT uses one pass).
    Total Inference Time is summed steady-state inference time over the shared
    evaluation set; Inference Speed is recomputed from the same timing. Baseline
    timing excludes model loading and uses consecutive prediction writes. WSM
    timing includes amortized loading and output decoding. All measured
    single-GPU runs use an RTX 8000; SenseNova FP32 timing uses four RTX 8000
    GPUs. Only award designations are colored red. $\dagger$ marks methods that additionally receive the same first-frame RGB reference.}
    \label{tab:external_evaluation_summary}
    \tiny
    \setlength{\tabcolsep}{2.5pt}
    \renewcommand{\arraystretch}{1.10}
    \definecolor{externalHeaderBg}{RGB}{239,235,248}
    \definecolor{externalOursBg}{RGB}{235,246,252}
    % Keep interface badges proportional to the surrounding table-caption font.
    \newcommand{\externaltag}[1]{%
        \raisebox{-0.15ex}{\resizebox{!}{1.65ex}{\numtag{#1}}}%
    }
    % Center every cell box on the row baseline without fixed-width m-columns.
    \newsavebox{\externalcellbox}
    \newcommand{\externalvcellstart}{\setbox\externalcellbox=\hbox\bgroup}
    \newcommand{\externalvcellend}{%
        \egroup
        \raisebox{\dimexpr\dp\externalcellbox/2-\ht\externalcellbox/2\relax}%
            {\usebox{\externalcellbox}}%
    }
    \newcolumntype{L}{>{\externalvcellstart}l<{\externalvcellend}}
    \newcolumntype{C}{>{\externalvcellstart}c<{\externalvcellend}}

    \vspace{1pt}
    {\small\noindent\textbf{\externaltag{A} Estimation and reconstruction: RGB$\rightarrow$Depth.}
    Native normalized depth in $[0,1]$ is used; external predictions receive a fixed
    2nd--98th-percentile linear map, depth direction is aligned before scoring,
    and WSM uses its direct output.\par}
    \resizebox{\linewidth}{!}{%
    \begin{tabular}{@{}L*{16}{C}@{}}
        \toprule
        \rowcolor{externalHeaderBg}
        Method & Venue & Parameters &
        \shortstack{Training Sample Budget\\(Updates$\times$Global Batch)} &
        \shortstack{Relative Sample\\Budget}$\downarrow$ &
        \shortstack{Training Time\\/ Hardware} &
        MAE$\downarrow$ & RMSE$\downarrow$ & PSNR$\uparrow$ &
        Acc$_{.05}\uparrow$ & Acc$_{.10}\uparrow$ & Acc$_{.20}\uparrow$ &
        Pearson$\uparrow$ & Spearman$\uparrow$ &
        \shortstack{Inference\\Steps}$\downarrow$ &
        \shortstack{Total Inference\\Time (s)}$\downarrow$ &
        \shortstack{Inference Speed\\(FPS)}$\uparrow$ \\
        \midrule
        Marigold~\cite{Ke_2024_CVPR} & \begin{tabular}[c]{@{}c@{}}CVPR 2024\\\textcolor{red}{Best Paper Award Candidate}\end{tabular} & $0.9$B & $18$K$\times32$ & $331.0$ & \shortstack{A few days\\1 GPU} & 0.247 & 0.291 & 11.22 & 0.183 & 0.293 & 0.441 & 0.890 & 0.875 & $1{\times}10$ & 105.47 & 0.2181 \\
        DMP~\cite{Lee_2024_CVPR} & CVPR 2024 & $0.86$B+LoRA & $50$K$\times8$ & $229.9$ & N/R & 0.277 & 0.324 & 10.14 & 0.119 & 0.211 & 0.386 & 0.736 & 0.693 & 5 & 24.38 & 0.9434 \\
        GeoWizard~\cite{fu2024geowizard} & ECCV 2024 & $0.9$B & $20$K$\times256$ & $2.94$K & \shortstack{Time N/R\\8 GPUs} & 0.308 & 0.384 & 8.72 & 0.237 & 0.345 & 0.440 & 0.818 & 0.795 & $50{\times}10$ & 7834.00 & 0.0029 \\
        Scaling Diffusion Perception~\cite{Ravishankar_2025_CVPR} & CVPR 2025 & $2.8$B ($1.0$B active) & \shortstack{Mixed-task phase N/R\\+$4$K MoE updates} & N/R & N/R & 0.232 & 0.274 & 12.22 & 0.235 & 0.344 & 0.508 & 0.868 & 0.825 & $50{\times}3$ & 833.92 & 0.0276 \\
        VGGT~\cite{Wang_2025_CVPR} & \begin{tabular}[c]{@{}c@{}}CVPR 2025\\\textcolor{red}{Best Paper Award}\end{tabular} & $1.2$B & $160$K$\times3{,}072$ frames & N/R & \shortstack{9 days/64 A100s\\13.8K A100-h} & 0.243 & 0.296 & 10.98 & 0.193 & 0.297 & 0.453 & 0.719 & 0.698 & \textbf{1 pass} & \textbf{4.83} & \textbf{4.7619} \\
        MODUS~\cite{ye2026modus} & ICML 2026 & $14$B ($7$B active) & $\sim68.8$K$\times64$ packed seq. & $\geq2.53$K & \shortstack{88.5 h/64 GH200s\\5,664 GH200-h} & 0.167 & 0.216 & 14.04 & 0.335 & 0.479 & 0.666 & 0.867 & 0.799 & 50 & 9025.88 & 0.0025 \\
        SenseNova-Vision~\cite{sensenova2026sensenovavision} & Technical Report & $7$B & $50$K steps (packed) & N/R & \shortstack{Time N/R\\$\geq16\times$80GB GPUs} & 0.171 & 0.217 & 13.90 & 0.314 & 0.444 & 0.637 & 0.856 & 0.834 & 50 & 3471.00 (4G) & 0.0066 \\
        \rowcolor{externalOursBg}
        \textbf{WSM} & Ours & $4$B+$12.5$M & $1740\times1$ & \textbf{1.0} & \shortstack{11.0 h\\1 A100} & \textbf{0.130} & \textbf{0.162} & \textbf{17.22} & \textbf{0.415} & \textbf{0.590} & \textbf{0.770} & \textbf{0.935} & \textbf{0.907} & 10 & 222.41 & 0.1034 \\
        \bottomrule
    \end{tabular}}

    \vspace{3pt}
    {\small\noindent\textbf{\externaltag{B} Conditional RGB generation: Edge$\rightarrow$RGB.}
    Canny thresholds are $100/200$.\par}
    \resizebox{\linewidth}{!}{%
    \begin{tabular}{@{}L*{13}{C}@{}}
        \toprule
        \rowcolor{externalHeaderBg}
        Method & Venue & Parameters &
        \shortstack{Training Sample Budget\\(Updates$\times$Global Batch)} &
        \shortstack{Relative Sample\\Budget}$\downarrow$ &
        \shortstack{Training Time\\/ Hardware} &
        \shortstack{Edge Density\\Error}$\downarrow$ & MAE$\downarrow$ & MSE$\downarrow$ &
        PSNR$\uparrow$ & SSIM$\uparrow$ &
        \shortstack{Inference\\Steps}$\downarrow$ &
        \shortstack{Total Inference\\Time (s)}$\downarrow$ &
        \shortstack{Inference Speed\\(FPS)}$\uparrow$ \\
        \midrule
        ControlNet~\cite{zhang2023adding} & \begin{tabular}[c]{@{}c@{}}ICCV 2023\\\textcolor{red}{Best Paper Award}\end{tabular} & $1.4$B & \shortstack{N/R$\times256$ (v1.1)} & N/R & \shortstack{3 days/8 A100s (v1.1)\\$\geq1{,}176$ A100-h total} & 0.0286 & 0.276 & 0.120 & 9.90 & 0.127 & 30 & \textbf{79.17} & \textbf{0.3032} \\
        \shortstack{IP-Adapter~\cite{ye2023ipadapter}\\(with ControlNet~\cite{zhang2023adding})}$^\dagger$ & Technical Report & $2.0$B & \shortstack{ControlNet: N/R$\times256$\\IP-Adapter: $1$M$\times64$} & $\geq145.5$K & \shortstack{ControlNet: 3 days/8 A100s\\IP-Adapter: time N/R/8 V100s} & 0.0387 & 0.242 & 0.100 & 10.72 & 0.271 & 30 & 87.21 & 0.2752 \\
        T2I-Adapter~\cite{mou2024t2iadapter} & AAAI 2024 & $2.6$B+$79$M & $20$K$\times256$ & $11.6$K & \shortstack{Time N/R\\4 A100s} & 0.0265 & 0.281 & 0.118 & 9.84 & 0.060 & 30 & 262.46 & 0.0914 \\
        UniControl~\cite{qin2023unicontrol} & NeurIPS 2023 & $1.5$B & $\sim900$K$\times64$ & $\sim130.9$K & \shortstack{$>5{,}000$ A100-h\\A100-40G} & 0.0298 & 0.293 & 0.129 & 9.36 & 0.058 & 50 & 265.64 & 0.0903 \\
        Uni-ControlNet~\cite{zhao2023unicontrolnet} & NeurIPS 2023 & $1.5$B & 1 epoch ($10$M pairs) & $22.7$K & N/R & 0.0597 & 0.284 & 0.118 & 9.77 & 0.083 & 50 & 381.89 & 0.0628 \\
        \rowcolor{externalOursBg}
        \textbf{WSM}$^\dagger$ & Ours & $4$B+$12.5$M & $440\times1$ & \textbf{1.0} & \shortstack{8.5 min\\1 A100} & \textbf{0.0157} & \textbf{0.087} & \textbf{0.021} & \textbf{19.51} & \textbf{0.762} & \textbf{10} & 202.56 & 0.1185 \\
        \bottomrule
    \end{tabular}}

    \vspace{3pt}
    {\small\noindent\textbf{\externaltag{C} Any-to-any control over representative structured signals.}
    Edge, relative depth, surface normals, and human-part semantics span
    geometric boundaries, spatial layout, local surface shape, and structured
    human understanding.\par}

    \vspace{1pt}
    \resizebox{\linewidth}{!}{%
    \begin{tabular}{@{}L*{12}{C}@{}}
        \toprule
        \rowcolor{externalHeaderBg}
        Method & Task & Venue & Parameters &
        \shortstack{Training Sample Budget\\(Updates$\times$Global Batch)} &
        \shortstack{Relative Sample\\Budget}$\downarrow$ &
        \shortstack{Training Time\\/ Hardware} &
        Accuracy$\uparrow$ & FPR$\downarrow$ & Specificity$\uparrow$ &
        \shortstack{Inference\\Steps}$\downarrow$ &
        \shortstack{Total Inference\\Time (s)}$\downarrow$ &
        \shortstack{Inference Speed\\(FPS)}$\uparrow$ \\
        \midrule
        MODUS~\cite{ye2026modus} & Normal$\rightarrow$Edge & ICML 2026 & $14$B ($7$B active) & $\sim68.8$K$\times64$ packed seq. & $\geq2.20$K & \shortstack{88.5 h/64 GH200s\\5,664 GH200-h} & 0.8671 & 0.0569 & 0.9431 & 50 & 12345.09 & 0.0019 \\
        \rowcolor{externalOursBg}
        \textbf{WSM} & Normal$\rightarrow$Edge & Ours & $4$B+$12.5$M & $2000\times1$ & \textbf{1.0} & \shortstack{9.3 h\\1 A100} & \textbf{0.8797} & \textbf{0.0296} & \textbf{0.9704} & \textbf{10} & \textbf{239.76} & \textbf{0.1001} \\
        \bottomrule
    \end{tabular}}

    \vspace{2pt}
    \resizebox{\linewidth}{!}{%
    \begin{tabular}{@{}L*{14}{C}@{}}
        \toprule
        \rowcolor{externalHeaderBg}
        Method & Task & Venue & Parameters &
        \shortstack{Training Sample Budget\\(Updates$\times$Global Batch)} &
        \shortstack{Relative Sample\\Budget}$\downarrow$ &
        \shortstack{Training Time\\/ Hardware} &
        \shortstack{Mean Angle\\($^\circ$)}$\downarrow$ &
        \shortstack{Median Angle\\($^\circ$)}$\downarrow$ &
        $11.25^\circ\uparrow$ & $22.5^\circ\uparrow$ & $30^\circ\uparrow$ &
        \shortstack{Inference\\Steps}$\downarrow$ &
        \shortstack{Total Inference\\Time (s)}$\downarrow$ &
        \shortstack{Inference Speed\\(FPS)}$\uparrow$ \\
        \midrule
        MODUS~\cite{ye2026modus} & \shortstack{Relative Depth\\$\rightarrow$Normal} & ICML 2026 & $14$B ($7$B active) & $\sim68.8$K$\times64$ packed seq. & $\geq2.25$K & \shortstack{88.5 h/64 GH200s\\5,664 GH200-h} & 35.19 & 31.13 & 0.222 & 0.431 & 0.512 & 50 & 12272.18 & 0.0020 \\
        \rowcolor{externalOursBg}
        \textbf{WSM} & \shortstack{Relative Depth\\$\rightarrow$Normal} & Ours & $4$B+$12.5$M & $1960\times1$ & \textbf{1.0} & \shortstack{9.6 h\\1 A100} & \textbf{26.40} & \textbf{22.48} & \textbf{0.231} & \textbf{0.527} & \textbf{0.664} & \textbf{10} & \textbf{1767.4} & \textbf{0.0136} \\
        \bottomrule
    \end{tabular}}

    \vspace{2pt}
    \resizebox{\linewidth}{!}{%
    \begin{tabular}{@{}L*{14}{C}@{}}
        \toprule
        \rowcolor{externalHeaderBg}
        Method & Task & Venue & Parameters &
        \shortstack{Training Sample Budget\\(Updates$\times$Global Batch)} &
        \shortstack{Relative Sample\\Budget}$\downarrow$ &
        \shortstack{Training Time\\/ Hardware} &
        \shortstack{Pixel\\Accuracy}$\uparrow$ & mIoU$\uparrow$ &
        \shortstack{Human\\IoU}$\uparrow$ & mAcc$\uparrow$ &
        Macro-F1$\uparrow$ &
        \shortstack{Inference\\Steps}$\downarrow$ &
        \shortstack{Total Inference\\Time (s)}$\downarrow$ &
        \shortstack{Inference Speed\\(FPS)}$\uparrow$ \\
        \midrule
        MODUS~\cite{ye2026modus} & \shortstack{Edge$\rightarrow$Human\\Segmentation} & ICML 2026 & $14$B ($7$B active) & $\sim68.8$K$\times64$ packed seq. & $\geq2.32$K & \shortstack{88.5 h/64 GH200s\\5,664 GH200-h} & 0.8431 & 0.6577 & 0.5148 & 0.7676 & 0.7630 & 50 & 9100.77 & 0.0026 \\
        \rowcolor{externalOursBg}
        \textbf{WSM} & \shortstack{Edge$\rightarrow$Human\\Segmentation} & Ours & $4$B+$12.5$M & $1900\times1$ & \textbf{1.0} & \shortstack{9.6 h\\1 A100} & \textbf{0.8862} & \textbf{0.6918} & \textbf{0.5391} & \textbf{0.7753} & \textbf{0.7780} & \textbf{10} & \textbf{1579.62} & \textbf{0.0152} \\
        \bottomrule
    \end{tabular}}
\end{table*}

\FloatBarrier
\subsection{External Comparison on the AVIS Evaluation Set}
\label{subsec:external_evaluation_comparison}

\paragraph{Comparison scope.}
Table~\ref{tab:external_evaluation_summary} covers five representative established design
paradigms: new trainable parameter blocks (ControlNet), the MoE-style
multi-expert adapter in UniControl, LoRA (DMP),
expert-based MoE-family architectures---the mixture-of-transformers in MODUS
and strict sparse-MoE diffusion in Scaling Diffusion Perception---and
IC-Wrapper methods (T2I-Adapter and Uni-ControlNet). WSM introduces internal
trajectory tunneling. The three blocks correspond to Interfaces \numtag{A}--\numtag{C}:
RGB-to-depth estimation, Edge-to-RGB generation, and any-to-any control.
We also include VGGT as a task-specific depth estimator.
Interface \numtag{C} evaluates transfers among edge, relative depth, surface
normal, and human-part segmentation signals. Both WSM and MODUS support a
broader transfer inventory; we select these tasks as a representative
cross-section of structured understanding, spanning object and region
boundaries, scene-scale spatial organization, local surface orientation, and
human semantics.
\textbf{Across all three externally comparable interfaces and all five representative
tasks, WSM leads every reported quality metric, demonstrating that its advantage
holds across all five adaptation paradigms.} Interface \numtag{D} combines first-frame RGB evidence
with a later auxiliary observation, for which we found no released image-level
method with matching inputs. Most remaining mappings enabled by this interface
similarly define new task settings without a like-for-like external comparator.

\paragraph{Evaluation data.}
Images from widely used benchmark datasets may have been included in the
web-scale pretraining corpora of large foundation models. We therefore use the
AVIS evaluation set, which is less commonly used for dense prediction, and generate
the task annotations evaluated here ourselves. This choice reduces the
likelihood of direct overlap with existing image--annotation pairs. None of the
evaluation images or annotations is used to train WSM. A semantic analysis of the
scene captions also finds almost no close match between the evaluation scenes and the
WSM training descriptions. Every method receives the same AVIS
inputs and targets. The external methods use their released checkpoints, and
WSM uses its trained TunnelingNets without further optimization. The WSM TunnelingNets for depth and Edge-to-RGB use
TunnelingNets 1 and 424; TunnelingNet 424 uses its recommended native-reference route.
Interface \numtag{C} uses TunnelingNets 275, 119, and 434 for Normal-to-Edge,
Relative-Depth-to-Normal, and Edge-to-Human-Segmentation, respectively.
\textbf{The comparison therefore measures all methods on the same unseen scenes
and annotations.}

\paragraph{Interface C metrics.}
For Normal$\rightarrow$Edge, accuracy, false-positive rate (FPR), and
specificity are computed from a shared binary edge threshold. Surface normals
are decoded into the same physical XYZ convention and evaluated only over
valid, unpadded pixels.
Mean and median errors are measured in degrees, and the three threshold
accuracies report the fraction of pixels below $11.25^\circ$, $22.5^\circ$, and
$30^\circ$. For Edge$\rightarrow$Human Segmentation, WSM predicts the complete
channel-103 target with background and five human-part labels. MODUS produces
a category-conditioned person mask, so the shared comparison merges the five
human-part labels into one human foreground class for both WSM and the target.
We report pixel accuracy, mean IoU (mIoU), human-foreground IoU, mean class
accuracy (mAcc), and macro-F1 under this common ontology.

\paragraph{Task parameters and training efficiency.}
Each WSM task adds a 12.5M-parameter TunnelingNet to the shared 4B model, only
0.31\% of the base model. Training Sample Budget measures optimizer updates
times global batch size. Across the eleven method--task comparisons with
recoverable budgets, external methods process 230--145.5K times as many training
examples as WSM; ratios for packed-sequence models remain lower bounds.
\textbf{WSM is 29.4K times more sample-efficient than the evaluated external
methods on average and 2.53K times more sample-efficient at the median, while
leading every reported quality metric.}

The reported training costs show the same scale of advantage. The Edge-to-RGB
TunnelingNet reaches its selected checkpoint in 8.5 minutes on one A100. ControlNet
Canny 1.1 alone continues training on eight A100s for three days, with more than
1,176 A100-hours across its complete training path, while UniControl reports
more than 5,000 A100-hours. The three any-to-any TunnelingNets reach their selected
checkpoints in 9.3--9.6 hours on one A100, compared with 5,664 GH200-hours for
MODUS; the depth TunnelingNet takes 11.0 A100-hours,
compared with 13.8K A100-hours for VGGT. These runs cover different task scopes,
so the table retains the original hardware and units. ControlNet does not
provide enough information to recover its optimizer-update count.
\textbf{WSM reaches its strongest checkpoints in minutes to hours on one GPU,
while several evaluated external methods report thousands of accelerator-hours.}

Across the 15 external method--task comparisons, WSM is 11.02 times faster
than the evaluated external methods on average and 1.89 times faster at the
median. This aggregate includes the specialized feed-forward depth estimators
that run faster than WSM.
\textbf{Across the complete comparison, WSM is 11.02 times faster than the
evaluated external methods on average and 1.89 times faster at the median.}

\paragraph{Results.}
\textbf{For relative depth, WSM leads all eight quality measures.} Its
MAE/RMSE are 0.130/0.162, compared with the next-best 0.167/0.216 from MODUS.
WSM also surpasses VGGT, the CVPR 2025 \textcolor{red}{Best Paper Award}
recipient, which obtains 0.243/0.296, and Marigold, a CVPR 2024
\textcolor{red}{Best Paper Award Candidate}, which obtains 0.247/0.291.
Scaling Diffusion Perception, a strict sparse-MoE DiT with eight experts and
top-2 routing, obtains 0.232/0.274. WSM reaches Pearson/Spearman correlations
of 0.935/0.907 and leads all three threshold accuracies. WSM predicts the
normalized depth target directly; external outputs receive the fixed
percentile-based linear mapping, with depth direction aligned before scoring.

\textbf{For conditional RGB, WSM leads all five quality measures against all
five external methods.} Against ControlNet, the ICCV 2023 \textcolor{red}{Best Paper
Award} recipient, WSM reduces MAE from 0.276 to 0.087 and MSE from 0.120 to
0.021, while increasing PSNR from 9.90 to 19.51 and SSIM from 0.127 to 0.762.
The interface-matched ControlNet+IP-Adapter composition improves ControlNet to
0.242 MAE and 0.271 SSIM after receiving the same first-frame RGB reference.
WSM still reduces its MAE by 64.1\% and raises SSIM by 0.491, while reducing
the best external edge-density error from 0.0265 to 0.0157 and using
10 inference steps, one third of ControlNet's 30 steps.
Reliable FID estimation requires a substantially larger evaluation pool and is
reserved for the large-scale evaluation.

\textbf{Interface \numtag{C} evaluates three representative structured transfers,
and WSM leads all 13 reported quality measures.} For Normal$\rightarrow$Edge,
WSM raises pixel accuracy from 0.8671 to 0.8797, reduces FPR from 0.0569 to
0.0296, and raises specificity from 0.9431 to 0.9704. For Relative
Depth$\rightarrow$Normal, the common physical-XYZ protocol reduces mean/median
angular error from $35.19^\circ/31.13^\circ$ to
$26.40^\circ/22.48^\circ$, while raising the $11.25^\circ$, $22.5^\circ$,
and $30^\circ$ accuracies from 0.222/0.431/0.512 to 0.231/0.527/0.664.
For Edge$\rightarrow$Human Segmentation, WSM raises pixel accuracy from 0.8431
to 0.8862 and mIoU from 0.6577 to 0.6918. It also improves human-foreground IoU
from 0.5148 to 0.5391, mAcc from 0.7676 to 0.7753, and macro-F1 from 0.7630 to
0.7780.
\textbf{These gains cover geometric boundaries, spatial layout, local surface
orientation, and human semantics while using five times fewer inference steps
and at least 2.2K times fewer training samples than MODUS, whose BAGEL backbone
uses a two-expert mixture-of-transformers architecture.}

\textbf{The central result is simple: one WSM design wins every reported quality
metric across five representative tasks, three interfaces, and five adaptation
paradigms. It is 29.4K times more sample-efficient and 11.02 times faster than the
evaluated external methods on average.} The gains include ControlNet, the ICCV 2023 \textcolor{red}{Best Paper
Award} recipient; VGGT, the CVPR 2025 \textcolor{red}{Best Paper Award}
recipient; and Marigold, a CVPR 2024 \textcolor{red}{Best Paper Award
Candidate}. The remaining WSM interface opens task combinations for which
comparable image-level systems do not yet exist.

\input{sec_preprint/5_postrelease}

% The snapshot deliberately preserves artifact provenance. For 1,076 TunnelingNets the
% evaluated run is the exact locally available recommendation. The remaining 72
% entries are connected to material channel 131 and used contract-compatible
% earlier runs because a later cluster-only retraining queue was absent from the
% evaluation mirror. We report the frozen snapshot here and retain the local
% re-evaluation as a separate paired evaluation, keeping model revisions separate.

\FloatBarrier

%% file: tables/evaluation_interface_coverage_normv2.tex
\begin{tabular}{lrrrr}
\toprule
Interface & Heads & Inputs & Targets & Evaluations \\
\midrule
RGB estimation / reconstruction & 38 & 1 & 38 & 692 \\
Text + WSG narrow generation & 38 & 38 & 1 & 704 \\
WSG conditional control & 536 & 37 & 37 & 8,411 \\
RGB-assisted inverse rendering & 536 & 37 & 37 & 8,384 \\
\midrule
Total & 1,148 & -- & 39 & 18,191 \\
\bottomrule
\end{tabular}

%% file: sec_preprint/5_postrelease.tex
% Table 7 and Section 5.3: post-release visual evaluation.
\begin{table*}[p]
    \centering
    \caption{\textbf{External comparison on post-release visual data.}
    Quality metrics are recomputed on frames drawn from recently published
    Wikimedia Commons videos. Model size, training budget, training cost,
    sampling steps, and controlled runtime measurements follow
    Table~\ref{tab:external_evaluation_summary}, since every method retains the
    same checkpoint and inference protocol. Training Sample Budget is the
    number of optimizer updates multiplied by the global batch size, and
    Relative Sample Budget is its ratio to the corresponding WSM TunnelingNet. N/R
    denotes a factor not reported by the paper or recoverable from the official
    code. Bold marks the best reported value in each arrow-directed column. $\dagger$
    marks methods that additionally receive the same first-frame RGB reference.}
    \label{tab:external_postrelease_summary}
    \tiny
    \setlength{\tabcolsep}{2.5pt}
    \renewcommand{\arraystretch}{1.10}
    \definecolor{postreleaseHeaderBg}{RGB}{239,235,248}
    \definecolor{postreleaseOursBg}{RGB}{235,246,252}
    \newcommand{\postreleasetag}[1]{%
        \raisebox{-0.15ex}{\resizebox{!}{1.65ex}{\numtag{#1}}}%
    }
    \newsavebox{\postreleasecellbox}
    \newcommand{\postreleasevcellstart}{\setbox\postreleasecellbox=\hbox\bgroup}
    \newcommand{\postreleasevcellend}{%
        \egroup
        \raisebox{\dimexpr\dp\postreleasecellbox/2-\ht\postreleasecellbox/2\relax}%
            {\usebox{\postreleasecellbox}}%
    }
    \newcolumntype{J}{>{\postreleasevcellstart}l<{\postreleasevcellend}}
    \newcolumntype{K}{>{\postreleasevcellstart}c<{\postreleasevcellend}}

    \vspace{1pt}
    {\small\noindent\textbf{\postreleasetag{A} Estimation and reconstruction: RGB$\rightarrow$Relative Depth.}\par}
    \resizebox{\linewidth}{!}{%
    \begin{tabular}{@{}J*{12}{K}@{}}
        \toprule
        \rowcolor{postreleaseHeaderBg}
        Method & Venue & Parameters &
        \shortstack{Training Sample Budget\\(Updates$\times$Global Batch)} &
        \shortstack{Relative Sample\\Budget}$\downarrow$ &
        \shortstack{Training Time\\/ Hardware} &
        RMSE$\downarrow$ & Acc$_{.20}\uparrow$ & Pearson$\uparrow$ & Spearman$\uparrow$ &
        \shortstack{Inference\\Steps}$\downarrow$ &
        \shortstack{Total Inference\\Time (s)}$\downarrow$ &
        \shortstack{Inference Speed\\(FPS)}$\uparrow$ \\
        \midrule
        Marigold~\cite{Ke_2024_CVPR} & \shortstack{CVPR 2024\\\textcolor{red}{Best Paper Award Candidate}} & $0.9$B & $18$K$\times32$ & $331.0$ & \shortstack{A few days\\1 GPU} & 0.474 & 0.224 & 0.627 & 0.814 & $1{\times}10$ & 105.47 & 0.2181 \\
        DMP~\cite{Lee_2024_CVPR} & CVPR 2024 & $0.86$B+LoRA & $50$K$\times8$ & $229.9$ & N/R & 0.427 & 0.279 & 0.593 & 0.698 & 5 & 24.38 & 0.9434 \\
        GeoWizard~\cite{fu2024geowizard} & ECCV 2024 & $0.9$B & $20$K$\times256$ & $2.94$K & \shortstack{Time N/R\\8 GPUs} & 0.508 & 0.304 & 0.484 & 0.569 & $50{\times}10$ & 7834.00 & 0.0029 \\
        Scaling Diffusion Perception~\cite{Ravishankar_2025_CVPR} & CVPR 2025 & $2.8$B ($1.0$B active) & \shortstack{Mixed-task phase N/R\\+$4$K MoE updates} & N/R & N/R & 0.370 & 0.359 & 0.736 & 0.741 & $50{\times}3$ & 833.92 & 0.0276 \\
        VGGT~\cite{Wang_2025_CVPR} & \shortstack{CVPR 2025\\\textcolor{red}{Best Paper Award}} & $1.2$B & $160$K$\times3{,}072$ frames & N/R & \shortstack{9 days/64 A100s\\13.8K A100-h} & 0.316 & 0.420 & 0.737 & 0.695 & \textbf{1 pass} & \textbf{4.83} & \textbf{4.7619} \\
        MODUS~\cite{ye2026modus} & ICML 2026 & $14$B ($7$B active) & $\sim68.8$K$\times64$ packed seq. & $\geq2.53$K & \shortstack{88.5 h/64 GH200s\\5,664 GH200-h} & 0.225 & 0.632 & 0.858 & 0.826 & 50 & 9025.88 & 0.0025 \\
        SenseNova-Vision~\cite{sensenova2026sensenovavision} & Technical Report & $7$B & $50$K steps (packed) & N/R & \shortstack{Time N/R\\$\geq16\times$80GB GPUs} & 0.213 & 0.628 & 0.827 & 0.808 & 50 & 3471.00 (4G) & 0.0066 \\
        \rowcolor{postreleaseOursBg}
        \textbf{WSM} & Ours & $4$B+$12.5$M & $1740\times1$ & \textbf{1.0} & \shortstack{11.0 h\\1 A100} & \textbf{0.207} & \textbf{0.654} & \textbf{0.866} & \textbf{0.839} & 10 & 222.41 & 0.1034 \\
        \bottomrule
    \end{tabular}}

    \vspace{3pt}
    {\small\noindent\textbf{\postreleasetag{B} Conditional RGB generation: Edge$\rightarrow$RGB.}\par}
    \resizebox{\linewidth}{!}{%
    \begin{tabular}{@{}J*{15}{K}@{}}
        \toprule
        \rowcolor{postreleaseHeaderBg}
        Method & Venue & Parameters &
        \shortstack{Training Sample Budget\\(Updates$\times$Global Batch)} &
        \shortstack{Relative Sample\\Budget}$\downarrow$ &
        \shortstack{Training Time\\/ Hardware} &
        \shortstack{Edge Density\\Error}$\downarrow$ & $F_{0.5}\uparrow$ & Precision$\uparrow$ & MAE$\downarrow$ & MSE$\downarrow$ &
        PSNR$\uparrow$ & SSIM$\uparrow$ &
        \shortstack{Inference\\Steps}$\downarrow$ &
        \shortstack{Total Inference\\Time (s)}$\downarrow$ &
        \shortstack{Inference Speed\\(FPS)}$\uparrow$ \\
        \midrule
        ControlNet~\cite{zhang2023adding} & \shortstack{ICCV 2023\\\textcolor{red}{Best Paper Award}} & $1.4$B & \shortstack{N/R$\times256$ (v1.1)} & N/R & \shortstack{3 days/8 A100s (v1.1)\\$\geq1{,}176$ A100-h total} & 0.0288 & 0.2821 & 0.2926 & 0.261 & 0.105 & 10.25 & 0.093 & 30 & \textbf{79.17} & \textbf{0.3032} \\
        \shortstack{IP-Adapter~\cite{ye2023ipadapter}\\(with ControlNet~\cite{zhang2023adding})}$^\dagger$ & Technical Report & $2.0$B & \shortstack{ControlNet: N/R$\times256$\\IP-Adapter: $1$M$\times64$} & $\geq145.5$K & \shortstack{ControlNet: 3 days/8 A100s\\IP-Adapter: time N/R/8 V100s} & 0.0186 & 0.3070 & 0.3094 & 0.218 & 0.079 & 11.47 & 0.226 & 30 & 87.21 & 0.2752 \\
        T2I-Adapter~\cite{mou2024t2iadapter} & AAAI 2024 & $2.6$B+$79$M & $20$K$\times256$ & $11.6$K & \shortstack{Time N/R\\4 A100s} & 0.0520 & 0.2377 & 0.3043 & 0.246 & 0.092 & 10.83 & 0.144 & 30 & 262.46 & 0.0914 \\
        UniControl~\cite{qin2023unicontrol} & NeurIPS 2023 & $1.5$B & $\sim900$K$\times64$ & $\sim130.9$K & \shortstack{$>5{,}000$ A100-h\\A100-40G} & 0.0505 & 0.2235 & 0.3208 & 0.246 & 0.091 & 10.94 & 0.044 & 50 & 265.64 & 0.0903 \\
        Uni-ControlNet~\cite{zhao2023unicontrolnet} & NeurIPS 2023 & $1.5$B & 1 epoch ($10$M pairs) & $22.7$K & N/R & 0.0816 & 0.1781 & 0.1708 & 0.264 & 0.106 & 10.22 & 0.073 & 50 & 381.89 & 0.0628 \\
        \rowcolor{postreleaseOursBg}
        \textbf{WSM}$^\dagger$ & Ours & $4$B+$12.5$M & $440\times1$ & \textbf{1.0} & \shortstack{8.5 min\\1 A100} & \textbf{0.0116} & \textbf{0.3107} & \textbf{0.3265} & \textbf{0.074} & \textbf{0.013} & \textbf{19.83} & \textbf{0.845} & \textbf{10} & 202.56 & 0.1185 \\
        \bottomrule
    \end{tabular}}

    \vspace{3pt}
    {\small\noindent\textbf{\postreleasetag{C} Any-to-any control over representative structured signals.}\par}
    \resizebox{\linewidth}{!}{%
    \begin{tabular}{@{}J*{16}{K}@{}}
        \toprule
        \rowcolor{postreleaseHeaderBg}
        Method & Task & Venue & Parameters &
        \shortstack{Training Sample Budget\\(Updates$\times$Global Batch)} &
        \shortstack{Relative Sample\\Budget}$\downarrow$ &
        \shortstack{Training Time\\/ Hardware} &
        \shortstack{Balanced\\Accuracy}$\uparrow$ & F1$\uparrow$ & F2$\uparrow$ &
        FNR$\downarrow$ & IoU$\uparrow$ & NPV$\uparrow$ & Recall$\uparrow$ &
        \shortstack{Inference\\Steps}$\downarrow$ &
        \shortstack{Total Inference\\Time (s)}$\downarrow$ &
        \shortstack{Inference Speed\\(FPS)}$\uparrow$ \\
        \midrule
        MODUS~\cite{ye2026modus} & Normal$\rightarrow$Edge & ICML 2026 & $14$B ($7$B active) & $\sim68.8$K$\times64$ packed seq. & $\geq2.20$K & \shortstack{88.5 h/64 GH200s\\5,664 GH200-h} & 0.5531 & 0.1697 & 0.1720 & 0.8192 & 0.0939 & 0.9212 & 0.1808 & 50 & 12345.09 & 0.0019 \\
        \rowcolor{postreleaseOursBg}
        \textbf{WSM} & Normal$\rightarrow$Edge & Ours & $4$B+$12.5$M & $2000\times1$ & \textbf{1.0} & \shortstack{9.3 h\\1 A100} & \textbf{0.5702} & \textbf{0.1831} & \textbf{0.2273} & \textbf{0.7068} & \textbf{0.1028} & \textbf{0.9253} & \textbf{0.2932} & \textbf{10} & \textbf{239.76} & \textbf{0.1001} \\
        \bottomrule
    \end{tabular}}

    \vspace{2pt}
    \resizebox{\linewidth}{!}{%
    \begin{tabular}{@{}J*{14}{K}@{}}
        \toprule
        \rowcolor{postreleaseHeaderBg}
        Method & Task & Venue & Parameters &
        \shortstack{Training Sample Budget\\(Updates$\times$Global Batch)} &
        \shortstack{Relative Sample\\Budget}$\downarrow$ &
        \shortstack{Training Time\\/ Hardware} &
        \shortstack{Mean Angle\\($^\circ$)}$\downarrow$ &
        \shortstack{Median Angle\\($^\circ$)}$\downarrow$ &
        $11.25^\circ\uparrow$ & $22.5^\circ\uparrow$ & $30^\circ\uparrow$ &
        \shortstack{Inference\\Steps}$\downarrow$ &
        \shortstack{Total Inference\\Time (s)}$\downarrow$ &
        \shortstack{Inference Speed\\(FPS)}$\uparrow$ \\
        \midrule
        MODUS~\cite{ye2026modus} & \shortstack{Relative Depth\\$\rightarrow$Normal} & ICML 2026 & $14$B ($7$B active) & $\sim68.8$K$\times64$ packed seq. & $\geq2.25$K & \shortstack{88.5 h/64 GH200s\\5,664 GH200-h} & 37.56 & 37.29 & 0.189 & 0.399 & 0.503 & 50 & 12272.18 & 0.0020 \\
        \rowcolor{postreleaseOursBg}
        \textbf{WSM} & \shortstack{Relative Depth\\$\rightarrow$Normal} & Ours & $4$B+$12.5$M & $1960\times1$ & \textbf{1.0} & \shortstack{9.6 h\\1 A100} & \textbf{25.13} & \textbf{22.31} & \textbf{0.228} & \textbf{0.553} & \textbf{0.677} & \textbf{10} & \textbf{1767.4} & \textbf{0.0136} \\
        \bottomrule
    \end{tabular}}

    \vspace{2pt}
    \resizebox{\linewidth}{!}{%
    \begin{tabular}{@{}J*{13}{K}@{}}
        \toprule
        \rowcolor{postreleaseHeaderBg}
        Method & Task & Venue & Parameters &
        \shortstack{Training Sample Budget\\(Updates$\times$Global Batch)} &
        \shortstack{Relative Sample\\Budget}$\downarrow$ &
        \shortstack{Training Time\\/ Hardware} &
        \shortstack{Pixel\\Accuracy}$\uparrow$ &
        \shortstack{Macro\\F1}$\uparrow$ &
        \shortstack{Mean Class\\Accuracy}$\uparrow$ &
        mIoU$\uparrow$ &
        \shortstack{Inference\\Steps}$\downarrow$ &
        \shortstack{Total Inference\\Time (s)}$\downarrow$ &
        \shortstack{Inference Speed\\(FPS)}$\uparrow$ \\
        \midrule
        MODUS~\cite{ye2026modus} & \shortstack{Edge$\rightarrow$Human\\Segmentation} & ICML 2026 & $14$B ($7$B active) & $\sim68.8$K$\times64$ packed seq. & $\geq2.32$K & \shortstack{88.5 h/64 GH200s\\5,664 GH200-h} & \textbf{0.929} & \textbf{0.917} & \textbf{0.920} & \textbf{0.856} & 50 & 9100.77 & 0.0026 \\
        \rowcolor{postreleaseOursBg}
        \textbf{WSM} & \shortstack{Edge$\rightarrow$Human\\Segmentation} & Ours & $4$B+$12.5$M & $1900\times1$ & \textbf{1.0} & \shortstack{9.6 h\\1 A100} & 0.910 & 0.883 & 0.874 & 0.804 & \textbf{10} & \textbf{1579.62} & \textbf{0.0152} \\
        \bottomrule
    \end{tabular}}
\end{table*}

\begin{figure*}[p]
    \centering
    \includegraphics[width=\textwidth,height=0.86\textheight,keepaspectratio]{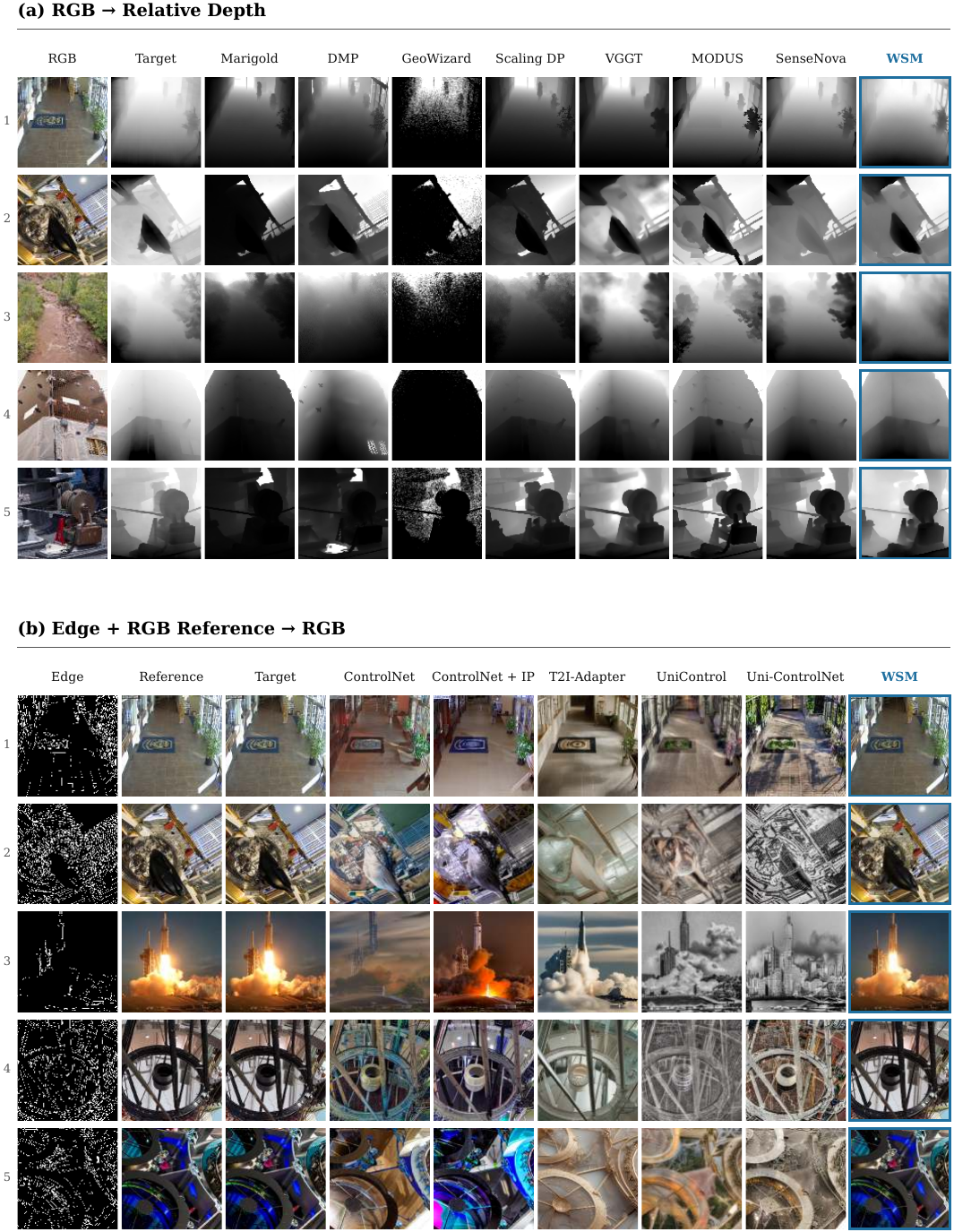}
    \caption{\textbf{Qualitative comparison for Interfaces \numtag{A} and
    \numtag{B} on the post-release evaluation set.}
    Each row contains a different crop-aligned example evaluated in
    Table~\ref{tab:external_postrelease_summary}. Panel (a) compares every
    relative-depth method, with depth maps displayed after the affine alignment
    used by the evaluation protocol. Panel (b) compares every conditional RGB
    method under identical edge conditions, RGB references, and targets. WSM is
    the rightmost column, outlined in blue.}
    \label{fig:external_postrelease_qualitative_ab}
\end{figure*}

\begin{figure*}[p]
    \centering
    \includegraphics[width=\textwidth,height=0.86\textheight,keepaspectratio]{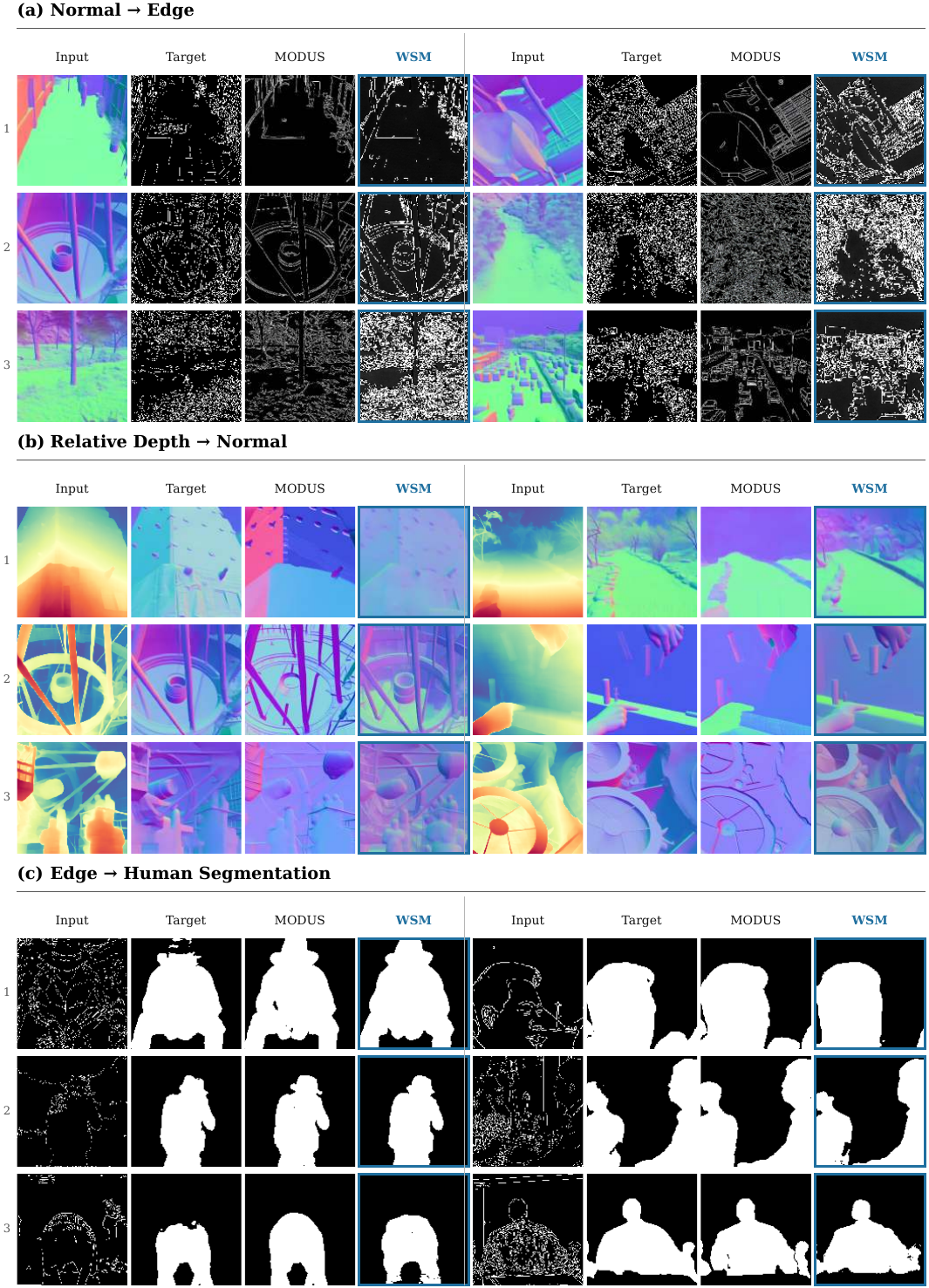}
    \caption{\textbf{Qualitative comparison for Interface \numtag{C} on the
    post-release evaluation set.}
    The three task blocks correspond to Normal$\rightarrow$Edge, Relative
    Depth$\rightarrow$Normal, and Edge$\rightarrow$Human Segmentation in
    Table~\ref{tab:external_postrelease_summary}. Within each four-column group,
    the input and target are followed by the MODUS and WSM predictions. WSM
    outputs are outlined in blue. The same spatial crop is applied to all four
    images within each group. For human segmentation, the target and both
    predictions are shown as binary human-foreground masks.}
    \label{fig:external_postrelease_qualitative_c}
\end{figure*}

\FloatBarrier
\subsection{Evaluation on Post-Release Visual Data}
\label{subsec:postrelease_visual_evaluation}

\paragraph{Evaluation data.}
To test whether the methods generalize to genuinely unseen images, we build a
second evaluation set from Wikimedia Commons videos released only after all evaluated
models had finished training. These videos became publicly available between
August 2 and September 4, 2026. Their frames therefore could not have been
included in training data collected from public sources. We extract frames and produce task annotations using the same
cropping and WSG pipeline as in Section~\ref{subsec:external_evaluation_comparison}.
Every method receives identical inputs and targets and is evaluated with its
trained model under the same inference protocol as
Table~\ref{tab:external_evaluation_summary}.

\paragraph{Comparison scope.}
Table~\ref{tab:external_postrelease_summary} preserves the three-interface
organization, method families, model sizes, and training-efficiency accounting
of Table~\ref{tab:external_evaluation_summary}. The quality columns are recomputed
on the post-release evaluation set. Interface \numtag{A} evaluates relative-depth
estimation, Interface \numtag{B} evaluates conditional RGB generation, and
Interface \numtag{C} evaluates transfers from surface normals to edges, from
relative depth to surface normals, and from edges to human-foreground
segmentation.

\paragraph{Qualitative comparison.}
Figures~\ref{fig:external_postrelease_qualitative_ab} and~\ref{fig:external_postrelease_qualitative_c} show the inputs, targets, and
predictions behind the aggregate scores. For relative depth, WSM preserves the
continuous floor-to-background ordering while retaining the plant and doorway
boundaries. For conditional RGB generation, it reconstructs the reference
scene without sacrificing the supplied edge layout; competing outputs introduce
larger changes in geometry, content, or appearance. The structured comparisons
show the same pattern across Interface \numtag{C}: WSM recovers a denser set of
target-aligned edges and maintains coherent surface orientation in the normal
map. In the human-segmentation examples, the common binary display isolates
foreground geometry: WSM preserves the dominant silhouette across close-up,
standing, seated, and bent poses without relying on part colors.
\textbf{The visual evidence therefore supports the quantitative result across
all three interfaces: WSM preserves scene-level structure and target geometry
on images released after training.}
\paragraph{Overall performance.}
\textbf{WSM ranks first on 23 quality measures across four of five
representative tasks, with leading results in every externally comparable
interface.} The advantage spans depth estimation, conditional RGB generation,
and structured signal transfer. Most notably, the same unified WSM framework
outperforms two \textcolor{red}{Best Paper Award} methods: VGGT on all four
reported depth metrics and ControlNet on all seven reported
conditional-generation metrics. WSM also exceeds the
\textcolor{red}{Best Paper Award Candidate} Marigold on all four reported depth
metrics.

\paragraph{Estimation and reconstruction.}
For RGB$\rightarrow$Relative Depth, WSM achieves the lowest RMSE and the highest
Acc$_{.20}$, Pearson, and Spearman scores among eight methods. Its RMSE of 0.207
improves over the next-best 0.213, while Acc$_{.20}$ rises from 0.632 to 0.654.
The simultaneous gains in linear and rank correlations show that WSM recovers
both relative depth magnitude and scene-wide depth ordering. \textbf{WSM is the
only method that leads every reported depth-quality measure on these
post-release images.}

\paragraph{Conditional RGB generation.}
For Edge$\rightarrow$RGB, WSM leads all seven reported quality measures. Relative
to the strongest competing value in each column, it reduces edge-density error
by 37.6\%, MAE by 66.1\%, and MSE by 83.5\%. It also improves PSNR by 8.36 dB and
raises SSIM from 0.226 to 0.845, while obtaining the highest $F_{0.5}$ and
precision. \textbf{These gains show that a compact WSM TunnelingNet simultaneously
preserves the supplied structure and reconstructs substantially more faithful
RGB content than the specialized conditional-generation baselines.}

\paragraph{Any-to-any control.}
The gains extend to both geometric transfers in Interface \numtag{C}. For
Normal$\rightarrow$Edge, WSM improves $F_2$ by 32.2\% and recall by 62.2\%, while
reducing the false-negative rate by 13.7\%. For Relative
Depth$\rightarrow$Normal, it reduces mean angular error by 33.1\% and median
angular error by 40.2\%; accuracy at $11.25^\circ$, $22.5^\circ$, and
$30^\circ$ improves by 20.6\%, 38.6\%, and 34.6\%, respectively. On the more
challenging Edge$\rightarrow$Human Segmentation task, WSM retains 94.0--98.0\%
of MODUS performance across Pixel Accuracy, Macro-F1, Mean Class Accuracy, and
mIoU. \textbf{WSM therefore achieves the strongest geometric-transfer quality
and remains close to MODUS on human-foreground segmentation.}

\paragraph{Training efficiency and generalization.}
Across comparisons with recoverable training budgets, each WSM TunnelingNet uses
approximately 230--145,500 times fewer training samples. On the three Interface
\numtag{C} tasks, WSM uses five times fewer inference steps than MODUS and runs
5.8--52.7 times faster. The human-segmentation TunnelingNet alone maintains a
2.32K-fold training-sample advantage while approaching MODUS quality.
\textbf{Together, the post-release results establish that WSM combines the
strongest overall quality with orders-of-magnitude faster task acquisition, and
that both advantages transfer to visual content that was unavailable during
training.}

\input{sec_preprint/5_independent_annotations}

%% file: sec_preprint/5_independent_annotations.tex
\begin{table*}[p]
    \centering
    \caption{\textbf{External comparison under independent public annotation conventions.}
    The images, crops, trained models, training-cost calculation, and inference
    protocols are identical to Table~\ref{tab:external_postrelease_summary}.
    Edge, relative-depth, and surface-normal annotations are generated with
    OpenCV Canny, Depth Anything V2, and Marigold Normals, respectively. WSG
    only serializes those independently generated conditions for WSM inference.
    Training Sample Budget is optimizer updates multiplied by global batch size;
    Relative Sample Budget is its ratio to the corresponding WSM TunnelingNet. N/R
    denotes information not reported or recoverable from official code. Bold
    marks the best value in each quality column; bold values marked with
    $\ast$ are second-best. $\dagger$ marks methods that
    additionally receive the same first-frame RGB reference.}
    \label{tab:external_independent_annotation_summary}
    \tiny
    \setlength{\tabcolsep}{2.5pt}
    \renewcommand{\arraystretch}{1.10}
    \definecolor{independentHeaderBg}{RGB}{239,235,248}
    \definecolor{independentOursBg}{RGB}{235,246,252}
    \newcommand{\independenttag}[1]{%
        \raisebox{-0.15ex}{\resizebox{!}{1.65ex}{\numtag{#1}}}%
    }
    \newsavebox{\independentcellbox}
    \newcommand{\independentvcellstart}{\setbox\independentcellbox=\hbox\bgroup}
    \newcommand{\independentvcellend}{%
        \egroup
        \raisebox{\dimexpr\dp\independentcellbox/2-\ht\independentcellbox/2\relax}%
            {\usebox{\independentcellbox}}%
    }
    \newcolumntype{J}{>{\independentvcellstart}l<{\independentvcellend}}
    \newcolumntype{K}{>{\independentvcellstart}c<{\independentvcellend}}

    \vspace{1pt}
    {\small\noindent\textbf{\independenttag{A} Estimation and reconstruction: RGB$\rightarrow$Relative Depth.}\par}
    \resizebox{\linewidth}{!}{%
    \begin{tabular}{@{}J*{13}{K}@{}}
        \toprule
        \rowcolor{independentHeaderBg}
        Method & Venue & Parameters &
        \shortstack{Training Sample Budget\\(Updates$\times$Global Batch)} &
        \shortstack{Relative Sample\\Budget}$\downarrow$ &
        \shortstack{Training Time\\/ Hardware} &
        \shortstack{Boundary Precision\\@2 px}$\uparrow$ &
        \shortstack{Boundary Precision\\@3 px}$\uparrow$ &
        \shortstack{Boundary\\FPR}$\downarrow$ &
        \shortstack{Gradient\\MAE}$\downarrow$ &
        \shortstack{Gradient\\RMSE}$\downarrow$ &
        \shortstack{Inference\\Steps}$\downarrow$ &
        \shortstack{Total Inference\\Time (s)}$\downarrow$ &
        \shortstack{Inference Speed\\(FPS)}$\uparrow$ \\
        \midrule
        Marigold~\cite{Ke_2024_CVPR} & \shortstack{CVPR 2024\\\textcolor{red}{Best Paper Award Candidate}} & $0.9$B & $18$K$\times32$ & $331.0$ & \shortstack{A few days\\1 GPU} & 0.736 & 0.768 & 0.0148 & 0.00575 & 0.0153 & $1{\times}10$ & 105.47 & 0.2181 \\
        DMP~\cite{Lee_2024_CVPR} & CVPR 2024 & $0.86$B+LoRA & $50$K$\times8$ & $229.9$ & N/R & 0.548 & 0.581 & 0.0183 & 0.00631 & 0.0169 & 5 & 24.38 & 0.9434 \\
        GeoWizard~\cite{fu2024geowizard} & ECCV 2024 & $0.9$B & $20$K$\times256$ & $2.94$K & \shortstack{Time N/R\\8 GPUs} & 0.203 & 0.234 & 0.2588 & 0.02707 & 0.0381 & $50{\times}10$ & 7834.00 & 0.0029 \\
        Scaling Diffusion Perception~\cite{Ravishankar_2025_CVPR} & CVPR 2025 & $2.8$B ($1.0$B active) & \shortstack{Mixed-task phase N/R\\+$4$K MoE updates} & N/R & N/R & 0.660 & 0.690 & 0.0164 & 0.00531 & 0.0139 & $50{\times}3$ & 833.92 & 0.0276 \\
        VGGT~\cite{Wang_2025_CVPR} & \shortstack{CVPR 2025\\\textcolor{red}{Best Paper Award}} & $1.2$B & $160$K$\times3{,}072$ frames & N/R & \shortstack{9 days/64 A100s\\13.8K A100-h} & 0.531 & 0.604 & 0.0183 & 0.00655 & 0.0149 & \textbf{1 pass} & \textbf{4.83} & \textbf{4.7619} \\
        MODUS~\cite{ye2026modus} & ICML 2026 & $14$B ($7$B active) & $\sim68.8$K$\times64$ packed seq. & $\geq2.53$K & \shortstack{88.5 h/64 GH200s\\5,664 GH200-h} & 0.703 & 0.723 & 0.0255 & 0.00582 & 0.0173 & 50 & 9025.88 & 0.0025 \\
        SenseNova-Vision~\cite{sensenova2026sensenovavision} & Technical Report & $7$B & $50$K steps (packed) & N/R & \shortstack{Time N/R\\$\geq16\times$80GB GPUs} & \textbf{0.807$^{\ast}$} & \textbf{0.846$^{\ast}$} & \textbf{0.0098} & \textbf{0.00418} & \textbf{0.0127} & 50 & 3471.00 (4G) & 0.0066 \\
        \rowcolor{independentOursBg}
        \textbf{WSM} & Ours & $4$B+$12.5$M & $1740\times1$ & \textbf{1.0} & \shortstack{11.0 h\\1 A100} & \textbf{0.815} & \textbf{0.848} & \textbf{0.0110$^{\ast}$} & \textbf{0.00469$^{\ast}$} & \textbf{0.0129$^{\ast}$} & 10 & 222.41 & 0.1034 \\
        \bottomrule
    \end{tabular}}

    \vspace{3pt}
    {\small\noindent\textbf{\independenttag{B} Conditional RGB generation: Edge$\rightarrow$RGB.}\par}
    \resizebox{\linewidth}{!}{%
    \begin{tabular}{@{}J*{15}{K}@{}}
        \toprule
        \rowcolor{independentHeaderBg}
        Method & Venue & Parameters &
        \shortstack{Training Sample Budget\\(Updates$\times$Global Batch)} &
        \shortstack{Relative Sample\\Budget}$\downarrow$ &
        \shortstack{Training Time\\/ Hardware} &
        \shortstack{Edge Density\\Error}$\downarrow$ & $F_{0.5}\uparrow$ &
        Precision$\uparrow$ & MAE$\downarrow$ & MSE$\downarrow$ & PSNR$\uparrow$ & SSIM$\uparrow$ &
        \shortstack{Inference\\Steps}$\downarrow$ &
        \shortstack{Total Inference\\Time (s)}$\downarrow$ &
        \shortstack{Inference Speed\\(FPS)}$\uparrow$ \\
        \midrule
        ControlNet~\cite{zhang2023adding} & \shortstack{ICCV 2023\\\textcolor{red}{Best Paper Award}} & $1.4$B & \shortstack{N/R$\times256$ (v1.1)} & N/R & \shortstack{3 days/8 A100s (v1.1)\\$\geq1{,}176$ A100-h total} & 0.0288 & 0.2821 & 0.2926 & 0.261 & 0.105 & 10.25 & 0.093 & 30 & \textbf{79.17} & \textbf{0.3032} \\
        \shortstack{IP-Adapter~\cite{ye2023ipadapter}\\(with ControlNet~\cite{zhang2023adding})}$^\dagger$ & Technical Report & $2.0$B & \shortstack{ControlNet: N/R$\times256$\\IP-Adapter: $1$M$\times64$} & $\geq145.5$K & \shortstack{ControlNet: 3 days/8 A100s\\IP-Adapter: time N/R/8 V100s} & 0.0186 & 0.3070 & 0.3094 & 0.218 & 0.079 & 11.47 & 0.226 & 30 & 87.21 & 0.2752 \\
        T2I-Adapter~\cite{mou2024t2iadapter} & AAAI 2024 & $2.6$B+$79$M & $20$K$\times256$ & $11.6$K & \shortstack{Time N/R\\4 A100s} & 0.0520 & 0.2377 & 0.3043 & 0.246 & 0.092 & 10.83 & 0.144 & 30 & 262.46 & 0.0914 \\
        UniControl~\cite{qin2023unicontrol} & NeurIPS 2023 & $1.5$B & $\sim900$K$\times64$ & $\sim130.9$K & \shortstack{$>5{,}000$ A100-h\\A100-40G} & 0.0505 & 0.2235 & 0.3208 & 0.246 & 0.091 & 10.94 & 0.044 & 50 & 265.64 & 0.0903 \\
        Uni-ControlNet~\cite{zhao2023unicontrolnet} & NeurIPS 2023 & $1.5$B & 1 epoch ($10$M pairs) & $22.7$K & N/R & 0.0816 & 0.1781 & 0.1708 & 0.264 & 0.106 & 10.22 & 0.073 & 50 & 381.89 & 0.0628 \\
        \rowcolor{independentOursBg}
        \textbf{WSM}$^\dagger$ & Ours & $4$B+$12.5$M & $440\times1$ & \textbf{1.0} & \shortstack{8.5 min\\1 A100} & \textbf{0.0116} & \textbf{0.3107} & \textbf{0.3265} & \textbf{0.074} & \textbf{0.013} & \textbf{19.83} & \textbf{0.845} & \textbf{10} & 202.56 & 0.1185 \\
        \bottomrule
    \end{tabular}}

    \vspace{3pt}
    {\small\noindent\textbf{\independenttag{C} Any-to-any control over three representative structured targets under independent annotations.}\par}

    \vspace{1pt}
    \resizebox{\linewidth}{!}{%
    \begin{tabular}{@{}J*{14}{K}@{}}
        \toprule
        \rowcolor{independentHeaderBg}
        Method & Task & Venue & Parameters &
        \shortstack{Training Sample Budget\\(Updates$\times$Global Batch)} &
        \shortstack{Relative Sample\\Budget}$\downarrow$ &
        \shortstack{Training Time\\/ Hardware} &
        \shortstack{Cosine\\Similarity}$\uparrow$ & \shortstack{XYZ\\RMSE}$\downarrow$ &
        \shortstack{Mean Angle\\($^\circ$)}$\downarrow$ &
        \shortstack{Angular RMSE\\($^\circ$)}$\downarrow$ &
        \shortstack{$90$th Percentile\\Angle ($^\circ$)}$\downarrow$ &
        \shortstack{Inference\\Steps}$\downarrow$ &
        \shortstack{Total Inference\\Time (s)}$\downarrow$ &
        \shortstack{Inference Speed\\(FPS)}$\uparrow$ \\
        \midrule
        MODUS~\cite{ye2026modus} & \shortstack{Relative Depth$\rightarrow$Normal} & ICML 2026 & $14$B ($7$B active) & $\sim68.8$K$\times64$ packed seq. & $\geq2.25$K & \shortstack{88.5 h/64 GH200s\\5,664 GH200-h} & 0.7263 & 0.3965 & 34.89 & 42.61 & 67.15 & 50 & 12272.18 & 0.0020 \\
        \rowcolor{independentOursBg}
        \textbf{WSM} & \shortstack{Relative Depth$\rightarrow$Normal} & Ours & $4$B+$12.5$M & $1960\times1$ & \textbf{1.0} & \shortstack{9.6 h\\1 A100} & \textbf{0.7894} & \textbf{0.3667} & \textbf{33.65} & \textbf{38.10} & \textbf{57.78} & \textbf{10} & \textbf{1767.40} & \textbf{0.0136} \\
        \bottomrule
    \end{tabular}}

    \vspace{2pt}
    \resizebox{\linewidth}{!}{%
    \begin{tabular}{@{}J*{13}{K}@{}}
        \toprule
        \rowcolor{independentHeaderBg}
        Method & Task & Venue & Parameters &
        \shortstack{Training Sample Budget\\(Updates$\times$Global Batch)} &
        \shortstack{Relative Sample\\Budget}$\downarrow$ &
        \shortstack{Training Time\\/ Hardware} &
        \shortstack{Pixel\\Accuracy}$\uparrow$ & mIoU$\uparrow$ &
        mAcc$\uparrow$ & Macro-F1$\uparrow$ &
        \shortstack{Inference\\Steps}$\downarrow$ &
        \shortstack{Total Inference\\Time (s)}$\downarrow$ &
        \shortstack{Inference Speed\\(FPS)}$\uparrow$ \\
        \midrule
        MODUS~\cite{ye2026modus} & \shortstack{Normal$\rightarrow$Human Segmentation} & ICML 2026 & $14$B ($7$B active) & $\sim68.8$K$\times64$ packed seq. & $\geq2.20$K & \shortstack{88.5 h/64 GH200s\\5,664 GH200-h} & 0.9126 & 0.8405 & 0.9233 & 0.8993 & 50 & 3105.40 & 0.0019 \\
        \rowcolor{independentOursBg}
        \textbf{WSM} & \shortstack{Normal$\rightarrow$Human Segmentation} & Ours & $4$B+$12.5$M & $2000\times1$ & \textbf{1.0} & \shortstack{9.6 h\\1 A100} & \textbf{0.9283} & \textbf{0.8488} & \textbf{0.9292} & \textbf{0.9131} & \textbf{10} & \textbf{243.13} & \textbf{0.0247} \\
        \bottomrule
    \end{tabular}}

    \vspace{2pt}
    \resizebox{\linewidth}{!}{%
    \begin{tabular}{@{}J*{14}{K}@{}}
        \toprule
        \rowcolor{independentHeaderBg}
        Method & Task & Venue & Parameters &
        \shortstack{Training Sample Budget\\(Updates$\times$Global Batch)} &
        \shortstack{Relative Sample\\Budget}$\downarrow$ &
        \shortstack{Training Time\/ Hardware} &
        \shortstack{Balanced\\Accuracy}$\uparrow$ & $F_1$ $\uparrow$ & $F_2$ $\uparrow$ & IoU$\uparrow$ & Recall$\uparrow$ &
        \shortstack{Inference\\Steps}$\downarrow$ &
        \shortstack{Total Inference\\Time (s)}$\downarrow$ &
        \shortstack{Inference Speed\\(FPS)}$\uparrow$ \\
        \midrule
        MODUS~\cite{ye2026modus} & \shortstack{Normal$\rightarrow$Edge} & ICML 2026 & $14$B ($7$B active) & $\sim68.8$K$\times64$ packed seq. & $\geq2.20$K & \shortstack{88.5 h/64 GH200s\\5,664 GH200-h} & 0.5705 & 0.1936 & 0.2092 & 0.1094 & 0.2320 & 50 & 16187.91 & 0.0015 \\
        \rowcolor{independentOursBg}
        \textbf{WSM} & \shortstack{Normal$\rightarrow$Edge} & Ours & $4$B+$12.5$M & $2000\times1$ & \textbf{1.0} & \shortstack{9.3 h\\1 A100} & \textbf{0.5846} & \textbf{0.1969} & \textbf{0.2835} & \textbf{0.1122} & \textbf{0.4549} & \textbf{10} & \textbf{1061.17} & \textbf{0.0226} \\
        \bottomrule
    \end{tabular}}
\end{table*}

\begin{figure*}[p]
    \centering
    \includegraphics[width=\textwidth,height=0.86\textheight,keepaspectratio]{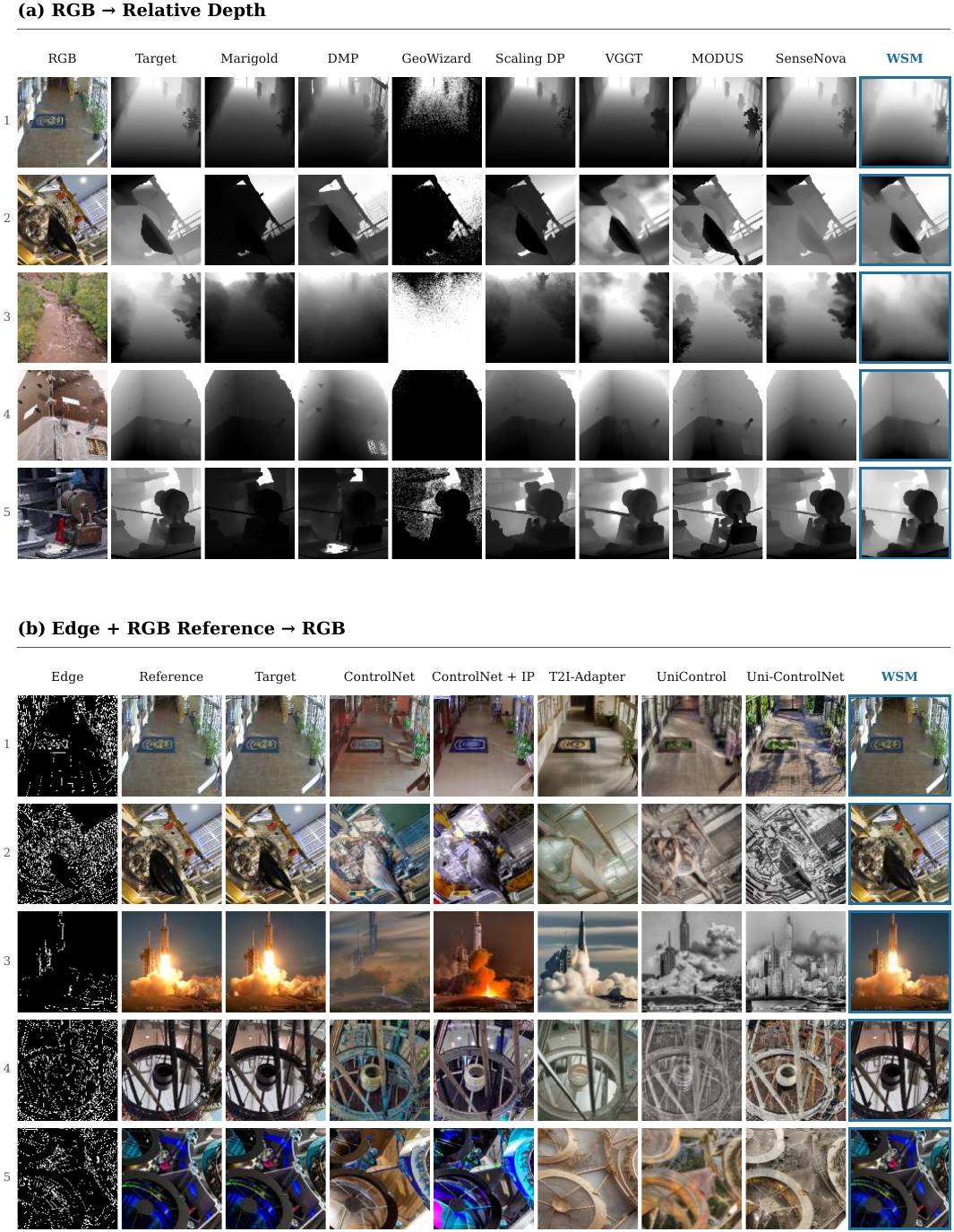}
    \caption{\textbf{Qualitative comparison for Interfaces \numtag{A} and
    \numtag{B} under independent public annotation conventions.}
    The rows use the same scene crops and order as
    Figure~\ref{fig:external_postrelease_qualitative_ab}. Panel (a) replaces
    the relative-depth target with Depth Anything V2; predictions use the same
    fixed method-level near--far convention and percentile normalization as
    the quantitative evaluation.
    Panel (b) is unchanged because the post-release experiment already used
    OpenCV Canny as the Edge condition. WSM is the rightmost column, outlined
    in blue.}
    \label{fig:external_independent_qualitative_ab}
\end{figure*}

\begin{figure*}[p]
    \centering
    \includegraphics[width=\textwidth,height=0.86\textheight,keepaspectratio]{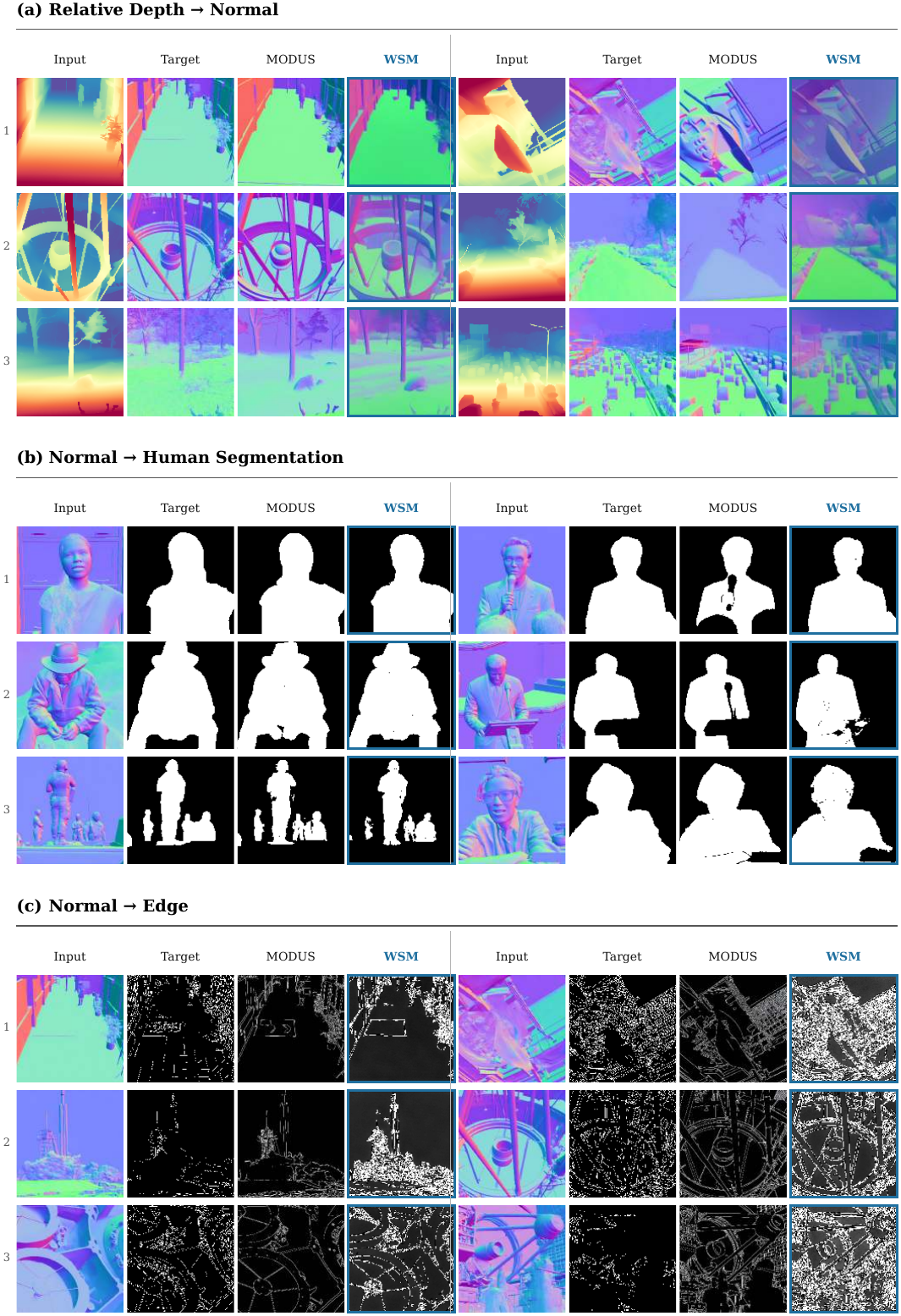}
    \caption{\textbf{Qualitative comparison for Interface \numtag{C} under
    independent public annotation conventions.}
    The three task blocks correspond to Relative Depth$\rightarrow$Normal,
    Normal$\rightarrow$Human Segmentation, and Normal$\rightarrow$Edge in
    Table~\ref{tab:external_independent_annotation_summary}. Each four-column
    group shows the independently annotated input and target followed by the
    MODUS and WSM predictions generated from that input. WSM outputs are
    outlined in blue.}
    \label{fig:external_independent_qualitative_c}
\end{figure*}

\FloatBarrier
\subsection{Evaluation with Independent Public Annotations}
\label{subsec:independent_annotation_evaluation}

\paragraph{Purpose and task setting.}
Sections~\ref{subsec:external_evaluation_comparison}
and~\ref{subsec:postrelease_visual_evaluation} compare WSM with prior methods
through the three interfaces for which released, input-matched baselines are
available. The post-release experiment additionally makes the evaluated images
unavailable during model training. Its geometric annotations, however, still
follow the WSG data-preparation convention used to train WSM. This leaves a
separate question: does WSM remain competitive when the desired outputs are
defined by unrelated public annotators?

This distinction is important because new images test visual transfer, whereas
new annotation conventions also test whether a method has learned the task
itself. We therefore combine both changes: every evaluated image postdates the
training of all compared models, and every geometric condition and target is
defined by a public annotator outside the WSG preparation pipeline. Success in
this setting requires the learned interface to survive changes in scene content,
numeric convention, and annotation family at the same time.

Table~\ref{tab:external_independent_annotation_summary} answers this question
on five representative tasks across the three reported interfaces. Interface
\numtag{A} uses RGB$\rightarrow$Relative Depth to test scene-scale spatial
estimation. Interface \numtag{B} uses Edge$\rightarrow$RGB to test whether a
method can synthesize appearance while respecting an explicit structural
condition. Interface \numtag{C} covers three target types: surface normals,
human foreground segmentation, and edges. Relative Depth$\rightarrow$Normal tests
surface recovery from scene-scale ordering; Normal$\rightarrow$Human
Segmentation tests semantic transfer; and Normal$\rightarrow$Edge tests
boundary reconstruction from local orientation. Each direction uses identical
input--output definitions for WSM and the released MODUS model. Interface
\numtag{D} is not included because, as discussed in
Section~\ref{subsec:external_evaluation_comparison}, no released image-level
method accepts its matched combination of observations.

\paragraph{Replacing annotations while keeping the evaluation fixed.}
We retain the post-release scene pool, image crops, trained models,
training-cost calculation, and official inference procedures. We replace the
geometric annotation pipeline: OpenCV Canny~\cite{4767851} produces binary
edges, Depth Anything V2 Large~\cite{yang2024depth} produces relative depth,
Marigold Normals v1.1~\cite{Ke_2024_CVPR} produces camera-space surface
normals, and Mask2Former-Swin-Large trained on ADE20K~\cite{cheng2021mask2former}
produces human foreground masks. Every method receives the same independently
generated conditions and targets. Relative-depth maps receive a fixed per-image 2nd--98th-percentile
linear map to $[0,1]$, and normal metrics operate directly on physical unit
vectors. WSG serves only as the storage and input-encoding format for WSM; it
does not generate any annotation used in this experiment.
No method is retrained, calibrated, or adapted to these annotations.
\textbf{The comparison therefore isolates transfer to an independent task
definition while keeping the evaluated images, crops, checkpoints, and
method-specific inference procedures fixed.}

\paragraph{Qualitative comparison.}
Figures~\ref{fig:external_independent_qualitative_ab}
and~\ref{fig:external_independent_qualitative_c} expose the predictions behind
Table~\ref{tab:external_independent_annotation_summary}. Under the independent
depth target, WSM preserves continuous scene ordering and prominent object
boundaries across indoor, outdoor, and object-centric scenes. The conditional
RGB panel reproduces the Canny-controlled comparison from
Figure~\ref{fig:external_postrelease_qualitative_ab}, since this condition was
already independent of WSG. In Interface \numtag{C}, WSM produces coherent surface-normal fields from
independently estimated relative depth, a clean human foreground mask from
surface normals, and structured edges from the same normal representation. These three
transfers cover geometric surfaces, human semantics, and boundaries. \textbf{The
visual comparisons confirm that WSM's advantage remains visible when both
structured inputs and evaluation targets are supplied by independent public
annotators.}

\paragraph{Overall performance.}
WSM leads every reported quality column in the three Interface C task blocks.
Together with the RGB-generation comparison, these results cover appearance,
surface geometry, human semantics, and boundary structure under independent
annotation conventions. The advantage is therefore not tied to a single target
space or one favorable metric; it persists across the three distinct any-to-any
targets shown in Table~\ref{tab:external_independent_annotation_summary}.

The same result is obtained with only 12.5M trainable target-side parameters per
TunnelingNet and 440--2,000 single-sample updates. All checkpoints and accounting rules
remain unchanged, so the training- and inference-efficiency advantages in
Section~\ref{subsec:postrelease_visual_evaluation} transfer directly to the
independent annotations. \textbf{WSM consequently delivers the strongest
overall quality across every externally comparable interface while preserving
its orders-of-magnitude task-acquisition advantage.}

\paragraph{Estimation and reconstruction.}
For RGB$\rightarrow$Relative Depth, WSM obtains the best boundary precision at
both two- and three-pixel tolerances (0.815/0.848). Its boundary false-positive
rate, gradient MAE, and gradient RMSE are all second-best, trailing SenseNova
only narrowly. WSM surpasses VGGT, the CVPR 2025
\textcolor{red}{Best Paper Award} recipient, on all five reported depth
measures. Relative to VGGT, WSM raises boundary precision by 53.5\% at two
pixels and 40.4\% at three pixels, while reducing boundary false positives by
39.9\%, gradient MAE by 28.4\%, and gradient RMSE by 13.4\%. It reaches this
result with 1,740 single-sample updates on one A100; the reported comparison
systems use sample budgets ranging from 229.9 to at least 2.94K times larger
where accounting is available. \textbf{WSM therefore combines the strongest
measured boundary preservation with exceptionally rapid acquisition, even when
the target-depth convention is defined by an unrelated public estimator.}

\paragraph{Conditional RGB generation.}
The Edge$\rightarrow$RGB condition is generated by OpenCV Canny for every
method. WSM leads all seven reported quality measures, covering edge
consistency, pixel error, PSNR, and SSIM. \textbf{Its advantage over
ControlNet, the ICCV 2023 \textcolor{red}{Best Paper Award} recipient, therefore
holds under a standard public edge convention rather than a WSG-derived
target.} The advantage also holds against the reference-enabled combination of
IP-Adapter and ControlNet: WSM reduces MAE by 66.1\% and MSE by 83.5\%, raises
PSNR from 11.47 to 19.83, and raises SSIM from 0.226 to 0.845. These gains are
learned in 440 updates---8.5 minutes on one A100---with a training-sample budget
at least 11.6K times smaller than every baseline for which a complete budget is
recoverable. \textbf{This is a complete quality sweep achieved by a minute-scale
TunnelingNet, including against both a Best Paper control architecture and its
reference-image extension.}

\paragraph{Any-to-any control.}
Interface \numtag{C} covers three complementary any-to-any transfers. For
Relative Depth$\rightarrow$Normal, WSM improves cosine similarity, XYZ RMSE,
mean angle, angular RMSE, and the 90th-percentile angular error; the latter
falls from 67.15$^\circ$ to 57.78$^\circ$. For Normal$\rightarrow$Human
Segmentation, WSM leads the four reported quality measures, including pixel
accuracy, mIoU, mean accuracy, and macro-F1. For Normal$\rightarrow$Edge, it
leads the selected balanced-accuracy, F1, F2, IoU, and recall measures. \textbf{WSM
therefore leads the reported quality measures across all three displayed target
types, covering surface orientation, human semantics, and boundary structure.}
The relative-depth normal transfer also uses five times fewer inference steps and
is 6.9 times faster than MODUS; the human-segmentation and edge transfers show
the same lightweight deployment pattern. Its single-sample training budget is
at least 2.20K times smaller. \textbf{The same tunneling mechanism therefore
exceeds a 14B MoE model across three distinct structured targets while using
substantially less training and inference computation.}

\paragraph{Comparison with other model families.}
The comparison covers substantially different routes to visual adaptation:
DMP uses LoRA-based diffusion perception; ControlNet introduces a trainable
parameter block; IP-Adapter and T2I-Adapter add image- or condition-specific
adapters; UniControl uses a MoE-style adapter; MODUS is a large sparse MoE; and
VGGT uses discriminative feed-forward regression. WSM is not specialized to any
one of their output domains. The same frozen backbone and the same tunneling
principle outperform these alternatives across depth, conditional RGB, and
condition-to-condition surface prediction. The award-winning comparisons make
this breadth especially concrete: WSM exceeds the CVPR 2025
\textcolor{red}{Best Paper Award} recipient VGGT on all five displayed depth
measures, and it exceeds the ICCV 2023 \textcolor{red}{Best Paper Award}
recipient ControlNet on all seven conditional-generation measures. It also
outperforms the reference-enabled IP-Adapter--ControlNet combination on those
seven measures. \textbf{The result is therefore a cross-paradigm improvement,
not an isolated gain over one architecture family.}

\paragraph{Conclusion.}
The three evaluations isolate complementary concerns. Section
\ref{subsec:external_evaluation_comparison} compares representative methods on
a common held-out set; Section~\ref{subsec:postrelease_visual_evaluation}
removes possible image-level pretraining overlap; and this section replaces the
WSG annotation convention with public annotators. In this final setting, WSM
remains strongest in conditional RGB generation and every structured transfer,
while retaining the best depth-boundary precision.
Across the five tasks, it outperforms award-winning specialized systems, modern
diffusion perception models, reference-conditioned adapters, and a 14B MoE
model, while adapting only a 12.5M-parameter TunnelingNet for each task.
\textbf{The 23 first-place and three second-place results show that WSM's
advantage persists across new visual content, independent target definitions,
three interfaces, five representative tasks, and fundamentally different
modeling paradigms. This directly demonstrates that TunnelingNet can make far
more effective and powerful use of a frozen generative backbone through
extremely lightweight training, thereby adding strong new capabilities at
scale. This represents an advance over the prior paradigm-level approaches
evaluated here.}

%% file: sec_preprint/6_qualitative.tex
\wsmnewphysicalpage

\section{Qualitative Evaluation}
\label{sec:qualitative}

\subsection{Scope and Organization}

In Section~\ref{sec:method}, we introduced four generative interfaces as
practical training configurations. They should not be confused with the four
high-level task families derived from the Unified Law---generation, estimation,
conditional control, and reconstruction. The task families describe what a
visual system is asked to do, whereas the interfaces specify how conditions and
targets are connected during training. Our experiments show that these four
interface configurations are sufficient to instantiate almost all evaluated
visual tasks.

As emphasized by the third sub-law, conditional control is itself not a single
homogeneous task: its condition and target may describe geometry, articulated
humans, physical properties, material appearance, or semantics. We therefore
organize the qualitative results into eleven concrete Parts. Part~I compares the
four interfaces directly. Parts~II--VI report RGB-conditioned estimation of
geometry and motion, human structure, physical properties, material and PBR
appearance, and segmentation and semantics. Parts~VII--VIII show narrow RGB
generation controlled respectively by geometry and human cues, and by physical,
semantic, and material cues. Part~IX presents WSG-to-WSG conditional control,
while Part~X presents RGB-assisted inverse rendering. Part~XI separately
presents categorical outputs with the categorical refiner enabled as an
integral component of every applicable TunnelingNet. Parts~II--X contain 131 visually
readable mappings. Part~I provides the interface overview, and Part~XI
separately presents the final categorical outputs. The remaining specialized
cross-family mappings are retained in the exhaustive atlases and the complete
quantitative inventory.

\begin{figure*}[p]
    \centering
    \includegraphics[width=0.91\textwidth]{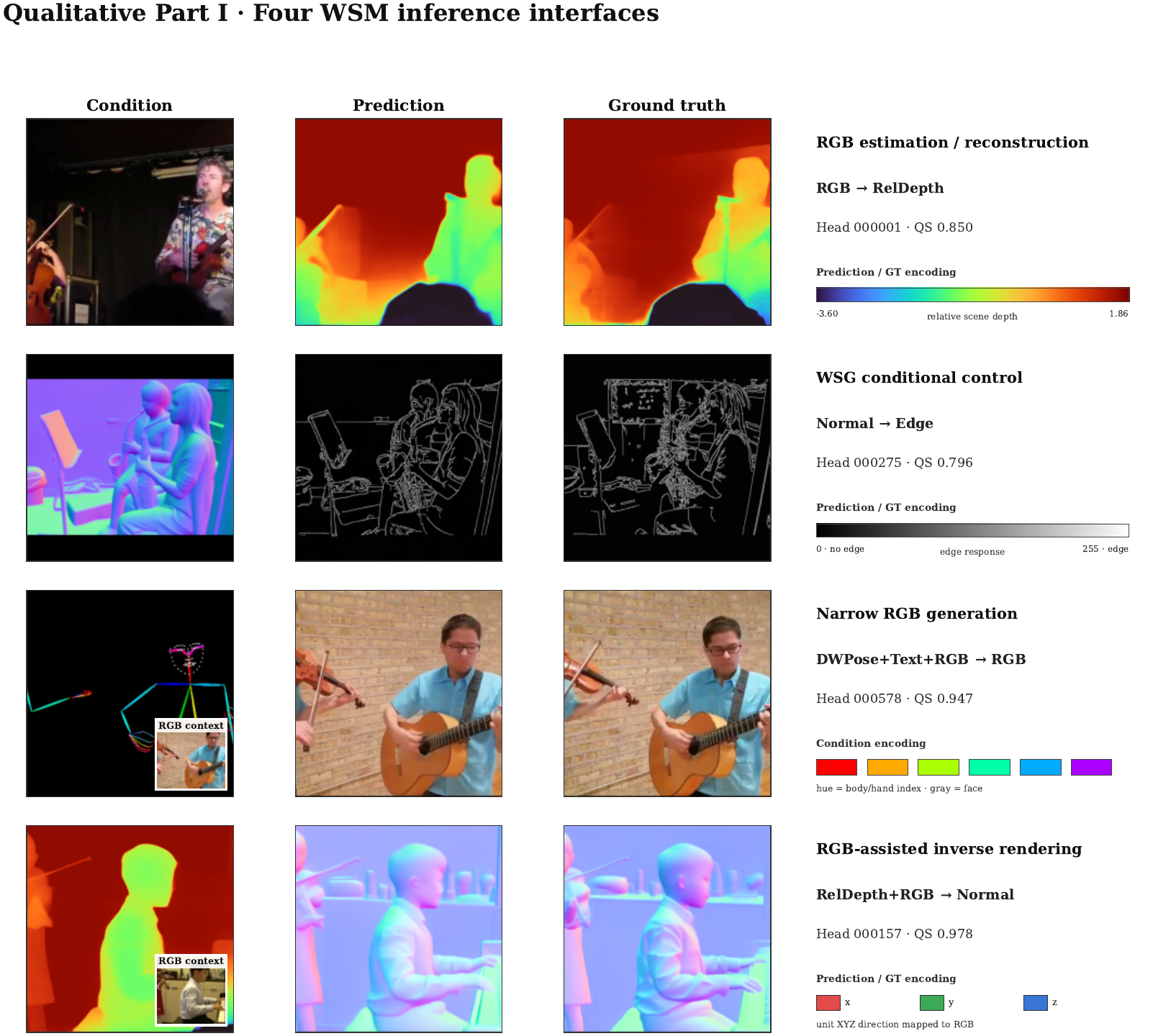}
    \caption{\textbf{Qualitative Part I: anatomy of the four WSM interfaces.}
    The four rows correspond to RGB-to-WSG estimation/reconstruction,
    WSG-to-WSG conditional control, text+WSG narrow generation, and RGB-assisted
    inverse rendering.  Each row retains the actual condition, prediction, and
    ground truth in a common three-column layout.  For interfaces that consume
    auxiliary RGB evidence, the RGB context appears as a labeled inset inside
    the primary condition.  All outputs are true 10-step samples initialized
    from Gaussian noise.}
    \label{fig:evaluation_four_interfaces}
\end{figure*}

\subsection{Part I: Four Inference Interfaces}

Part~I introduces one end-to-end inference example from each interface in a shared
condition--prediction--target layout.  The rows establish how RGB estimation,
narrow RGB generation, condition-only control, and RGB-assisted inverse
rendering are represented in the galleries that follow.  Their qualitative
implications are analyzed after all figures in
Section~\ref{subsec:qualitative_interface_analysis}.

\begin{figure*}[p]
    \centering
    \includegraphics[width=0.96\textwidth]{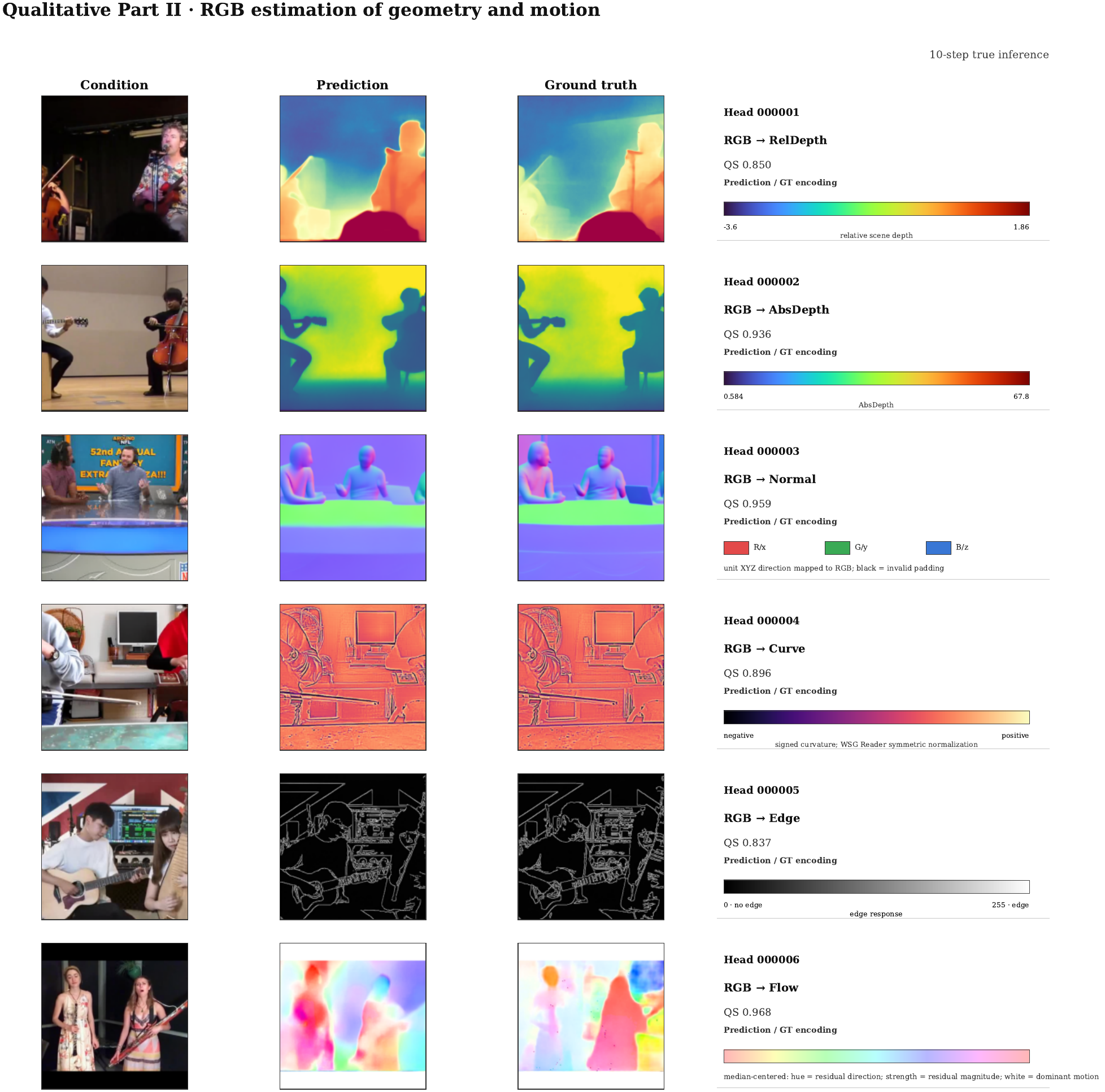}
    \vspace{-4pt}
    \caption{\textbf{Qualitative Part II: RGB estimation of geometry and motion.}
    Six TunnelingNets recover relative and absolute depth, normals, curvature, edges,
    and optical flow.  Curvature is shown explicitly rather than being absorbed
    into a generic geometry example; the row-local keys make scalar, vector,
    and boundary encodings directly interpretable.}
    \label{fig:gallery_geometry_motion}
\end{figure*}

\begin{figure*}[p]
    \centering
    \includegraphics[width=0.96\textwidth]{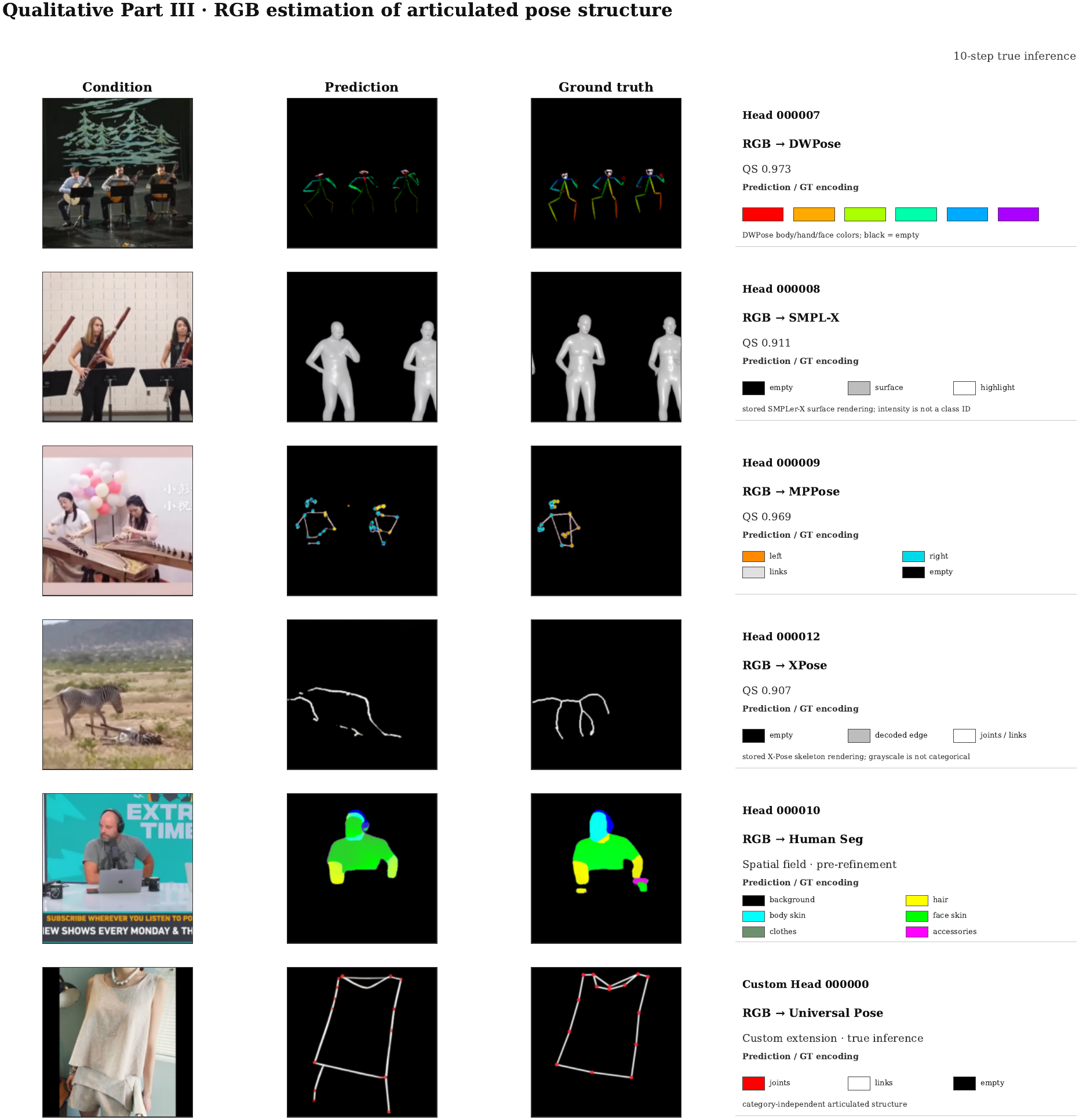}
    \vspace{-4pt}
    \caption{\textbf{Qualitative Part III: RGB estimation of articulated pose structure.}
    DWPose, SMPL-X, MediaPipe Pose, XPose, MediaPipe Seg, and Universal
    Pose expose complementary skeleton, surface, region, and category-independent
    articulation representations. The examples retain sparse limb topology and
    body extent rather than judging articulated structure only through dense
    pixel similarity. \textcolor{red}{For MediaPipe Pose (3D pose), we
    deliberately retain an example with incomplete pseudo ground truth: the
    generated result uses its learned prior to recover missing structure and
    can therefore surpass the automatic annotation.}}
    \label{fig:gallery_pose_structure}
\end{figure*}

\begin{figure*}[p]
    \centering
    \includegraphics[width=0.96\textwidth]{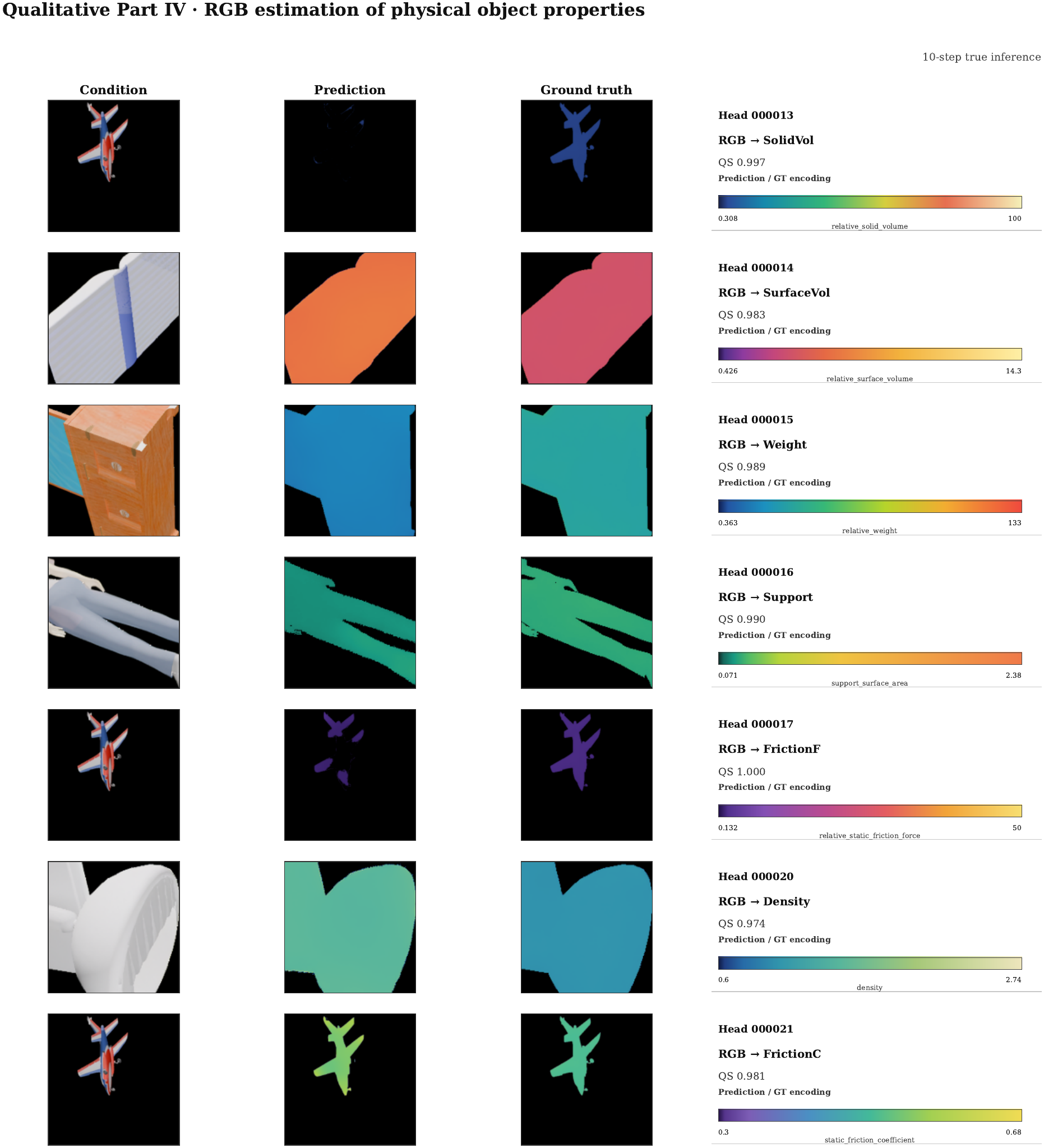}
    \vspace{-4pt}
    \caption{\textbf{Qualitative Part IV: RGB estimation of physical object properties.}
    Seven object-aligned fields cover solid and surface volume, weight, support,
    static friction force, density, and friction coefficient.  These outputs
    test whether spatial support remains stable while the predicted physical
    quantity changes.}
    \label{fig:gallery_physical_properties}
\end{figure*}

\begin{figure*}[p]
    \centering
    \includegraphics[width=0.96\textwidth]{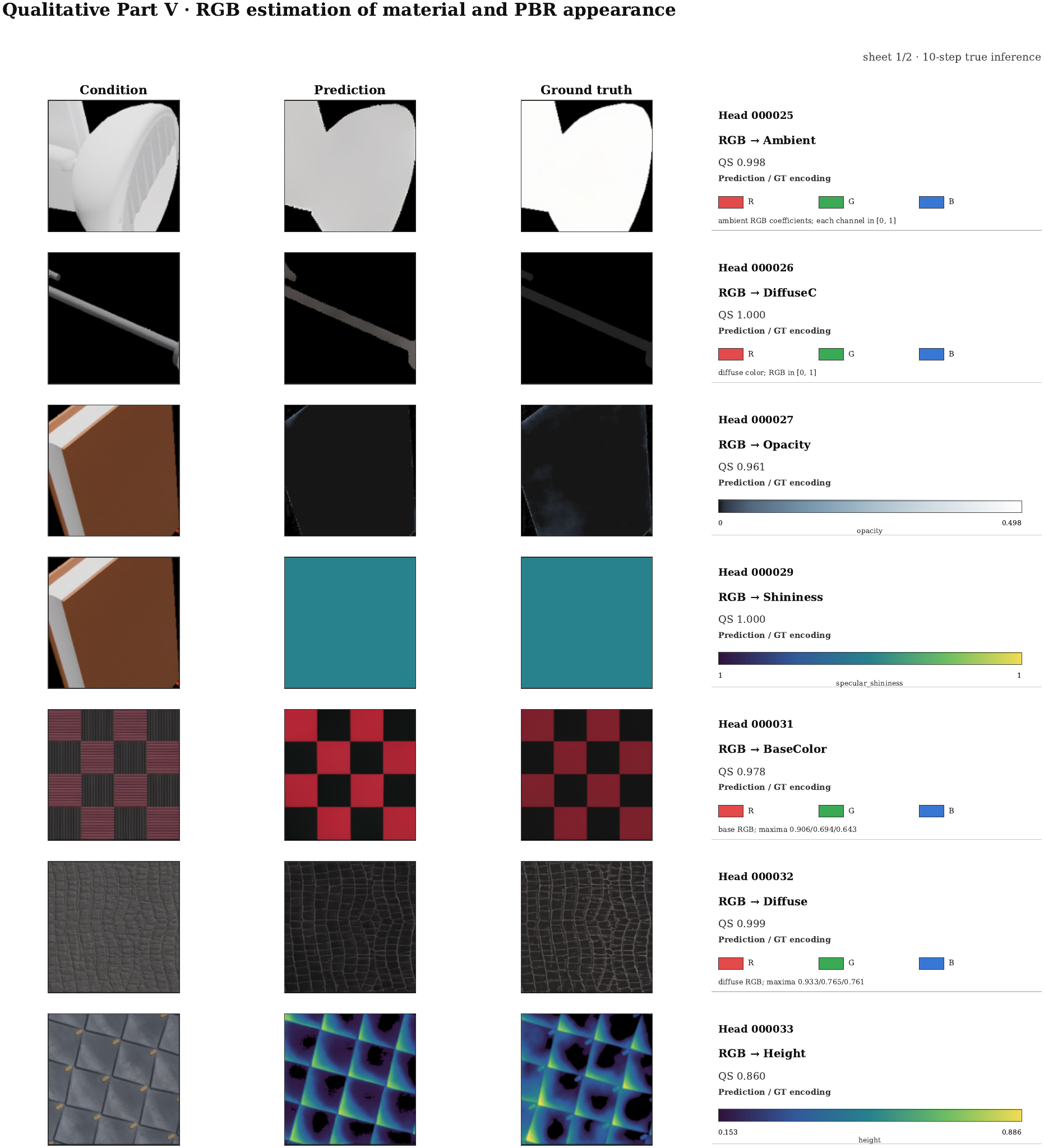}
    \vspace{-4pt}
    \caption{\textbf{Qualitative Part V: RGB estimation of material and PBR appearance (1/2).}
    Ambient and diffuse color, opacity, specular color, shininess, base color,
    and diffuse response separate spatial material support from the value
    assigned within it.}
    \label{fig:gallery_material_pbr_i}
\end{figure*}

\begin{figure*}[p]
    \centering
    \includegraphics[width=0.96\textwidth]{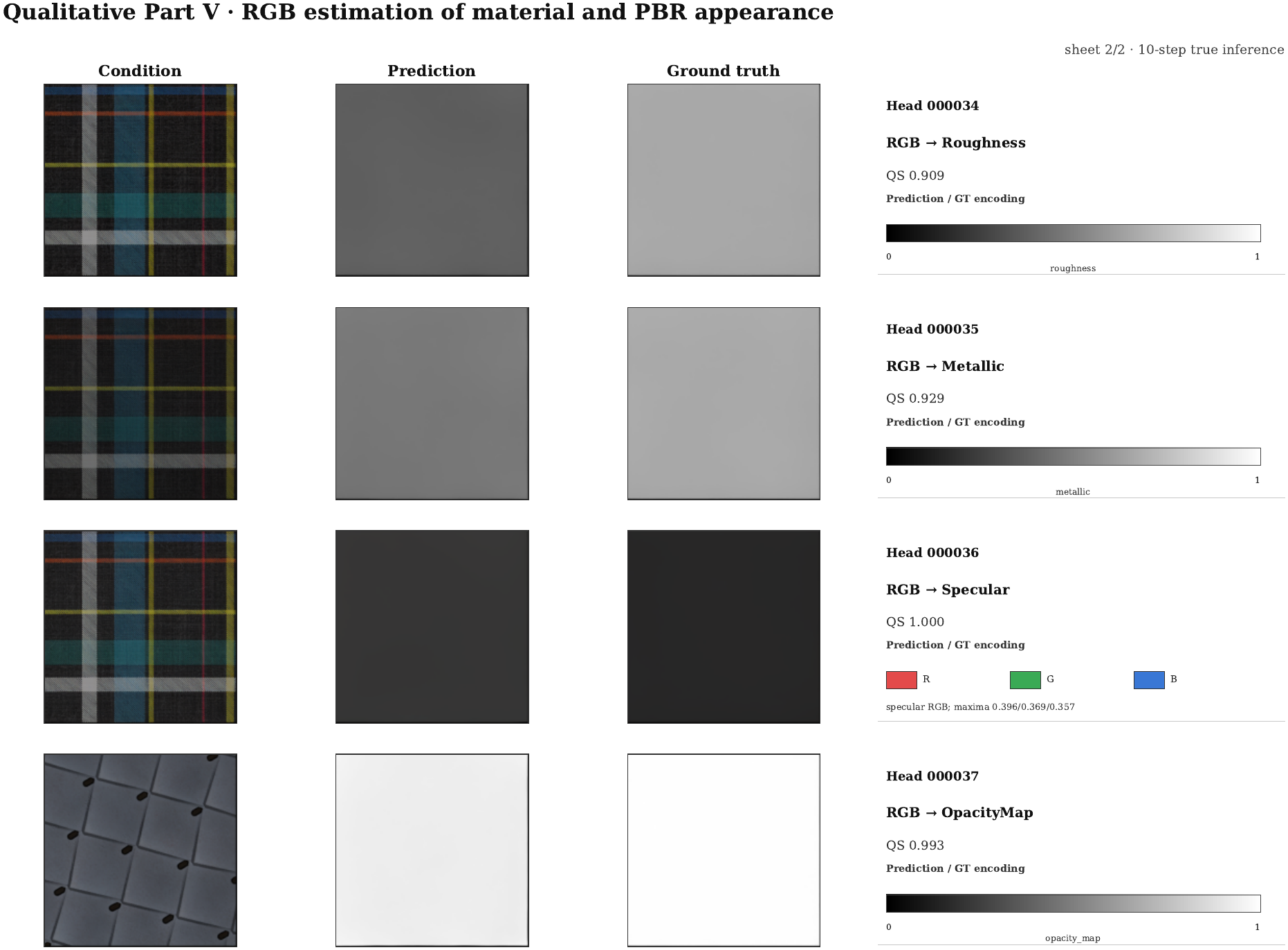}
    \vspace{-4pt}
    \caption{\textbf{Qualitative Part V: RGB estimation of material and PBR appearance (2/2).}
    Height, roughness, metallicity, specular response, and opacity maps
    complete the PBR interface. Prediction and ground truth use the same
    channel-specific encoding in every row.}
    \label{fig:gallery_material_pbr_ii}
\end{figure*}

\begin{figure*}[p]
    \centering
    \includegraphics[width=0.96\textwidth]{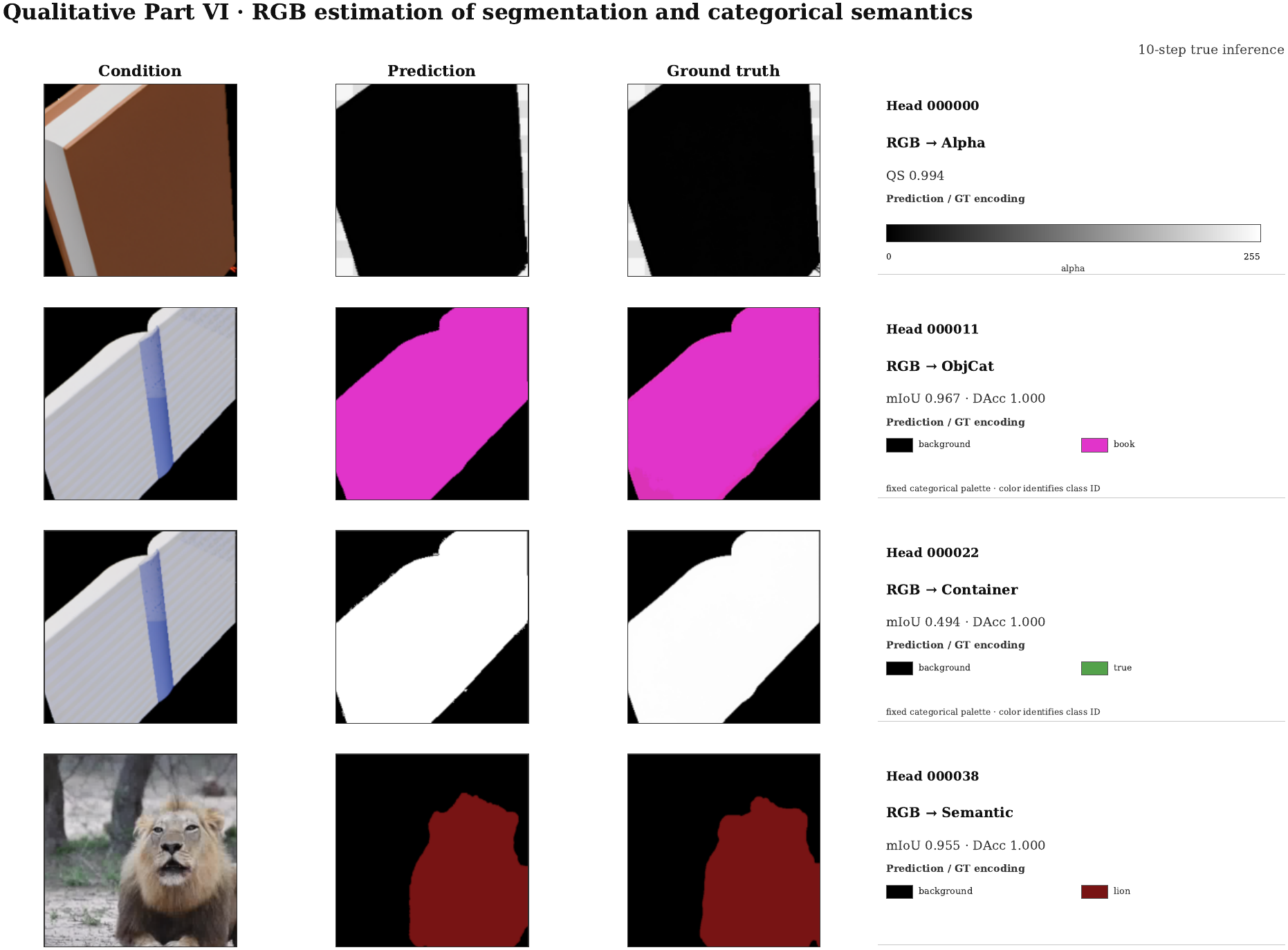}
    \vspace{-4pt}
    \caption{\textbf{Qualitative Part VI: RGB estimation of segmentation and semantics.}
    Alpha support, object category, container, and multi-class semantic masks
    isolate the spatial and categorical aspects
    of segmentation. The categorical rows show selected successful cases;
    aggregate accuracy remains reported by the quantitative matrices.}
    \label{fig:gallery_segmentation_semantics}
\end{figure*}

\subsection{Parts II--X: Task-Family Coverage}

Parts~II--VI cover 33 readable RGB-estimation outputs rather than using a
small proxy set. They separate geometric differential structure, articulated
pose, physical quantities, PBR channels, and segmentation because success in
one family does not imply success in another. In particular, the curvature and
mask rows test thin or discrete structure that a depth-only gallery would miss.

\begin{figure*}[p]
    \centering
    \includegraphics[width=0.96\textwidth]{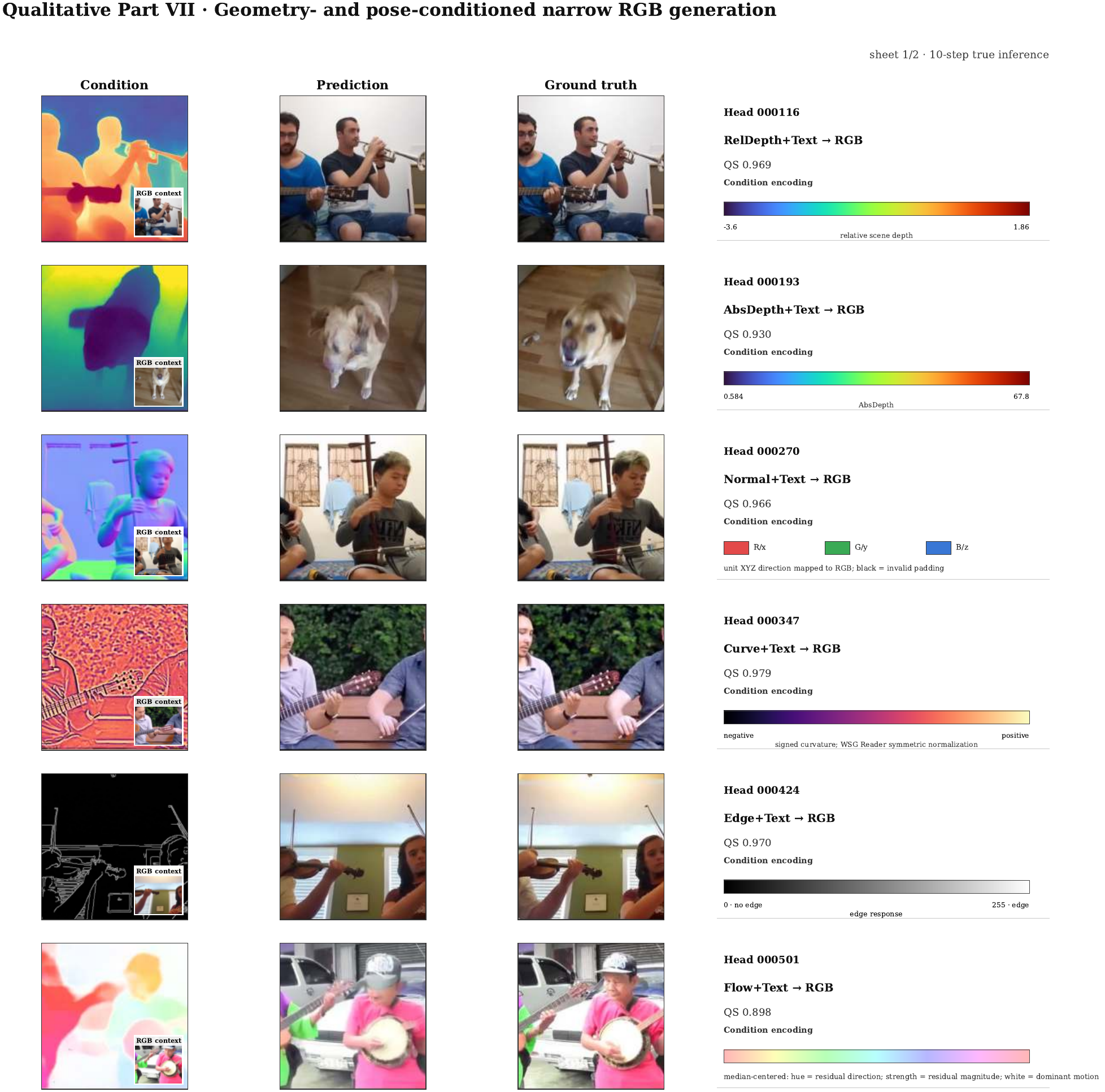}
    \vspace{-4pt}
    \caption{\textbf{Qualitative Part VII: geometry-conditioned narrow RGB generation (1/2).}
    Relative depth, absolute depth, normals, curvature, and edges supply the
    target-frame structure.  Historical RGB context is embedded inside the
    condition panel and never substitutes for the evaluation target frame.}
    \label{fig:gallery_narrow_geometry}
\end{figure*}

\begin{figure*}[p]
    \centering
    \includegraphics[width=0.96\textwidth]{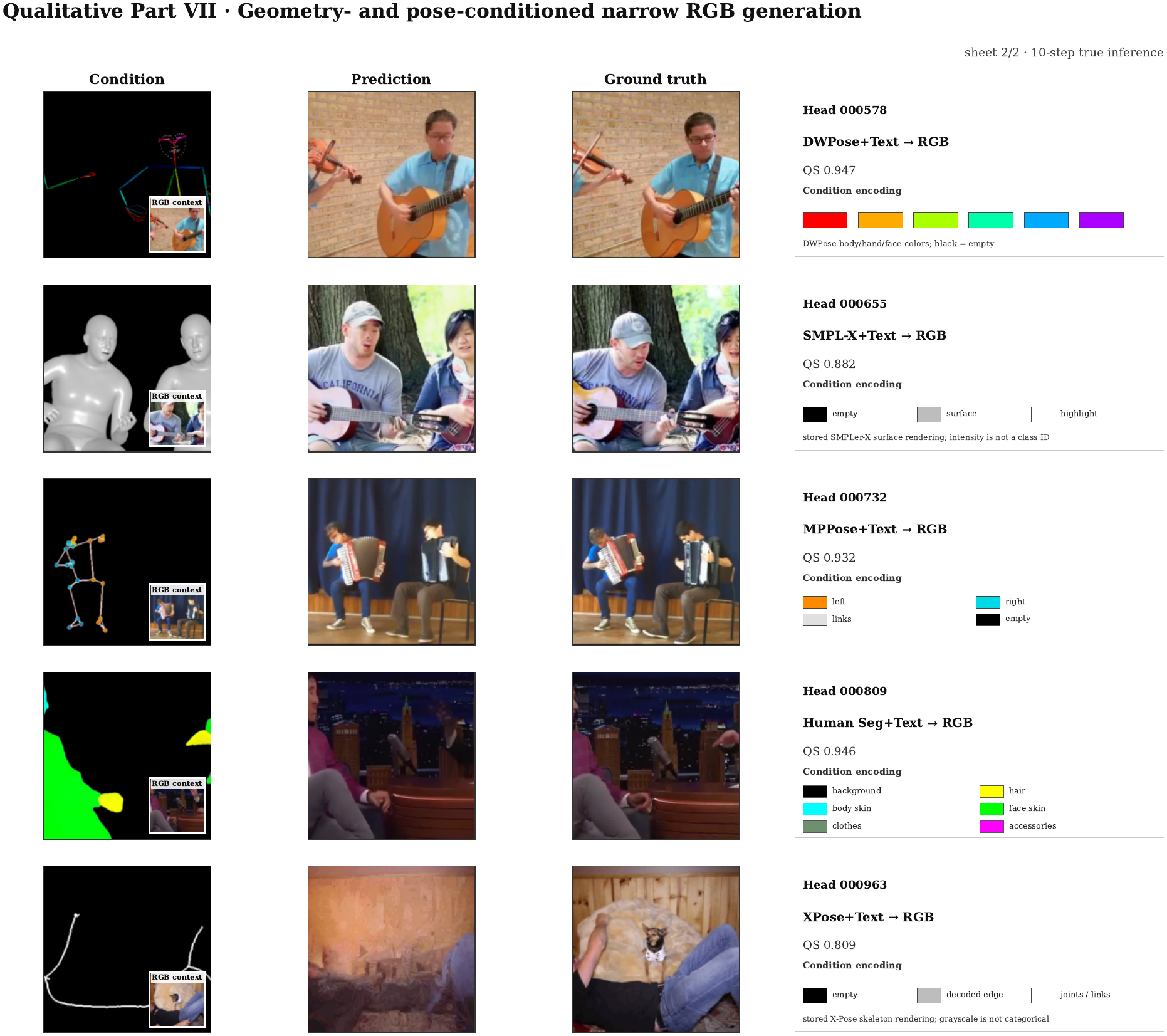}
    \vspace{-4pt}
    \caption{\textbf{Qualitative Part VII: motion- and pose-conditioned narrow RGB generation (2/2).}
    DWPose, SMPL-X, MediaPipe pose and MediaPipe Seg, and XPose test whether sparse,
    articulated, or semantic pose conditions redirect the generated frame while
    text and the historical reference preserve scene identity.}
    \label{fig:gallery_narrow_pose}
\end{figure*}

\begin{figure*}[p]
    \centering
    \includegraphics[width=0.96\textwidth]{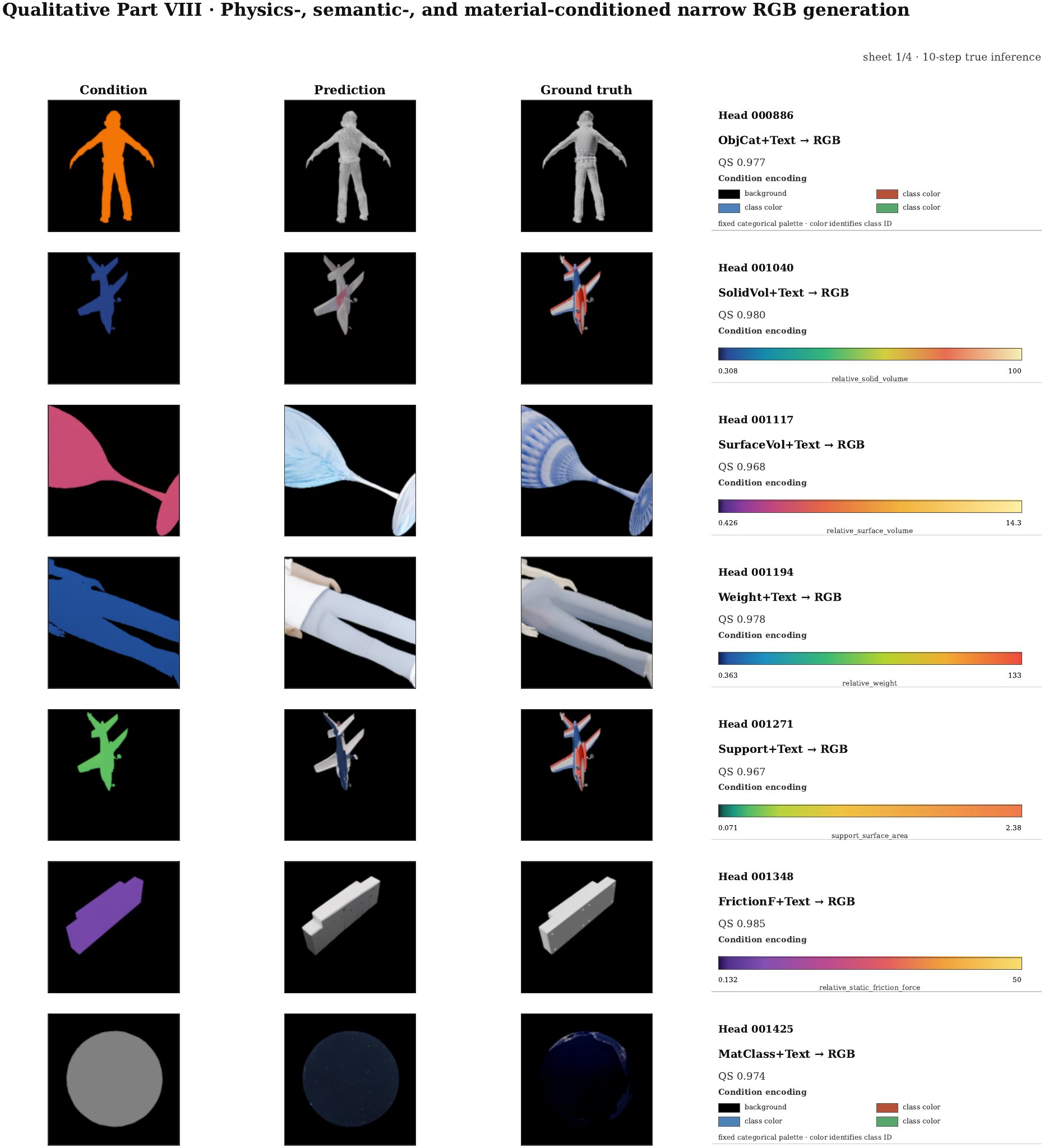}
    \vspace{-4pt}
    \caption{\textbf{Qualitative Part VIII: physics-, semantic-, and material-conditioned narrow RGB generation (1/4).}
    Object class, volume, weight, support, and friction fields begin the
    exhaustive non-geometric source sweep.}
    \label{fig:gallery_narrow_properties_i}
\end{figure*}

\begin{figure*}[p]
    \centering
    \includegraphics[width=0.96\textwidth]{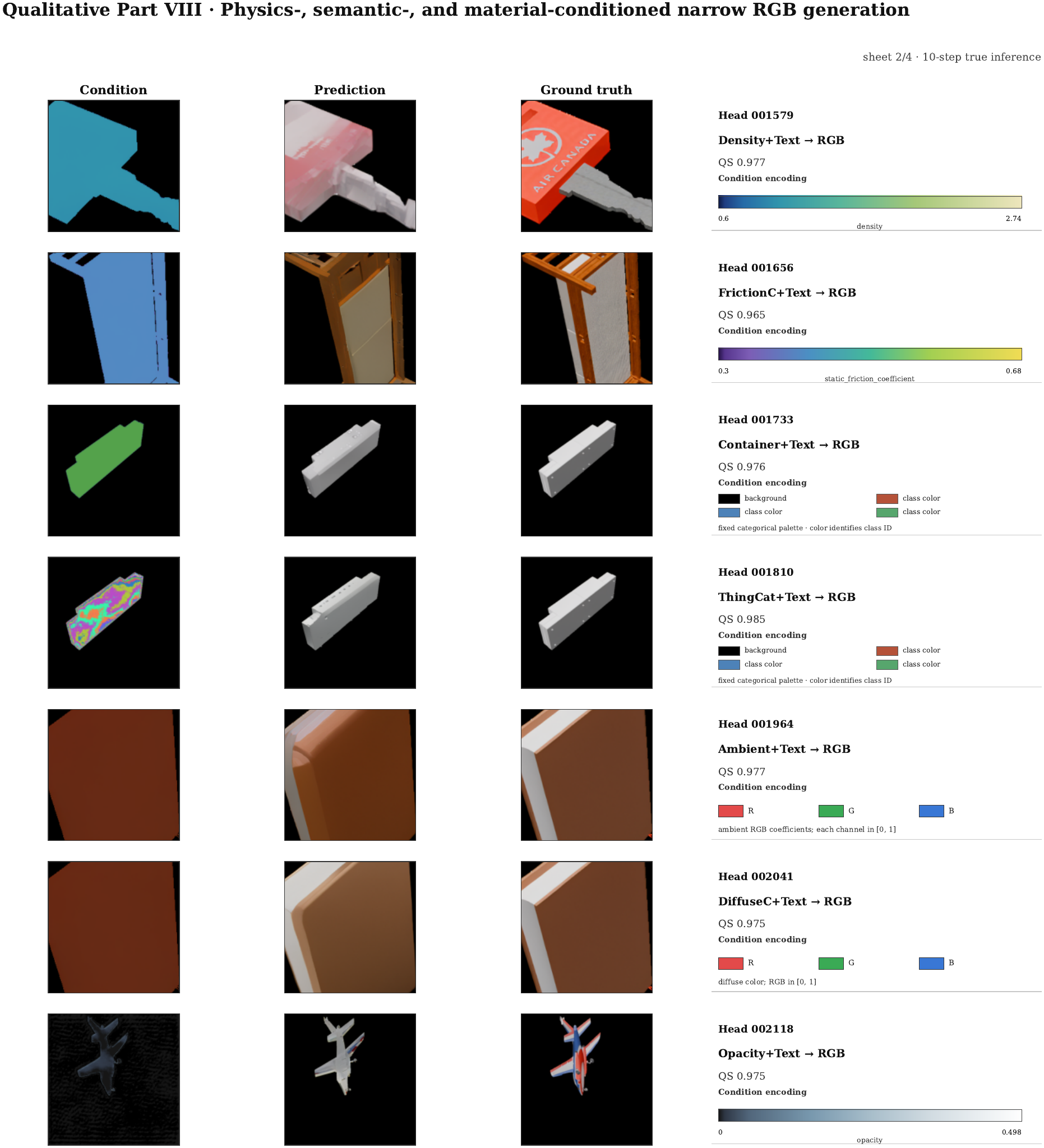}
    \vspace{-4pt}
    \caption{\textbf{Qualitative Part VIII: physics-, semantic-, and material-conditioned narrow RGB generation (2/4).}
    Material class, density, friction coefficient, container, entity category,
    ambient color, and diffuse color test controls whose information is
    concentrated inside object support rather than distributed across the image.}
    \label{fig:gallery_narrow_properties_ii}
\end{figure*}

\begin{figure*}[p]
    \centering
    \includegraphics[width=0.96\textwidth]{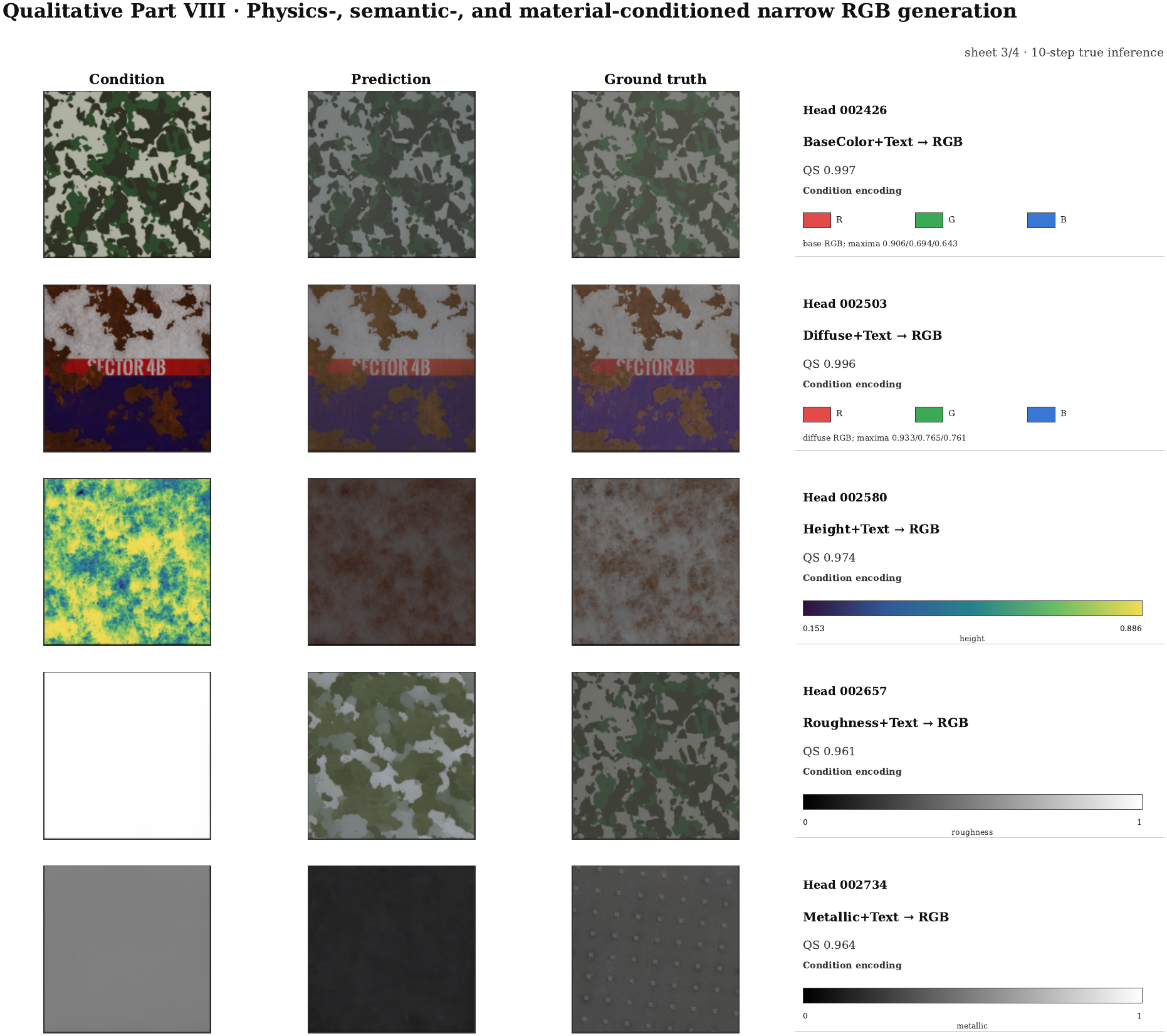}
    \vspace{-4pt}
    \caption{\textbf{Qualitative Part VIII: physics-, semantic-, and material-conditioned narrow RGB generation (3/4).}
    Base color, diffuse response, height, roughness, and metallicity show how
    appearance controls interact with text and RGB context.}
    \label{fig:gallery_narrow_material_i}
\end{figure*}

\begin{figure*}[p]
    \centering
    \includegraphics[width=0.96\textwidth]{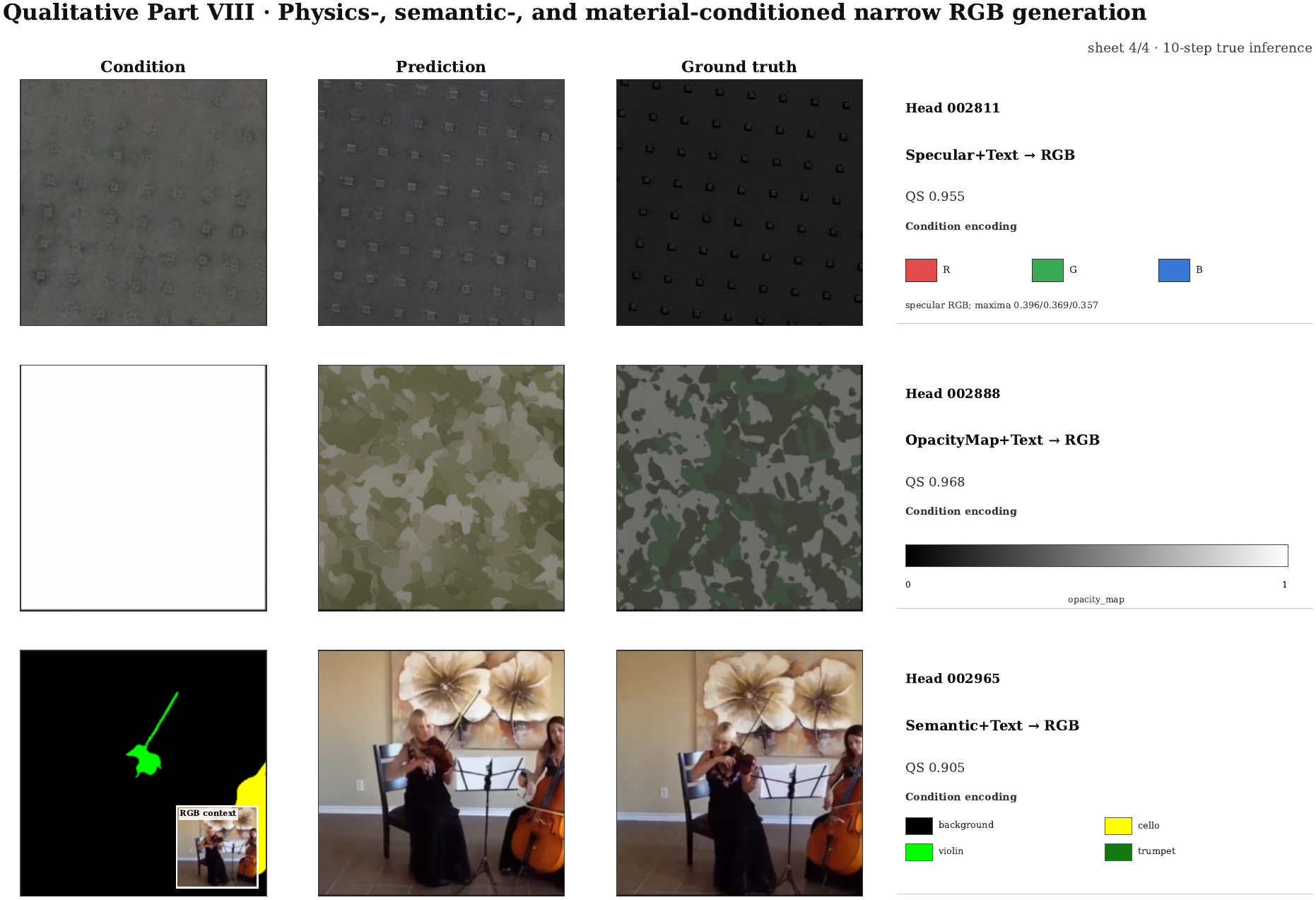}
    \vspace{-4pt}
    \caption{\textbf{Qualitative Part VIII: physics-, semantic-, and material-conditioned narrow RGB generation (4/4).}
    Specular response, opacity maps, and semantic masks complete the 33
    visually readable narrow-generation source modalities.}
    \label{fig:gallery_narrow_material_ii}
\end{figure*}

Parts~VII--VIII reverse the estimation interface for 33 visually meaningful
source modalities. WSG channels specify the structure or property to preserve, while
text and, when available, a same-video historical frame carry appearance.  RGB
context is shown as a picture-in-picture inset; static examples without a valid
predecessor simply omit the inset rather than displaying an empty panel.

\begin{figure*}[p]
    \centering
    \includegraphics[width=0.96\textwidth]{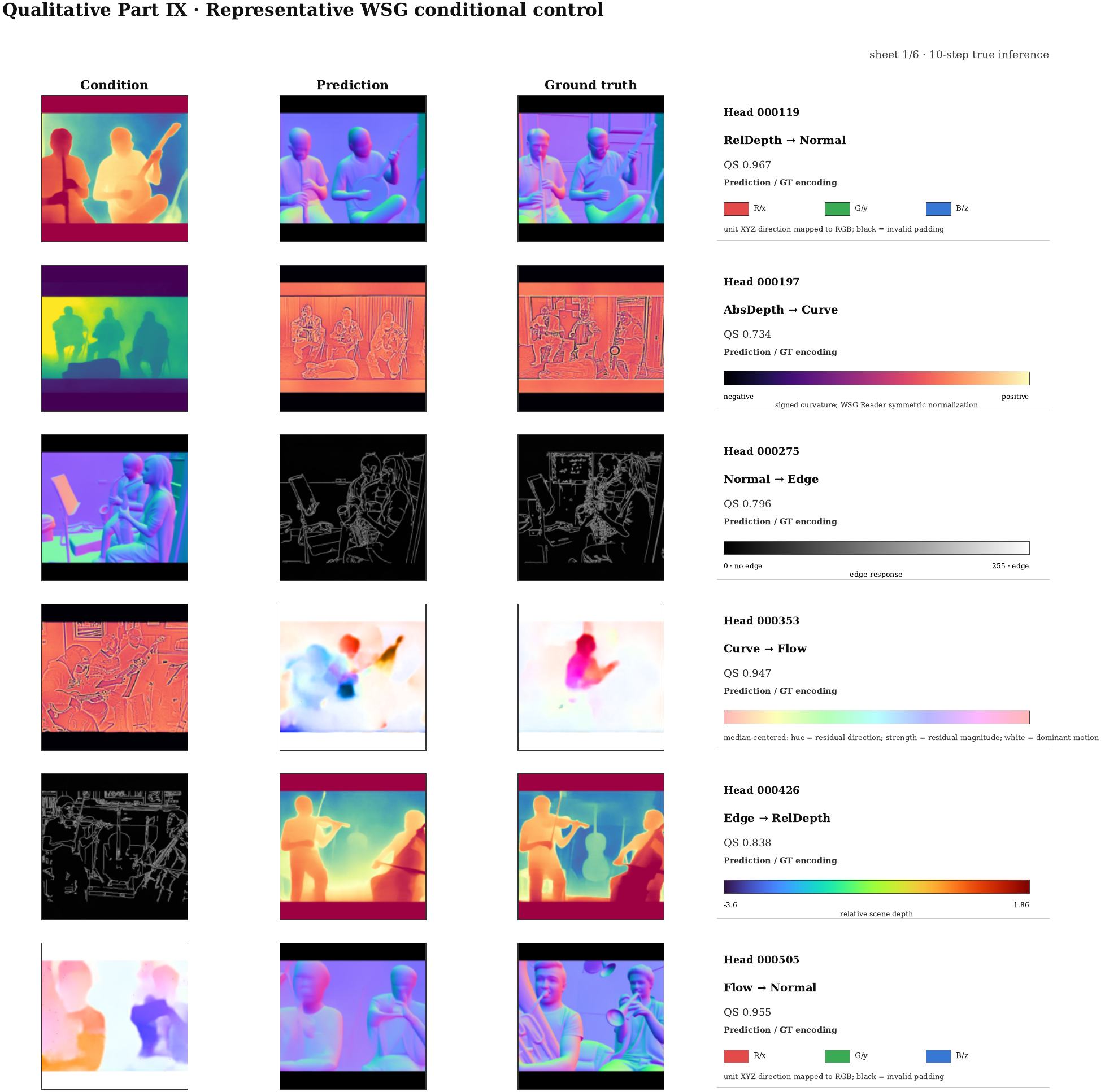}
    \vspace{-4pt}
    \caption{\textbf{Qualitative Part IX: representative WSG conditional control (1/6).}
    Six geometry-to-geometry transfers cover depth, normals, curvature, edges,
    and flow without auxiliary RGB evidence.}
    \label{fig:gallery_control_i}
\end{figure*}

\begin{figure*}[p]
    \centering
    \includegraphics[width=0.96\textwidth]{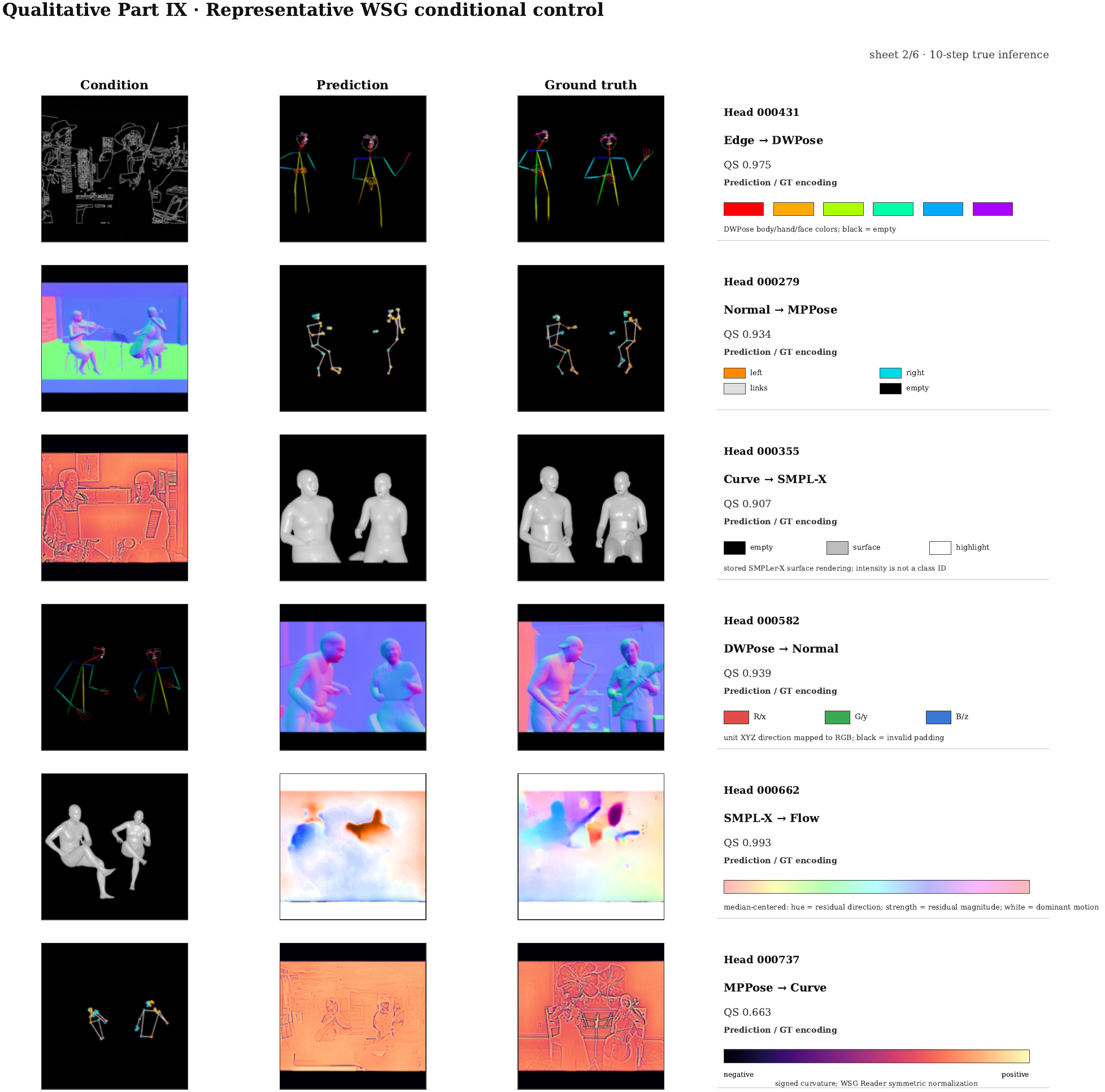}
    \vspace{-4pt}
    \caption{\textbf{Qualitative Part IX: representative WSG conditional control (2/6).}
    Geometry-to-pose and pose-to-geometry mappings test whether dense surface
    structure and sparse articulated structure can supervise one another.}
    \label{fig:gallery_control_ii}
\end{figure*}

\begin{figure*}[p]
    \centering
    \includegraphics[width=0.96\textwidth]{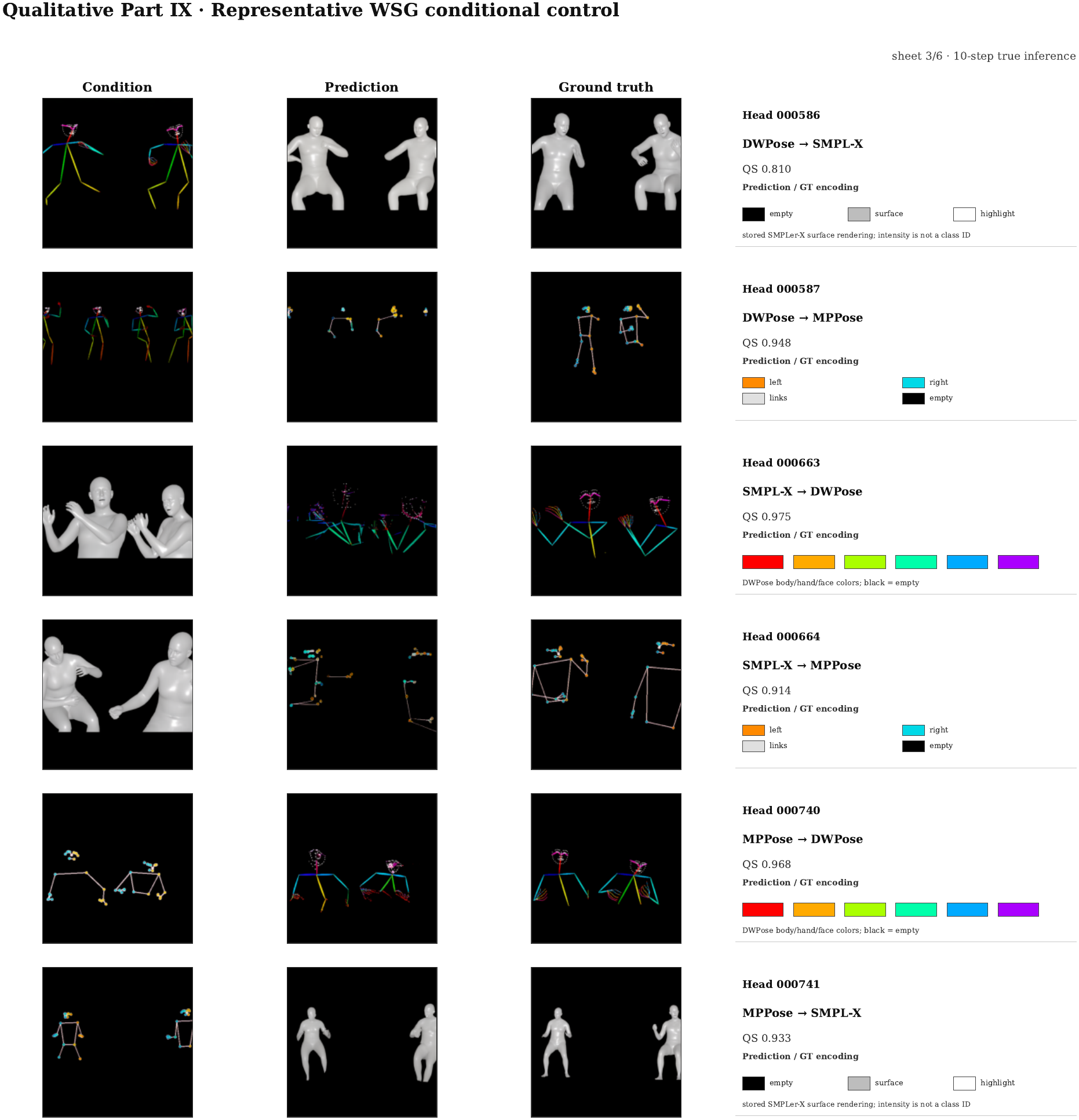}
    \vspace{-4pt}
    \caption{\textbf{Qualitative Part IX: representative WSG conditional control (3/6).}
    All six directed conversions among DWPose, SMPL-X, and MediaPipe pose are
    shown, exposing the common articulated structure beneath distinct encodings.}
    \label{fig:gallery_control_iii}
\end{figure*}

\begin{figure*}[p]
    \centering
    \includegraphics[width=0.96\textwidth]{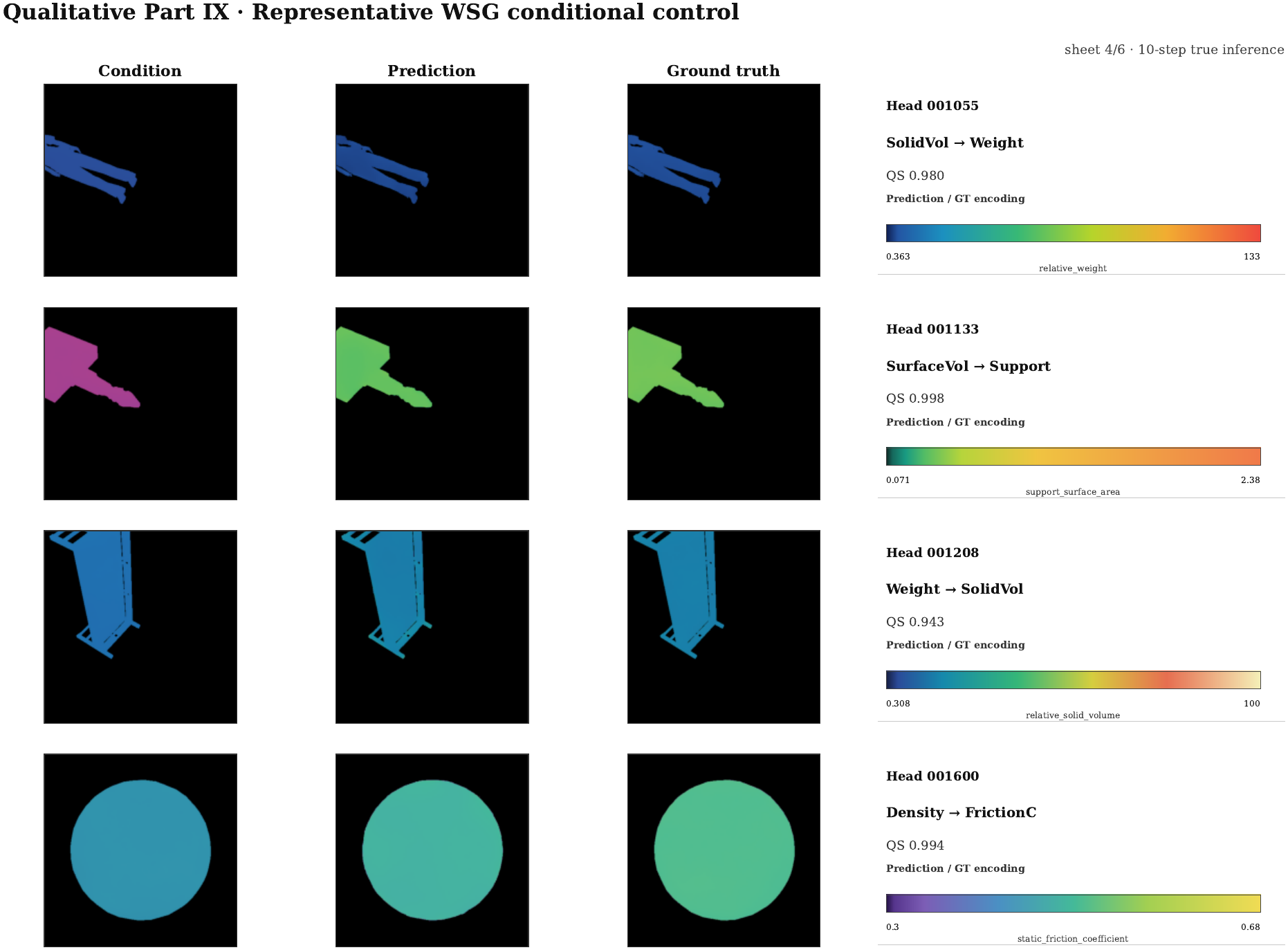}
    \vspace{-4pt}
    \caption{\textbf{Qualitative Part IX: representative WSG conditional control (4/6).}
    Four physical-property transfers probe volume, weight, support, density,
    and friction while preserving object-aligned spatial support.}
    \label{fig:gallery_control_iv}
\end{figure*}

\begin{figure*}[p]
    \centering
    \includegraphics[width=0.96\textwidth]{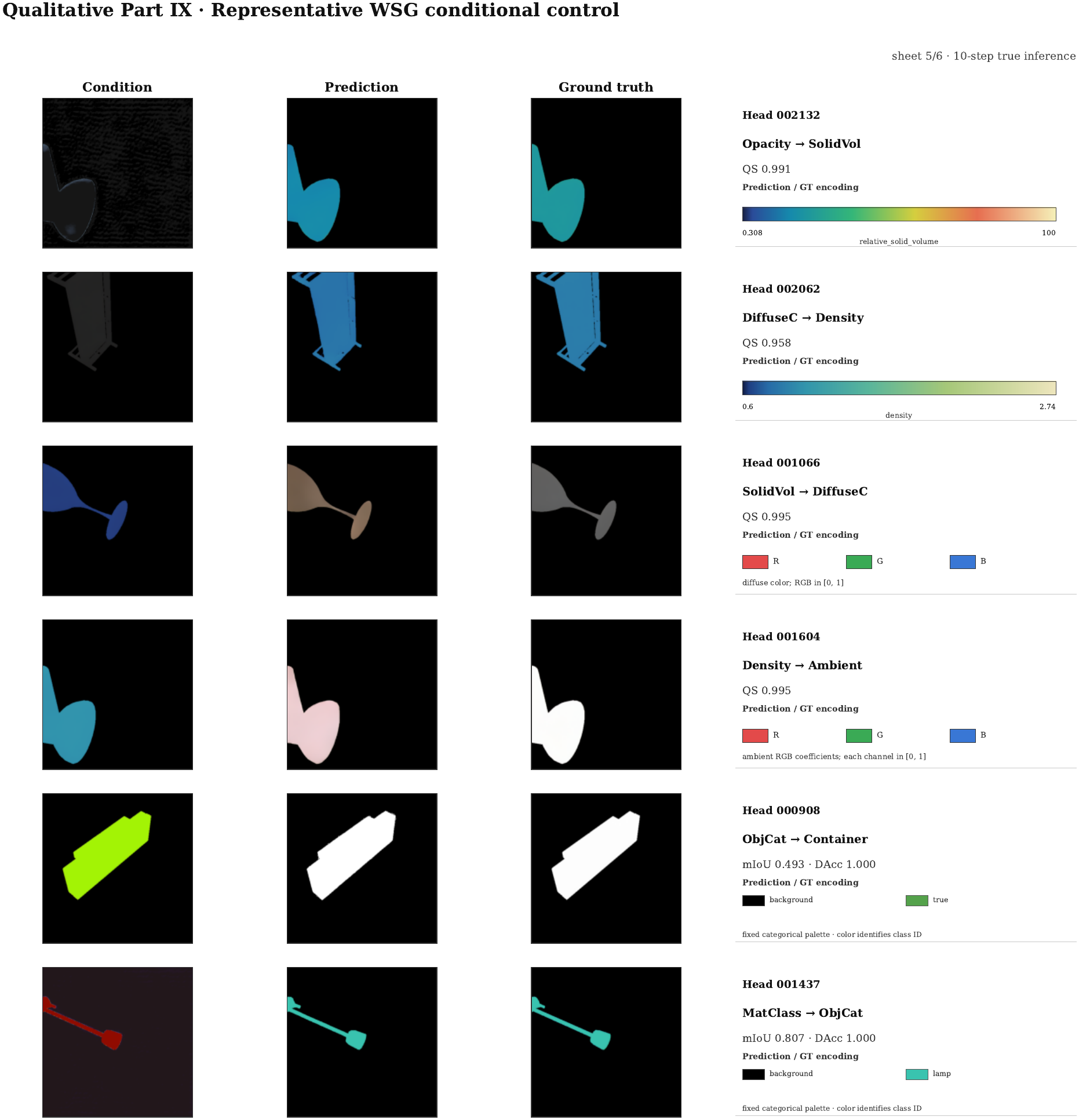}
    \vspace{-4pt}
    \caption{\textbf{Qualitative Part IX: representative WSG conditional control (5/6).}
    Four material-to-physics and physics-to-material mappings test whether
    shared object support is sufficient for translation between distinct value
    spaces; two categorical rows test label-aware transport.}
    \label{fig:gallery_control_v}
\end{figure*}

\begin{figure*}[p]
    \centering
    \includegraphics[width=0.96\textwidth]{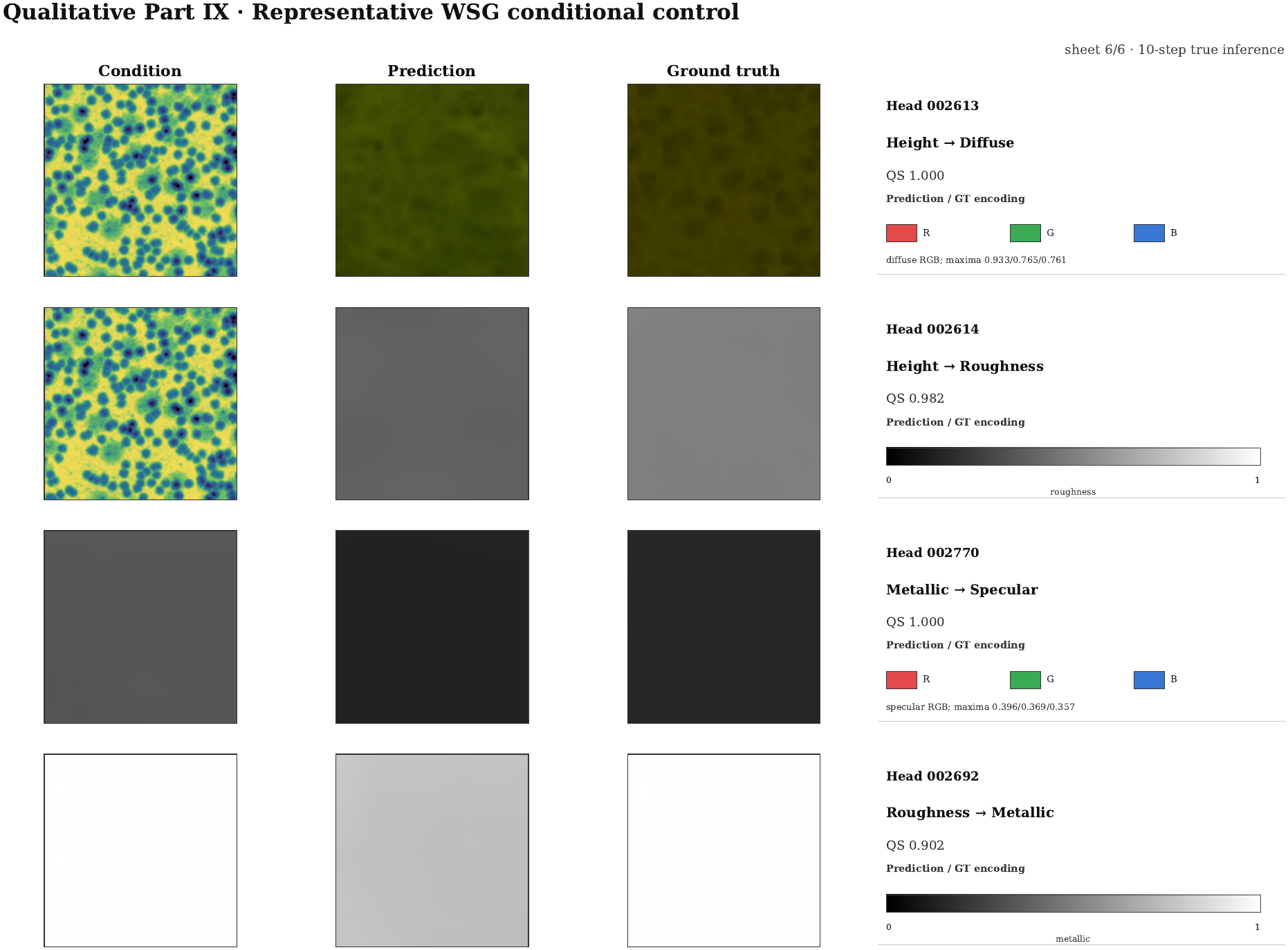}
    \vspace{-4pt}
    \caption{\textbf{Qualitative Part IX: representative WSG conditional control (6/6).}
    Four PBR conversions isolate translation among base color, diffuse,
    roughness, metallicity, and specular response.}
    \label{fig:gallery_control_vi}
\end{figure*}

\begin{figure*}[p]
    \centering
    \includegraphics[width=0.96\textwidth]{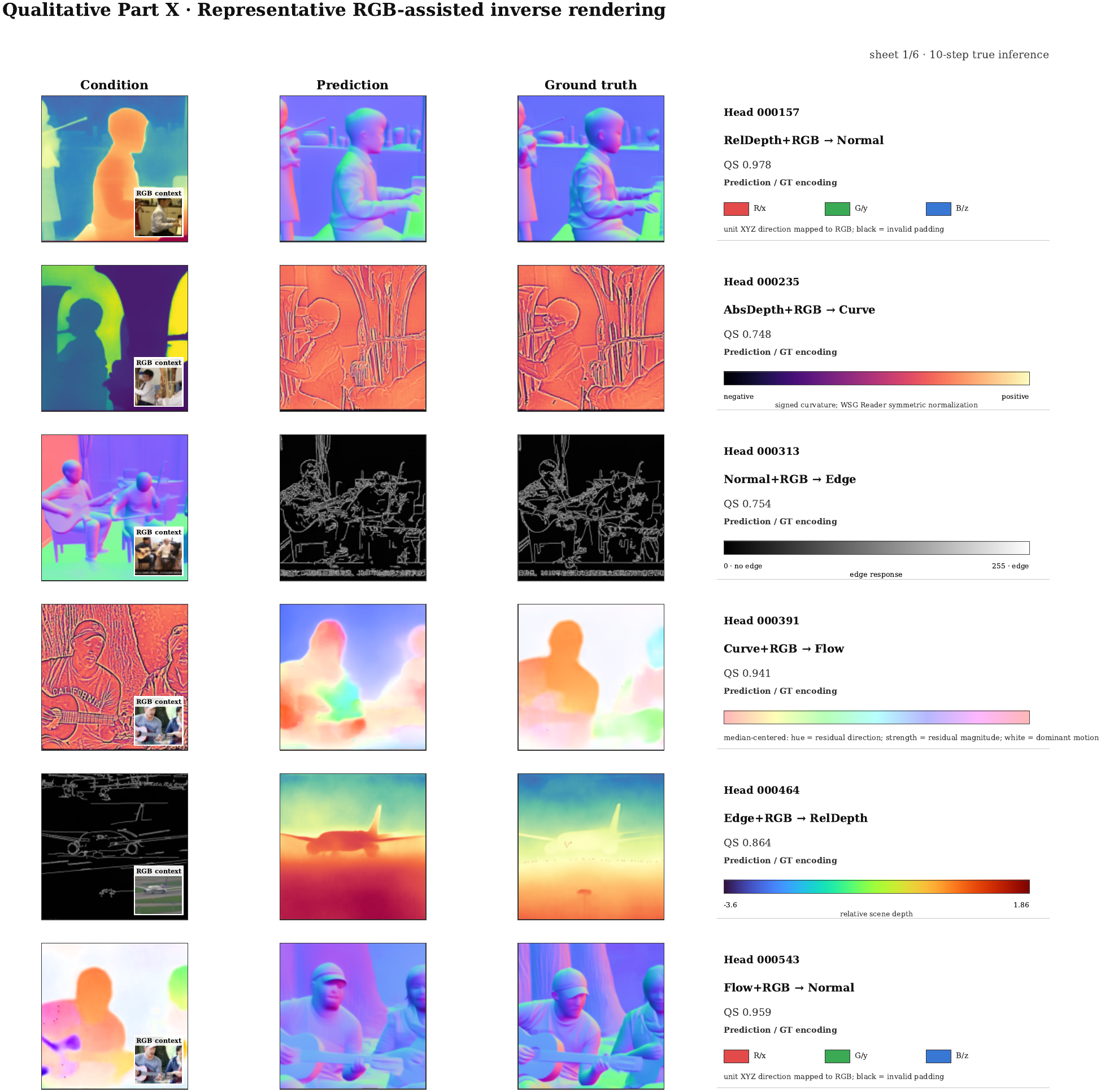}
    \vspace{-4pt}
    \caption{\textbf{Qualitative Part X: representative RGB-assisted inverse rendering (1/6).}
    Geometry-to-geometry transfers repeat Part~IX with observed RGB evidence,
    making scene-specific boundaries and identity available to the TunnelingNet.}
    \label{fig:gallery_inverse_i}
\end{figure*}

\begin{figure*}[p]
    \centering
    \includegraphics[width=0.96\textwidth]{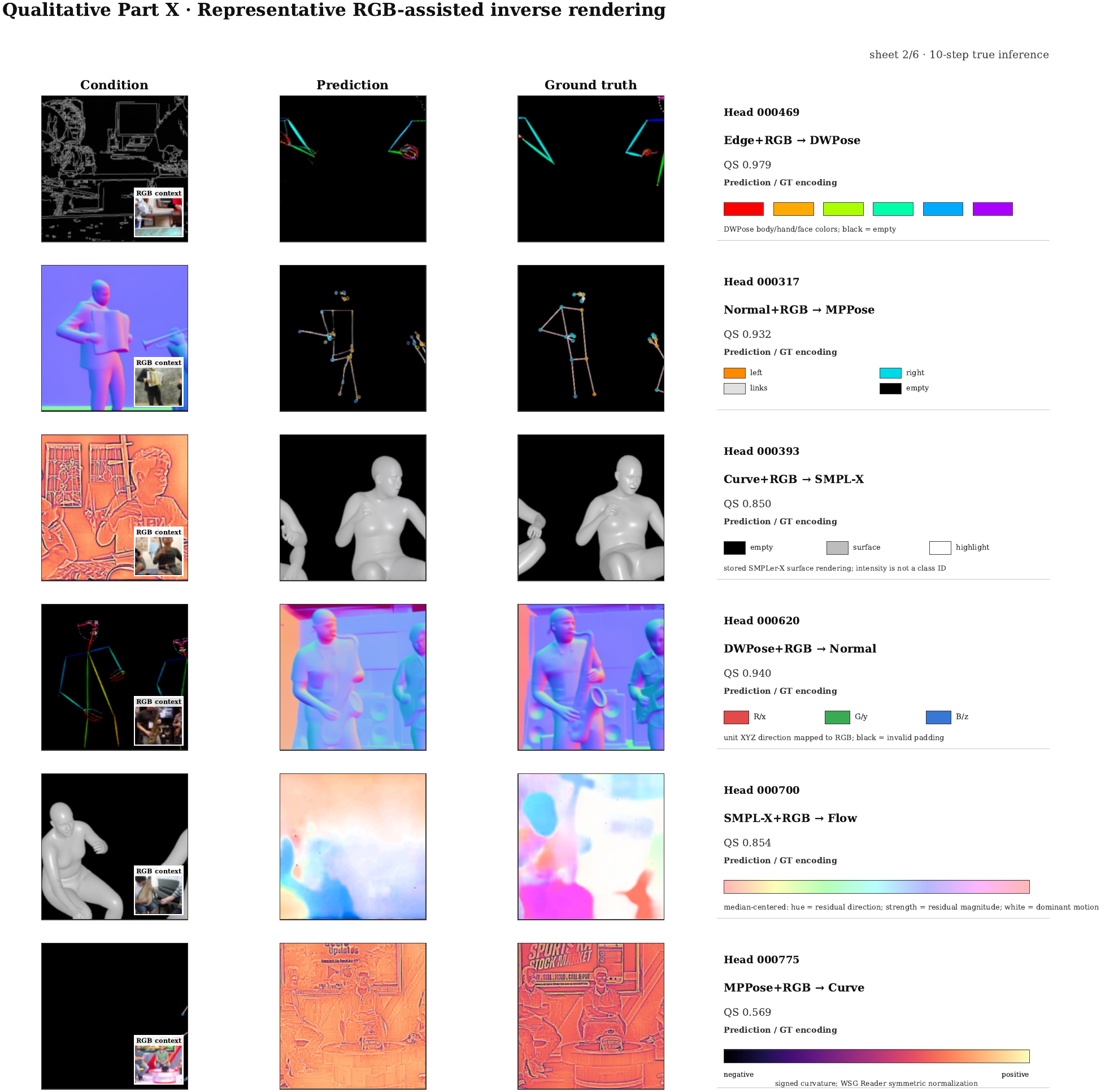}
    \vspace{-4pt}
    \caption{\textbf{Qualitative Part X: representative RGB-assisted inverse rendering (2/6).}
    Geometry-to-pose and pose-to-geometry mappings show how RGB evidence resolves
    structure that is underdetermined by the first condition alone.}
    \label{fig:gallery_inverse_ii}
\end{figure*}

\begin{figure*}[p]
    \centering
    \includegraphics[width=0.96\textwidth]{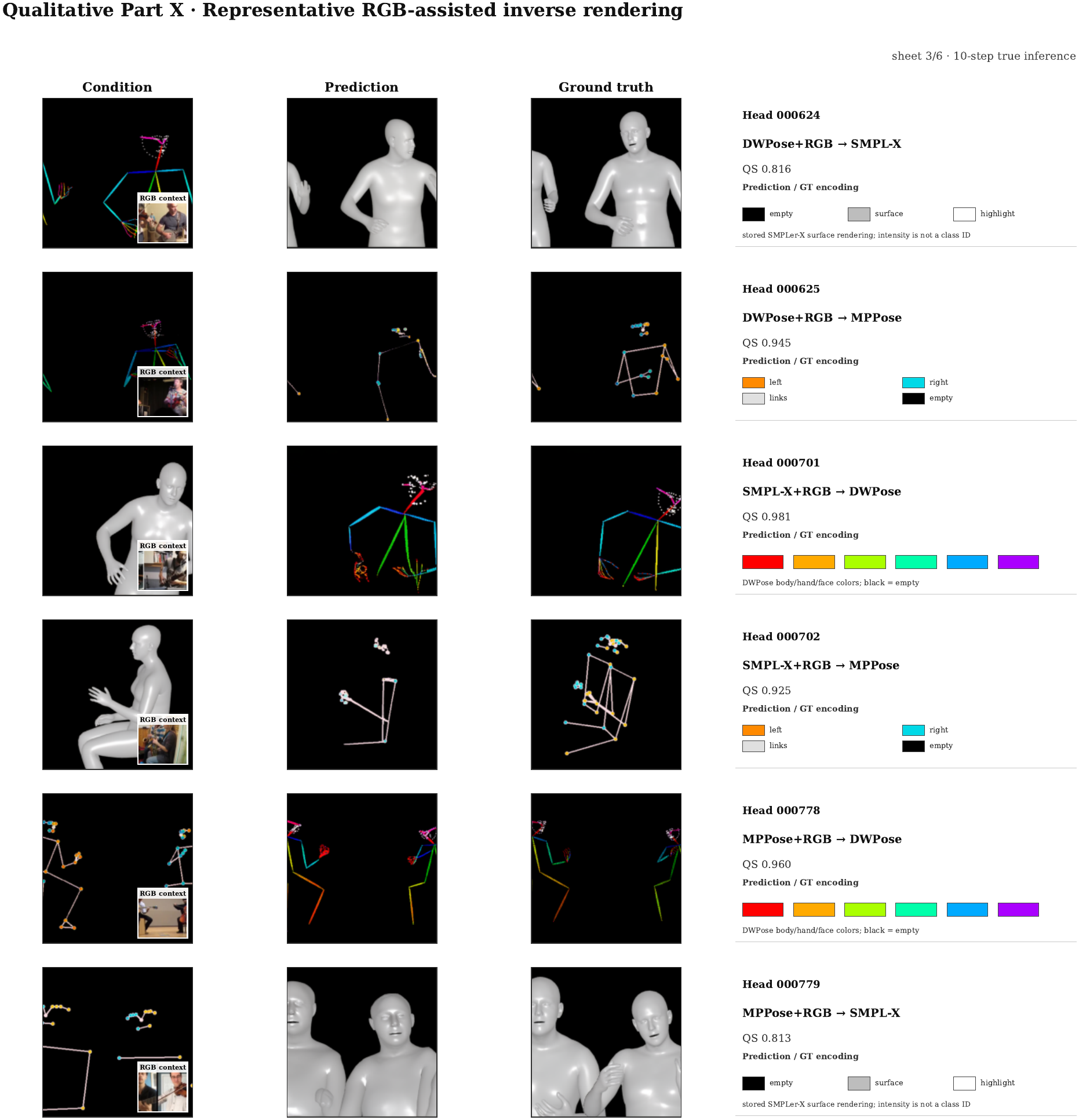}
    \vspace{-4pt}
    \caption{\textbf{Qualitative Part X: representative RGB-assisted inverse rendering (3/6).}
    Six directed pose conversions retain their explicit condition while using
    the RGB observation to recover scene-specific body extent and visibility.}
    \label{fig:gallery_inverse_iii}
\end{figure*}

\begin{figure*}[p]
    \centering
    \includegraphics[width=0.96\textwidth]{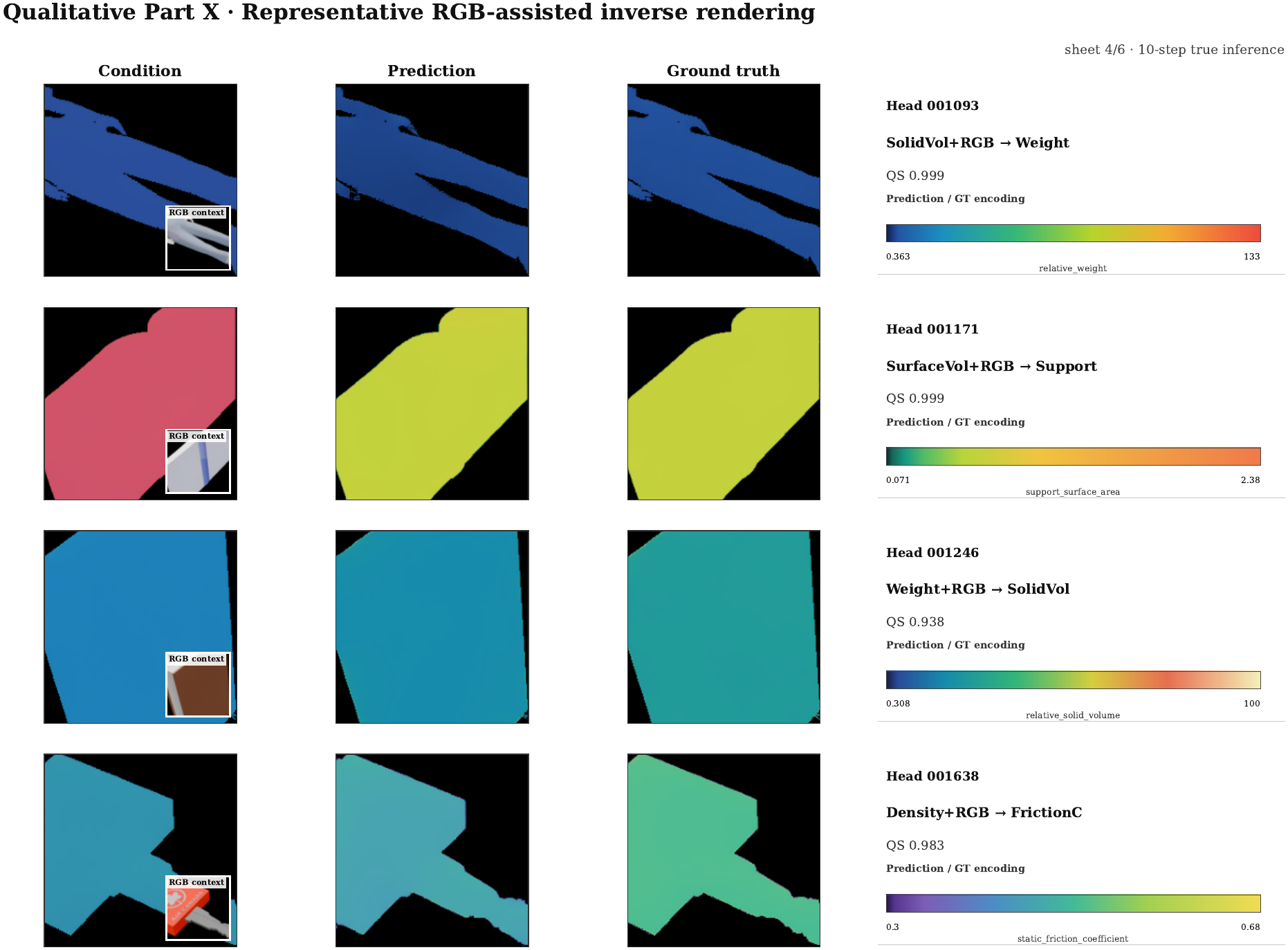}
    \vspace{-4pt}
    \caption{\textbf{Qualitative Part X: representative RGB-assisted inverse rendering (4/6).}
    Physical-property mappings combine object-aligned structured evidence with
    RGB boundaries to stabilize support and predicted value.}
    \label{fig:gallery_inverse_iv}
\end{figure*}

\begin{figure*}[p]
    \centering
    \includegraphics[width=0.96\textwidth]{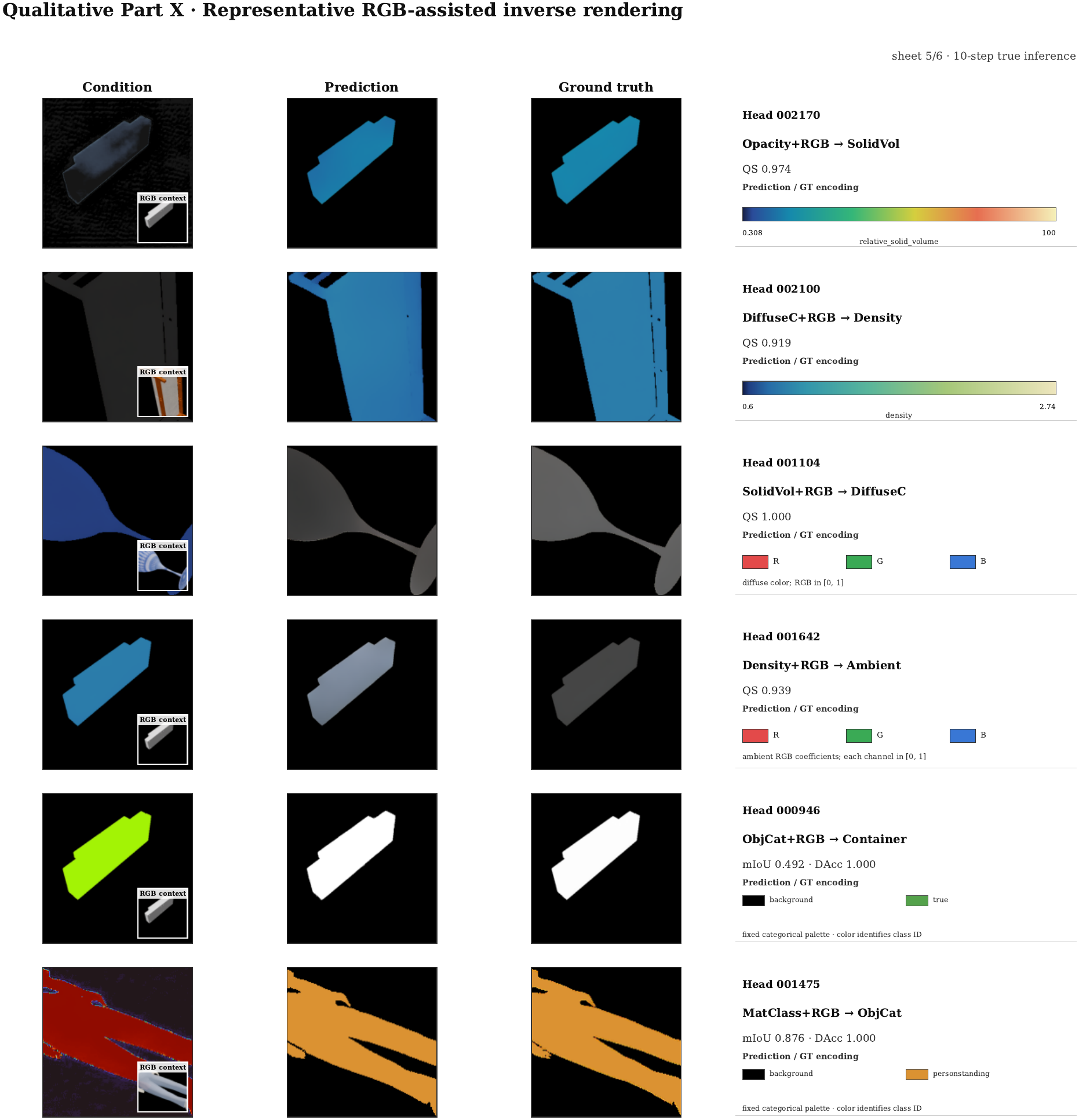}
    \vspace{-4pt}
    \caption{\textbf{Qualitative Part X: representative RGB-assisted inverse rendering (5/6).}
    Four cross-family transfers show where visible object identity complements
    the structured input; two categorical rows retain label-aware evaluation.}
    \label{fig:gallery_inverse_v}
\end{figure*}

\begin{figure*}[p]
    \centering
    \includegraphics[width=0.96\textwidth]{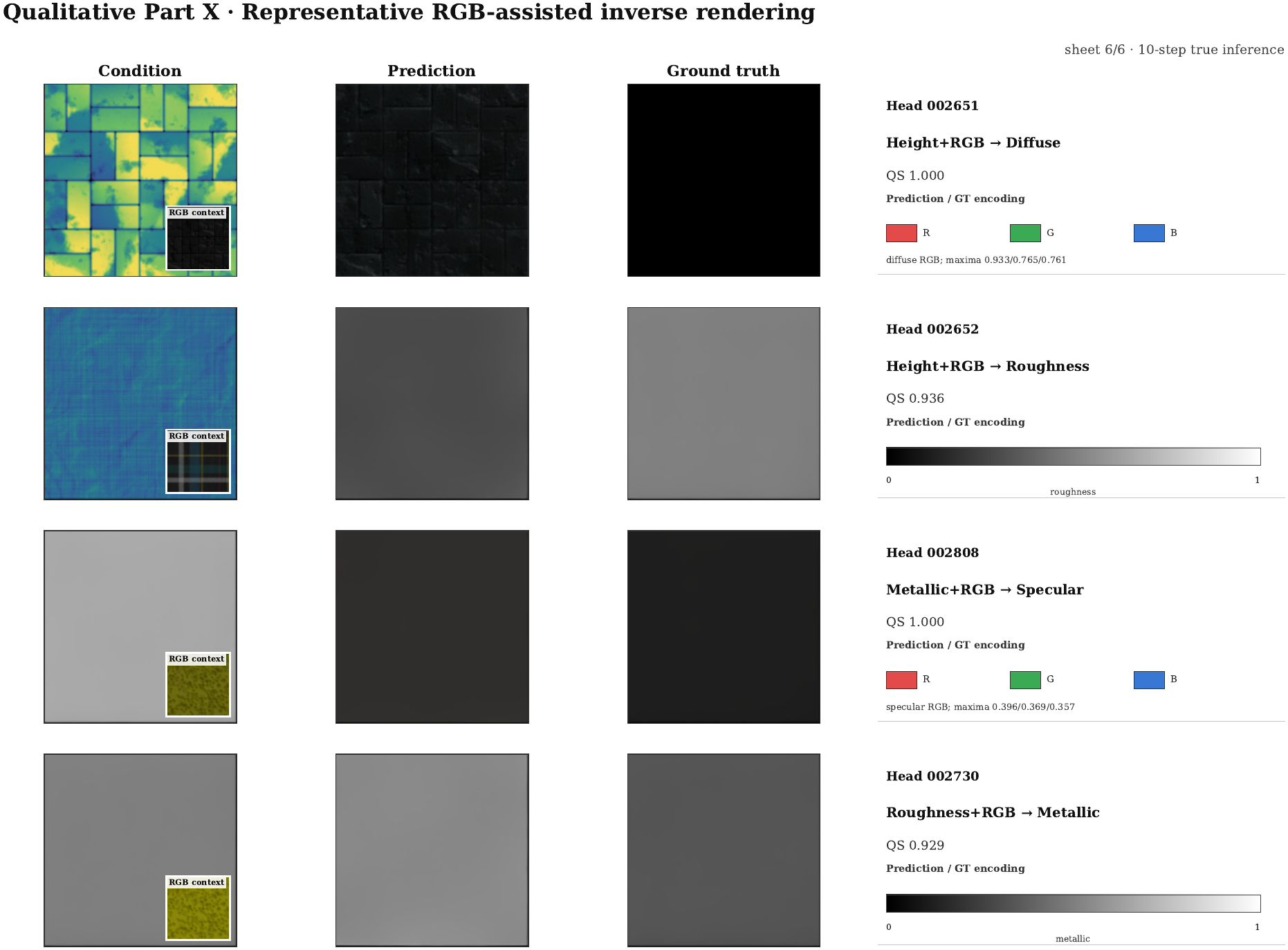}
    \vspace{-4pt}
    \caption{\textbf{Qualitative Part X: representative RGB-assisted inverse rendering (6/6).}
    Four PBR conversions complete the 32-mapping RGB-assisted slice of the
    family of 536 TunnelingNets.}
    \label{fig:gallery_inverse_vi}
\end{figure*}

The paired control and inverse-rendering Parts reveal where auxiliary RGB
evidence matters.  Conversions between closely related representations, such
as pose systems or compatible PBR channels, remain strong without RGB.
Ambiguous mappings benefit from observation because it supplies visible
boundaries, appearance, and object identity that are absent from the structured
condition.  This comparison separates genuine cross-variable translation from
evidence contributed by the scene image.

\paragraph{Visualization conventions.}
Each row contains a compact key for the modality whose colors must be
interpreted, and prediction and ground truth share that key. Continuous scalar
fields use the modality-level range defined by WSG Reader and recorded in the
artifact metadata; the displayed prediction and target are never normalized
independently. Fixed grayscale fields retain their numeric endpoints. Normal
and flow keys state component, direction, and magnitude semantics; direct PBR
colors retain their RGB meaning; categorical squares name the labels represented
by the palette. Consequently, the figures show both spatial agreement and the
meaning of the predicted value.

\FloatBarrier

\begin{figure*}[p]
    \centering
    \includegraphics[width=0.96\textwidth]{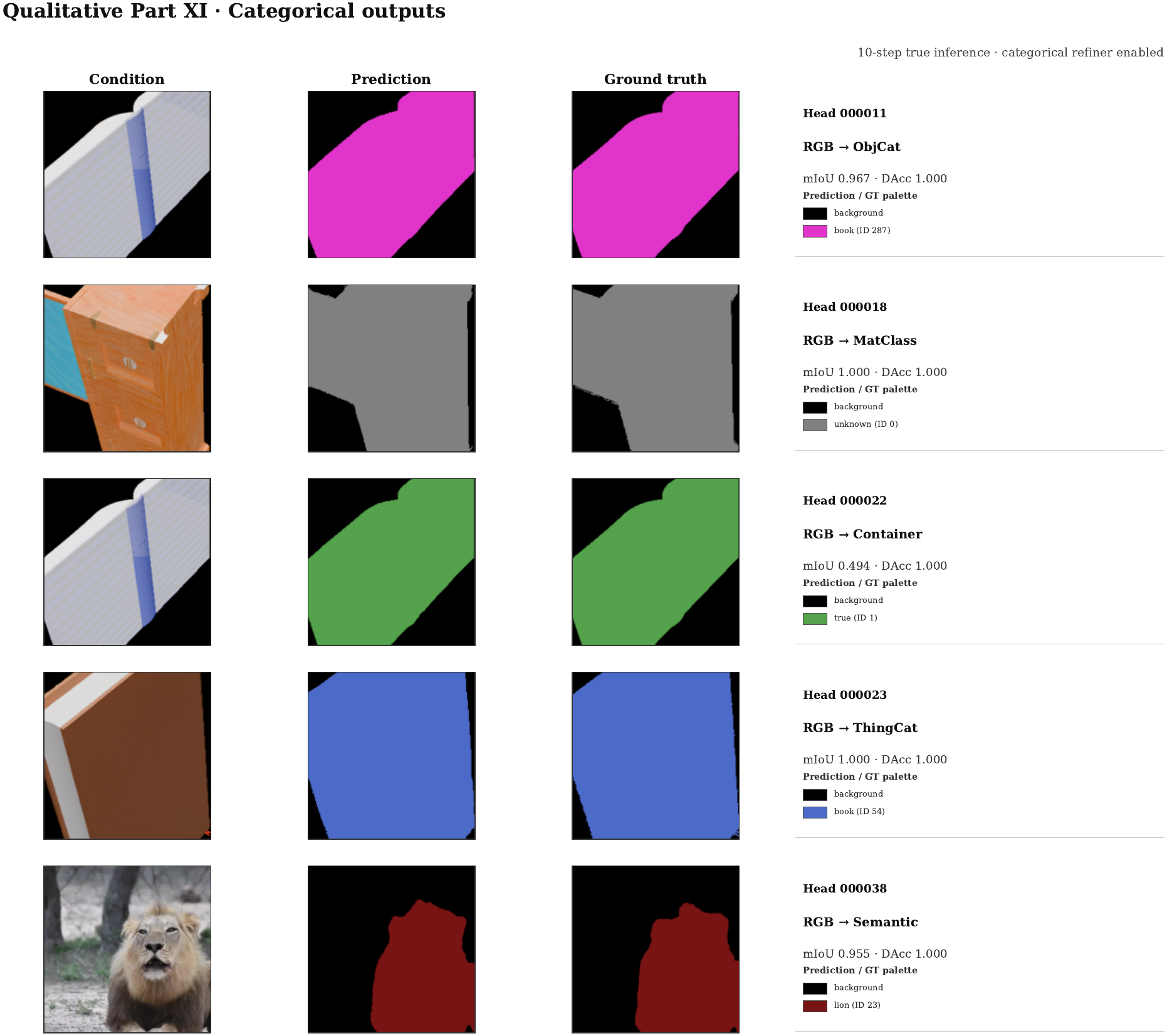}
    \caption{\textbf{Qualitative Part XI: categorical outputs.}
    One verified example is retained for each of the five
    categorical targets. Each row shows the condition, the final output of the
    complete categorical TunnelingNet, and the corresponding ground truth.}
    \label{fig:evaluation_categorical_examples}
\end{figure*}

% Failure-case diagnostics are retained in the source for future supplementary
% analysis, but are not part of the main-paper qualitative presentation.
\iffalse
\begin{figure*}[p]
    \centering
    \includegraphics[width=0.78\textwidth]{fig/evaluation_categorical_failure_cases_normv2.pdf}
    \caption{\textbf{Explicitly retained categorical failure cases.}
    These lowest-mIoU samples remain in the main paper rather than being hidden
    by aggregate metrics. They show that the categorical branch assigns labels
    within generated support but cannot reconstruct a region absent from the
    spatial prediction.}
    \label{fig:evaluation_categorical_failures}
\end{figure*}
\fi

\FloatBarrier

\subsection{Interface Comparison}
\label{subsec:qualitative_interface_analysis}

Part~I exposes an evidence asymmetry that is consistent with human geometric
reasoning.  In the Normal-to-Edge row of
Figure~\ref{fig:evaluation_four_interfaces}, the TunnelingNet produces a visually
cleaner boundary map than the pseudo-ground-truth rendering: smooth surface
normals retain geometric discontinuities while suppressing texture- and
illumination-specific high-frequency variation.  The objective is therefore
not to reproduce every incidental detail of the source representation, but to
transport the evidence needed by the requested target.  Conversely, when a
target depends on boundaries or appearance not determined by a structured
condition alone, RGB-assisted inverse rendering restores scene-specific cues
from the observation.  Omitting RGB yields a geometry-supported abstraction;
supplying RGB restores observation-specific detail.  We next examine every
remaining qualitative figure in order, separating spatial fidelity, value
calibration, and categorical assignment.

\subsection{RGB Estimation}

\paragraph{Geometry and motion.}
Figure~\ref{fig:gallery_geometry_motion} shows that the same RGB condition can
be translated into six geometrically distinct representations. Normals,
curvature, and absolute depth are especially stable (QS 0.959, 0.896, and
0.936). Relative depth and the direction-consistent flow example retain the
dominant scene layout (0.850 and 0.968), while Edge is less uniform (0.837) because thin
boundaries are only weakly determined by local appearance.  The contrast is consistent with
human geometric reasoning---surface orientation can remain clear even when an
exact metric distance or boundary convention is ambiguous.

\paragraph{Articulated structure.}
Figure~\ref{fig:gallery_pose_structure} tests DWPose, SMPL-X, MediaPipe Pose,
XPose, MediaPipe Seg, and Universal Pose rather than six renderings of one dense
field.
Their common success
is structural: joints remain connected in a recognizable body layout. In the
third row, the prediction occasionally completes joints that are missing
or malformed in the automatically produced MediaPipe pseudo ground truth.  This
does not make the proxy annotation infallible; instead, it shows that the
learned human prior can denoise imperfect supervision.  The fourth-row XPose
targets are likewise produced by an automatic detector rather than manual
annotation, and their quality remains a limitation of the present dataset.
Nevertheless, their scale supplies enough weak supervision for the TunnelingNet to
recover characteristic skeletal structure.  The result is evidence that the
proposed paradigm can discover useful structure from large, imperfectly
annotated corpora, while also identifying where better labels would improve
the model. The MediaPipe Seg row complements these skeletal encodings with a
dense estimate of human support. Its official six-region palette distinguishes
background, hair, body skin, face skin, clothes, and accessories. We render
the TunnelingNet's generated field through this discrete palette before categorical-refiner
correction, so the figure exposes both region support and direct class assignment.
The final row extends the same image-to-structure formulation beyond fixed human
representations: a custom TunnelingNet recovers the category-independent red-joint,
white-link Universal Pose used by MP-100 from an RGB observation.

\paragraph{Physical properties.}
Figure~\ref{fig:gallery_physical_properties} contains seven object-aligned
fields whose QS values lie between 0.974 and 1.000.  Solid and surface volume,
weight, support, friction, and density preserve nearly the
same object support while changing the value carried inside it.  This result
should not be read as direct measurement of an object's physical constants;
rather, it shows that the TunnelingNet learns the dataset's encoded physical estimate
without losing spatial correspondence to the observed object.

\paragraph{Material and PBR appearance I.}
Figure~\ref{fig:gallery_material_pbr_i} compares ambient and diffuse color,
opacity, specular color, shininess, base color, and diffuse response. The strong
opacity, specular-color, shininess, and base-color examples show that broad,
piecewise-smooth appearance attributes can be transported with little spatial
drift. Ambient and diffuse color remain more sensitive to illumination, making
them less directly recoverable than support or intrinsic appearance.

\paragraph{Material and PBR appearance II.}
Figure~\ref{fig:gallery_material_pbr_ii} exposes the harder part of the PBR
interface. Height, metallicity, and specular response remain strong, whereas
roughness and opacity maps deviate more visibly from the target. These channels
require calibrated local intensity rather than only a correct object mask;
consequently, a prediction may retain the right region while assigning an
imperfect material value.

\paragraph{Segmentation and semantics.}
Figure~\ref{fig:gallery_segmentation_semantics} demonstrates that continuous
Alpha and categorical masks can share the same generative interface.  The
selected material, entity, object, container, and semantic examples have high
mIoU and dominant-class accuracy, so both the generated support and principal palette label are
correct in these cases.  They are deliberately readable successful examples,
not a substitute for the aggregate matrices:
Figure~\ref{fig:evaluation_categorical_examples} later presents the final
outputs of the complete categorical TunnelingNets.

\subsection{Narrow RGB Generation}

\paragraph{Geometric control.}
Figure~\ref{fig:gallery_narrow_geometry} reverses the RGB-estimation direction.
Depth, normals, curvature, edges, and flow now constrain RGB synthesis while
text and optional historical RGB provide appearance.  The six examples range
from QS 0.898 to 0.979; curvature and normal control are particularly stable,
while flow is weakest.  This ordering reflects how directly a condition fixes
the target frame: static geometry constrains most pixels, whereas flow must be
interpreted jointly with temporal and appearance context.

\paragraph{Human control.}
Figure~\ref{fig:gallery_narrow_pose} replaces dense geometry with articulated
human representations.  DWPose, SMPL-X, MediaPipe Pose/Class, and XPose all
produce coherent RGB frames, with QS spanning 0.809--0.947.  The lower SMPL-X
and XPose scores expose the difficulty of reconstructing texture and
background from sparse pose evidence.  QS measures target-frame reconstruction
and therefore complements, rather than replaces, visual inspection of whether
the requested pose actually redirects the person.

\paragraph{Object-level physical control.}
Figure~\ref{fig:gallery_narrow_properties_i} covers object category, volume,
weight, support, and friction conditions. Despite their low visual bandwidth,
these fields yield recognizable frames because the physical and category
channels narrow the set of compatible appearances.

\paragraph{Class and material control.}
Figure~\ref{fig:gallery_narrow_properties_ii} continues with material class,
density, friction coefficient, container, entity category, and
ambient color. These conditions remain useful even when their nonzero
information occupies only a small part of the canvas. Text and historical RGB
remain necessary because the structured channel determines what must be
preserved but does not fully specify photometric appearance.

\paragraph{Appearance control.}
Figure~\ref{fig:gallery_narrow_material_i} separates direct color fields from
derived PBR controls. Base color and diffuse response retain broad object support, while height, roughness, and metallicity constrain
appearance more indirectly. The comparison shows how text and historical RGB
resolve the many RGB images compatible with a low-bandwidth material field.

\paragraph{PBR and semantic control.}
Figure~\ref{fig:gallery_narrow_material_ii} completes the readable source
sweep with specular response, opacity map, and semantic masks. The PBR fields
preserve continuous appearance structure; a semantic mask constrains identity
and support but intentionally leaves within-class texture and illumination
underdetermined.

\subsection{Conditional Control}

\paragraph{Geometry-to-geometry transport.}
Figure~\ref{fig:gallery_control_i} spans QS 0.734--0.967, revealing that
geometric channels are not interchangeable merely because they share image
coordinates.  Relative-depth-to-Normal and Flow-to-Normal are strong, whereas
Normal-to-Edge is lower because smooth normals suppress texture boundaries
present in the target edge convention.  The cleaner predicted edge is thus a
meaningful geometric abstraction, even when it does not minimize every target
pixel error.

\paragraph{Geometry and pose.}
Figure~\ref{fig:gallery_control_ii} crosses dense surfaces with sparse body
structure.  Edge-to-DWPose and DWPose-to-Normal reach 0.975 and 0.939, showing that
visible boundaries can localize a body and a skeleton can guide coarse surface
orientation. Curve-to-SMPL-X and MediaPipe-to-Curvature reach 0.907 and 0.663,
as they require recovering quantities absent from the source rather than
translating a shared structure.

\paragraph{Pose-system conversion.}
Figure~\ref{fig:gallery_control_iii} isolates six directed conversions among
DWPose, SMPL-X, and MediaPipe Pose. All six retain strong joint organization,
while DWPose-to-SMPL-X is the hardest at 0.810.  The asymmetry is important:
two pose systems can describe the same person yet expose different joints,
surface priors, and occlusion conventions, so reversing an edge is not the same
learning problem.

\paragraph{Physical-property transport.}
Figure~\ref{fig:gallery_control_iv} is close to a ceiling for volume, weight,
support, density, and friction mappings (0.943--0.998).  These tasks share object-aligned
support and often vary mainly in the value assigned inside that support.  The
result explains the coherent high-score block in the quantitative matrix
without implying that arbitrary physical variables are causally equivalent.

\paragraph{Cross-family transport.}
Figure~\ref{fig:gallery_control_v} tests deliberately less homogeneous pairs.
Opacity-to-SolidVolume and DiffuseColor-to-Density remain strong, while
Density-to-Ambient is harder. Spatial support transfers readily across
families, but an appearance value cannot always be inferred from a physical
scalar. The two categorical rows additionally test whether shared support can
carry a discrete label rather than only a continuous value.

\paragraph{PBR transport.}
Figure~\ref{fig:gallery_control_vi} isolates four PBR conversions. Base-color
and diffuse relations remain strongly constrained by shared material support.
The fourth row, roughness-to-metallic, exposes a fundamentally weaker form of
evidence: roughness does not determine metallicity, so the TunnelingNet can only exploit
learned visual regularities---for example, that metallic surfaces are often
smooth---to form a plausible estimate. This is not deterministic recovery, and
the resulting degradation is expected. Nevertheless, the prediction goes
beyond copying aligned pixels: it reveals a cross-attribute visual association
that can be strengthened as supervision and coverage improve.

\subsection{RGB-Assisted Inverse Rendering}

\paragraph{Geometry with observed appearance.}
Figure~\ref{fig:gallery_inverse_i} is the RGB-assisted counterpart of
Figure~\ref{fig:gallery_control_i}.  RGB improves
Edge-to-RelativeDepth (0.838 to 0.864) and modestly improves both Normal outputs,
but Normal-to-Edge remains difficult and decreases from 0.796 to 0.754 in this sample.  The
reference therefore contributes missing scene evidence; it does not guarantee
a monotonic gain when the target convention conflicts with the structural
condition.

\paragraph{Geometry--pose inversion.}
Figure~\ref{fig:gallery_inverse_ii} shows the same mixed but interpretable
behavior.  RGB yields 0.932 for Normal-to-MediaPipe and 0.850 for
Curvature-to-SMPL-X, while MediaPipe-to-Curvature remains the weakest example at
0.569.
Visible appearance resolves body extent and occlusion, but a single RGB frame
still cannot uniquely determine every differential surface quantity.

\paragraph{Pose inversion.}
Figure~\ref{fig:gallery_inverse_iii} provides the cleanest evidence that the
reference image complements, rather than replaces, the structured condition.
Compared directly with the condition-only conversions in
Figure~\ref{fig:gallery_control_iii}, the RGB-assisted results are visibly more
complete and stable: the observation resolves person count, visible body
extent, and occlusion cues that a sparse pose representation leaves ambiguous.
Three of the six directed conversions improve quantitatively;
DWPose-to-SMPL-X rises from 0.810 to 0.816 and SMPL-X-to-MediaPipe from 0.914
to 0.925. The pose channel specifies the requested representation, while RGB
restores the scene evidence needed to instantiate it accurately.

\paragraph{Physical inversion.}
Figure~\ref{fig:gallery_inverse_iv} retains the near-ceiling behavior of the
physical block, now with RGB evidence. All four displayed mappings remain
between 0.938 and 0.999 QS. Differences are small because object
support was already available in the structured condition; RGB mainly sharpens
visible boundaries.

\paragraph{Cross-family inversion.}
Figure~\ref{fig:gallery_inverse_v} shows where observation is most useful.
Density-to-Ambient improves substantially, while opacity-, diffuse-, and
solid-volume-conditioned mappings remain stable. The categorical rows show the
same principle for discrete targets: RGB can clarify visible object identity,
but cannot repair a missing spatial support region. The evidence supports
selective assistance rather than a claim that RGB universally dominates
structured transport.

\paragraph{PBR inversion.}
Figure~\ref{fig:gallery_inverse_vi} isolates four RGB-assisted PBR conversions.
The largest gain occurs for roughness-to-metallic, where visible appearance
supplies evidence absent from roughness alone. Directly related color and
specular fields change less because their structured conditions already carry
most of the required support and value information. This contrast localizes
when observation repairs a cross-modal ambiguity.

\subsection{Categorical Output Analysis}

For categorical targets, the categorical refiner is an integral component of
the corresponding TunnelingNet and is enabled by default during inference.
Figure~\ref{fig:evaluation_categorical_examples} presents the resulting
end-to-end outputs for Semantic, ThingCat, Container, MatClass, and ObjCat.
The shared palettes make both spatial extent and class identity directly
comparable with the ground truth.

\subsection{Sparse Digitization and Extended Tasks}

WSM can admit named nodes outside the fixed 256-channel WSG schema,
provided that the new variable has an aligned, machine-readable representation.
This gives sparse annotations and newly digitized structures the same operational
status as an existing image-like endpoint without renumbering the released dense
graph or retraining the shared generator.

Universal Pose is the first explicit example of this extension path. We render
the category-independent MP-100 landmark annotations as sparse red joints and
white links on a black canvas, register RGB and Universal Pose as named nodes in
a custom task graph, and train one additional TunnelingNet. The final row of
Figure~\ref{fig:gallery_pose_structure} is a 10-step generation result on the evaluation set
from that TunnelingNet. It extends the fixed human-pose representations above it to the
articulated structure of animals and objects while preserving the same
condition--prediction--target interface.

To keep quantitative claims stable as the system grows, extended tasks remain
separate from the frozen 1,148-task aggregate until they pass the same
evaluation protocol. Subsequent additions can be reported in this subsection
with their data source, task definition, selected model, metrics,
and qualitative artifact. The task bank can therefore grow continually while
each released result remains reproducible and verifiable.

\FloatBarrier

%% file: sec_preprint/7_discussion.tex
\wsmnewphysicalpage

\section{Discussion}
\label{sec:discussion}

\subsection{Experimental Evidence}
\label{sec:discussion_evidence}

\paragraph{What our three experimental sections establish.}
Our ablation study shows how TunnelingNet acquires a new visual route. We find
that attention topology, evidence routing, tunnel placement, and high-noise
trajectory supervision all shape how quickly a task-specific TunnelingNet learns its target
representation while preserving the RGB prior. The resulting configuration
produces recognizable structure within tens of updates and reaches up to
approximately $100\times$ greater training efficiency than the
parameter-matched native-attention baseline. Our quantitative study then asks
how widely this recipe holds: it evaluates 1,148 visual tasks, with one
independently deployable TunnelingNet per task, across 39 output groups, four
interfaces, and 18,191 end-to-end inference evaluations. Our qualitative study lets us inspect the visual relationships behind
those numbers across geometry, articulated humans, motion, physical
properties, materials, PBR appearance, segmentation, categorical identity,
and RGB synthesis. Together, our experiments give a consistent picture: a
shared generator can learn a wide vocabulary of World Snapshot transformations
when each route receives focused parameters and clearly routed evidence.

\paragraph{Scale and strength of our baseline.}
We summarize continuous outputs with the higher-is-better quality score QS and
categorical outputs with dominant-class accuracy (DAcc), abbreviated together
as QS/DAcc. Across the 1,147 tasks with a valid primary score, 779 (67.9\%) exceed
$\mathrm{QS/DAcc}=0.8$. Within the 969 continuous tasks, 744 (76.8\%) exceed
$\mathrm{QS}=0.8$, 537 (55.4\%) exceed $\mathrm{QS}=0.9$, and the mean QS is
0.860. Strong results appear across all four interfaces: 37/38 narrow-generation
tasks, 367/449 RGB-assisted inverse-rendering tasks, 312/449 condition-only
control tasks, and 28/33 continuous estimation or reconstruction tasks exceed
0.8. A conservative sensitivity check that removes the two near-constant
output groups flagged by the artifact analysis still gives 666/891 continuous
tasks (74.7\%) above 0.8. These measurements give us confidence that the strong
region extends across hundreds of heterogeneous mappings. The paired matrices
also show a natural dependence on the visual evidence shared by the condition
and target.

\subsection{Unified Task Graph and Model Scale}
\label{sec:discussion_graph_scale}

\paragraph{Why the directed graph matters.}
Each WSM task is a directed path from available evidence to a World Snapshot
target. Moving a variable from the condition side to the target side asks the
model to generate that variable. RGB therefore becomes one node in a much
larger generative domain that also includes geometry, pose, motion, material,
physics, semantics, and audio. This construction makes training naturally
parallel: every edge shares the frozen backbone, while its small tunnel can be
scheduled, optimized, and validated independently. It also makes capabilities
compositional. A generated variable can become evidence for the next edge, so
longer visual chains can be assembled from learned paths.

The complete modular WSM comprises one shared backbone, a resident
collection of trained TunnelingNets and an explicit task router. Independent TunnelingNet
optimization and physical sharding of the TunnelingNet files enable parallel training and modular
distribution without changing this model boundary. All TunnelingNets remain
registered together, while a task key sparsely activates one edge and a visual
chain activates a selected sequence. This resident-versus-active organization
is analogous to sparsely activated expert models. WSM routes with an explicit
visual-task key; conventional expert models commonly learn their gating. We verified that each of the 1,148 frozen dense TunnelingNets
contains exactly 12,510,208 tunnel parameters, giving 14.362B TunnelingNet parameters
in total. Together with the 3.9B visual backbone and 0.282B parameters from the
180 deployed categorical refiners, the complete resident visual parameter bank
is approximately 18.54B parameters. In this complete deployment form, WSM v1
Augustus is an 18B-class model with 1,148 selectable visual tasks.\footnote{This count excludes the shared text encoder and VAE.} A single inference route activates one 12.51M
tunnel, so its active visual parameter count remains approximately 3.913B;
categorical routes add only 1.54--1.60M active refiner parameters. Replicating
the 3.9B backbone once for every evaluated task would already require
approximately 4.48T visual parameters. Sharing one backbone therefore reduces
this task-library parameterization by about $241\times$, while retaining a
dedicated generative route for every task.

\subsection{Mathematical Scalability of the Unified Law}
\label{sec:discussion_scaling}

\begin{figure*}[t]
  \centering
  \includegraphics[width=0.98\textwidth]{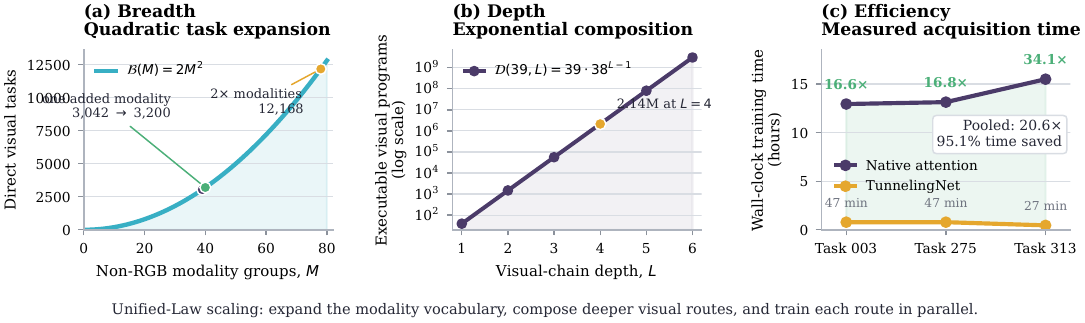}
  \caption{\textbf{Mathematical scalability of the Unified Law.} (a) The four-interface task graph grows as
  $\mathcal{B}(M)=2M^2$; the current 39 non-RGB modality groups define 3,042
  dense mappings. (b) Composing directed routes produces exponentially more
  executable visual programs as chain depth increases. (c) Directly logged
  wall-clock measurements on three matched tasks show $16.6$--$34.1\times$
  faster route acquisition with TunnelingNet, with a pooled $20.6\times$
  speedup and 95.1\% time reduction. The three panels measure complementary
  properties: the width of the visual-task graph, the depth of reasoning over
  that graph, and the cost of instantiating its learned routes.}
  \label{fig:unified_law_scaling}
\end{figure*}

\paragraph{Toward a law-scale foundation for vision.}
Most computer-vision models are built around one task, and broadly capable
systems usually collect a small set of task-specific interfaces. WSM v1 places 1,148 evaluated tasks inside one generative system, while
its released dense task graph configures 3,042 possible mappings. Relative to the conventional scope of one to ten tasks, this expands
the task count by roughly two to three orders of magnitude. It also changes the
unit of progress: we can grow one directed generative graph by adding World
Snapshot variables and routes, and every new route becomes part of the same
model.

The central result therefore reaches beyond any individual benchmark score. A
scientific law is useful when it compresses many apparently separate phenomena
into one practical rule. Our experiments provide large-scale evidence for
such a compression in computer vision: geometry, pose, motion, material,
physics, semantics, audio, and appearance become World Snapshots, and
generation becomes their common computational operation. TunnelingNet makes
this principle trainable in parallel, the directed graph makes it compositional,
and the shared backbone makes thousands of routes deployable as one WSM. This
is the law-scale ambition of the Unified Law for Visual Tasks: to organize a
growing field of visual capabilities through one reusable generative
principle. WSM v1 Augustus is its first large-scale realization, and the 3,042 mappings in its dense task graph provide an immediate path for expansion.

The scalability implied by the Unified Law admits a compact mathematical
form. We characterize it along three measurable axes: the breadth of directly
trained tasks, the depth of composed visual programs, and the cost of acquiring
the learned routes. Together, these expressions formalize how one unified
generative principle expands into a growing visual-task system.

\noindent\textbf{Unified-Law scalability formula: breadth and depth.}
Let $M$ be the number of non-RGB World Snapshot modality groups. The
four-interface dense graph contains
\begin{equation}
\begin{aligned}
\mathcal{B}(M)\equiv T_{\mathrm{direct}}(M)
&= M + M + M(M-1) + M(M-1) \\
&= 2M^2 ,
\end{aligned}
\label{eq:unified_law_task_expansion}
\end{equation}
covering RGB estimation, narrow RGB generation, directed conditional control,
and RGB-assisted inverse rendering. The current $M=39$ configuration gives
exactly 3,042 dense mappings. Once a new modality is aligned frame-by-frame
with the existing World Snapshots, it can appear as condition and target
throughout the graph. For $K$ added modalities, the direct expansion is
\begin{equation}
\Delta_K T
= T(M+K)-T(M)
=4MK+2K^2 .
\label{eq:unified_law_modality_growth}
\end{equation}
At the present scale, one new modality creates 158 direct tasks and ten create
1,760. Doubling the modality vocabulary from 39 to 78 quadruples the dense
task graph from 3,042 to 12,168 mappings.

Composition grows even faster. Starting from RGB, the condition-only graph
already admits
\begin{equation}
\mathcal{D}(M,L)\equiv P_L=M(M-1)^{L-1}
\label{eq:unified_law_chain_growth}
\end{equation}
visual chains of length $L$. For $M=39$, lengths one through four yield 39,
1,482, 56,316, and 2,140,008 possible directed paths. Equations
\ref{eq:unified_law_task_expansion}--\ref{eq:unified_law_chain_growth} form the
breadth--depth expansion rule induced by the Unified Law. The breadth
$\mathcal{B}(M)$ grows quadratically with the modality vocabulary, while the
reasoning depth $\mathcal{D}(M,L)$ grows exponentially with visual-chain
length. These axes reinforce each other: frame-aligned modalities raise the
branching base available at every reasoning step, and longer visual chains
raise its exponent. WSM can therefore expand the variety of visual questions
and the depth of the programs used to answer them at the same time.

The third scaling axis is the cost of acquiring these routes. Let $t_i$ denote
the measured wall-clock training time of task route $i$ and let $G$ be the
number of parallel workers. Because every TunnelingNet shares a frozen backbone
and can be optimized independently, the wall-clock cost of constructing an
$N$-task bank is approximately
\begin{equation}
\mathcal{C}_{\mathrm{train}}(N,G)
\simeq \frac{1}{G}\sum_{i=1}^{N} t_i
\simeq \frac{N\bar{t}}{G}.
\label{eq:unified_law_training_cost}
\end{equation}
This expression makes the practical value of the graph explicit: capability
breadth can grow across tasks while acquisition time decreases almost linearly
with available parallelism. In our controlled wall-clock records for Tasks
003, 275, and 313, native attention required 12h56m32s, 13h08m09s, and
15h29m53s, whereas TunnelingNet required 46m55s, 47m01s, and 27m18s at the
selected milestones. The corresponding speedups are $16.6\times$,
$16.8\times$, and $34.1\times$. Pooled across the three tasks, training falls
from 41.58 to 2.02 GPU-hours, a measured $20.6\times$ acceleration and a
95.1\% reduction in acquisition time. This wall-clock result complements the
$19.0$--$38.5\times$ update-count acceleration and the $77$--$100\times$
earlier emergence of recognizable structure in Sec.~\ref{sec:ablation}.

\subsection{Unifying and Expanding Computer Vision}
\label{sec:discussion_task_generative}

This section tests the central scope claim of the Unified Law. We ask two
questions. First, can the established tasks of computer vision be expressed
within the WSM interface system? Second, after those established tasks are
removed, how many tasks evaluated by WSM define genuinely new visual
problems? The first question measures \emph{unification}; the second measures
\emph{expansion}.

We use Papers with Code as the historical reference because it maintained one
of the broadest public catalogs of computer-vision research, connecting named
tasks with papers, datasets, code, and benchmarks. Its final public records
contain 1,477 visual or vision-related tasks, which we use as a representative
inventory of the established field~\cite{worldsnapshot2025pwcarchive}. For
unification, we examine whether each historical task can be written as a route
through the WSM interfaces. For expansion, we compare the 1,148 tasks evaluated
by WSM v1 with this inventory and count the evaluated tasks that have no
historical counterpart. The complete matching rules and task-level records are
provided in Sec.~\ref{sec:supp_pwc_task_analysis}.

\begin{figure*}[t]
  \centering
  \includegraphics[width=0.98\textwidth]{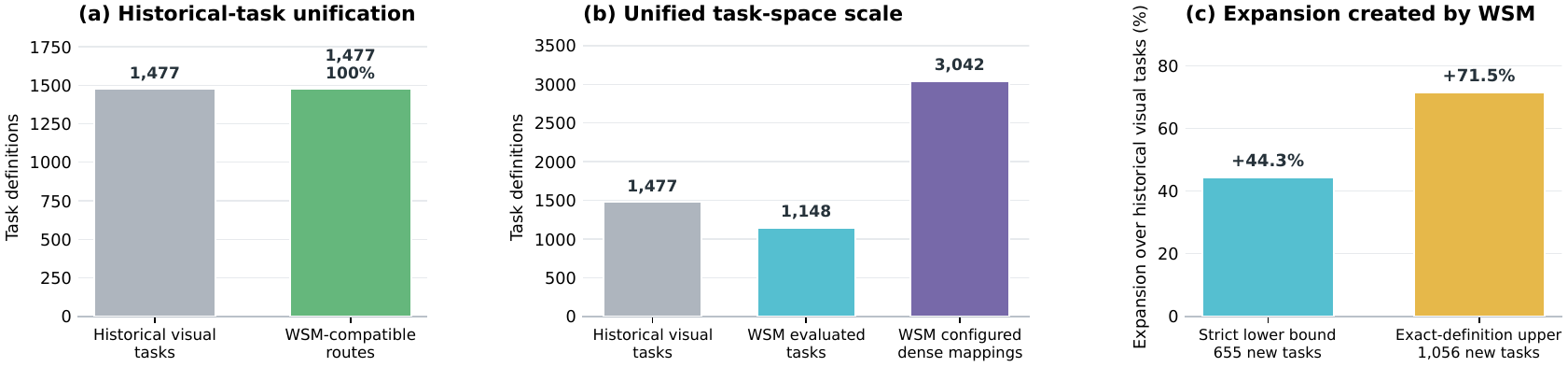}
  \caption{\textbf{WSM unifies and expands computer vision.}
  Papers with Code recorded 1,477 visual or vision-related tasks, providing a
  representative inventory of the existing field. (a) WSM can express all
  1,477 tasks within its unified interfaces. (b) WSM v1 evaluates 1,148 tasks
  within a dense graph of 3,042 configured mappings. (c) After removing overlaps with existing
  tasks, 655--1,056 evaluated WSM tasks are new, expanding the established
  computer-vision task inventory by 44.3--71.5\%. Detailed calculations are
  provided in Sec.~\ref{sec:supp_pwc_task_analysis}.}
  \label{fig:pwc_historical_task_scope}
\end{figure*}

The result is direct. All 1,477 historical visual tasks can be expressed within
the WSM interfaces, giving 100\% compatibility with the representative
historical inventory. After historical overlaps are removed, 655--1,056 of the
1,148 evaluated WSM tasks are new. Relative to the 1,477 established visual
tasks, WSM therefore expands the computer-vision task space by 44.3--71.5\%.
Figure~\ref{fig:pwc_historical_task_scope} presents these two results.

Together, these findings establish both parts of the claim. WSM provides one
generative interface system for the established field of computer vision, and
it substantially enlarges that field with executable task definitions and
trained baselines. Its dense graph already contains 3,042 mappings, and
each newly aligned modality creates further families of estimation, generation,
control, and reconstruction tasks. We call this direction
\emph{Task-Generative Computer Vision}: a visual framework that unifies the
existing task space and systematically creates new tasks beyond it.

\noindent\textbf{Position and interpretation of evidence.}
WSM makes three complementary claims, each supported at its appropriate
experimental scale. First, WSG and the four interfaces provide one operational
rule for expressing heterogeneous visual computations as directed generation.
Second, WSM v1 turns that rule into an executable system: 1,148 distinct
input--output tasks have their own trained TunnelingNets and end-to-end inference
results within one shared-backbone model.
Here, \emph{unified} refers to the shared representation, generative backbone,
route-acquisition rule and router. Independently
optimized TunnelingNets remain sparsely activated components within that model, not
replications of the full backbone.
Third, the historical analysis shows that the four interfaces cover 1,477
historical visual tasks, while the 1,148 figure separately measures tasks that
were actually trained and evaluated. Keeping these quantities distinct makes
the breadth claim directly auditable.

The validation scale was deliberately increased from tens of representative
routes, to hundreds of cross-family mappings, and finally to 1,148 frozen
tasks. Prior generalist vision systems commonly remain at tens of tasks, and
reports beyond one hundred are rare. WSM therefore moves the empirical frontier
to 1,148 separately trained and evaluated tasks: more than an order of
magnitude beyond even a 100-task system, $29$--$38\times$ beyond a
30--40-task system, and two to three orders beyond the conventional one-to-ten
task setting. These levels answer different questions: matched MTA--NJA ablations
isolate TunnelingNet acquisition behavior; QS/DAcc and 18,191 generated outputs
measure system-wide coverage; task-native metrics establish target-specific
quality; and AVIS, post-release imagery, and independent public annotations
establish transfer after acquisition. In this last setting, \emph{zero-shot}
and \emph{training-free} mean that a route is trained once and then applied to
unseen datasets and annotation conventions without dataset-specific
retraining. Likewise, the reported budget ratios measure marginal task
acquisition after the respective pretrained priors, whereas the matched
attention ablations isolate the contribution of TunnelingNet itself. This
layered protocol evaluates scale without equating coverage with identical
ceiling accuracy: the complete score distributions and disclosed failure
regions identify which TunnelingNets merit further improvement while preserving the
system-level evidence that one method can acquire, register, and execute more
than one thousand visual tasks.

Finally, the breadth and chain equations describe the tasks available in the graph.
An input--output mapping is a distinct task because changing either its available evidence
or its target changes the learned conditional distribution; opposite
directions are therefore different visual problems. The chain count measures
the compositional paths made available by trained edges, while the directly
reported claims remain anchored to trained TunnelingNets and observed outputs.
Together, these definitions separate demonstrated execution,
cross-dataset transfer, interface-level scope, and future composition without
weakening the central result: WSM is the first realization of this principle at
thousand-task scale.

\subsection{Formal Scope and Falsifiability of the Unified Law}
\label{pre:sec:formal_unified_law}

\noindent\textbf{Why a law, and why vision now?}
Empirical laws in artificial intelligence capture reproducible regularities over defined regimes. The canonical study \emph{Scaling Laws for Neural Language Models}~\cite{kaplan2020scalinglawsneurallanguage} examines cross-entropy under one autoregressive language-modeling
objective: the models are trained on WebText2, and the analysis establishes
predictive power laws over model size, dataset size, and compute, with trends
spanning six to eight orders of magnitude. Its importance does not come from
separately validating every NLP task, but from discovering a stable empirical
relation that compresses observations and predicts behavior outside individual
runs. In task diversity, that evidence centers on one learning objective; WSM
instead realizes one visual organizing principle through 1,148 separately trained
visual tasks, 39 output groups, four interfaces, and 18,191 generated outputs.
On a count of learning objectives,
one is less than 0.1\% of 1,148. The two laws are validated along different scientific axes:
a scaling law is tested by depth along model, data, and compute scales, whereas
a unifying law is tested by breadth across heterogeneous visual tasks. The
breadth of 1,148 tasks directly supports the unification claim. The broader
scientific precedent is clear: an empirical law in AI is established by a
reproducible regularity over a defined regime.

\noindent\textbf{The law and its empirical realization.}
The Unified Law has two deliberately separated levels. The law itself is a
representation statement: once the information required by a visual task can
cross the model boundary without losing a task-relevant distinction, the task
can be written as conditional generation of requested world fields from known
world fields. WSM and TunnelingNet provide an empirical realization of this
statement: a shared generator can acquire many such directed relations through
small, independently installed routes. The first level does not depend on a
particular parameterization; the measured parameter counts and training times
belong to the second.

Formally, let $T$ map known fields $C$ to requested fields $Y$ under data
distribution $\mathcal{D}_T$, let $s=(s_C,s_Y)$ be a fixed serialization, and
write $y_a\sim_T y_b$ when two targets are equivalent under the task's native
answer and evaluation. Define the \emph{preservation condition}
$\pi(T,s)\in\{0,1\}$ by
\begin{equation}
\begin{aligned}
\pi(T,s)=1 \quad\Longleftrightarrow\quad
&s_C(c_a)=s_C(c_b)\\
&\quad\Rightarrow p_T(Y\mid c_a)=p_T(Y\mid c_b),\\
&s_Y(y_a)=s_Y(y_b)\\
&\quad\Rightarrow y_a\sim_T y_b.
\end{aligned}
\label{pre:eq:preservation_condition}
\end{equation}
Thus the condition code preserves every input distinction that can change the
correct answer distribution, and the target code preserves every distinction
that the task evaluates. For a model that observes only this fixed
serialization, $\pi=0$ is an information boundary: no later architecture,
objective, or amount of training can recover a distinction that $s$ has
already erased. A different serialization may restore it; the impossibility
claim is therefore about the declared $s$, not about the task in isolation.

Let $\kappa(T,G,s)\in\{0,1\}$ denote the separate \emph{coverage condition}:
$G$ can consume $s_C(C)$, emit and decode $s_Y(Y)$, and its pretrained target
prior has support for valid structures in that serialized target domain.
Interface compatibility and target knowledge are both required: placing a
field in an independently trained latent space may preserve it perfectly while
still giving the frozen generator no meaning for its coordinates. When
$\pi\kappa=1$, the original task has the generative reduction
\begin{equation}
\hat z_Y\sim p_{\theta}\!\left(z_Y\mid s_C(C),t\right),
\qquad \hat Y=d_Y(\hat z_Y),
\label{pre:eq:unified_law_reduction}
\end{equation}
where $t$ is optional text and $d_Y$ is the declared target decoder. Solving
this serialized conditional-generation problem solves $T$ up to $\sim_T$.
This is the formal content of the Unified Law; generation, estimation,
control, reconstruction, and cross-property translation differ in the fields
placed on the two sides of Eq.~\ref{pre:eq:unified_law_reduction}.

\noindent\textbf{Backbone scope and learning prediction.}
The known fields $C$, requested fields $Y$, and directed relation
$T:C\!\rightarrow\!Y$ describe the task's basic behavior. They do not change
when a model uses autoregression, diffusion, flow matching, or another
prediction mechanism. A backbone $G$ enters through target coverage and the
measured quality and acquisition cost. For another backbone that can consume
the condition and generate the target representation, the same comparison
applies, although its absolute quality, cost, and best route may change.

Let $Q_T(B;G,s)$ be native task quality under acquisition budget $B$, and let
$\mathcal{C}_T(q;G,s)$ be the cost of reaching comparison quality $q$. If
$s^{+}$ preserves more task-relevant information and better matches target
knowledge already present in $G$ than $s^{-}$, the law predicts
\begin{equation}
\begin{aligned}
Q_T(B;G,s^{+})&\gtrsim Q_T(B;G,s^{-}),\\
\mathcal{C}_T(q;G,s^{+})&\lesssim \mathcal{C}_T(q;G,s^{-}).
\end{aligned}
\label{pre:eq:comparative_learning_prediction}
\end{equation}
At fixed budget the aligned design should learn more; at matched quality it
should cost less. The comparison keeps the task, generator, data, and evaluator
fixed, and each task retains its native quality measure. The size of the
advantage varies with target abstraction, information missing from the
condition, and target knowledge already present in the backbone. The modality
experiments and their plain-language interpretation appear in
Sec.~\ref{pre:subsec:memoir_modalities}.

\noindent\textbf{Three complementary levels of evidence.}
Section~\ref{subsec:three_level_law_validation} supports the law at three
levels. The first asks whether a rule formed from four initial Tunnels still
works when applied at scale; it yields 1,148 trained and evaluated tasks. The
second asks whether these routes follow each individual input; deliberately
pairing their fixed predictions with the wrong targets increases error in every
interface. The third asks why representation choice changes learning
difficulty. We write the same surface-normal target in five lossless coordinate
systems. All five reconstruct exactly the same image, while only the native VAE
coordinates use the organization learned by the frozen generator during
pretraining. Native coordinates reduce checkpoint-averaged angular error by
20.1\%--26.6\%, reach the lowest final error, and win 59 of 64 final
sample--transformation comparisons.

The result gives Eq.~\ref{pre:eq:comparative_learning_prediction} a direct
experimental meaning: complete information is necessary, and a representation
already understood by the backbone makes that information substantially easier
to learn. The law therefore provides both a unified description of visual tasks
and a practical rule for selecting their representations. Native visual
coordinates are the first choice for dense visual targets; sparse and abstract
targets call for a visual reorganization or a suitable output boundary.

The WSM realization additionally installs each route independently around
frozen shared parameters. Dense TunnelingNets use a 12.5M-parameter target-side
tunnel in the reported configuration, while sparse and categorical targets add
their declared native output components. Complete inference through the native
target interface measures $Q_T$; a private codec round trip or partial preview
does not establish output quality.

\noindent\textbf{Formal consequences of the experiments.}
The preservation and coverage conditions above summarize a larger
parameterized knowledge system derived in
Sec.~\ref{pre:subsec:memoir_parameterized_knowledge}. Four consequences are
especially important for interpreting the law.

First, representational accessibility and operational mastery are different
quantities. For field $X_i$, define accessibility at tolerance $\epsilon_i$ as
\begin{equation}
A_i(\epsilon_i)=\mathbf{1}\!\left[r_i\leq\epsilon_i\right],
\end{equation}
and summarize operational mastery over a declared task set $\mathcal{T}_i$ by
the task-quality profile
\begin{equation}
\mathbf{M}_i(\mathcal{T}_i)=
\left(Q_T\right)_{T\in\mathcal{T}_i},
\end{equation}
where $X_i$ participates as a condition, target, or intermediate field in
one-to-one, one-to-many, many-to-one, or composed conversions. Accessibility is
the first requirement; operational mastery adds measured relations with other
fields. An accurate round trip does not imply high $Q_T$. The full
experimental interpretation is given in
Sec.~\ref{pre:subsec:memoir_modalities}, especially
the distinction stated at
page~\pageref{pre:par:memoir_accessibility_mastery}.

This distinction includes the earlier observation that reconstruction and
compatibility are different quantities. The private
round-trip error $r_i$ of a codec can approach zero while the directed
cross-field error $e_{i\rightarrow j}$ remains high; equal dimensions and
matched marginal statistics do not make two latent coordinates synonymous
(Eq.~\ref{pre:eq:memoir_modal_compatibility}). Compatibility requires the
preservation condition plus a sample-level relation and usable target
semantics. The abstraction discarded by a codec is measured by
$\alpha(Z;X)=H(X\mid Z)$, while the unresolved information in a directed task is
$\mu(C\!\rightarrow\!Y\mid t)=H(Y\mid C,t)$. These quantities explain why the
useful objective changes from continuous reconstruction toward symbolic
prediction as abstraction increases, and why reversing condition and target
creates a different task. Text, noise, another field, or a pretrained prior
must supply choices not determined by the condition
(Eqs.~\ref{pre:eq:memoir_abstraction}--\ref{pre:eq:memoir_text_gain}).

Second, representation design determines which distinctions survive. For a
sparse field with occupied fraction $\rho(Y)$, an empty image can have pixel
loss $\mathcal{O}(\rho(Y))$ while missing every task-relevant object. Dense
image codecs and sparse structured boundaries are therefore not interchangeable
(Eq.~\ref{pre:eq:memoir_sparse_background}). After information survives the
boundary, path geometry and target knowledge remain separate resources: a
shorter or straighter path does not imply a better decoded target when the
frozen generator lacks support for valid target structures
(Eq.~\ref{pre:eq:memoir_path_knowledge}).

Third, scalability is defined through marginal installation and terminal
behavior. The interference quantity $\Delta Q_{i\leftarrow j}$ measures whether adding
route $j$ changes installed task $i$; independent installation requires it to
remain zero within evaluation tolerance. Resident parameters grow with the
number of trained routes. Each request activates task-specific parameters from
only the selected route
(Eqs.~\ref{pre:eq:memoir_interference}--\ref{pre:eq:memoir_resident_active}).
Fourth, decreasing a surrogate training loss does not imply that the complete
decoded output improves in its native task space
(Eq.~\ref{pre:eq:memoir_terminal_success}). These distinctions turn the
Unified Law from a verbal analogy into a sequence of measurable claims about
information, direction, prior knowledge, interference, cost, and final output.

\noindent\textbf{Experimental origin of the conditions.}
The experiments above led to two law conditions and one installation
criterion. Preservation arose from
sparse-pixel, semantic-color, shared-VAE, and
object-structure failures; Sec.~\ref{pre:subsec:memoir_preservation} gives the
information argument, and Sec.~\ref{pre:subsec:memoir_modalities} explains how
abstraction and direction change the useful learning form. Coverage arose when
private latent spaces reconstructed themselves yet remained unreadable to a
frozen generator, and when optimal transport, reflow, and stronger bridges
changed the path without supplying target knowledge. Independent installation
arose when joint lanes, fused conditions, packed LoRAs, and shared-target LoRAs
coupled otherwise distinct capabilities. The complete chronological account is
given in Sec.~\ref{pre:subsec:memoir_experimental_logic};
Tables~\ref{tab:bridge_variant_record} and
\ref{tab:bridge_variant_level_record} record the corresponding families
and representative variants. Table~\ref{pre:tab:formal_family_traceability} gives the
family-level map, while Table~\ref{pre:tab:formal_variant_traceability} lists
the 198 currently indexed named variants separately with their direct observations and formal
links. Section~\ref{pre:subsec:memoir_summary} summarizes the full record as
eight core experimental principles.

These boundaries were recovered across 60 experimental families, 198 named
variants, and at least 2,293 documented training attempts, then applied in WSM v1.
All 1,148 reported tasks were trained and evaluated through the same task-keyed
acquisition framework, with interface-specific components declared explicitly.
The matched MTA--NJA experiments are mechanism-level evidence within this
realization---$19.0$--$38.5\times$ faster structured acquisition under MTA and
an 11.0\% lower RGB error under NJA---not part of the law's definition.
Likewise, the independent-annotation evaluation tests whether acquired routes
transfer beyond the annotation convention without retraining.

\phantomsection\label{pre:par:unified_law_refutation}
\noindent\textbf{Refutation criteria.}
The representational statement would fail if a task satisfying the declared
preservation and coverage conditions could not be expressed and evaluated
through Eq.~\ref{pre:eq:unified_law_reduction}. The learning prediction would
fail if a matched comparison reliably reversed the ordering in
Eq.~\ref{pre:eq:comparative_learning_prediction}. Independent installation would fail if
adding a route degraded an existing route. Its claimed breadth would fail if
such counterexamples accumulated throughout the stated visual regime. A failure
at a disclosed boundary---erased information,
an unreadable target space, an uncovered target prior, or absent paired
supervision identifies which premise is missing; it does not silently redefine
the task as a successful case. These criteria make the law, its empirical
realization, and the limits of WSM v1 separately testable.

Computer vision is ready for an organizing law of its own. After decades of
progress measured mainly as separate models and benchmarks, the field needs a
principle that explains how those tasks relate and predicts how new ones can be
constructed. Scaling laws describe how performance changes as resources grow;
the Unified Law addresses what can be scaled---the modular visual capability space itself. The four interfaces make a concrete and refutable prediction: once task-relevant information is encoded as World Snapshots, heterogeneous visual computations and newly aligned visual information can be incorporated into the same generative formulation. WSM v1 provides the first thousand-task realization and unusually broad evidence for this principle. More broadly, WSM and LLMs can serve as peer foundations. An LLM unifies language and abstract reasoning; WSM unifies perception, generation, reconstruction, and control over explicit visual world states. Coupling these foundations could lead toward a general-purpose multimodal foundation model for universal task execution, in which linguistic intelligence and visual world intelligence operate as coequal, mutually conditioning capabilities. By these criteria, the Unified Law is a testable law-scale hypothesis supported by the first thousand-task realization. It opens a path from task-specific models toward general-purpose visual world intelligence.

\FloatBarrier

%% file: sec_preprint/8_conclusion.tex
\clearpage

\section{Conclusion}
\label{sec:conclusion}

We presented the Unified Law for Visual Tasks, which expresses visual tasks as
transformations among image-like World Snapshots. World Snapshot Graphics
encodes RGB and auxiliary attributes in a representation with up to 256 channels, while
four interfaces express generation, estimation, conditional control, and
reconstruction in one generative formulation. Together, these interfaces and
TunnelingNet realize 1,148 evaluated visual tasks within a dense graph of 3,042
configured mappings. We validate both the method and the Unified Law at this scale, establishing
their effectiveness across more than one thousand tasks.

Our experiments show that TunnelingNet preserves the pretrained RGB-synthesis
prior while efficiently acquiring structured visual transformations, reaching
up to approximately $100\times$ greater training efficiency than the
parameter-matched native-attention baseline. Across all evaluated tasks,
WSM also achieves consistently strong performance, surpassing specialized prior
methods---including \textcolor{red}{Best Paper} and \textcolor{red}{Best Paper Candidate} systems---on the
reported metrics. The mathematical scalability of
the Unified Law further connects model growth to task growth: aligned
modalities expand direct-task breadth quadratically, visual chains expand
compositional depth exponentially, and independent task routes can be trained
in parallel. These routes are registered parameters of one modular WSM and are
sparsely activated by an explicit task key; independent optimization and
physical sharding do not make them standalone models. WSM therefore provides one extensible generative foundation for a
large and continually growing vocabulary of visual tasks. Its longer-term
trajectory is to unify, expand, and ultimately transcend the fixed task
boundaries of conventional Computer Vision.

In addition, we compare WSM with the visual tasks established before this work
and obtain two findings. First, among the tasks recorded by Papers with Code,
1,477 are visual or vision-related tasks, broadly representing the vast
established field of Computer Vision. All 1,477 tasks are compatible with the WSM interfaces and can be
expressed by the same system. \textbf{After half a century of development, we bring the
vast field of Computer Vision into a unified generative framework.}
Second, \textbf{WSM creates 655--1,056 entirely new visual tasks and expands
Computer Vision by 44.3--71.5\%.} Every evaluated task is accompanied by a
trained baseline. Together, these results establish WSM as a task-generative
foundation that consolidates the historical field and opens a substantially
larger visual-task space.

\clearpage

%% file: sec_preprint/X_suppl.tex
\clearpage
\maketitlesupplementary

\section{Appendix: Implementation Details}
\label{sec:supp_implementation}

\textbf{Task definitions.} Two JSON files define the dense and sparse tasks.
Each task specifies its TunnelingNet index, task type, input modalities, target
modality, implementation type, and the availability of paired training data. The
dense file configures 3,042 visual task mappings around one shared
generative backbone, while the sparse graph contains 11 additional digitized
mappings. Sparse TunnelingNets are reported separately in this appendix and are not included
in the main-paper count of 1,148 tasks. The frozen benchmark provides checkpoints and
complete results on the evaluation set for 1,148 dense TunnelingNets. The
remaining mappings use the same input--output format and define the next set of
tasks to train and evaluate. The training program reads these definitions,
prioritizes selected task types, and trains each TunnelingNet on one GPU.

\textbf{Dense WSG path.} Dense channels are trained from latent caches. For each
TunnelingNet, the launcher passes the target modality, condition modality list,
TunnelingNet index, and task type into the TunnelingNet trainer. The dataset intersects
target and condition sample IDs, applies optional dataset-family whitelists, and
builds aligned tuples of target latent, condition latent, reference latent,
caption embedding, frame index, video id, and source metadata. Reference routing
is task dependent. The selected text-guided narrow-generation recipe retains NJA
and routes first-frame RGB through the native appearance-reference slot.
Inverse rendering routes RGB as an extra condition. Audio-conditioned dense masks append projected AST audio
features as an additional condition lane.

\textbf{Task recipes.} The implementation resolves a recipe from the task type.
Estimation/reconstruction, conditional control, and inverse rendering use
Modality Tunneling Attention with a trainable target lane and frozen condition
lanes. Text-guided narrow generation uses Native Joint Attention; in the
selected recipe, first-frame RGB enters the native appearance-reference slot.
Each experiment saves its final attention mode, trainable lanes, branches,
normalization rules, optimization and validation settings, and TunnelingNet
metadata in \texttt{resolved\_runtime\_config.json}.

\textbf{Optimization and monitoring.} Dense TunnelingNets are optimized with AdamW on
the trainable tunnel parameters only. Training uses bfloat16, gradient
checkpointing when requested, gradient clipping, fixed validation timestep grids,
validation-loss monitoring, and early stopping based on validation improvement.
Preview generation decodes the
predicted target, the condition, and the ground truth through the VAE. For WSG
modalities, the preview path writes decoded values back to a 256-channel WSG
array and invokes the WSG reader so that depth, normal, pose, mask, material, and
audio-related targets are inspected in their native visualization format.

\textbf{Normalization and reversibility.} Each WSG config carries a portable
set of normalization rules with channel ranges and per-modality modes. Training uses
normalized tensors, while visualization and export invert these rules
before invoking WSG readers. This keeps the learning problem numerically stable
and preserves the original physical or semantic units at the interface boundary.

\textbf{Sparse WSG path.} Sparse TunnelingNets follow the same input--output task definition but
use task-specific adapters before returning to the WSG convention. The sparse
launcher resolves entries from the sparse JSON graph and dispatches by
\texttt{adapter\_task}. Human structure TunnelingNets build RGB-crop datasets and cache
PARE features for SMPL/SMPLer-X, DWPose, and MediaPipe targets. The bounding-box
TunnelingNet uses full RGB images with visibility and area filters. The object-structure
TunnelingNet converts WSG object geometry into normalized point clouds and predicts a
breadth-first skeleton tree with parent slots and relative or absolute
coordinates. Audio TunnelingNets build frame-aligned audio features from WSG audio
channels; the visual-to-audio branch can adapt MusicGen with image-prefix
conditioning through the same compact-tunnel principle. The sparse system therefore handles lists, boxes,
keypoints, skeletons, body parameters, and audio tokens as digitized targets in
the same directed WSG graph.

\textbf{Parallel training.} Dense TunnelingNets can be trained independently:
the scheduler assigns them to available GPUs, uses one process per TunnelingNet, saves
each experiment in \texttt{train\_log/Channel\_Head\_xxxxxx}, and resumes or skips
completed TunnelingNets. Sparse training uses the same TunnelingNet indexing scheme.
Each adapter owns its data builder, cache, and trainer. Dense WSG targets exploit
the shared rectified-flow generator; sparse digitized targets use compact
supervised adapters when their native output space is structured rather than
image-like.

\begin{figure*}[!t]
    \centering
    \includegraphics[width=0.82\textwidth]{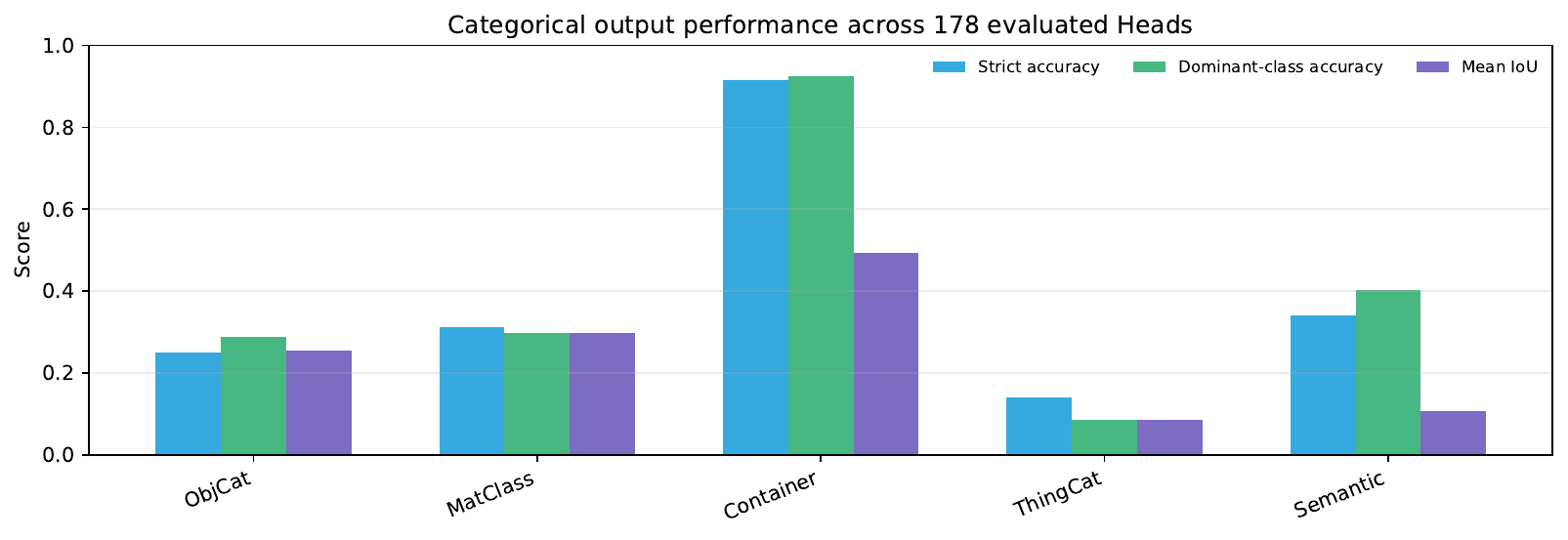}
    \caption{\textbf{Categorical output performance across 178 complete TunnelingNets.}
    Bars report the final composed output for strict pixel accuracy,
    dominant-class accuracy, and mean IoU across all five semantic categorical
    targets.}
    \label{fig:categorical_output_metrics}
\end{figure*}

\begin{figure}[!b]
    \centering
    \includegraphics[width=\columnwidth]{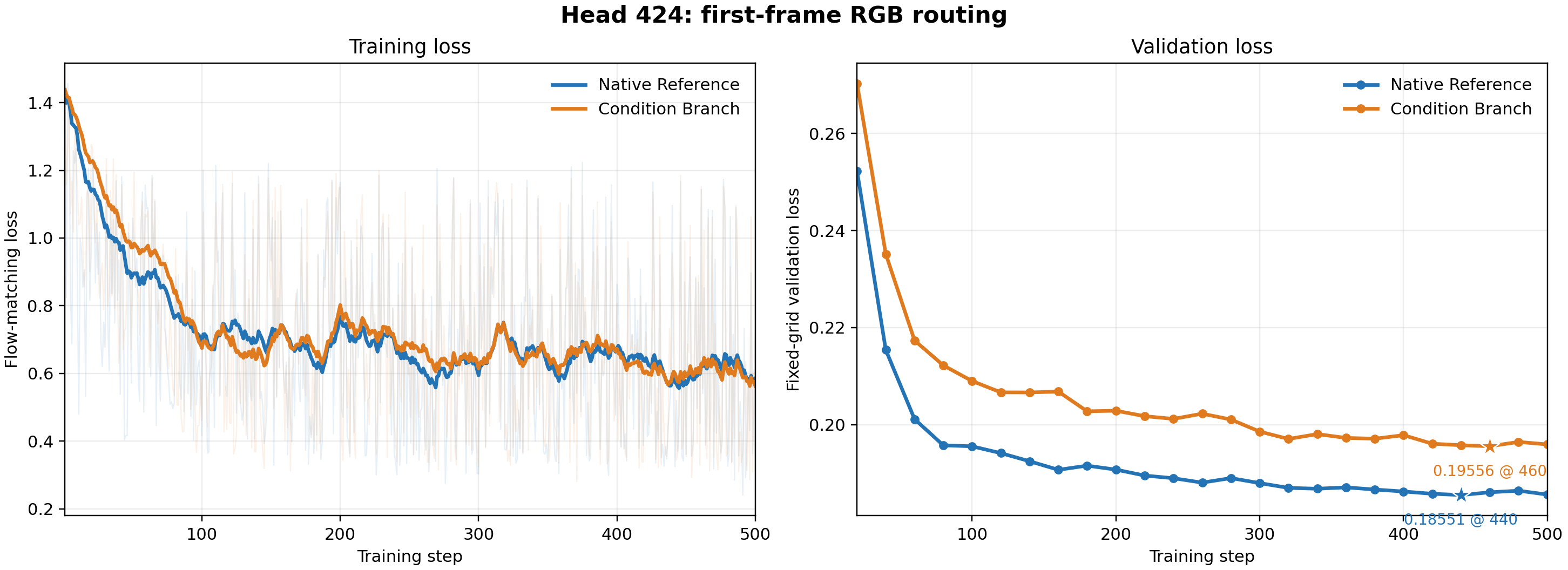}
    \caption{\textbf{First-frame RGB routing for Edge-to-RGB generation.}
    TunnelingNet 424 routes the same first-frame RGB through either the native FLUX
    reference branch or the WSM first-frame condition branch. The two routes
    follow similar optimization trajectories. At update 500, native-reference
    routing obtains a validation loss of 0.1856, compared with 0.1959 for the
    condition branch, while their smoothed training losses are 0.5838 and
    0.5633, respectively. The recommended TunnelingNet 424 therefore uses the native
    reference route as its final training setting.}
    \label{fig:head424_reference_routing_loss}
\end{figure}

\begin{table}[t]
    \centering
    \caption{\textbf{Final categorical output metrics by target.}
    Strict accuracy, dominant-class accuracy (DAcc), and mIoU are measured on
    all evaluated categorical TunnelingNets.}
    \label{tab:categorical_output_targets}
    \footnotesize
    \resizebox{\columnwidth}{!}{%
        \input{tables/evaluation_categorical_output_by_target_normv2}    }
\end{table}

\begin{table*}[!t]
    \centering
    \caption{\textbf{Evaluation of the seven default native sparse TunnelingNets.} Each TunnelingNet is evaluated in its native output space, so its metrics follow the predicted structure. All values use the selected model for that TunnelingNet.}
    \label{tab:native_sparse_heads}
    \small
    \setlength{\tabcolsep}{4.5pt}
    \begin{tabular}{c l l l}
        \toprule
        TunnelingNet & Mapping & Native metrics & Result \tabularnewline
        \midrule
        000000 & RGB crop $\rightarrow$ SMPL pose and shape & SMPL shape-coefficient MAE $\downarrow$ & 0.0373 \tabularnewline
        000001 & RGB $\rightarrow$ person boxes & Precision / Recall / F1 $\uparrow$ & 0.820 / 0.770 / 0.794 \tabularnewline
        000002 & RGB crop $\rightarrow$ DWPose & Median XY error $\downarrow$ / PCK@0.10 $\uparrow$ / PCK@0.20 $\uparrow$ & 0.0925 / 0.526 / 0.725 \tabularnewline
        000003 & RGB crop $\rightarrow$ MediaPipe pose & Median XY error $\downarrow$ / PCK@0.10 $\uparrow$ / PCK@0.20 $\uparrow$ & 0.0852 / 0.568 / 0.797 \tabularnewline
        000005 & Audio + RGB $\rightarrow$ source box & Mean IoU / Macro-category IoU / IoU@0.5 $\uparrow$ & 0.325 / 0.325 / 0.351 \tabularnewline
        000006 & Audio $\rightarrow$ semantic category & Top-1 / Balanced accuracy $\uparrow$ & 0.671 / 0.602 \tabularnewline
        000010 & RGB + Audio $\rightarrow$ sounding regions & Region-selection F1 / Semantic-category accuracy $\uparrow$ & 0.768 / 0.796 \tabularnewline
        \bottomrule
    \end{tabular}
\end{table*}

\begin{figure*}[t]
    \centering
    \includegraphics[width=0.90\textwidth]{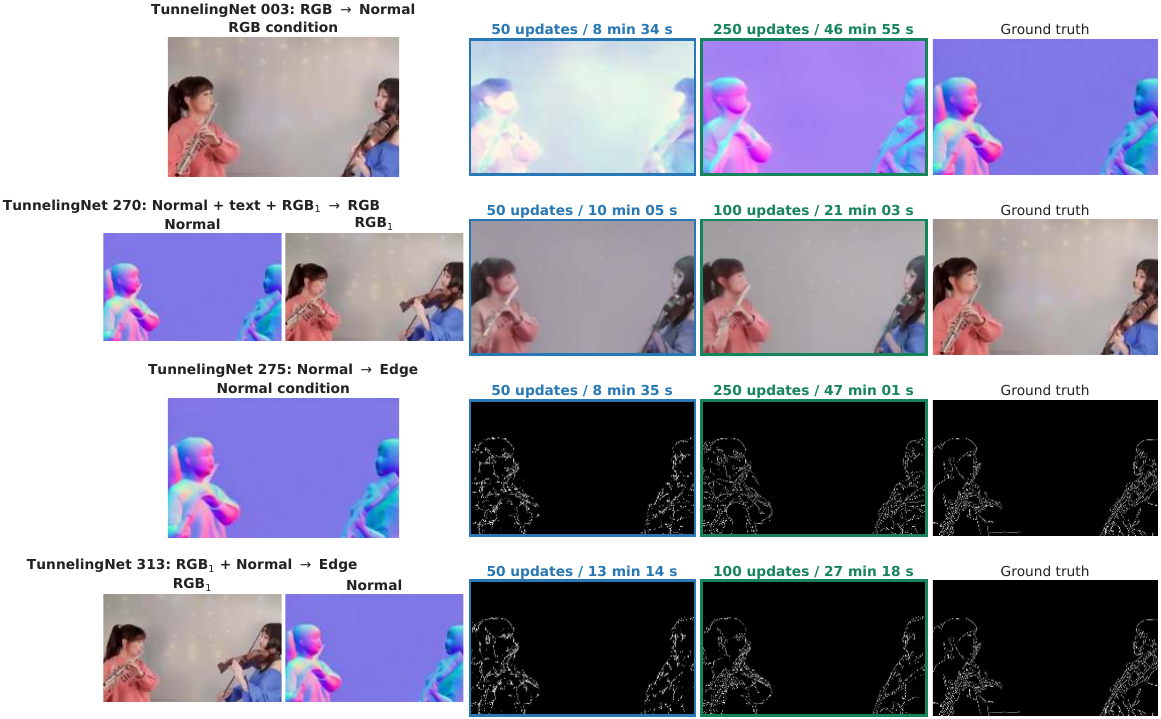}
    \caption{\textbf{Extended minute-scale acquisition evidence.}
    Each row shows the condition, the first saved MTA output at update 50,
    a later selected output, and the ground truth.
    With batch size one, update 50 corresponds to only 50 sampled examples and
    is reached in 8 minutes 34 seconds, 10 minutes 5 seconds, 8 minutes
    35 seconds, and 13 minutes 14 seconds for TunnelingNets 003, 270, 275, and 313.
    The selected 250-, 100-, 250-, and 100-update outputs require 46 minutes
    55 seconds, 21 minutes 3 seconds, 47 minutes 1 second, and 27 minutes
    18 seconds, respectively. Blue and green borders mark the first saved
    and selected outputs.}
    \label{fig:minute_scale_detailed}
\end{figure*}

\section{Appendix: Extended Experimental Analysis}
\label{sec:supp_extended_experiments}

\subsection{Ablations and Categorical Results}
\label{subsec:supp_categorical_limitations}

\paragraph{Categorical output protocol.}
Categorical targets expose a stricter requirement than continuous channels:
the generated support must be spatially correct and every supported region must
also receive the correct semantic identity. The classification branch is
therefore an integral component of each categorical TunnelingNet. We evaluate
only the final composed prediction---generated spatial support followed by
category assignment---rather than treating the intermediate pseudo-color field
as a separate model variant. Across five categorical targets, this evaluation covers
178 complete TunnelingNets and 2,374 evaluation outputs.

\paragraph{Additional RGB-routing ablation.}
Figure~\ref{fig:head424_reference_routing_loss} complements the main-paper
routing study with TunnelingNet 424, which generates RGB from text, Edge, and
first-frame RGB. We isolate how the first-frame RGB enters the model while
holding the 500-update training budget fixed.

\paragraph{Aggregate categorical performance.}
Figure~\ref{fig:categorical_output_metrics} and
Table~\ref{tab:categorical_output_targets} report the complete categorical
results. The binary container target is already reliable, reaching 0.915 strict
accuracy and 0.925 dominant-class accuracy. Object category and material class obtain mIoU values of 0.255 and
0.298. Fine-grained object-class and semantic-segmentation
outputs remain harder, with mIoU values of 0.085 and 0.107. Strict accuracy,
dominant-class accuracy, and mIoU are reported together because they measure
overall pixel agreement, the identity of the principal semantic region, and
both missed and spurious semantic support, respectively.

\paragraph{Errors introduced by categorical representations.}
The qualitative artifact analysis applies an additional, representation-level
filter. We omit MatRegion from qualitative panels because its colors encode
per-object local region IDs without cross-sample semantic identity.

\paragraph{Categorical accuracy across input modalities.}
End-to-end categorical reliability is not yet uniformly strong and stable across
114 cross-modal mappings.
This set is not a list of failed models: several mappings already provide useful
or strong predictions, but their reliability remains sensitive to the semantic
evidence available in the input. It contains 57 conditional-control and 57
RGB-assisted inverse-rendering TunnelingNets, with 38 mappings for each of Targets 111,
131, and 136.

\paragraph{Outputs that need further improvement.}
Targets 111, 131, and 136 require more robust separation of closely
related categories across heterogeneous input modalities. Target 195 is retained
in the reported categorical results after the region-handling correction, while
Target 103 is evaluated as a dense human-segmentation map rather than as a
categorical-refiner output. Future work will prioritize the three withheld target
families through better-balanced classification, stronger region-level evidence,
and systematic re-evaluation under the same evaluation protocol.

% \begin{table*}[t]
%     \centering
%     \caption{\textbf{Categorical TunnelingNet inventory prioritized for further improvement.}
%     The table lists the reviewed conditional-control and RGB-assisted
%     inverse-rendering TunnelingNets for each categorical target. Inclusion denotes that
%     a mapping is not yet consistently strong enough for the current release; it
%     does not imply that every listed TunnelingNet fails.}
%     \label{tab:categorical_upgrade_inventory}
%     \scriptsize
%     \input{tables/evaluation_categorical_upgrade_inventory}
% \end{table*}
%
% \FloatBarrier

\subsection{More Results}
\label{subsec:supp_more_results}

\paragraph{Minute-Scale Rapid Learning.}
\label{subsec:extended_minute_scale}

The compact main-paper sequence isolates learning during the first 10--50 updates
trajectory and its 200-update endpoint. Figure~\ref{fig:minute_scale_detailed} retains the complementary
evidence used during model development: the input condition, the first
saved MTA output, a later selected model, and the ground truth.
Because batch size is one, update count is also the number of sampled training
examples. A TunnelingNet develops visible task structure after only tens of
examples and reaches the selected quality within at most 250 examples and
47 minutes. Qualitative change was observed interactively at approximately
five minutes, but no decoded checkpoint was written at that time; the figure
therefore reports only saved outputs and measured wall-clock times.

\paragraph{Sparse-task results.}
The default sparse launcher selects seven verified TunnelingNets: SMPL pose and shape,
person boxes, DWPose, MediaPipe pose, sound-source localization, audio category
recognition, and audio-selected semantic regions.
Table~\ref{tab:native_sparse_heads} evaluates each output using its native
geometry or classification metric. Optional object and visual-to-audio
experiments are not included in this default set. Since the sparse and custom
graphs are compact, each validated TunnelingNet stores the path of its selected model in
its task entry; the 1,148 dense TunnelingNets use a separate model list.

\FloatBarrier

\subsection{Capability Showpiece: Speedrunning Celebrated Visual Systems}
\label{subsec:custom_capability_showpiece}
\begin{figure*}[t]
    \centering
    \includegraphics[width=0.98\textwidth]{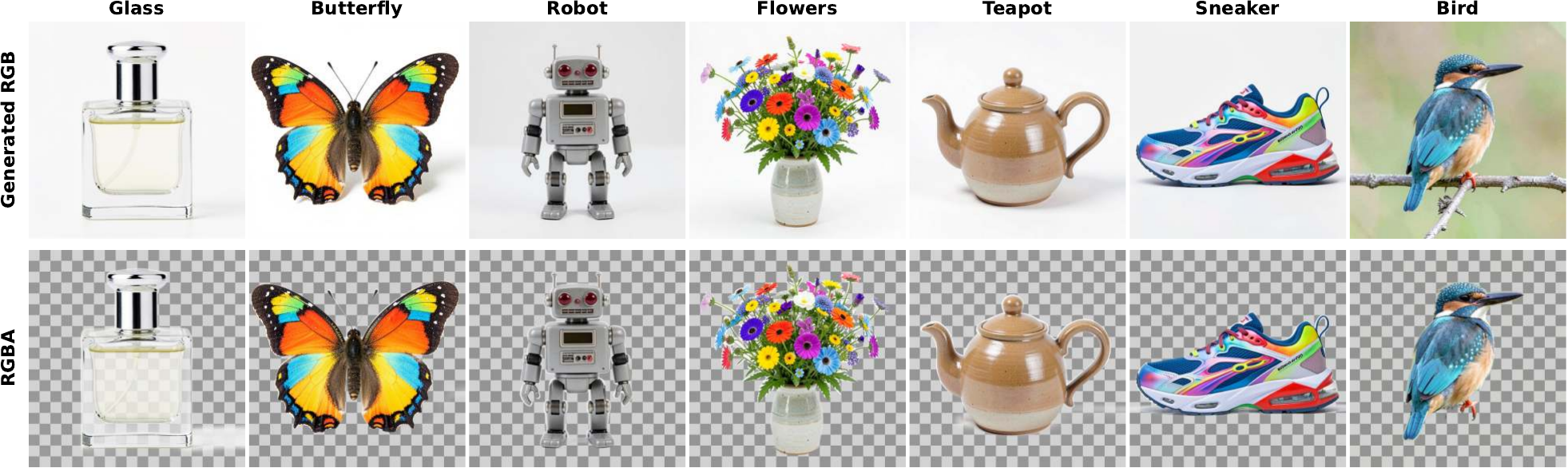}
    \caption{\textbf{Text-to-RGBA generation through a custom transparency TunnelingNet.}
    Each column executes the complete chain rather than reading a cached
    transparent asset: WSM first generates the RGB image from text, predicts
    its alpha matte with the custom LayerDiffuse TunnelingNet, and composes
    the resulting RGBA image over a checkerboard. Fine structures such as
    antennae, flower stems, handles, laces, and feathers remain visible in the
    predicted transparency.}
    \label{fig:custom_rgba_chain}
\end{figure*}

\begin{figure*}[t]
    \centering
    \includegraphics[width=0.98\textwidth]{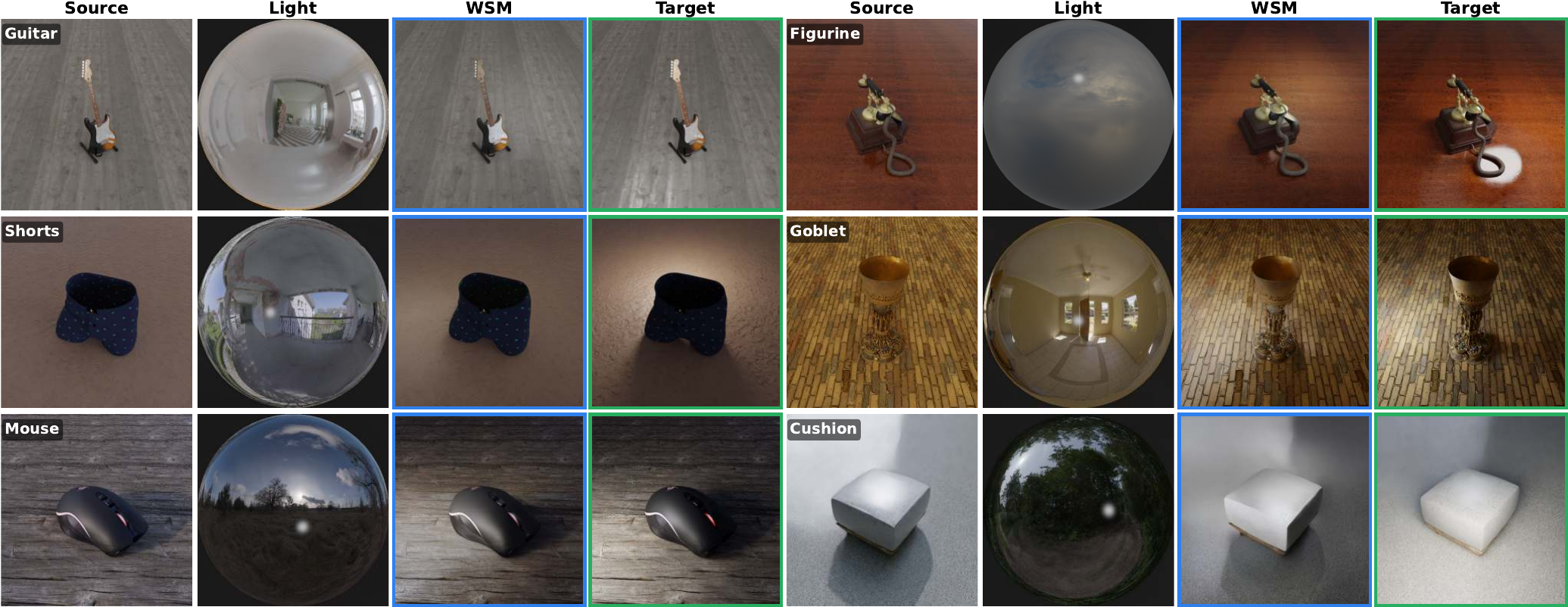}
    \caption{\textbf{Rapid acquisition of the full-score ICLR 2025 IC-Light
    relighting task.} Each group shows the source image, the spatial light
    probe, the WSM prediction, and the held-out target. Blue and green borders
    mark the WSM prediction and target, respectively. A single custom TunnelingNet
    redirects the frozen backbone according to the requested illumination while
    retaining object identity, material, texture, and geometry.}
    \label{fig:custom_iclight_relighting}
\end{figure*}

The preceding 1,148 TunnelingNets already establish the scalability of TunnelingNet.
We therefore end not with another routine mapping, but with a deliberately
ambitious showpiece: two signature capabilities that previously motivated
celebrated standalone systems are added to the same frozen 4B backbone as
ordinary lightweight TunnelingNets. We target the transparent-asset capability
of LayerDiffuse~\cite{10.1145/3658150} and the illumination-controlled
relighting task of the full-score ICLR 2025 IC-Light
paper~\cite{zhang2025scaling}. The transparency TunnelingNet requires only 1,000
updates with batch size four (4,000 sampled training examples), and the
relighting TunnelingNet only 2,800 single-sample updates. Figures~\ref{fig:custom_rgba_chain}
and~\ref{fig:custom_iclight_relighting} show held-out 10-step inference: WSM
completes a text-to-RGB-to-RGBA generation chain and performs spatially
controlled relighting while preserving scene content. In TunnelingNet,
headline capabilities that previously justified dedicated systems and papers
become lightweight additions to one shared backbone. This is the practical
power of deep tunneling: a strong new visual capability no longer requires
building another model from scratch.

\FloatBarrier

\input{sec_preprint/X_controlworld_guide}

\clearpage
\section{Appendix: Historical Task Interface Analysis}
\label{sec:supp_pwc_task_analysis}

To compare WSM with the tasks studied in previous Computer Vision research,
we manually reviewed 1,947 task entries collected from the final Papers with
Code archive~\cite{worldsnapshot2025pwcarchive}. For each task, we read its name
and description and checked the listed categories, benchmarks, and metrics. We
count a task as visual when visual data appears in its input or output, or when
visual evidence is necessary to solve it. This includes vision--language and
audiovisual tasks, but excludes entries based only on text, audio, graphs, or
physiological signals.

The analysis separates two questions. The A--D assessment asks whether each
historical visual task can be represented by the WSM dense or sparse interface
system; all 1,477 visually grounded entries receive such a route. The second
assessment asks whether each of 1,072 WSM cross-variable tasks had already been
defined with the same input evidence and output target. It measures new task
creation and does not measure compatibility with the historical catalog.

We organize the analysis into four ordered difficulty levels, from \textbf{A} (hardest) to \textbf{D} (easiest). \textbf{A} requires additional structured, temporal, interactive, or language-output development, for which WSM's sparse mode is the intended extension mechanism. \textbf{B} requires a minor interface, loss, label, data-adapter, or decoder adjustment. \textbf{C} keeps the method unchanged but requires a new aligned modality or dataset and corresponding training. \textbf{D} is directly compatible with an existing interface and represents an application, protocol, style, or preference scenario. The easiest D-level cases can be executed by the existing model without changing the loss or introducing a new task-specific training set. \textbf{X} marks an archive cross-listing with no visual role and is excluded from the difficulty ordering. Thus, $\mathrm{A}>\mathrm{B}>\mathrm{C}>\mathrm{D}$ orders the tasks by the engineering effort needed to realize them in WSM. Table~\ref{tab:pwc_task_scope_summary} summarizes this review.

\begin{table}[t]
  \centering
  \caption{\textbf{Manual difficulty assessment of the final archived
  Computer Vision task inventory.} Counts come from manual review of 1,947 archive entries. The 470 X entries
  do not involve vision and are excluded from the visual-task percentages.}
  \label{tab:pwc_task_scope_summary}
  \small
  \input{tables/pwc_task_scope/pwc_cv_task_scope_summary}
\end{table}

Among the 1,477 visual or vision-related tasks, 231 already fit an existing WSM
interface, 255 require a new aligned modality or dataset, 286 require a minor
interface or output adjustment, and 705 require a structured sparse extension.
The central result is that every reviewed visual task has a route through the
dense or sparse WSM interface system; the four levels describe how much
additional work that route requires.

\subsection{Novelty of evaluated WSM tasks}
We separately examine how many of the 1,072 evaluated cross-variable tasks
introduce an input--output task absent from the historical literature. Of these tasks, 655 have no historical
exact task match, while 401 remain conservative ambiguities because the published
task name does not specify its input evidence, and 16 have a clear match with
the same input and output. The first count is the strict lower bound on new WSM
tasks; adding the underspecified cases gives the upper estimate of 1,056. Table~\ref{tab:wsm_historical_overlap_summary} summarizes these results.

\begin{table}[t]
  \centering
  \caption{\textbf{Historical overlap of the 1,072 evaluated WSM
  cross-variable tasks.} Rows group mappings according to whether the same
  input--output task appears in the historical literature.}
  \label{tab:wsm_historical_overlap_summary}
  \small
  \input{tables/pwc_task_scope/wsm_historical_overlap_summary}
\end{table}

Only 16 of the 1,072 mappings have a clear historical counterpart. At least
655 mappings are therefore new, and the estimate rises to 1,056 when the
historically underspecified cases are included. Relative to the 1,477
established visual tasks, this expands the Computer Vision task space by
44.3--71.5\%. Together, Tables~\ref{tab:pwc_task_scope_summary} and
\ref{tab:wsm_historical_overlap_summary} show that WSM both provides a common
route for the established field and opens a substantially larger visual-task
space.

\clearpage
\onecolumn
\subsection{Historical Tasks Covered by the Four Interfaces}
Table~\ref{tab:pwc_task_scope_full} provides the task-level evidence behind the
aggregate result in Table~\ref{tab:pwc_task_scope_summary}. Each row records the
archived task definition and our judgment of how it relates to WSM. An A--D label
identifies the route required to express a visual task through the dense or sparse
WSM interface system, while X marks an entry that does not involve vision. All
1,477 visual or vision-related tasks receive an A--D route, and the remaining 470
nonvisual cross-listings receive X. The table therefore supports the central
result of Table~\ref{tab:pwc_task_scope_summary}: the WSM interface system covers
the complete reviewed visual-task set.

\input{tables/pwc_task_scope/pwc_cv_task_scope_full_inventory}

\input{sec_preprint/X_tutorial}

\input{sec_preprint/X_empirical_derivation}

%% file: tables/evaluation_categorical_output_by_target_normv2.tex
\begin{tabular}{llrrrrr}
\toprule
Target & Name & Heads & Samples & Strict & DAcc & mIoU \\
\midrule
111 & Object category & 39 & 497 & 0.250 & 0.288 & 0.255 \\
131 & Material class & 39 & 460 & 0.311 & 0.298 & 0.298 \\
135 & Container flag & 39 & 429 & 0.915 & 0.925 & 0.493 \\
136 & Object class & 39 & 460 & 0.141 & 0.085 & 0.085 \\
195 & Semantic segmentation & 22 & 528 & 0.341 & 0.401 & 0.107 \\
\bottomrule
\end{tabular}

%% file: sec_preprint/X_controlworld_guide.tex
\clearpage
\onecolumn
\twocolumn
\raggedbottom
\section{Appendix: ControlWorld User Guide}
\label{sec:controlworld-user-guide}
\noindent\begin{minipage}{\columnwidth}
ControlWorld is the reference interface for selecting and running WSM tasks. A user selects a trained mapping in the WSG Task Graph, supplies its image and text inputs, selects the corresponding trained model, and generates the output. The output panel then applies the target modality's WSG Reader visualization and legend. The desktop captures below show the same released 10-step Euler results used in the qualitative evaluation, with identical tasks, inputs, models, seeds, and outputs.
\end{minipage}

\noindent\textbf{Reading each capture.} The upper task graph identifies the selected input--output route. The input cards show the supplied conditions, the central controls expose checkpoint and sampling settings, and the output card shows the decoded result. Parts I, II, VII, and IX are reproduced here; for Part IX, only the first three source figures are included.

\subsection{The Four WSM Interfaces}
Part I exposes one desktop workflow for each of the four interface definitions.

\begin{figure*}[p]
  \centering
  \begin{minipage}[t]{0.492\textwidth}
    \centering
    \includegraphics[width=\linewidth,height=1.94514\linewidth]{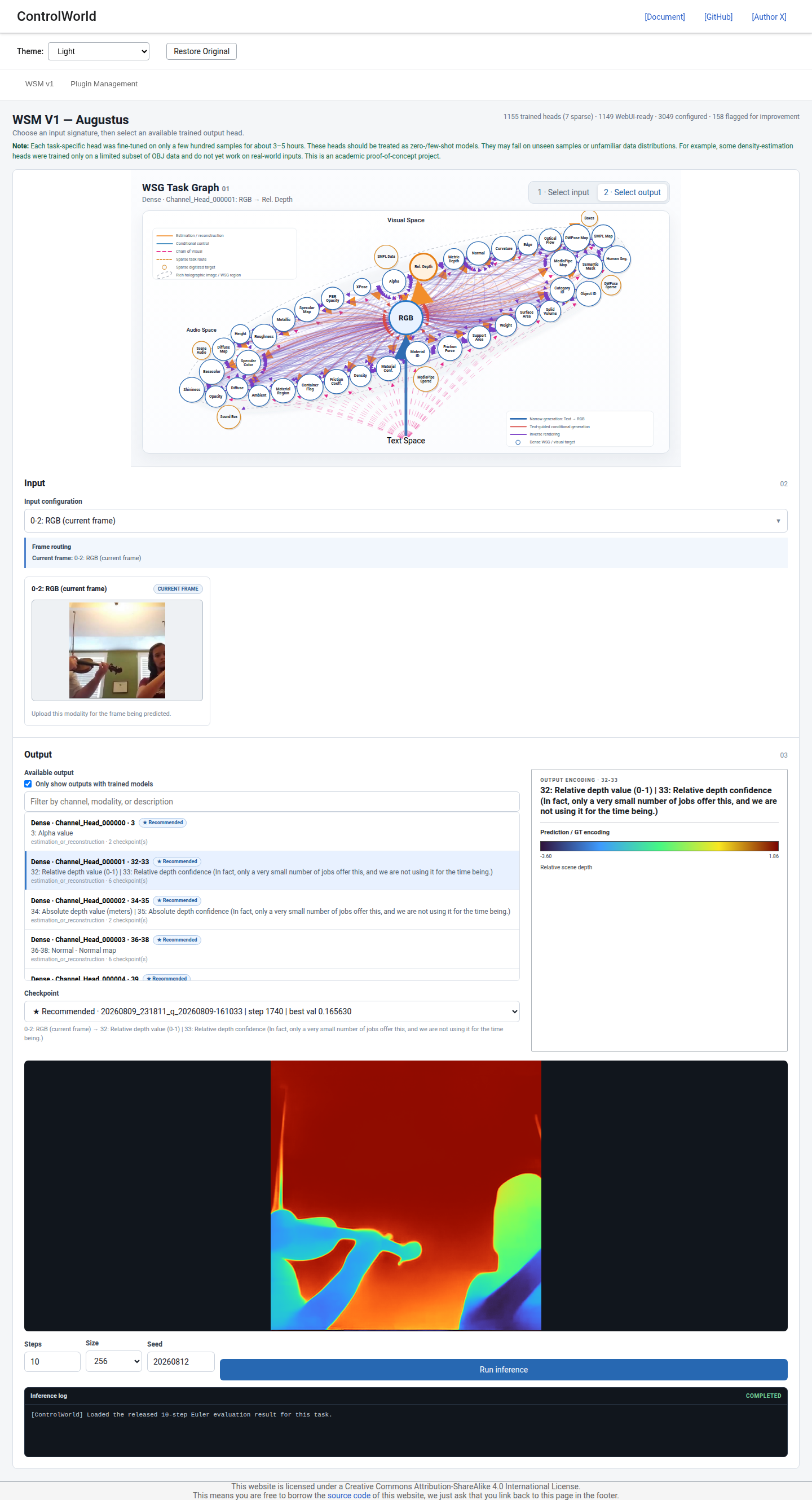}
    \vspace{-4pt}
    {\small (a) \textbf{TunnelingNet 000001:} RGB $\rightarrow$ RelDepth.}
  \end{minipage}
  \hfill
  \begin{minipage}[t]{0.492\textwidth}
    \centering
    \includegraphics[width=\linewidth,height=1.94514\linewidth]{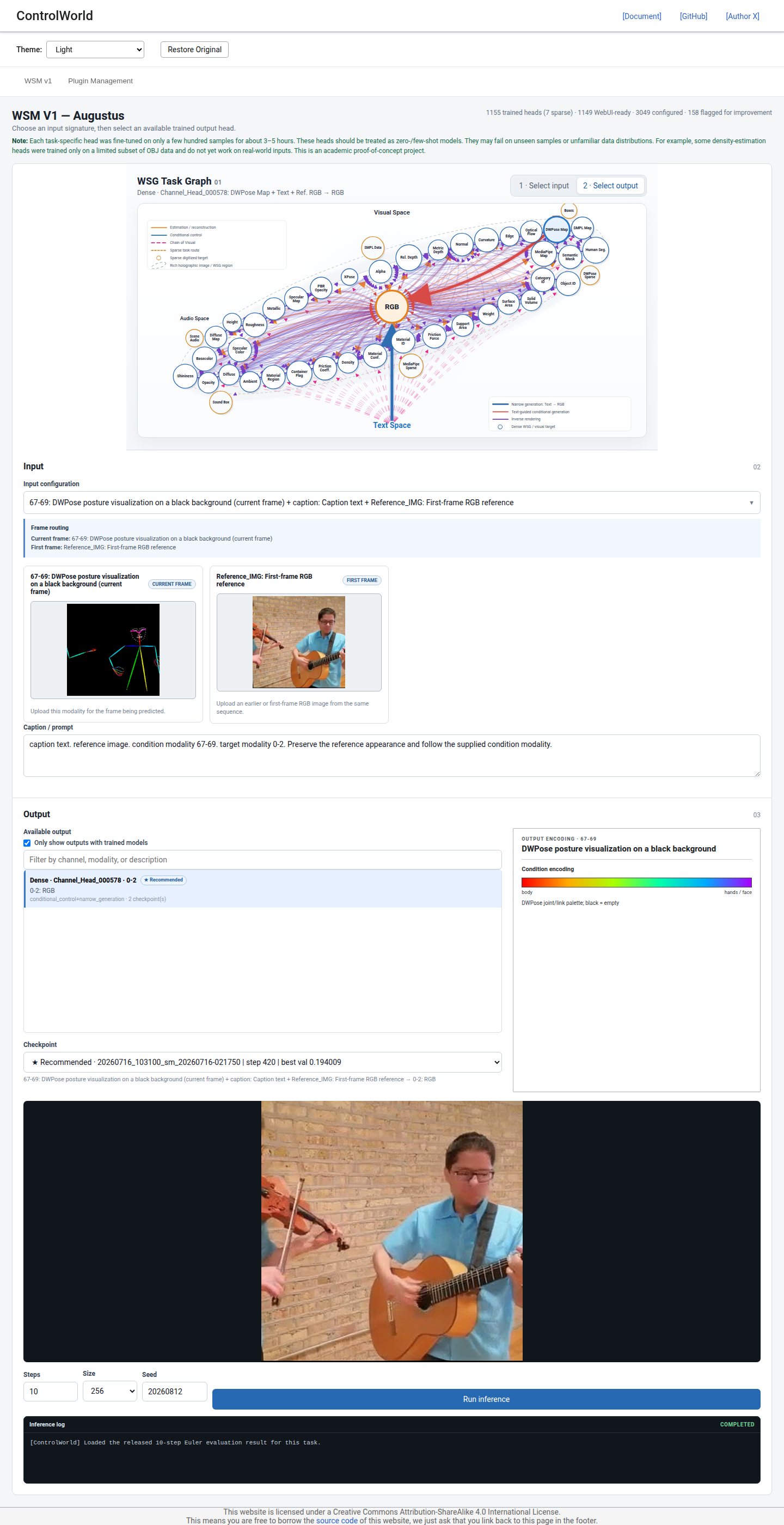}
    \vspace{-4pt}
    {\small (b) \textbf{TunnelingNet 000578:} DWPose+Text+RGB1 $\rightarrow$ RGB.}
  \end{minipage}
  \caption{ControlWorld desktop workflows for Qualitative Part I: TunnelingNet 000001 (RGB to RelDepth) and TunnelingNet 000578 (DWPose+Text+RGB1 to RGB). Each panel preserves the full task graph, input, sampling, and decoded-output interface.}
  \label{fig:controlworld-guide-i-01}
\end{figure*}

\begin{figure*}[p]
  \centering
  \begin{minipage}[t]{0.492\textwidth}
    \centering
    \includegraphics[width=\linewidth,height=1.89375\linewidth]{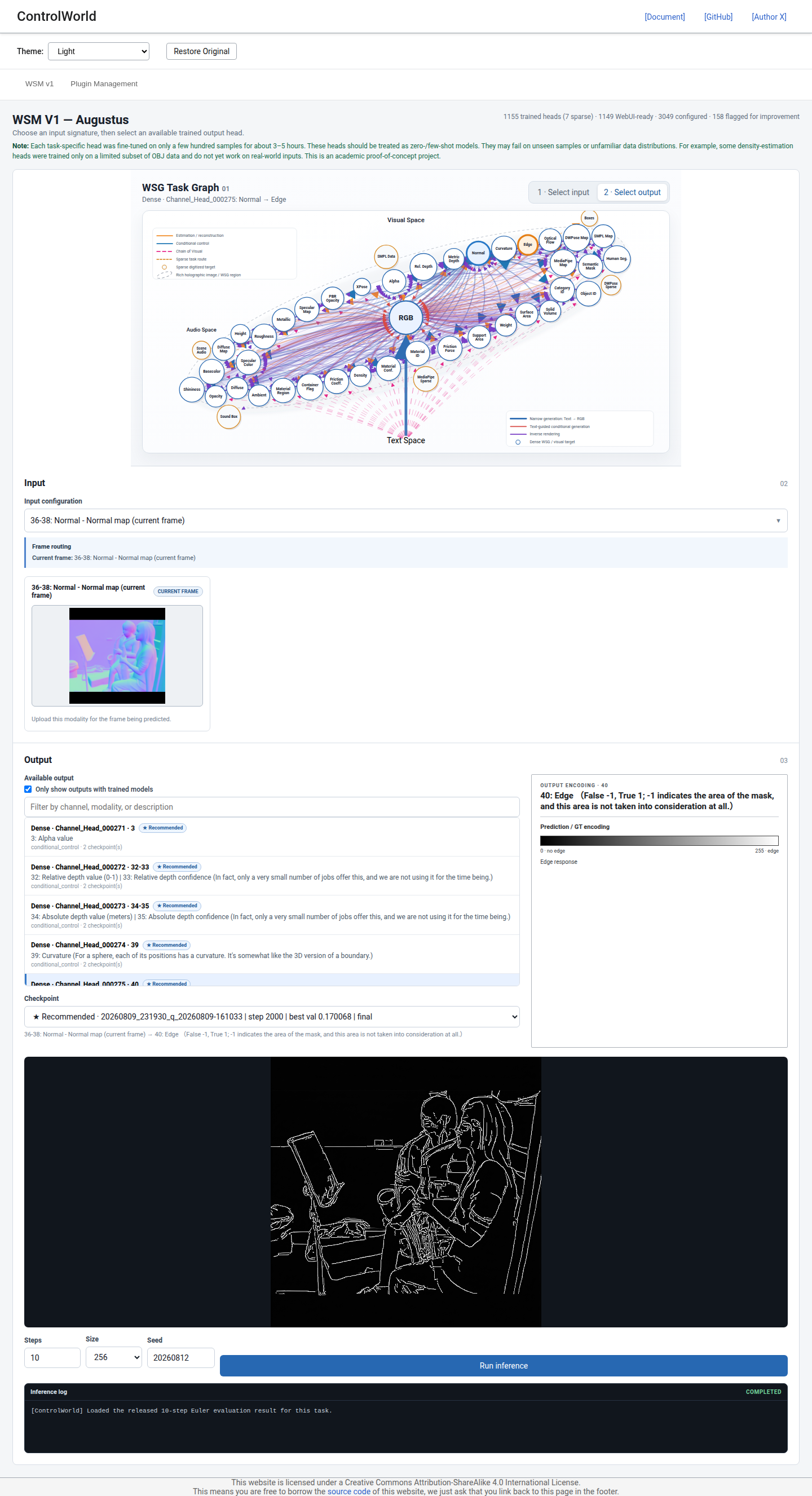}
    \vspace{-4pt}
    {\small (a) \textbf{TunnelingNet 000275:} Normal $\rightarrow$ Edge.}
  \end{minipage}
  \hfill
  \begin{minipage}[t]{0.492\textwidth}
    \centering
    \includegraphics[width=\linewidth,height=1.89375\linewidth]{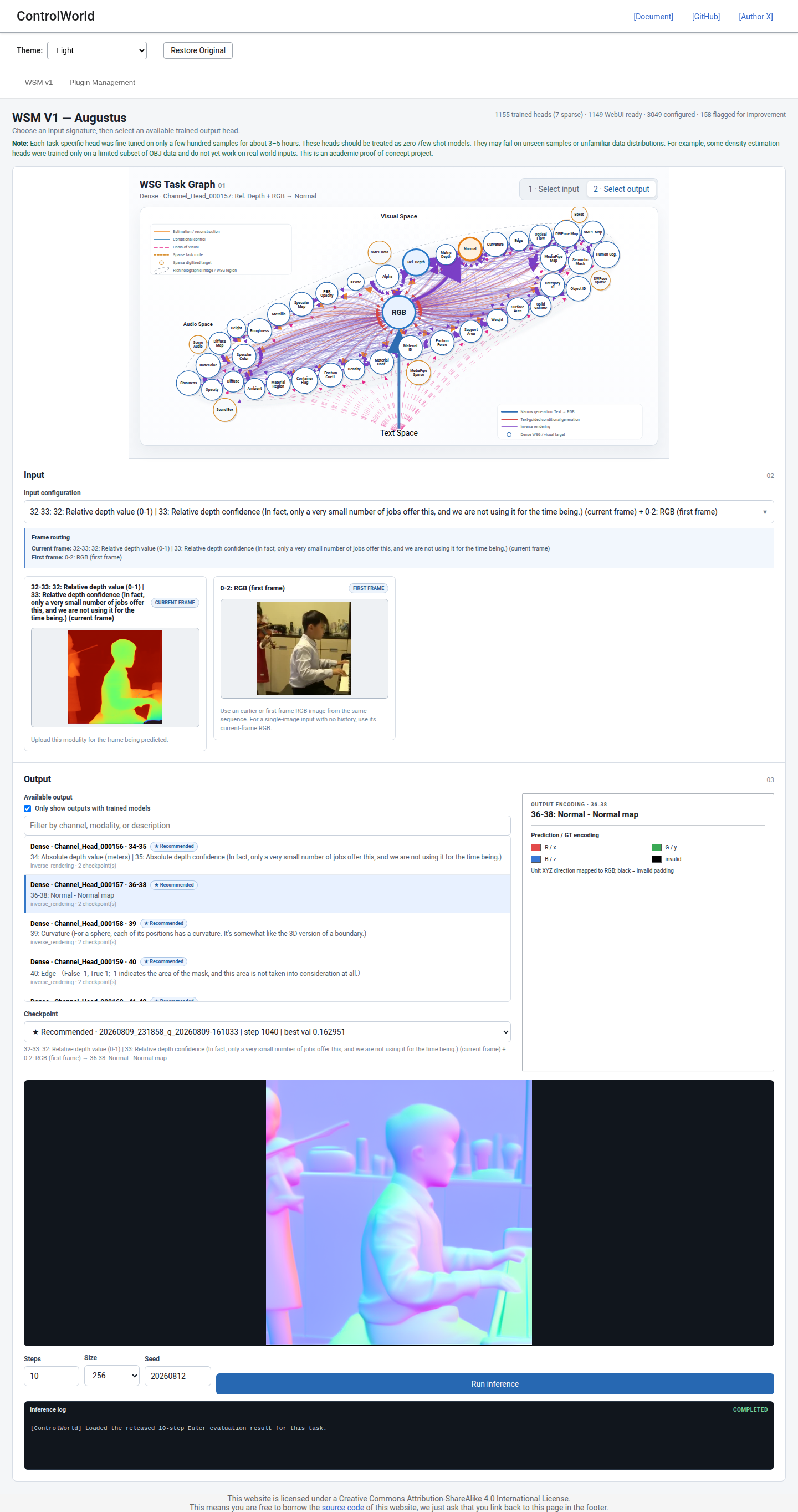}
    \vspace{-4pt}
    {\small (b) \textbf{TunnelingNet 000157:} RelDepth+RGB $\rightarrow$ Normal.}
  \end{minipage}
  \caption{ControlWorld desktop workflows for Qualitative Part I: TunnelingNet 000275 (Normal to Edge) and TunnelingNet 000157 (RelDepth+RGB to Normal). Each panel preserves the full task graph, input, sampling, and decoded-output interface.}
  \label{fig:controlworld-guide-i-02}
\end{figure*}

\subsection{RGB Estimation of Geometry and Motion}
Part II demonstrates the complete RGB-centered geometry and motion sheet.

\begin{figure*}[p]
  \centering
  \begin{minipage}[t]{0.492\textwidth}
    \centering
    \includegraphics[width=\linewidth,height=1.84653\linewidth]{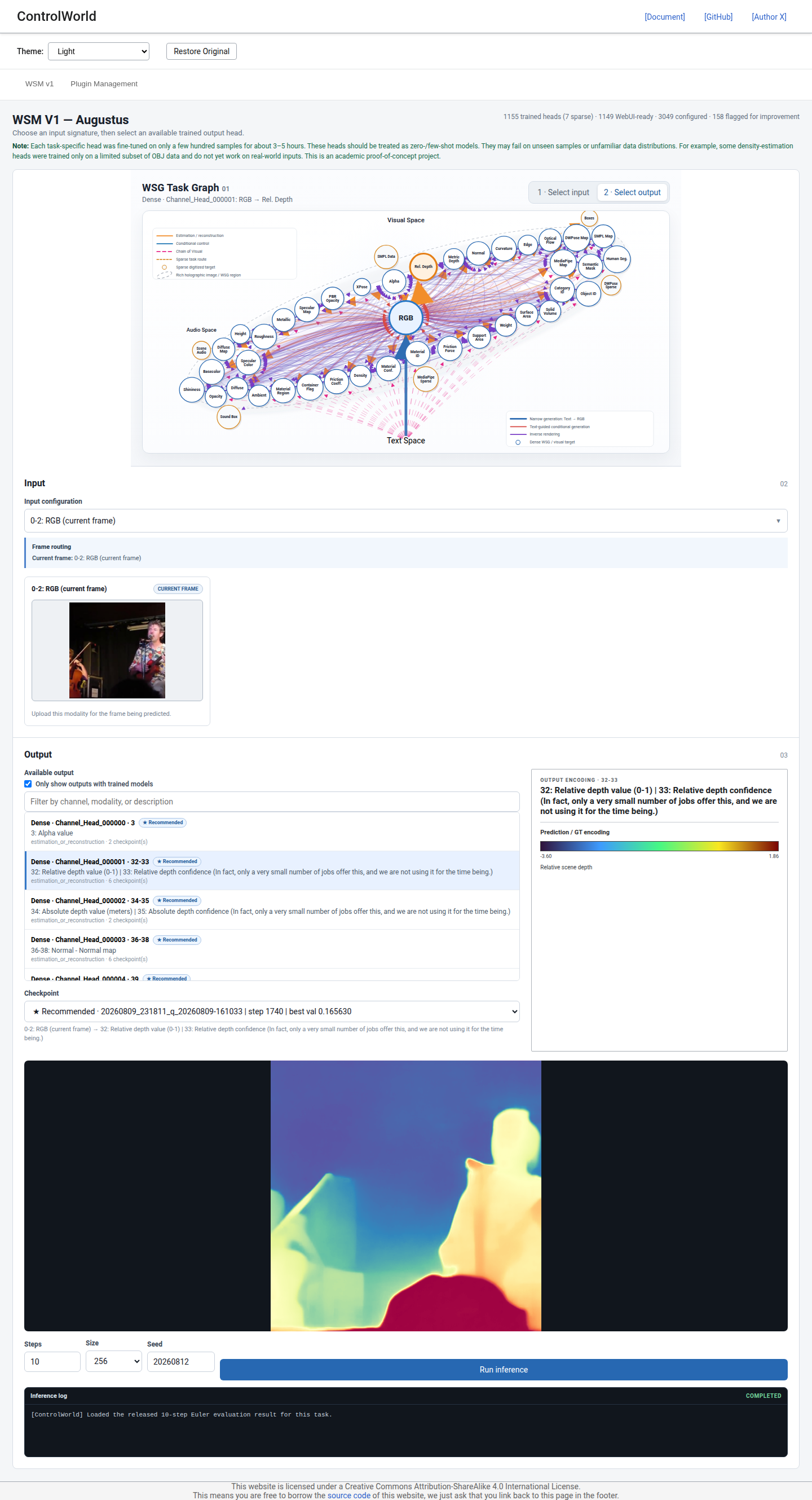}
    \vspace{-4pt}
    {\small (a) \textbf{TunnelingNet 000001:} RGB $\rightarrow$ RelDepth.}
  \end{minipage}
  \hfill
  \begin{minipage}[t]{0.492\textwidth}
    \centering
    \includegraphics[width=\linewidth,height=1.84653\linewidth]{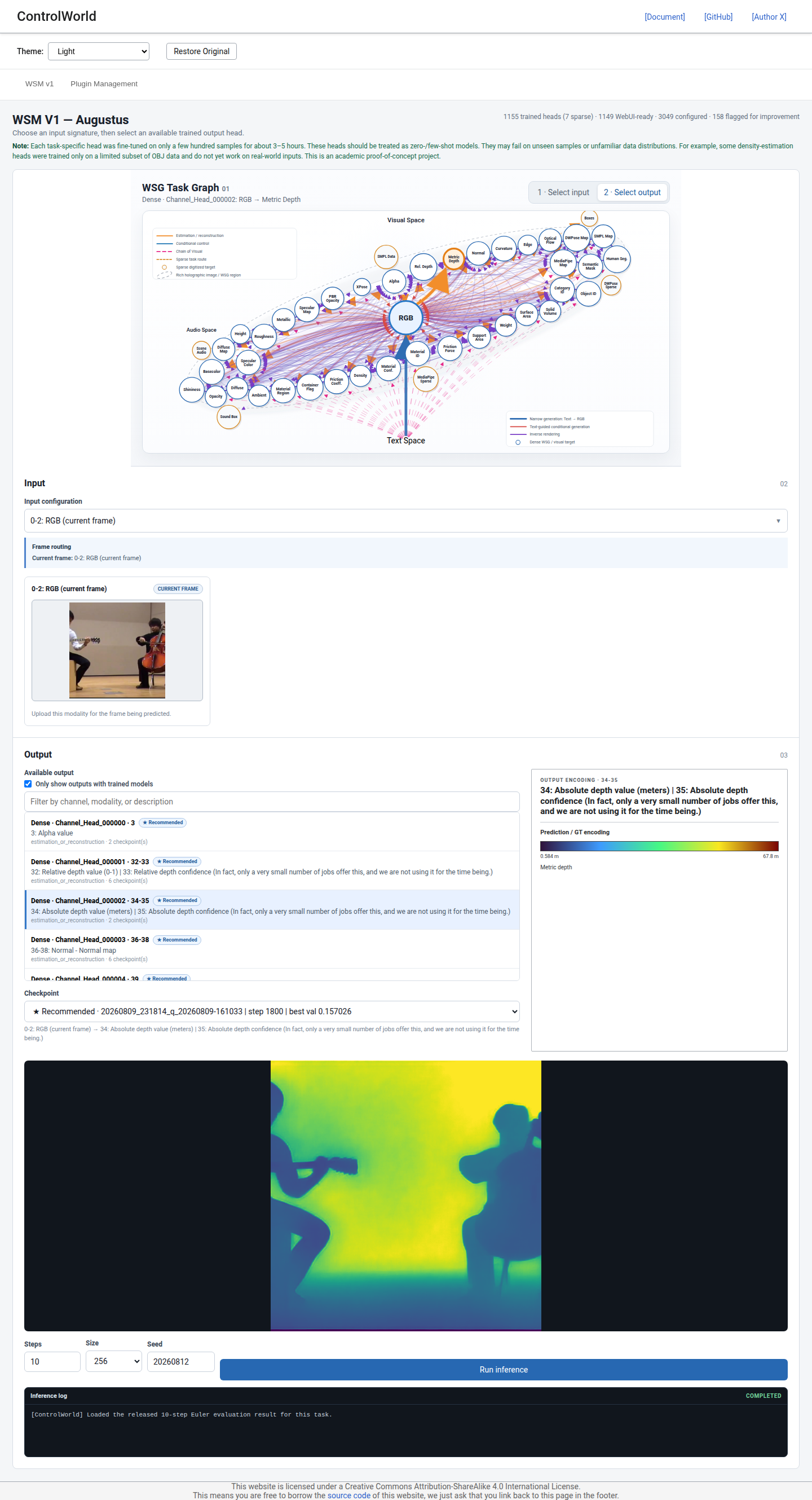}
    \vspace{-4pt}
    {\small (b) \textbf{TunnelingNet 000002:} RGB $\rightarrow$ AbsDepth.}
  \end{minipage}
  \caption{ControlWorld desktop workflows for Qualitative Part II: TunnelingNet 000001 (RGB to RelDepth) and TunnelingNet 000002 (RGB to AbsDepth). Each panel preserves the full task graph, input, sampling, and decoded-output interface.}
  \label{fig:controlworld-guide-ii-03}
\end{figure*}

\begin{figure*}[p]
  \centering
  \begin{minipage}[t]{0.492\textwidth}
    \centering
    \includegraphics[width=\linewidth,height=1.83542\linewidth]{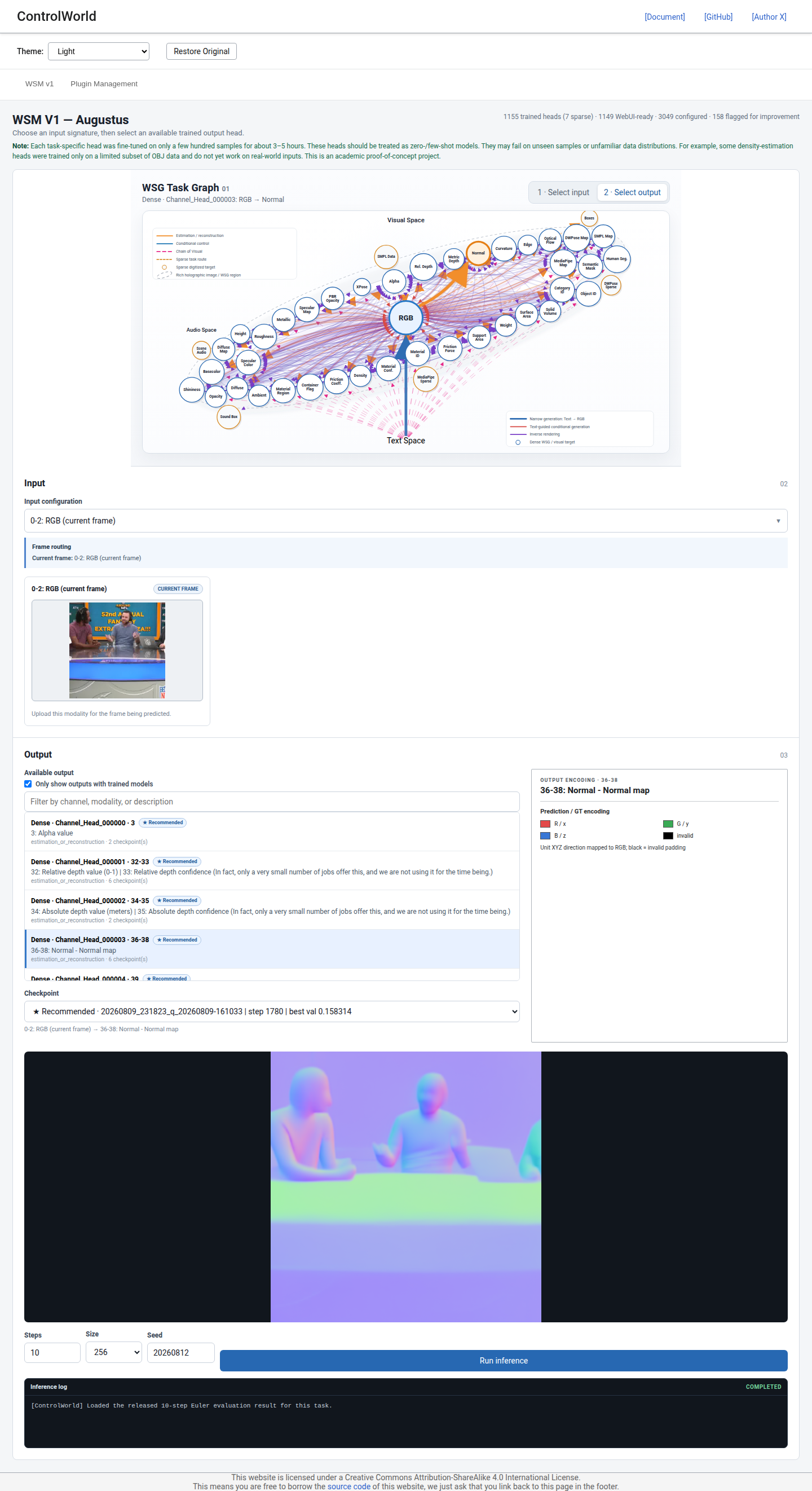}
    \vspace{-4pt}
    {\small (a) \textbf{TunnelingNet 000003:} RGB $\rightarrow$ Normal.}
  \end{minipage}
  \hfill
  \begin{minipage}[t]{0.492\textwidth}
    \centering
    \includegraphics[width=\linewidth,height=1.83542\linewidth]{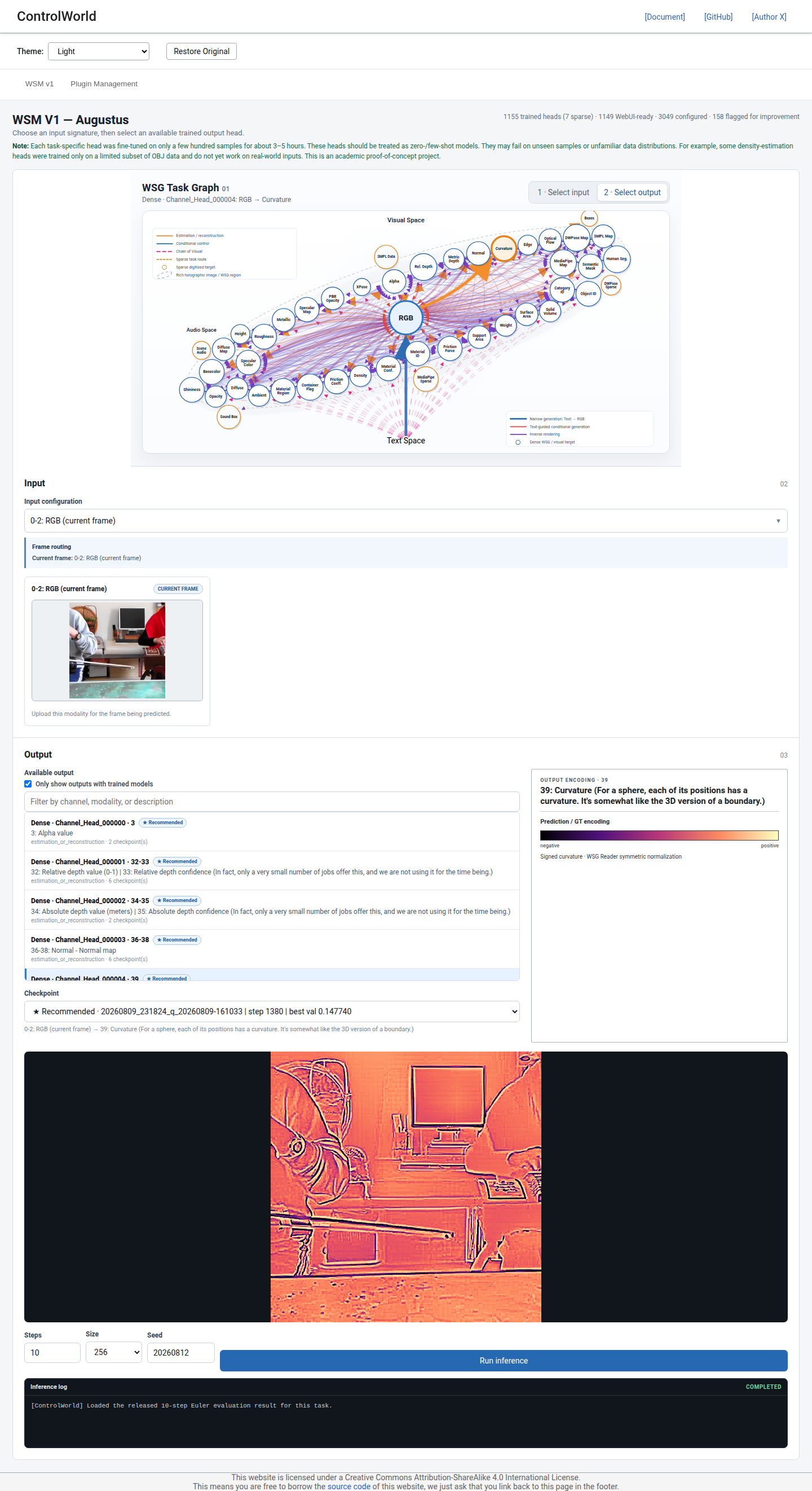}
    \vspace{-4pt}
    {\small (b) \textbf{TunnelingNet 000004:} RGB $\rightarrow$ Curve.}
  \end{minipage}
  \caption{ControlWorld desktop workflows for Qualitative Part II: TunnelingNet 000003 (RGB to Normal) and TunnelingNet 000004 (RGB to Curve). Each panel preserves the full task graph, input, sampling, and decoded-output interface.}
  \label{fig:controlworld-guide-ii-04}
\end{figure*}

\begin{figure*}[p]
  \centering
  \begin{minipage}[t]{0.492\textwidth}
    \centering
    \includegraphics[width=\linewidth,height=1.83542\linewidth]{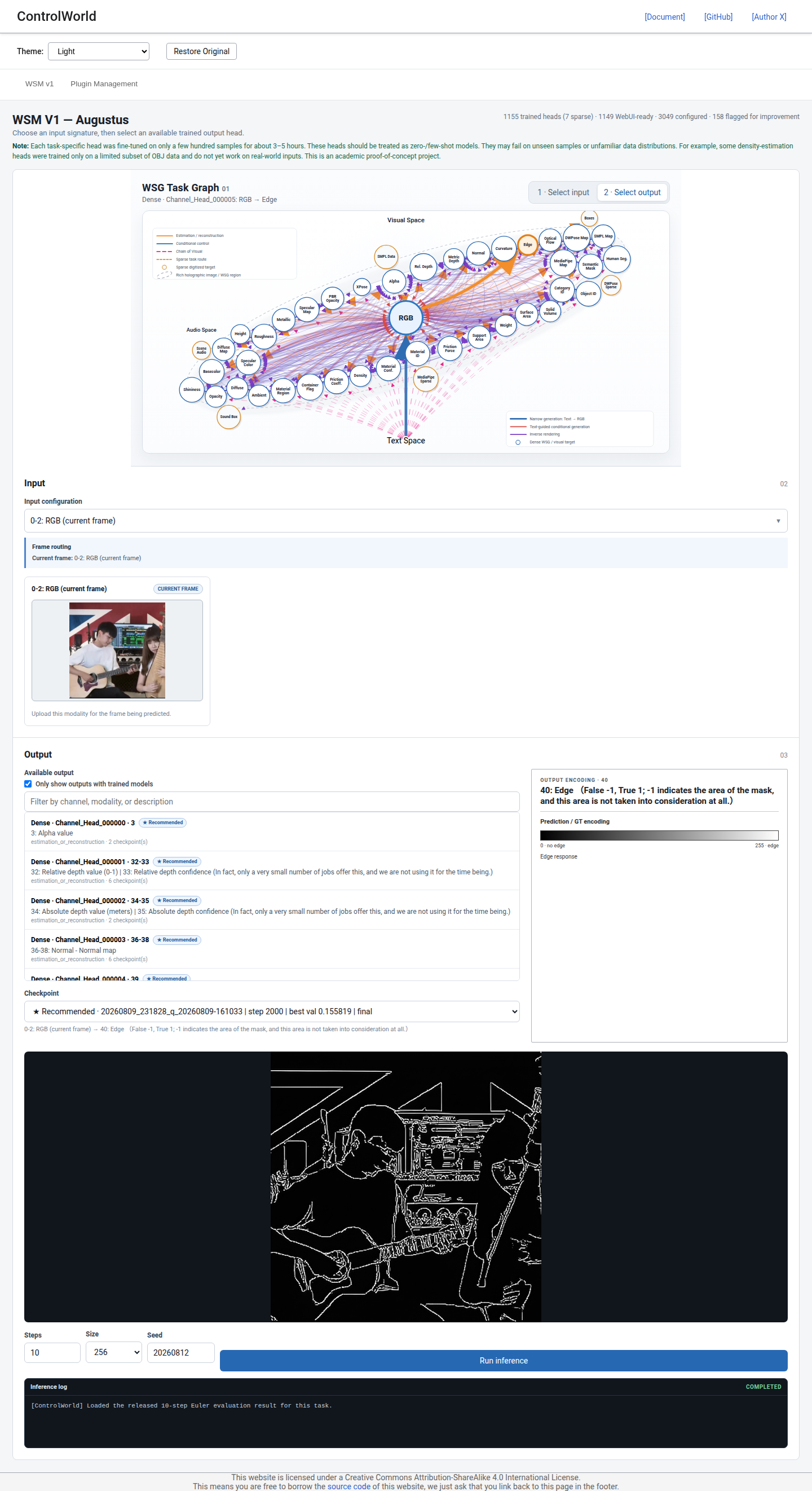}
    \vspace{-4pt}
    {\small (a) \textbf{TunnelingNet 000005:} RGB $\rightarrow$ Edge.}
  \end{minipage}
  \hfill
  \begin{minipage}[t]{0.492\textwidth}
    \centering
    \includegraphics[width=\linewidth,height=1.83542\linewidth]{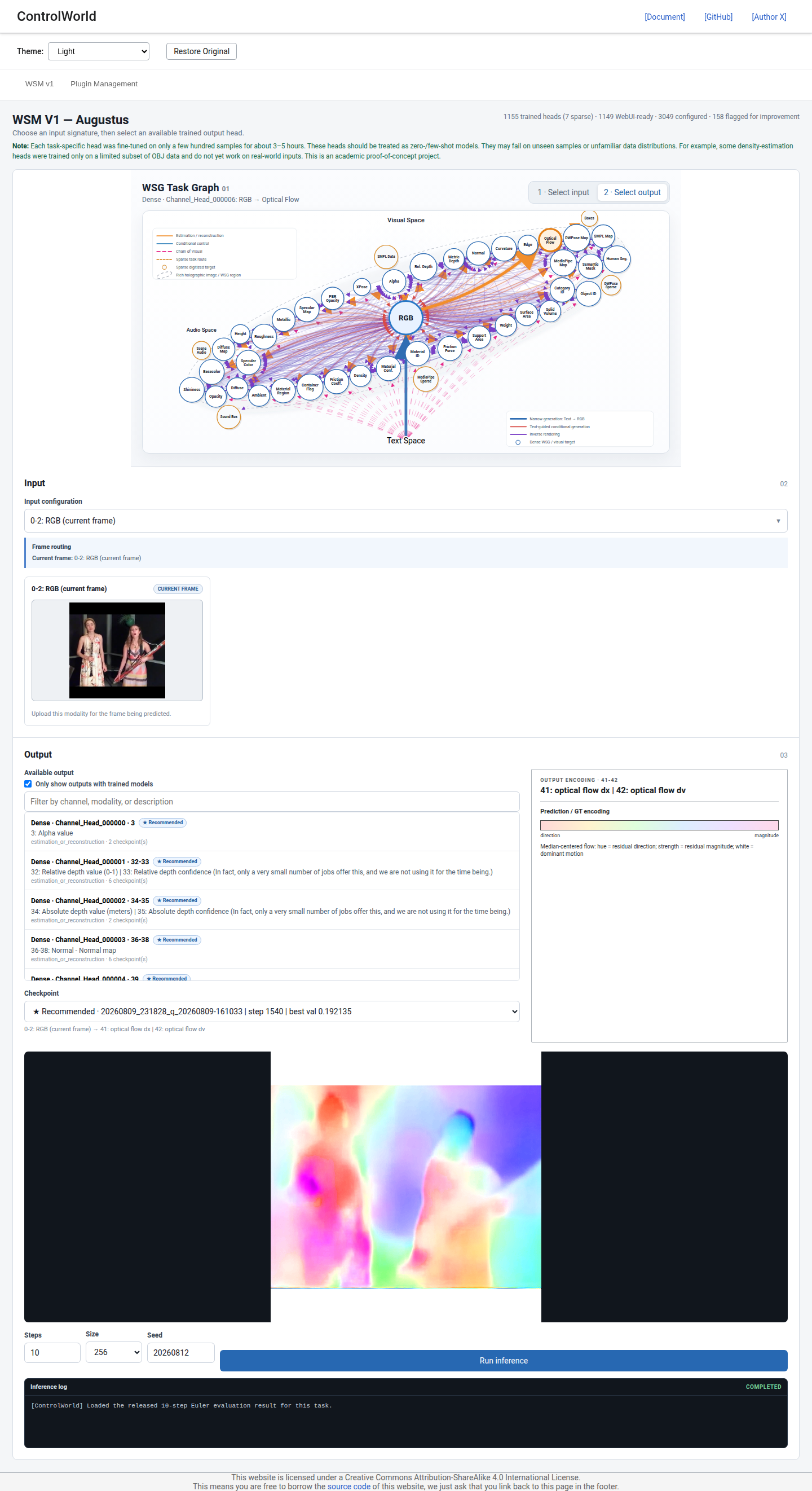}
    \vspace{-4pt}
    {\small (b) \textbf{TunnelingNet 000006:} RGB $\rightarrow$ Flow.}
  \end{minipage}
  \caption{ControlWorld desktop workflows for Qualitative Part II: TunnelingNet 000005 (RGB to Edge) and TunnelingNet 000006 (RGB to Flow). Each panel preserves the full task graph, input, sampling, and decoded-output interface.}
  \label{fig:controlworld-guide-ii-05}
\end{figure*}

\subsection{Narrow Generation from Geometry and Human Structure}
Part VII uses the first four released narrow-generation examples, as a compact walkthrough of geometry-to-RGB operation.

\begin{figure*}[p]
  \centering
  \begin{minipage}[t]{0.492\textwidth}
    \centering
    \includegraphics[width=\linewidth,height=1.99097\linewidth]{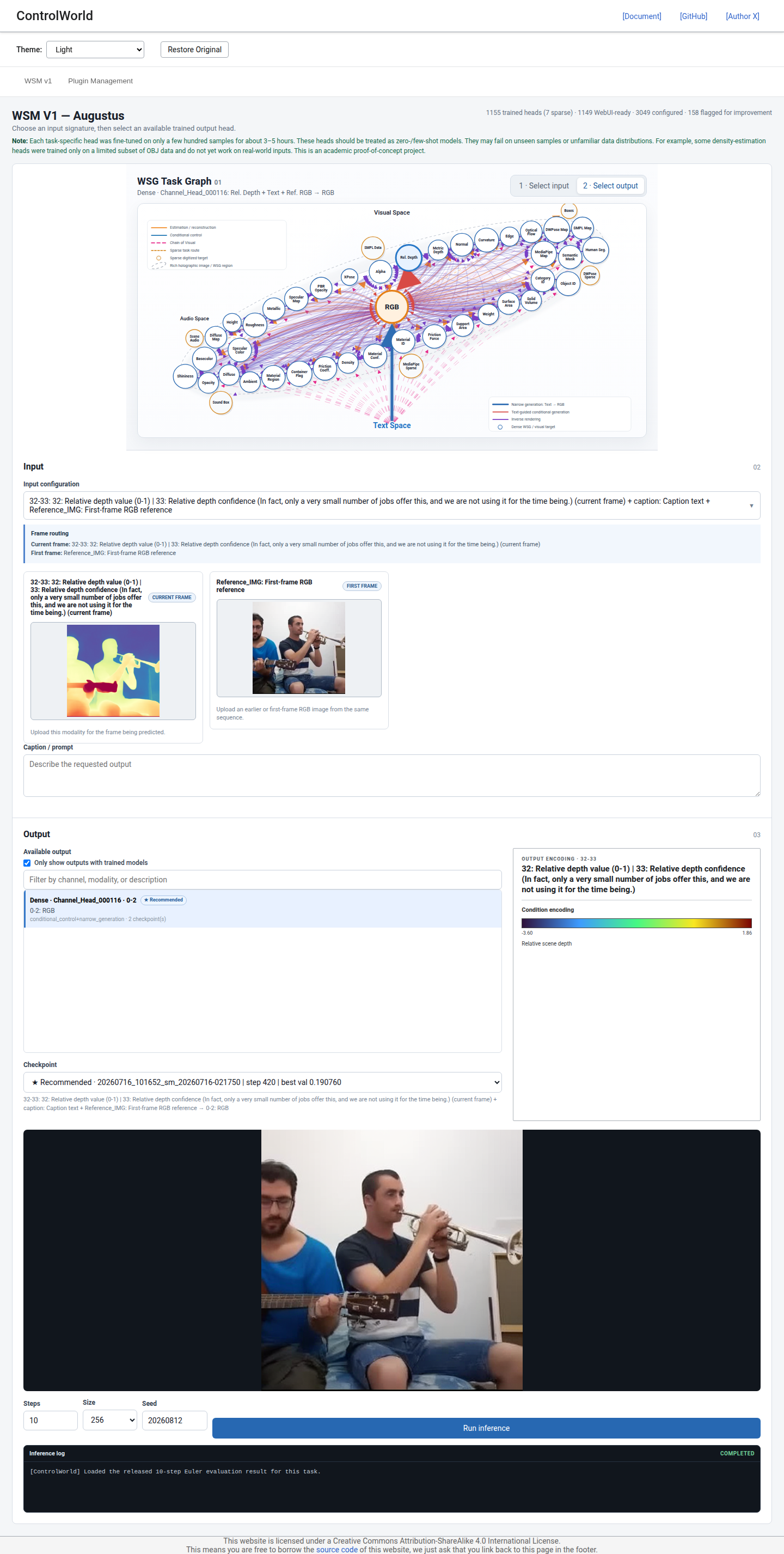}
    \vspace{-4pt}
    {\small (a) \textbf{TunnelingNet 000116:} RelDepth+Text+RGB1 $\rightarrow$ RGB.}
  \end{minipage}
  \hfill
  \begin{minipage}[t]{0.492\textwidth}
    \centering
    \includegraphics[width=\linewidth,height=1.99097\linewidth]{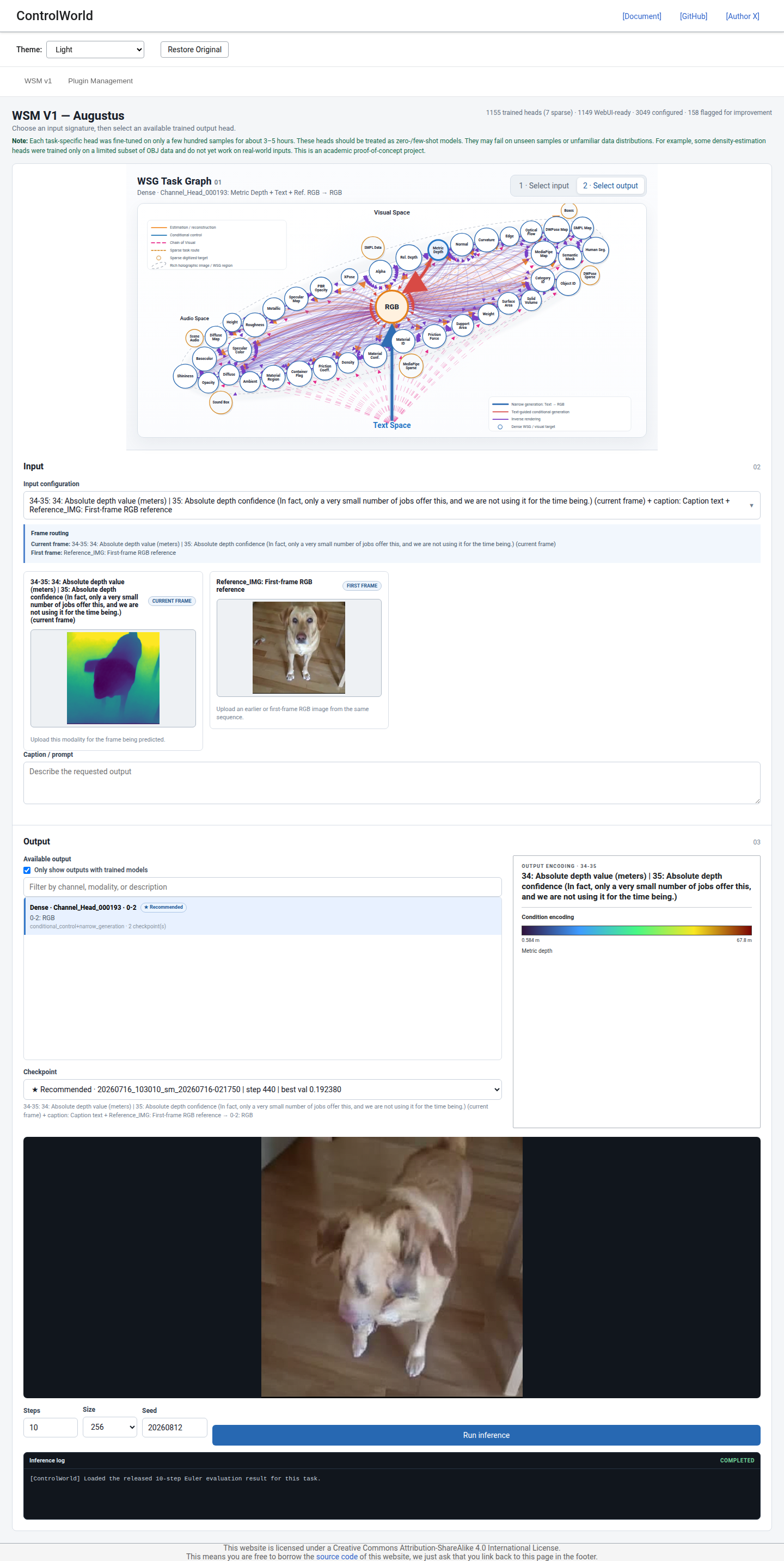}
    \vspace{-4pt}
    {\small (b) \textbf{TunnelingNet 000193:} AbsDepth+Text+RGB1 $\rightarrow$ RGB.}
  \end{minipage}
  \caption{ControlWorld desktop workflows for Qualitative Part VII: TunnelingNet 000116 (RelDepth+Text+RGB1 to RGB) and TunnelingNet 000193 (AbsDepth+Text+RGB1 to RGB). Each panel preserves the full task graph, input, sampling, and decoded-output interface.}
  \label{fig:controlworld-guide-vii-06}
\end{figure*}

\begin{figure*}[p]
  \centering
  \begin{minipage}[t]{0.492\textwidth}
    \centering
    \includegraphics[width=\linewidth,height=1.96458\linewidth]{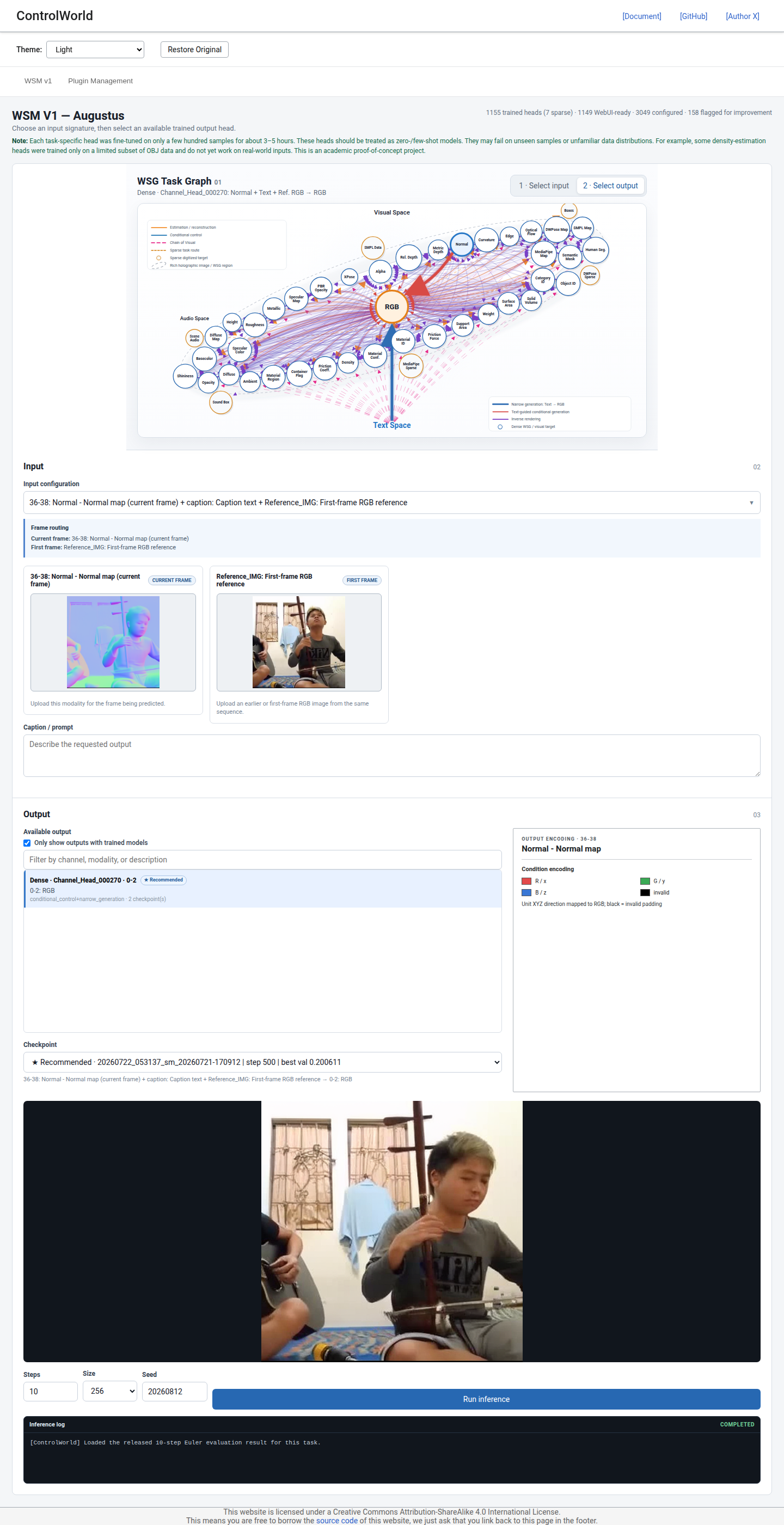}
    \vspace{-4pt}
    {\small (a) \textbf{TunnelingNet 000270:} Normal+Text+RGB1 $\rightarrow$ RGB.}
  \end{minipage}
  \hfill
  \begin{minipage}[t]{0.492\textwidth}
    \centering
    \includegraphics[width=\linewidth,height=1.96458\linewidth]{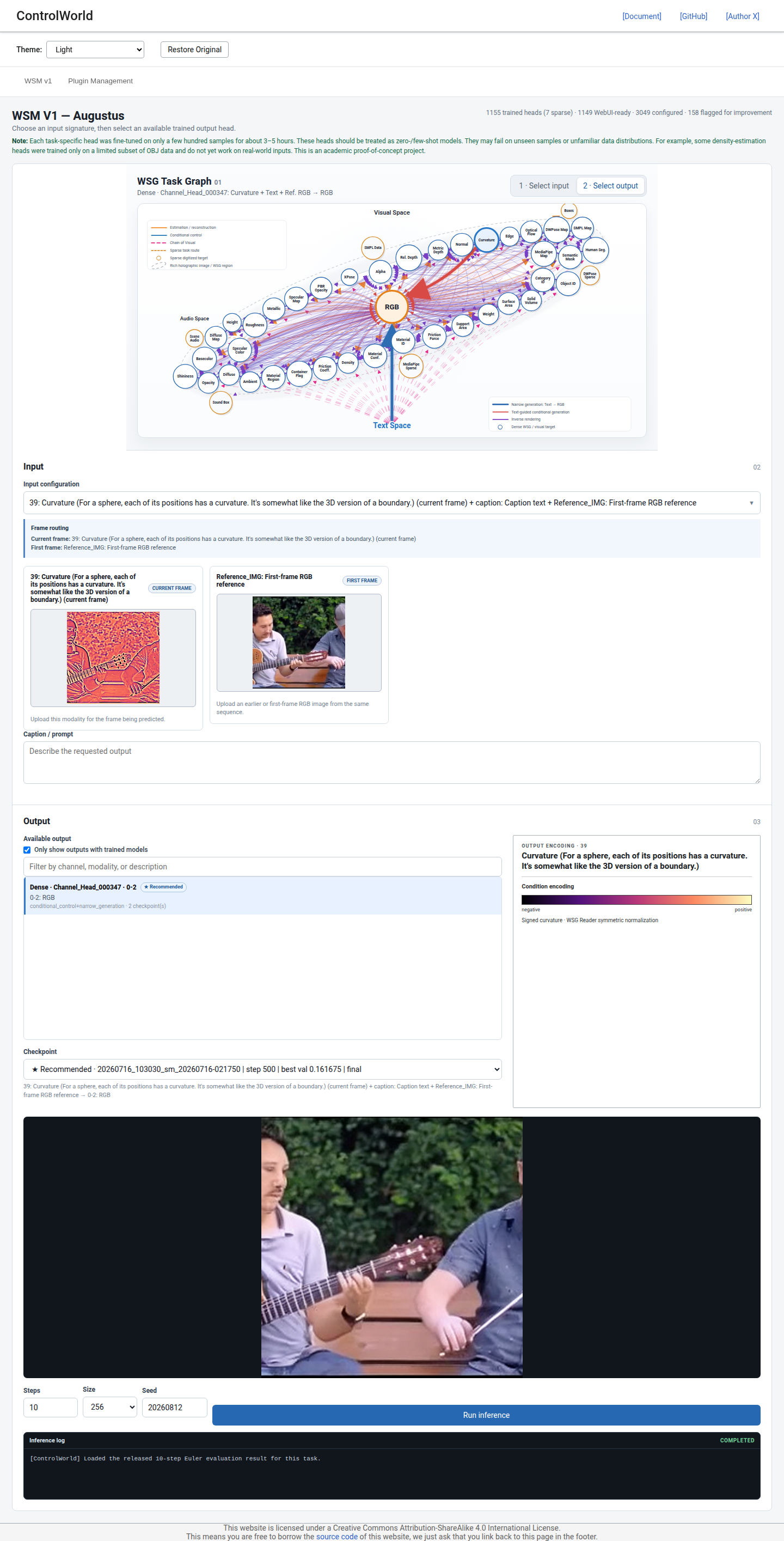}
    \vspace{-4pt}
    {\small (b) \textbf{TunnelingNet 000347:} Curve+Text+RGB1 $\rightarrow$ RGB.}
  \end{minipage}
  \caption{ControlWorld desktop workflows for Qualitative Part VII: TunnelingNet 000270 (Normal+Text+RGB1 to RGB) and TunnelingNet 000347 (Curve+Text+RGB1 to RGB). Each panel preserves the full task graph, input, sampling, and decoded-output interface.}
  \label{fig:controlworld-guide-vii-07}
\end{figure*}

\subsection{Conditional Control}
Part IX uses every example from its first three qualitative sheets (18 tasks) and omits sheets four through six.

\begin{figure*}[p]
  \centering
  \begin{minipage}[t]{0.492\textwidth}
    \centering
    \includegraphics[width=\linewidth,height=1.89236\linewidth]{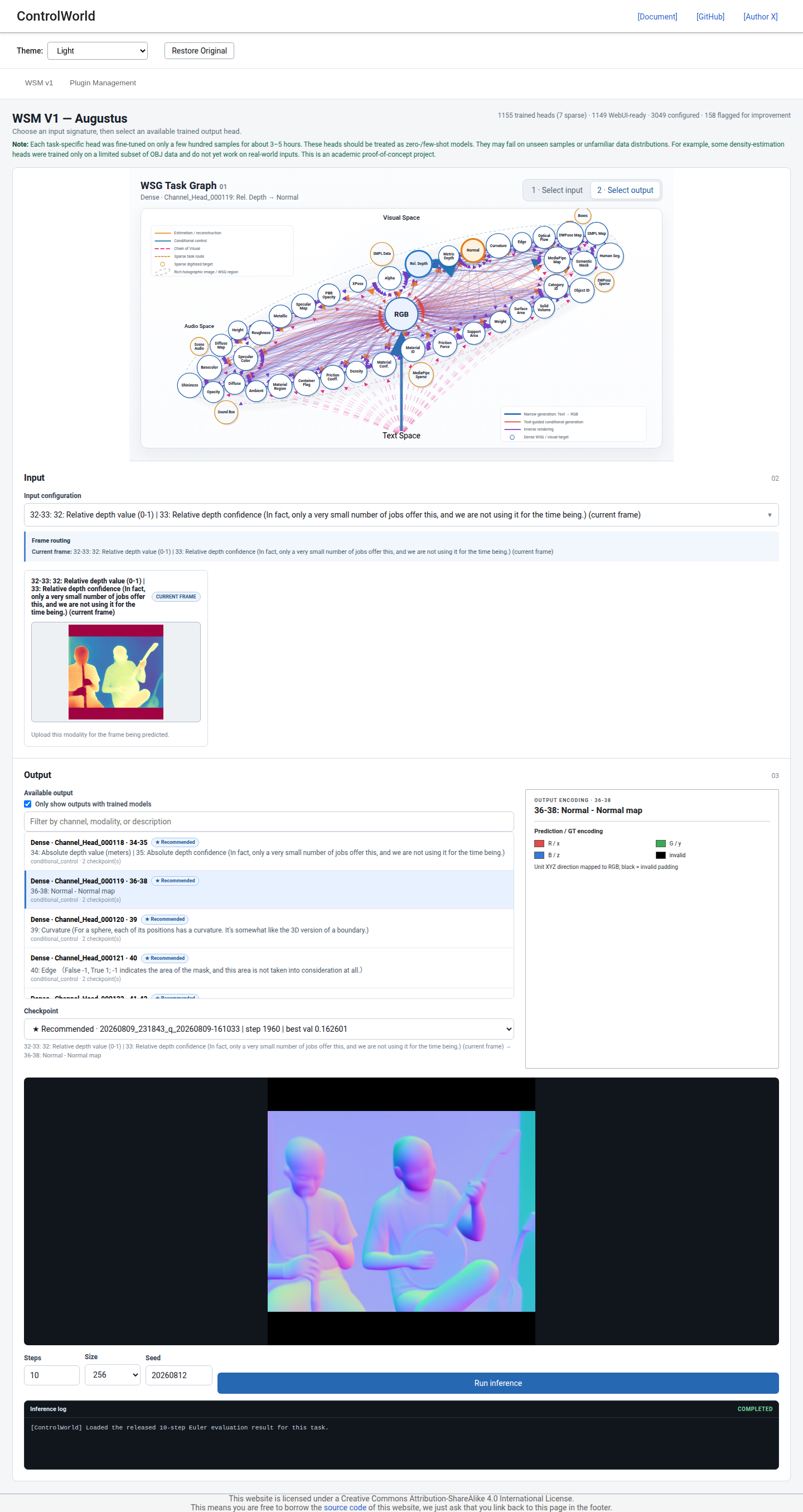}
    \vspace{-4pt}
    {\small (a) \textbf{TunnelingNet 000119:} RelDepth $\rightarrow$ Normal.}
  \end{minipage}
  \hfill
  \begin{minipage}[t]{0.492\textwidth}
    \centering
    \includegraphics[width=\linewidth,height=1.89236\linewidth]{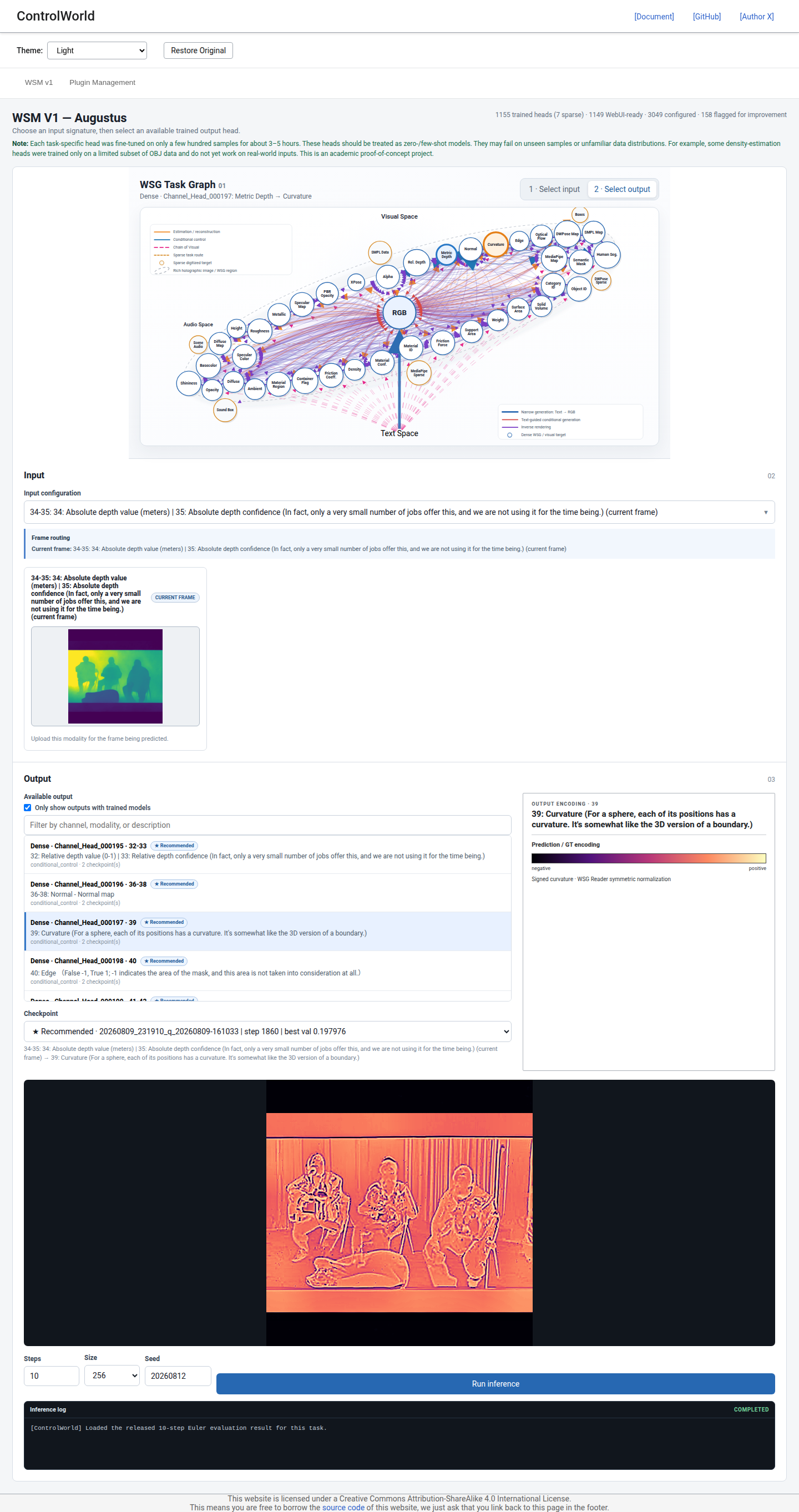}
    \vspace{-4pt}
    {\small (b) \textbf{TunnelingNet 000197:} AbsDepth $\rightarrow$ Curve.}
  \end{minipage}
  \caption{ControlWorld desktop workflows for Qualitative Part IX: TunnelingNet 000119 (RelDepth to Normal) and TunnelingNet 000197 (AbsDepth to Curve). Each panel preserves the full task graph, input, sampling, and decoded-output interface.}
  \label{fig:controlworld-guide-ix-08}
\end{figure*}

\begin{figure*}[p]
  \centering
  \begin{minipage}[t]{0.492\textwidth}
    \centering
    \includegraphics[width=\linewidth,height=1.86181\linewidth]{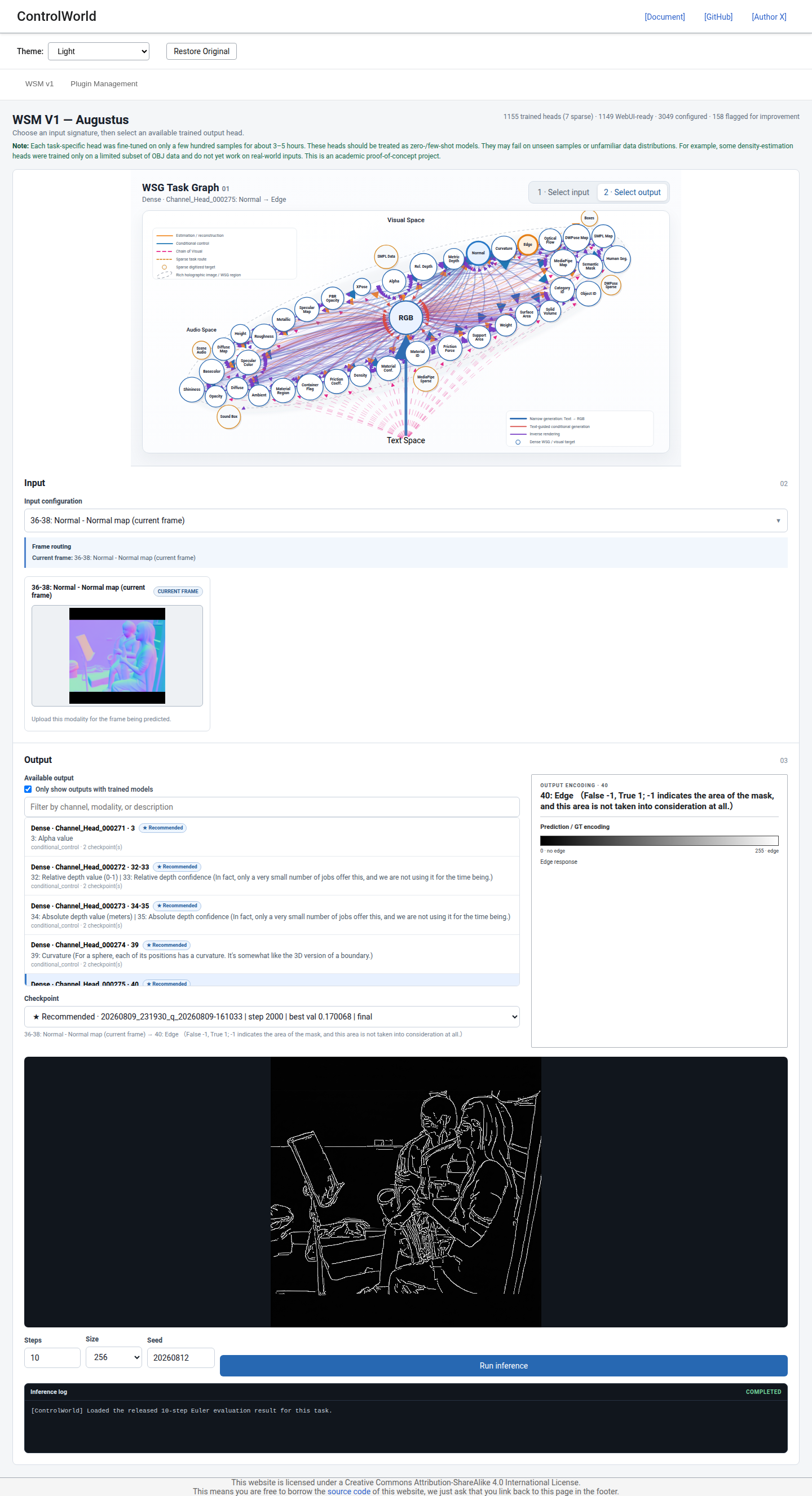}
    \vspace{-4pt}
    {\small (a) \textbf{TunnelingNet 000275:} Normal $\rightarrow$ Edge.}
  \end{minipage}
  \hfill
  \begin{minipage}[t]{0.492\textwidth}
    \centering
    \includegraphics[width=\linewidth,height=1.86181\linewidth]{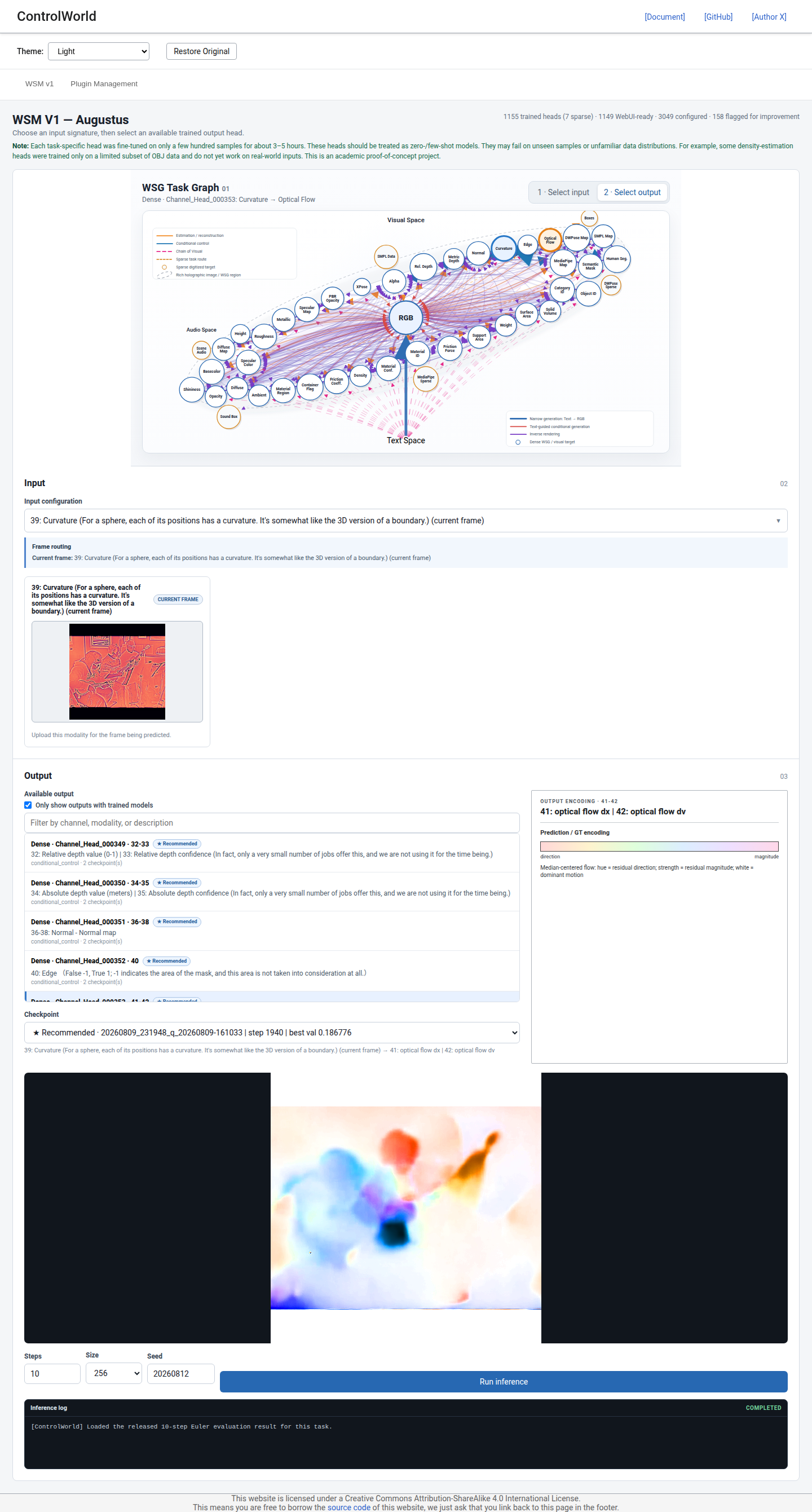}
    \vspace{-4pt}
    {\small (b) \textbf{TunnelingNet 000353:} Curve $\rightarrow$ Flow.}
  \end{minipage}
  \caption{ControlWorld desktop workflows for Qualitative Part IX: TunnelingNet 000275 (Normal to Edge) and TunnelingNet 000353 (Curve to Flow). Each panel preserves the full task graph, input, sampling, and decoded-output interface.}
  \label{fig:controlworld-guide-ix-09}
\end{figure*}

\begin{figure*}[p]
  \centering
  \begin{minipage}[t]{0.492\textwidth}
    \centering
    \includegraphics[width=\linewidth,height=1.87708\linewidth]{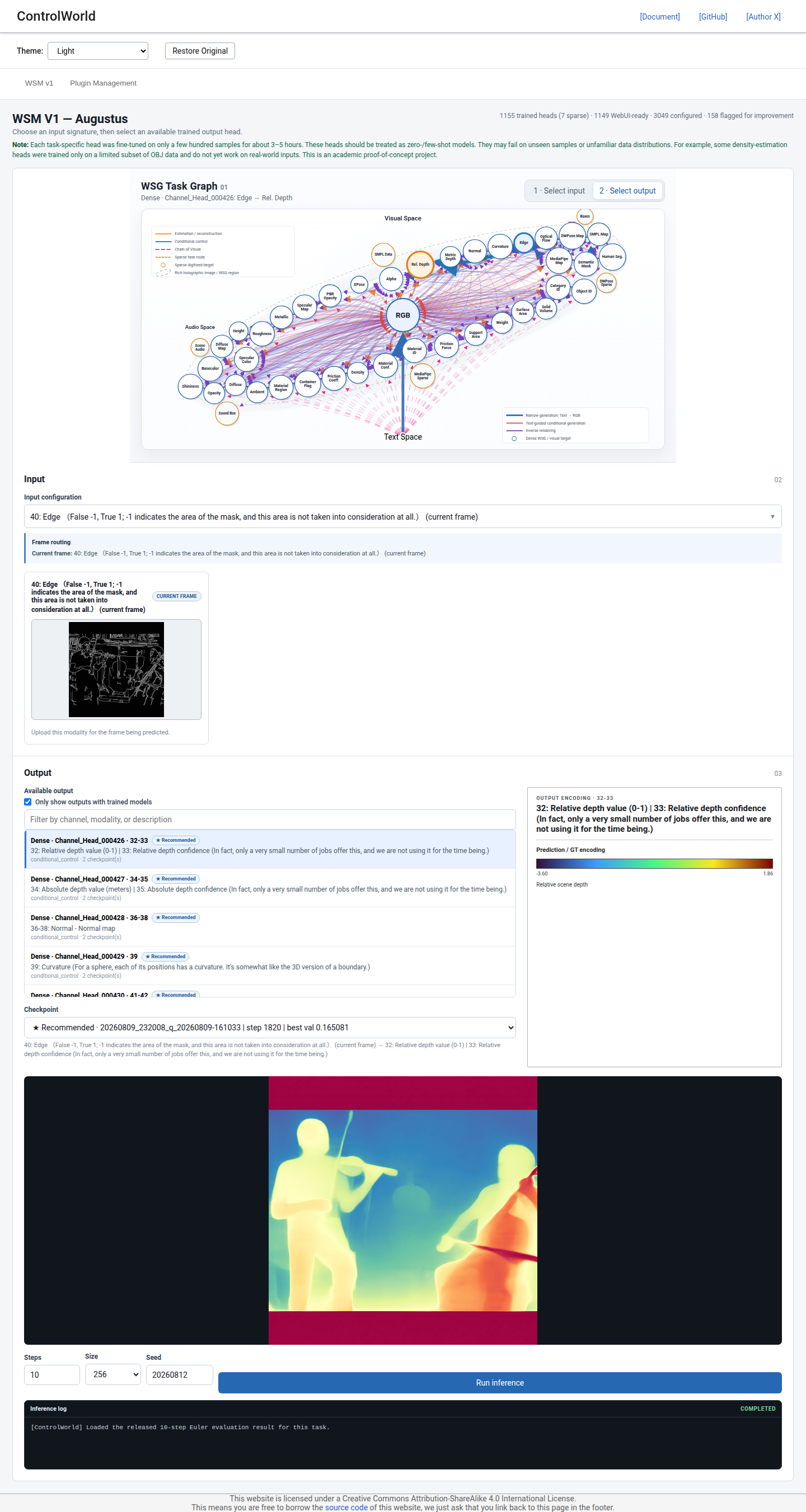}
    \vspace{-4pt}
    {\small (a) \textbf{TunnelingNet 000426:} Edge $\rightarrow$ RelDepth.}
  \end{minipage}
  \hfill
  \begin{minipage}[t]{0.492\textwidth}
    \centering
    \includegraphics[width=\linewidth,height=1.87708\linewidth]{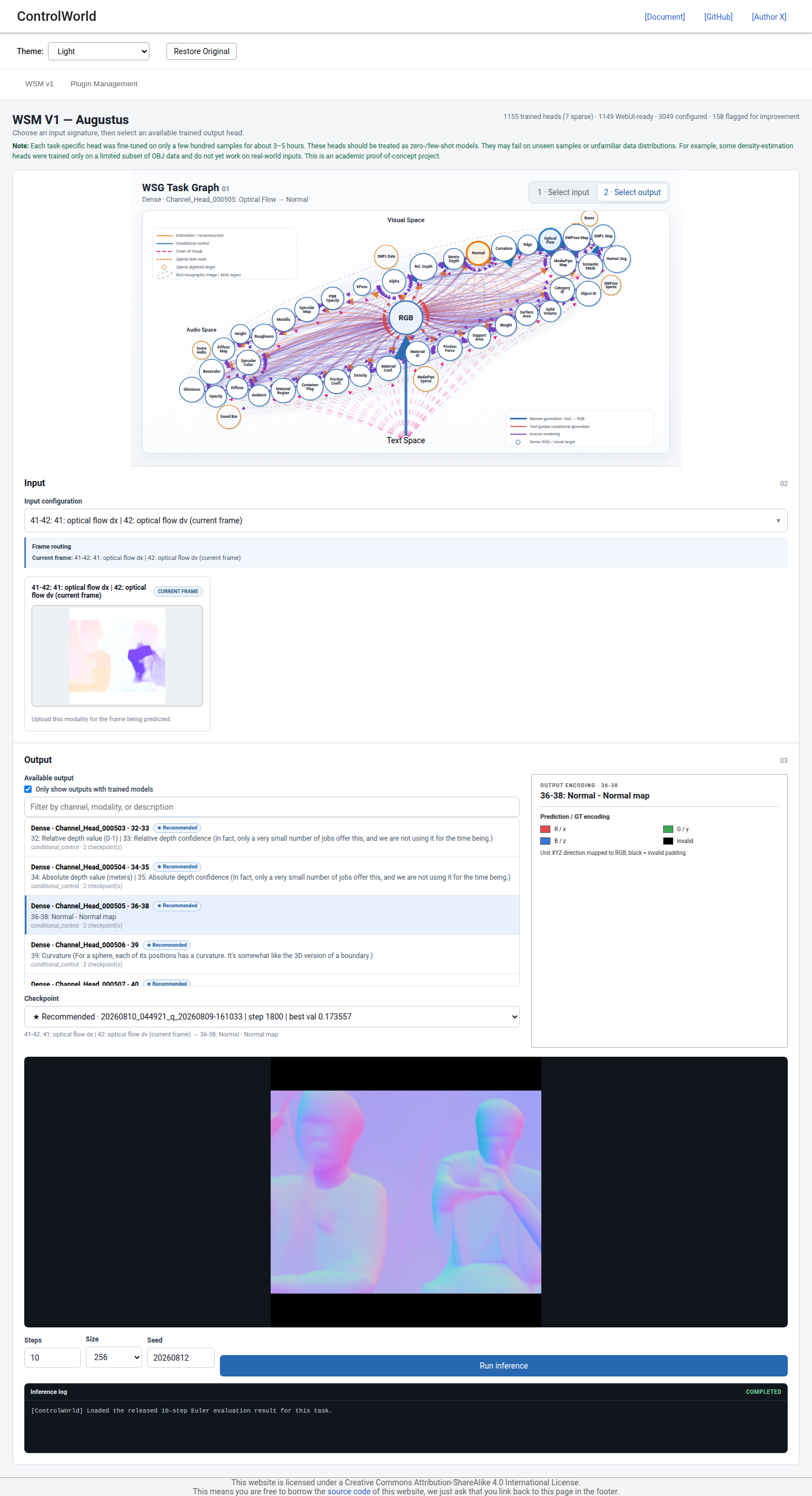}
    \vspace{-4pt}
    {\small (b) \textbf{TunnelingNet 000505:} Flow $\rightarrow$ Normal.}
  \end{minipage}
  \caption{ControlWorld desktop workflows for Qualitative Part IX: TunnelingNet 000426 (Edge to RelDepth) and TunnelingNet 000505 (Flow to Normal). Each panel preserves the full task graph, input, sampling, and decoded-output interface.}
  \label{fig:controlworld-guide-ix-10}
\end{figure*}

\begin{figure*}[p]
  \centering
  \begin{minipage}[t]{0.492\textwidth}
    \centering
    \includegraphics[width=\linewidth,height=1.87708\linewidth]{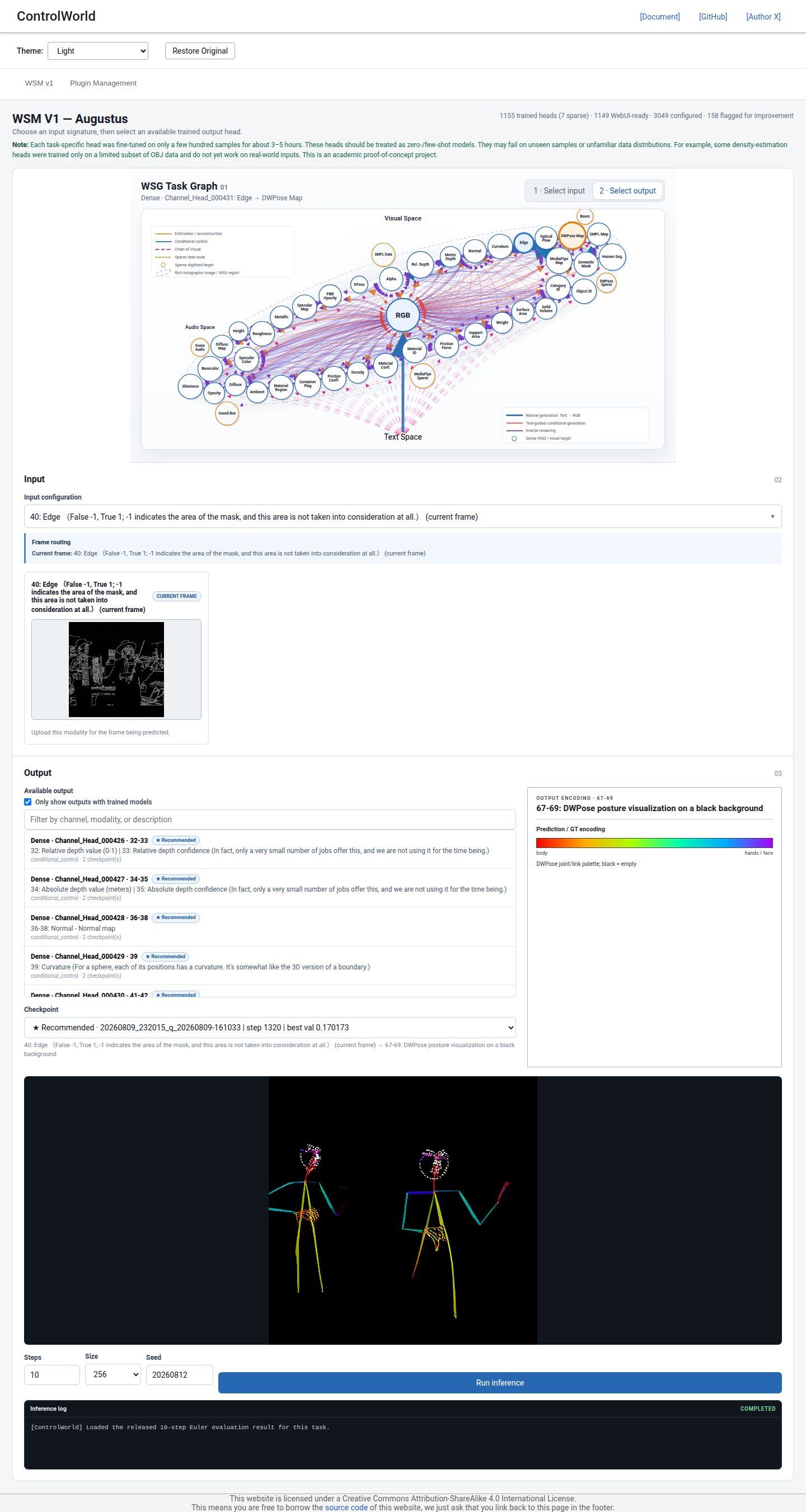}
    \vspace{-4pt}
    {\small (a) \textbf{TunnelingNet 000431:} Edge $\rightarrow$ DWPose.}
  \end{minipage}
  \hfill
  \begin{minipage}[t]{0.492\textwidth}
    \centering
    \includegraphics[width=\linewidth,height=1.87708\linewidth]{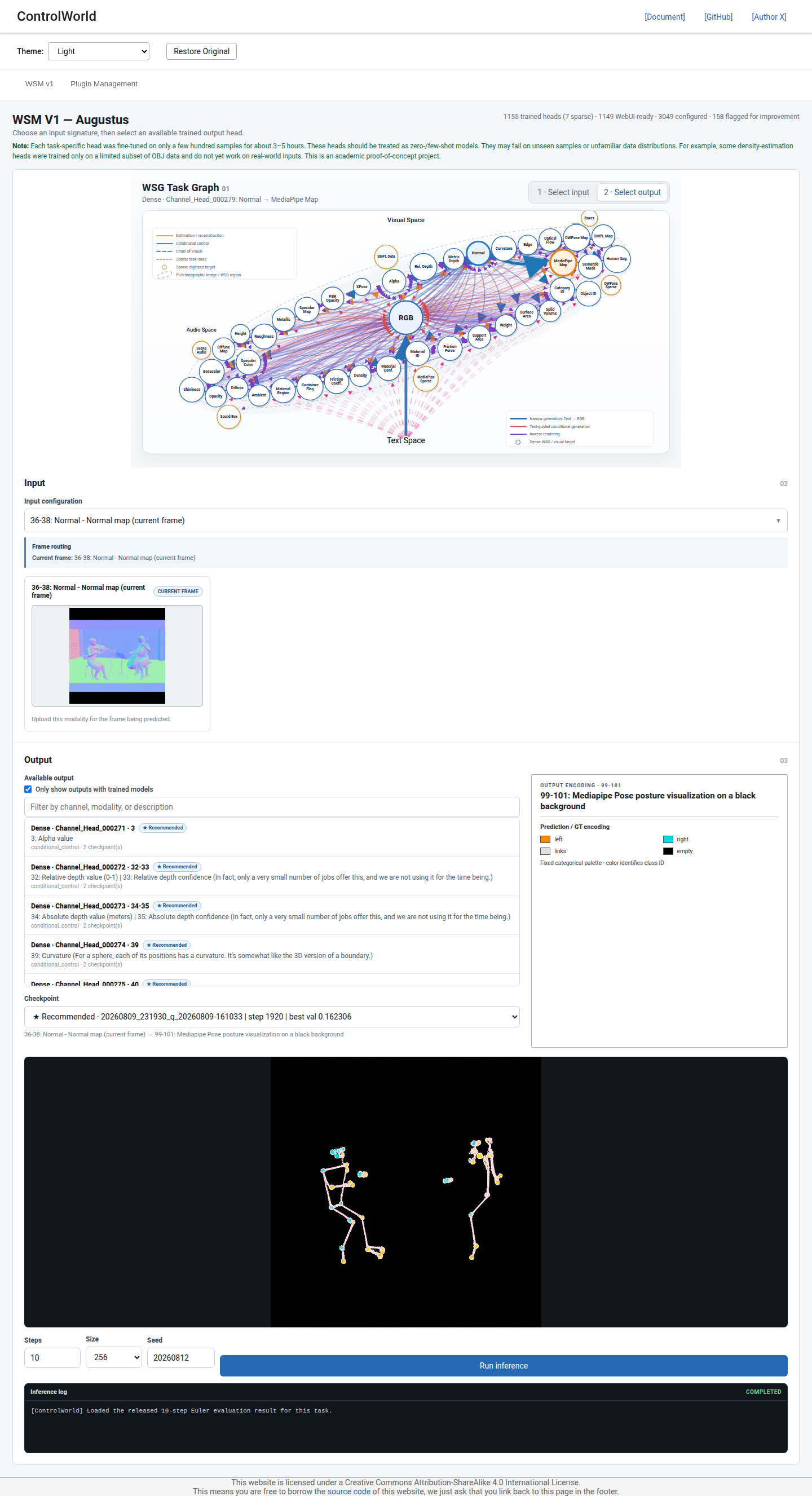}
    \vspace{-4pt}
    {\small (b) \textbf{TunnelingNet 000279:} Normal $\rightarrow$ MPPose.}
  \end{minipage}
  \caption{ControlWorld desktop workflows for Qualitative Part IX: TunnelingNet 000431 (Edge to DWPose) and TunnelingNet 000279 (Normal to MPPose). Each panel preserves the full task graph, input, sampling, and decoded-output interface.}
  \label{fig:controlworld-guide-ix-11}
\end{figure*}

\begin{figure*}[p]
  \centering
  \begin{minipage}[t]{0.492\textwidth}
    \centering
    \includegraphics[width=\linewidth,height=1.87292\linewidth]{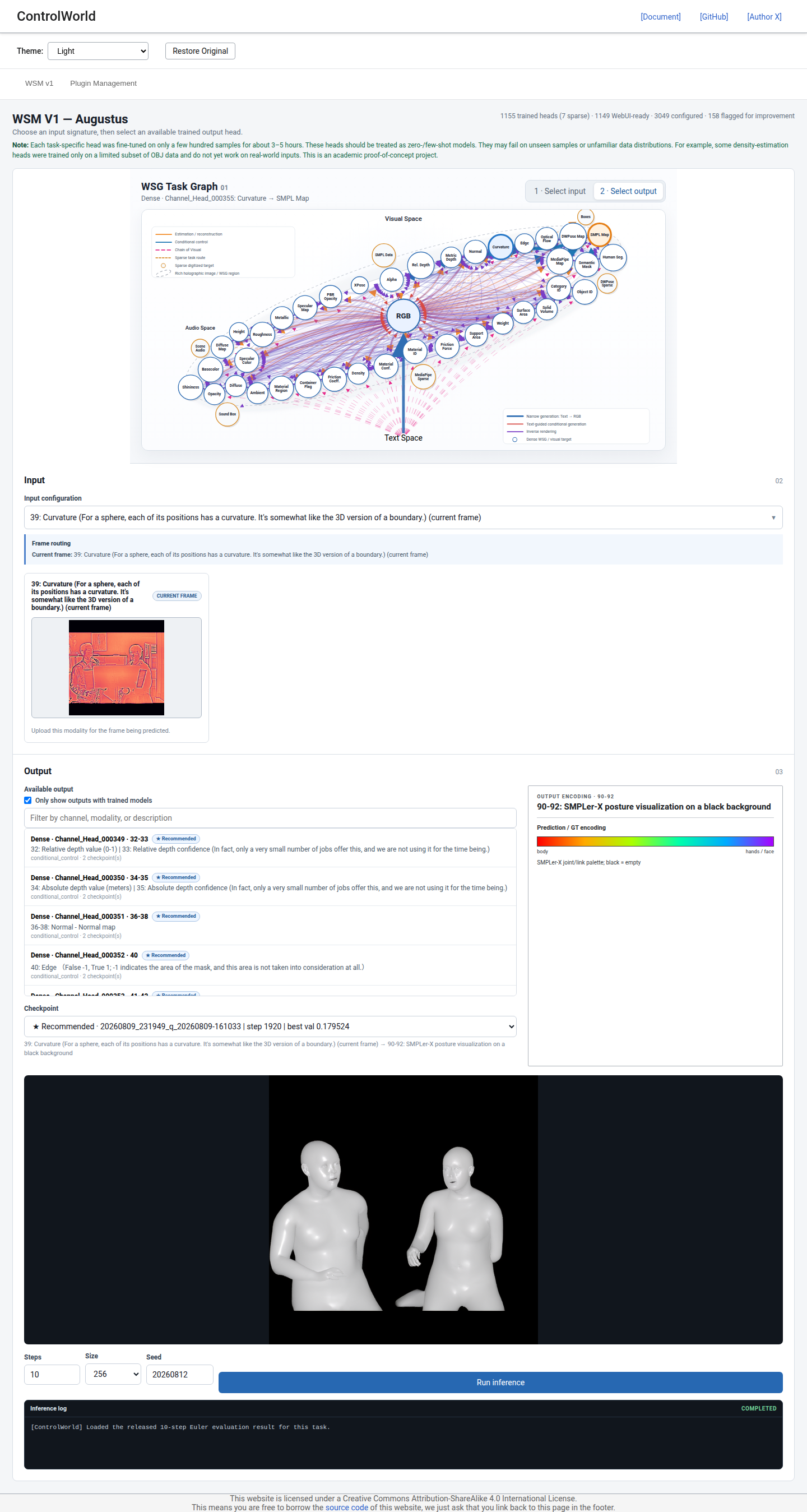}
    \vspace{-4pt}
    {\small (a) \textbf{TunnelingNet 000355:} Curve $\rightarrow$ SMPL-X.}
  \end{minipage}
  \hfill
  \begin{minipage}[t]{0.492\textwidth}
    \centering
    \includegraphics[width=\linewidth,height=1.87292\linewidth]{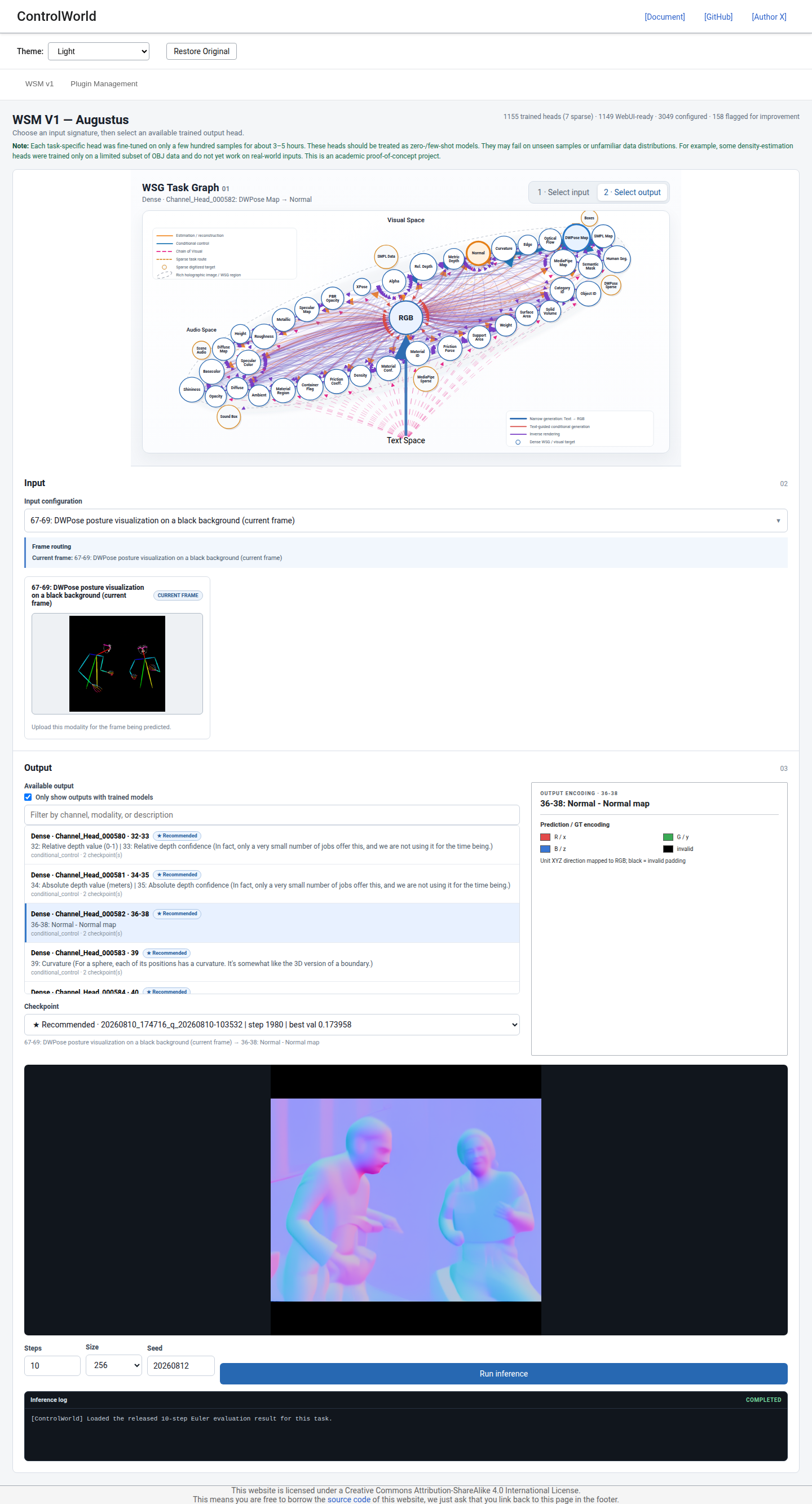}
    \vspace{-4pt}
    {\small (b) \textbf{TunnelingNet 000582:} DWPose $\rightarrow$ Normal.}
  \end{minipage}
  \caption{ControlWorld desktop workflows for Qualitative Part IX: TunnelingNet 000355 (Curve to SMPL-X) and TunnelingNet 000582 (DWPose to Normal). Each panel preserves the full task graph, input, sampling, and decoded-output interface.}
  \label{fig:controlworld-guide-ix-12}
\end{figure*}

\begin{figure*}[p]
  \centering
  \begin{minipage}[t]{0.492\textwidth}
    \centering
    \includegraphics[width=\linewidth,height=1.86319\linewidth]{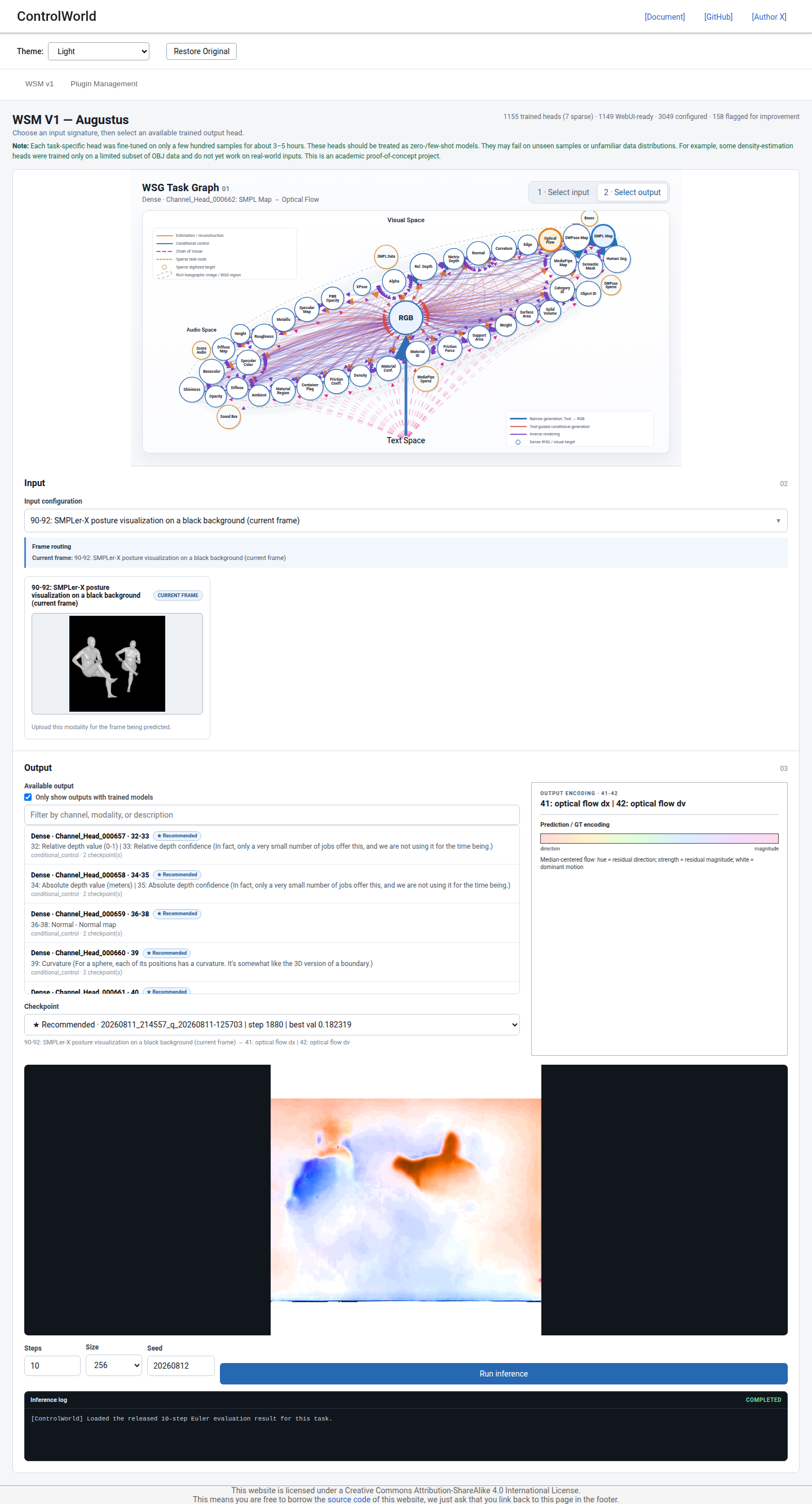}
    \vspace{-4pt}
    {\small (a) \textbf{TunnelingNet 000662:} SMPL-X $\rightarrow$ Flow.}
  \end{minipage}
  \hfill
  \begin{minipage}[t]{0.492\textwidth}
    \centering
    \includegraphics[width=\linewidth,height=1.86319\linewidth]{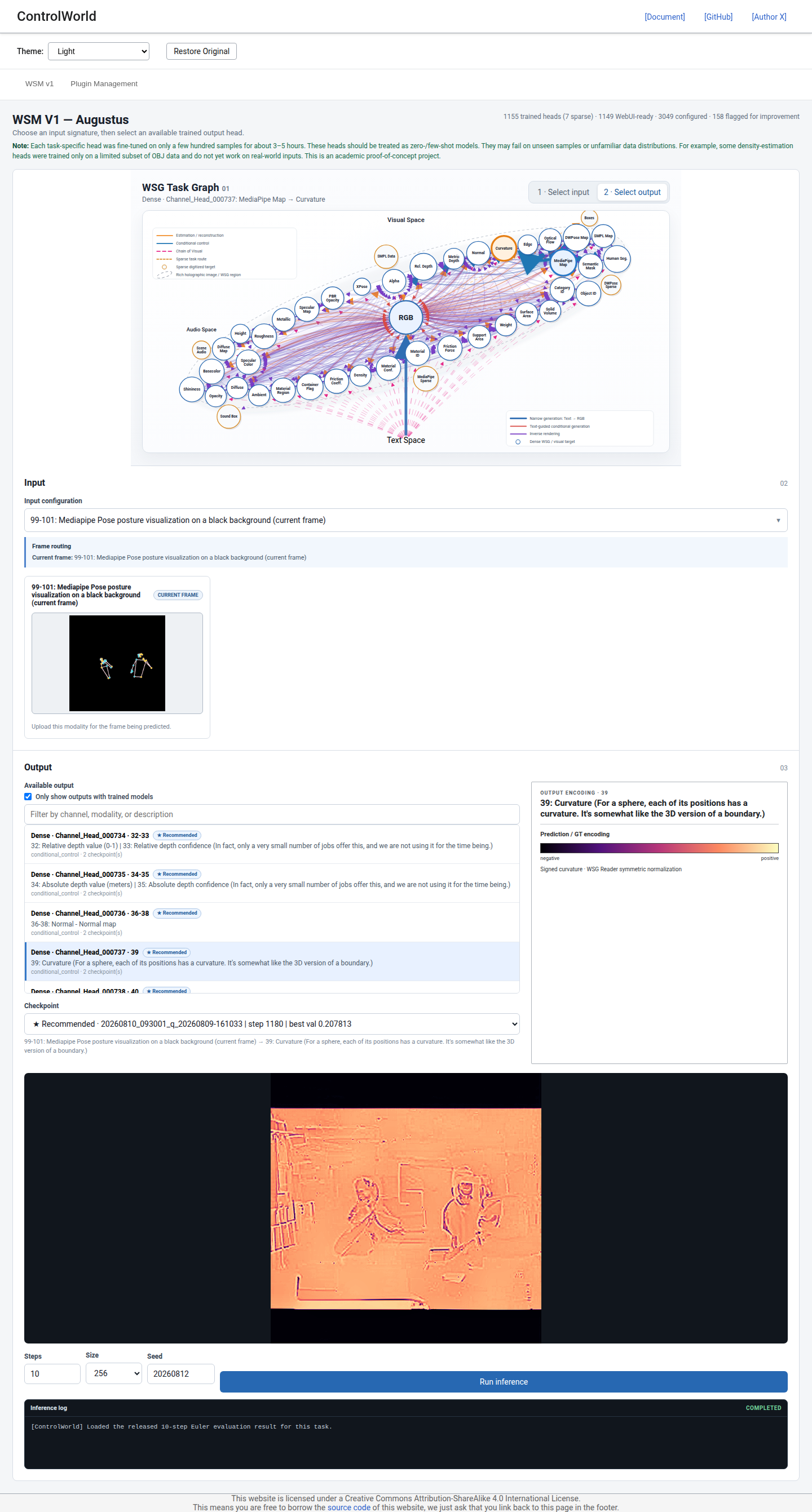}
    \vspace{-4pt}
    {\small (b) \textbf{TunnelingNet 000737:} MPPose $\rightarrow$ Curve.}
  \end{minipage}
  \caption{ControlWorld desktop workflows for Qualitative Part IX: TunnelingNet 000662 (SMPL-X to Flow) and TunnelingNet 000737 (MPPose to Curve). Each panel preserves the full task graph, input, sampling, and decoded-output interface.}
  \label{fig:controlworld-guide-ix-13}
\end{figure*}

\begin{figure*}[p]
  \centering
  \begin{minipage}[t]{0.492\textwidth}
    \centering
    \includegraphics[width=\linewidth,height=1.85208\linewidth]{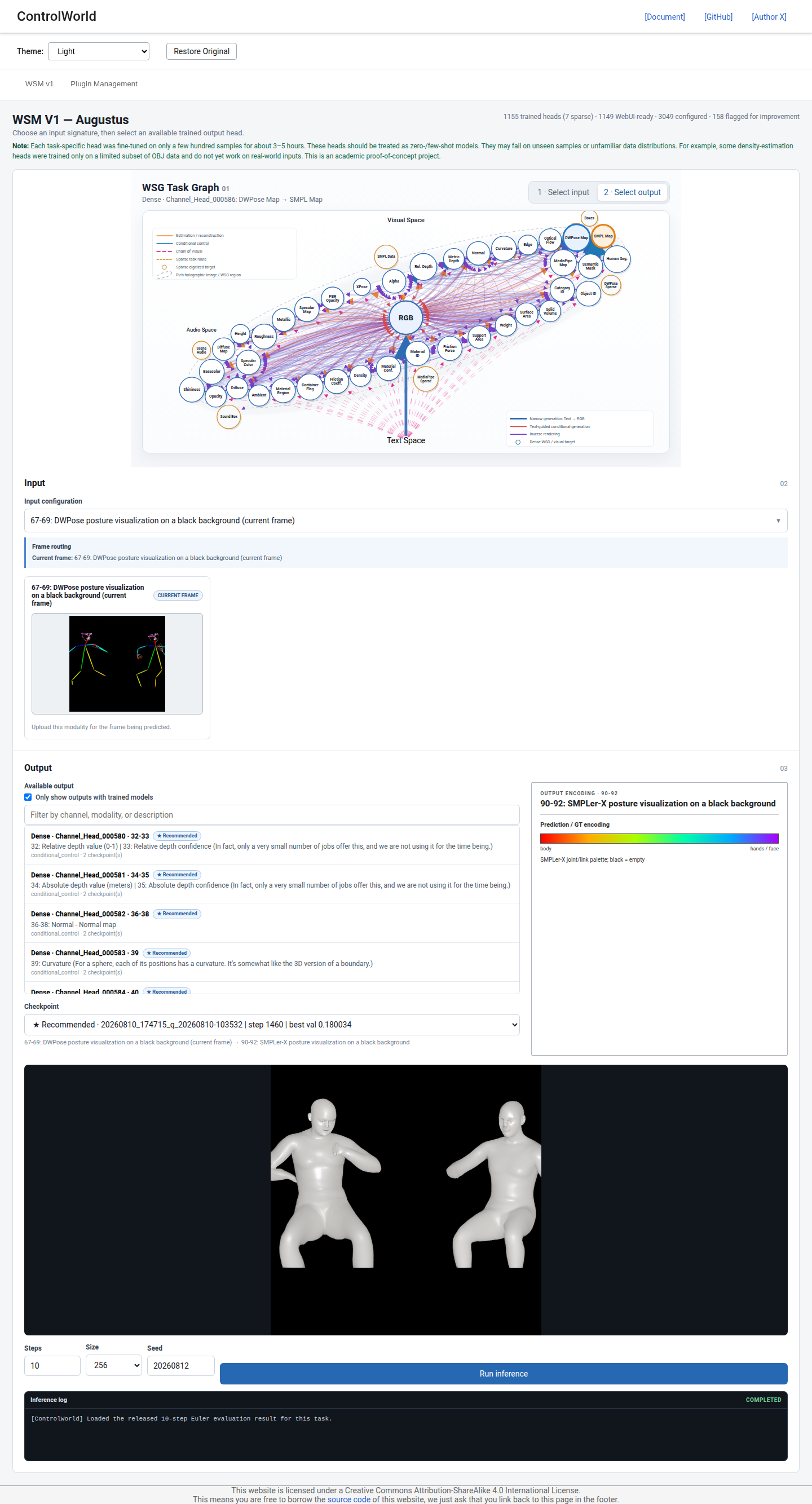}
    \vspace{-4pt}
    {\small (a) \textbf{TunnelingNet 000586:} DWPose $\rightarrow$ SMPL-X.}
  \end{minipage}
  \hfill
  \begin{minipage}[t]{0.492\textwidth}
    \centering
    \includegraphics[width=\linewidth,height=1.85208\linewidth]{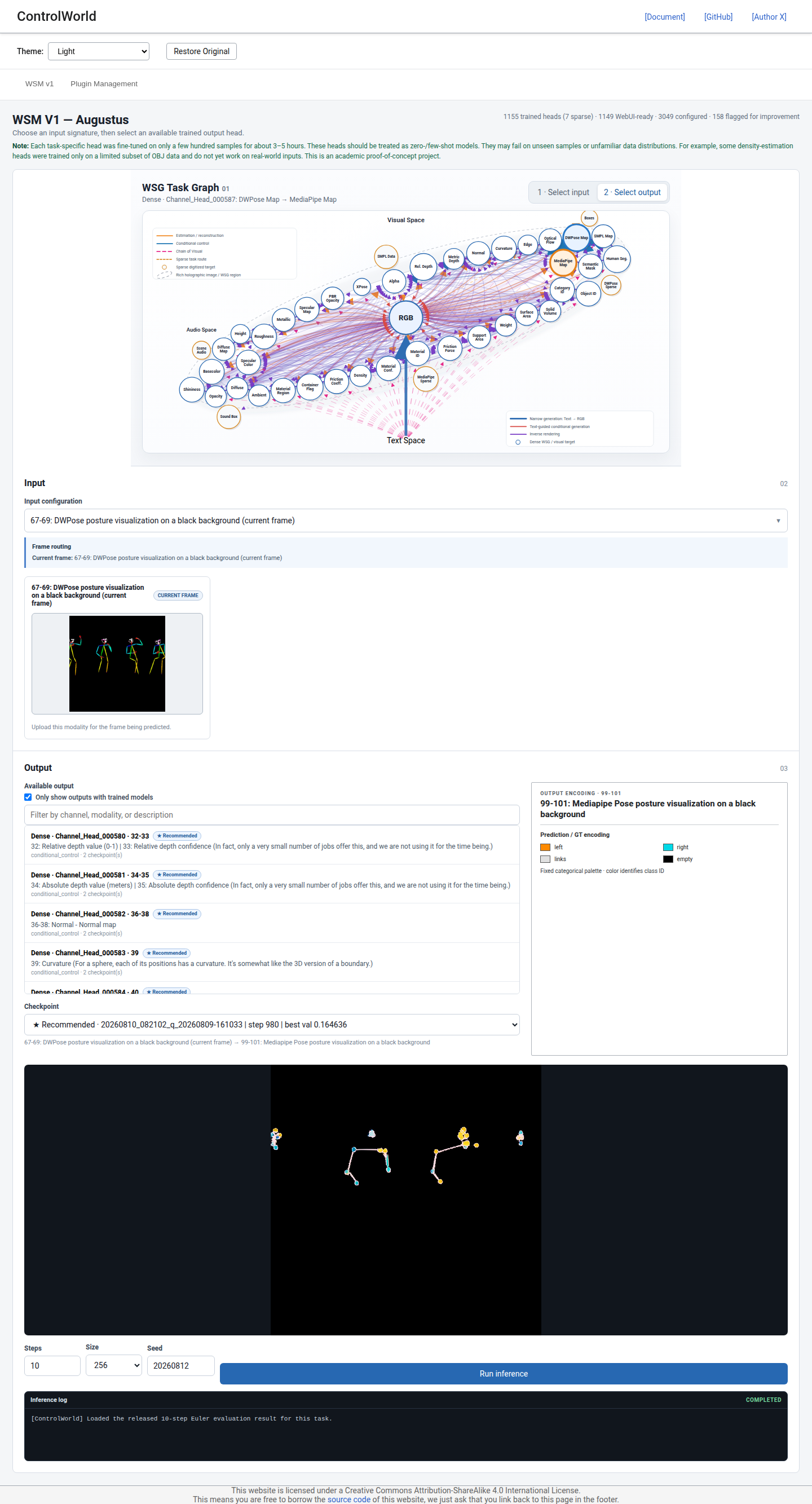}
    \vspace{-4pt}
    {\small (b) \textbf{TunnelingNet 000587:} DWPose $\rightarrow$ MPPose.}
  \end{minipage}
  \caption{ControlWorld desktop workflows for Qualitative Part IX: TunnelingNet 000586 (DWPose to SMPL-X) and TunnelingNet 000587 (DWPose to MPPose). Each panel preserves the full task graph, input, sampling, and decoded-output interface.}
  \label{fig:controlworld-guide-ix-14}
\end{figure*}

\begin{figure*}[p]
  \centering
  \begin{minipage}[t]{0.492\textwidth}
    \centering
    \includegraphics[width=\linewidth,height=1.85208\linewidth]{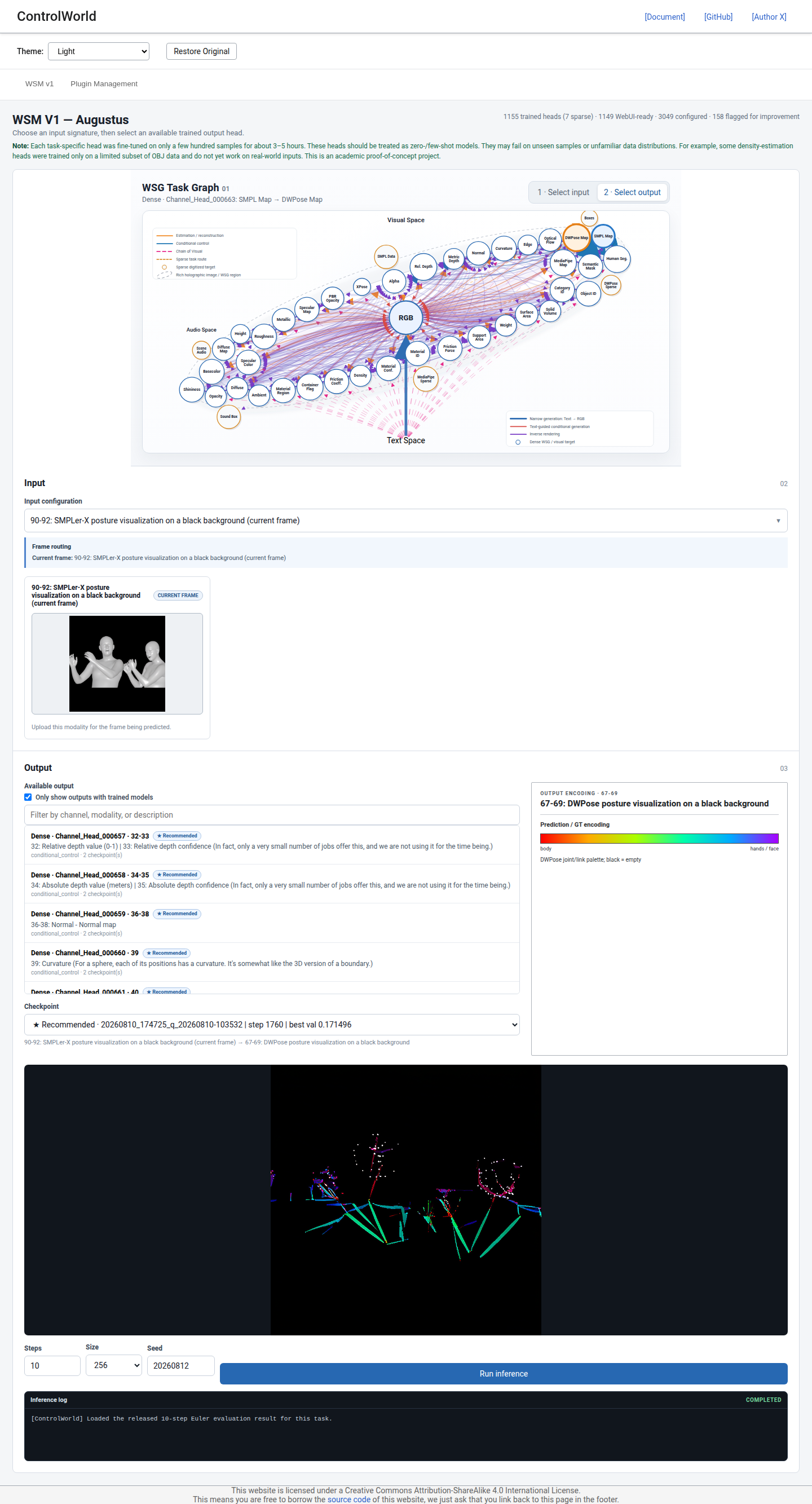}
    \vspace{-4pt}
    {\small (a) \textbf{TunnelingNet 000663:} SMPL-X $\rightarrow$ DWPose.}
  \end{minipage}
  \hfill
  \begin{minipage}[t]{0.492\textwidth}
    \centering
    \includegraphics[width=\linewidth,height=1.85208\linewidth]{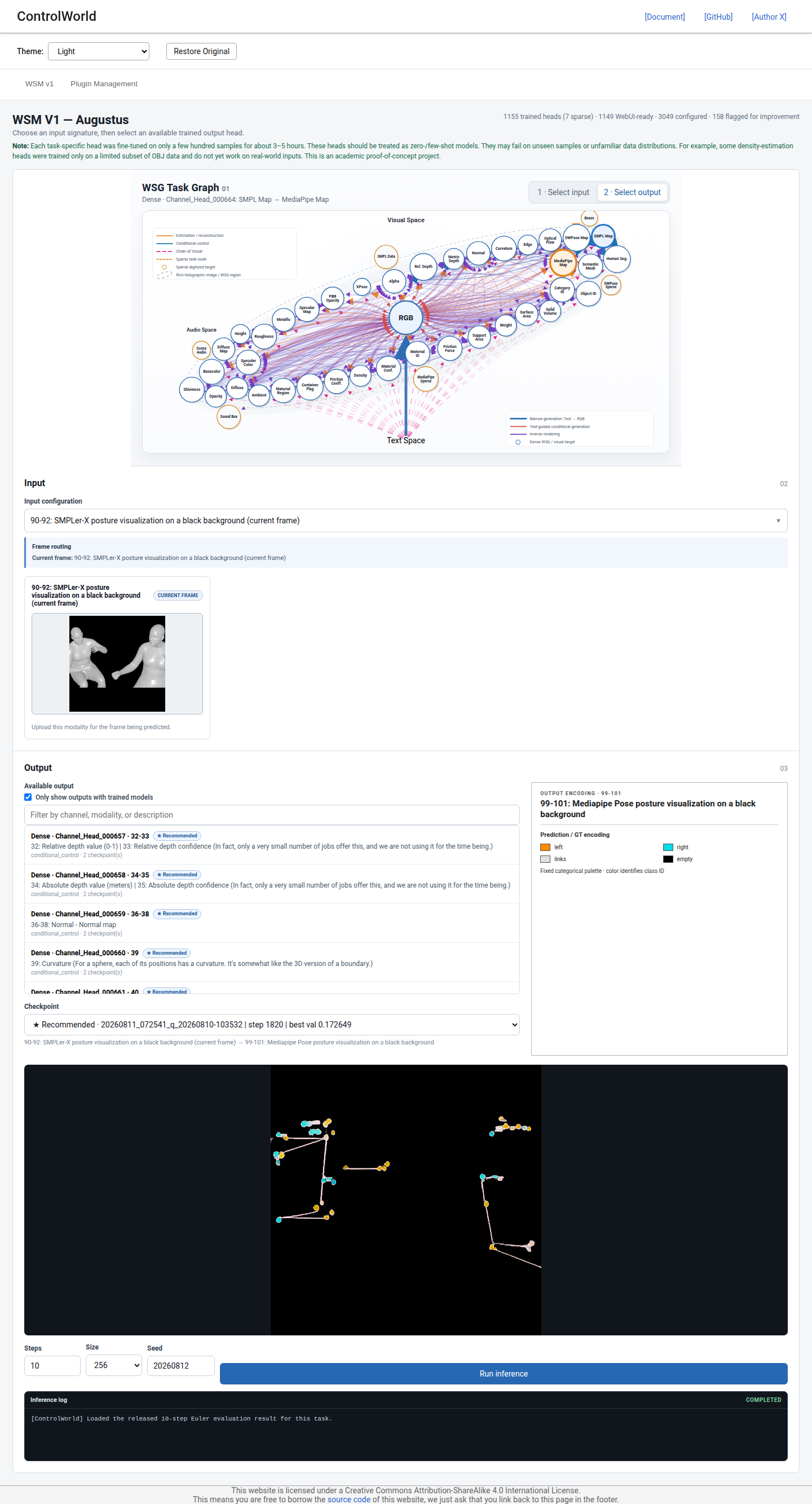}
    \vspace{-4pt}
    {\small (b) \textbf{TunnelingNet 000664:} SMPL-X $\rightarrow$ MPPose.}
  \end{minipage}
  \caption{ControlWorld desktop workflows for Qualitative Part IX: TunnelingNet 000663 (SMPL-X to DWPose) and TunnelingNet 000664 (SMPL-X to MPPose). Each panel preserves the full task graph, input, sampling, and decoded-output interface.}
  \label{fig:controlworld-guide-ix-15}
\end{figure*}

\begin{figure*}[p]
  \centering
  \begin{minipage}[t]{0.492\textwidth}
    \centering
    \includegraphics[width=\linewidth,height=1.85208\linewidth]{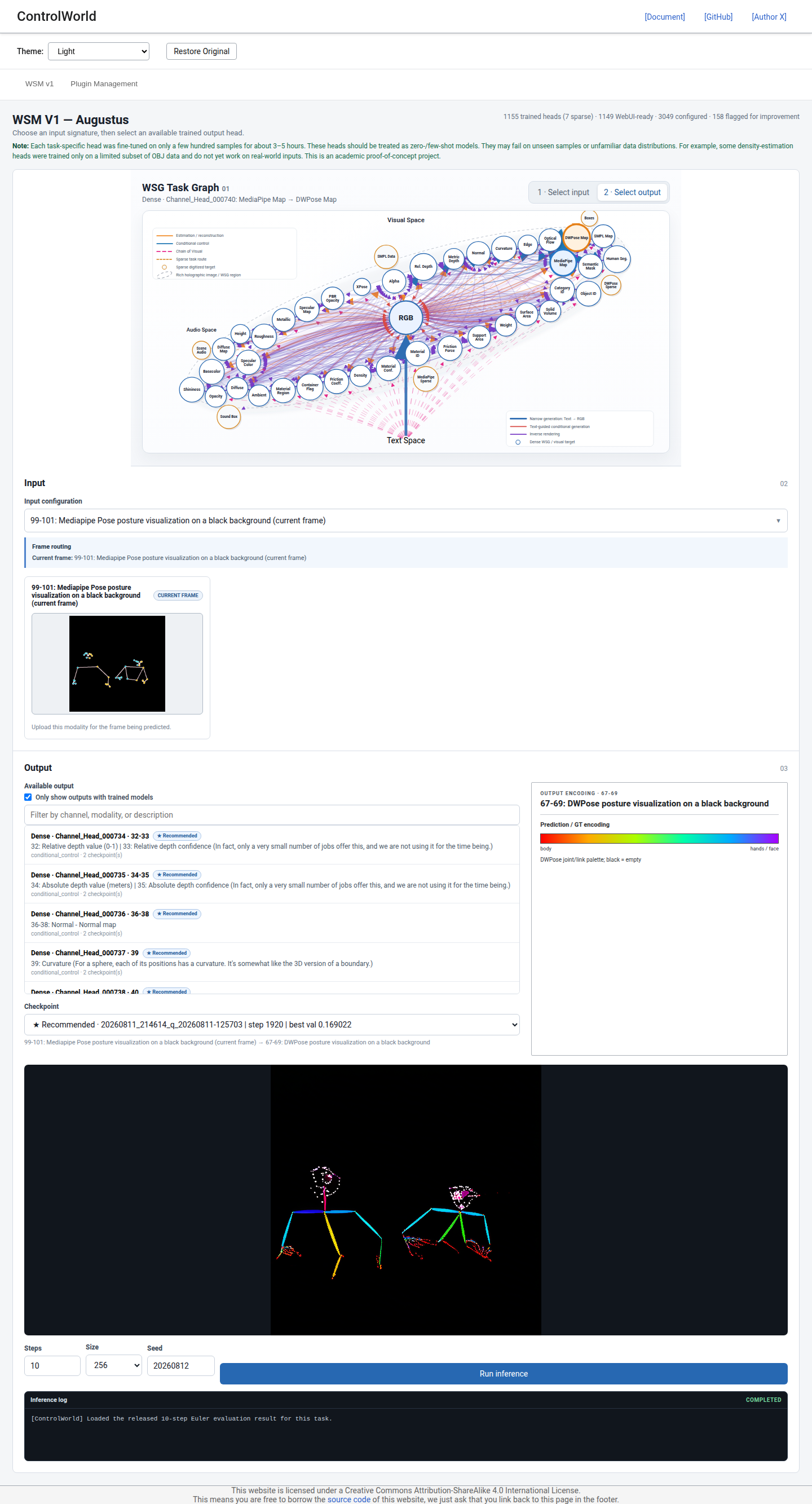}
    \vspace{-4pt}
    {\small (a) \textbf{TunnelingNet 000740:} MPPose $\rightarrow$ DWPose.}
  \end{minipage}
  \hfill
  \begin{minipage}[t]{0.492\textwidth}
    \centering
    \includegraphics[width=\linewidth,height=1.85208\linewidth]{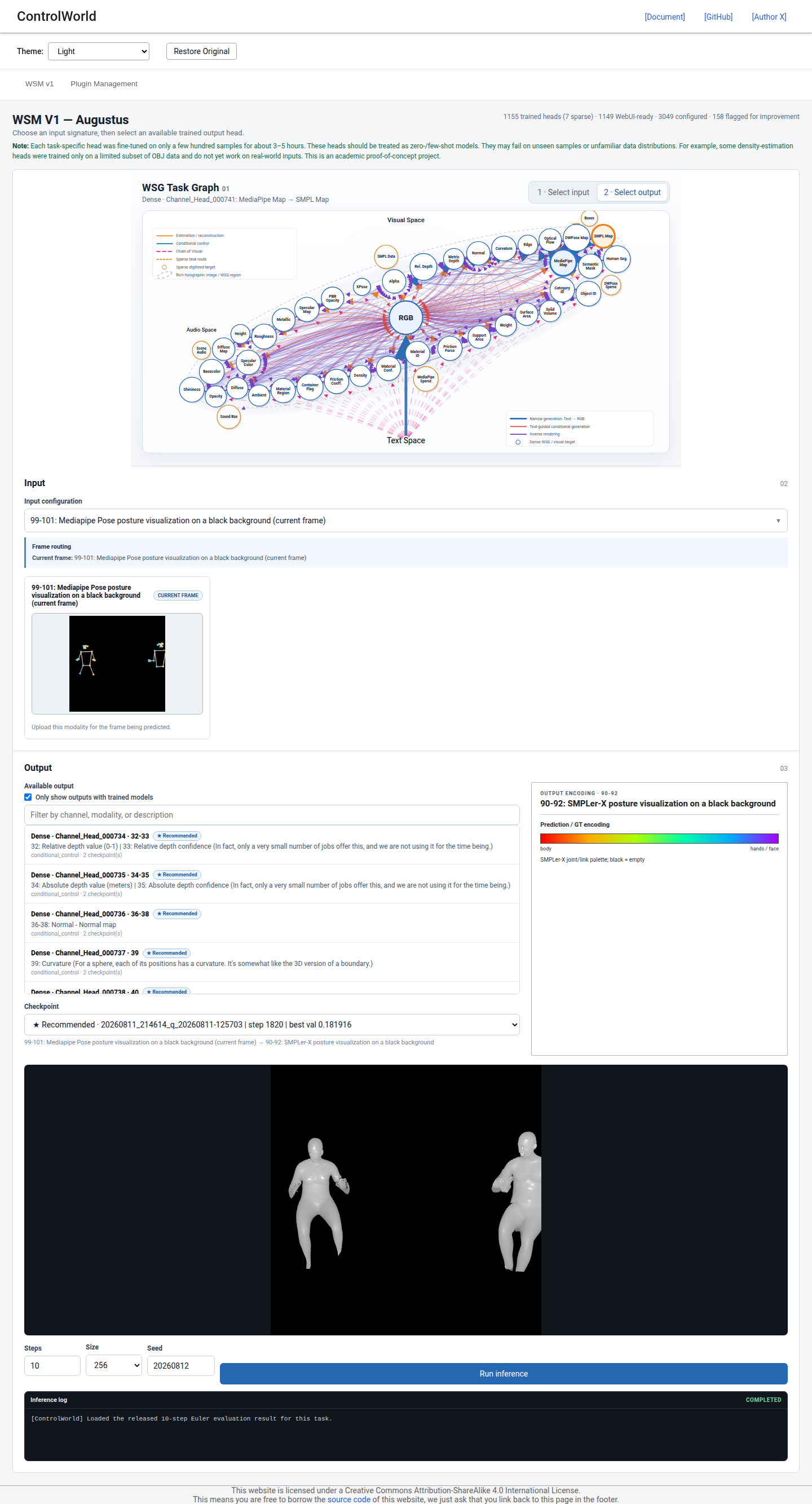}
    \vspace{-4pt}
    {\small (b) \textbf{TunnelingNet 000741:} MPPose $\rightarrow$ SMPL-X.}
  \end{minipage}
  \caption{ControlWorld desktop workflows for Qualitative Part IX: TunnelingNet 000740 (MPPose to DWPose) and TunnelingNet 000741 (MPPose to SMPL-X). Each panel preserves the full task graph, input, sampling, and decoded-output interface.}
  \label{fig:controlworld-guide-ix-16}
\end{figure*}

\FloatBarrier
\clearpage
\flushbottom

%% file: tables/pwc_task_scope/pwc_cv_task_scope_summary.tex
\begin{tabularx}{\columnwidth}{Xrr}
\toprule
Difficulty assessment & Tasks & Archive share \\
\midrule
Difficulty A: structured development via WSM sparse mode & 705 & 36.2\% \\
Difficulty B: minor interface, loss, label, or decoder adaptation & 286 & 14.7\% \\
Difficulty C: aligned-modality data and training & 255 & 13.1\% \\
Difficulty D: directly compatible application or scenario & 231 & 11.9\% \\
Non-visual archive cross-listing & 470 & 24.1\% \\
\midrule
Total archived CV entries & 1,947 & 100.0\% \\
\bottomrule
\end{tabularx}

%% file: tables/pwc_task_scope/wsm_historical_overlap_summary.tex
\begin{tabularx}{\columnwidth}{Xrr}
\toprule
Historical-overlap judgment & Tasks & Share of 1,072 \\
\midrule
No historical full-contract match (unambiguously new) & 655 & 61.1\% \\
Historical umbrella is underspecified (conservative ambiguity) & 401 & 37.4\% \\
Clear historical full-contract overlap & 16 & 1.5\% \\
\bottomrule
\end{tabularx}

%% file: sec_preprint/X_tutorial.tex
\clearpage
\onecolumn
\twocolumn
\raggedbottom
\section{Appendix: Tutorial}
\label{sec:supp_tutorial}

% Local visual hierarchy reconstructed from the original tutorial correspondence.
% Optional subsection titles keep the table of contents and PDF bookmarks plain.
\definecolor{TutorialRed}{HTML}{FF0000}
\definecolor{TutorialBlue}{HTML}{0000FF}
\definecolor{TutorialGreen}{HTML}{FF0000}
\newcommand{\TutorialKey}[1]{\textcolor{TutorialRed}{\textbf{#1}}}
\newcommand{\TutorialTerm}[1]{\textcolor{TutorialBlue}{\textbf{#1}}}
\newcommand{\TutorialResult}[1]{\textcolor{TutorialGreen}{\textbf{#1}}}
\newcommand{\TutorialEmph}[1]{\textbf{#1}}

This tutorial is written for AI beginners and assumes no prior knowledge of
multimodal learning or multimodal capability extension. It begins with the basic
problem, introduces the major extension paradigms in a shared visual language,
and then explains the relationship among WSG, WSM, and TunnelingNet. The goal is
to let a new reader understand both the field and the historical significance of
organizing visual capabilities as a large graph of directed, independently
manageable, and composable tasks.

We view this appendix as the opening module of a broader, multi-course and
multi-level tutorial program. Introductory modules establish the foundations of
AI and multimodal learning; intermediate modules compare capability-extension
paradigms; and advanced modules study task graphs, trajectory redirection,
large-scale evaluation, and the construction of general-purpose multimodal
systems. The intended path therefore extends from a reader's first encounter
with the field to its research frontier.

In this sense, the tutorial is also a pre-textbook draft: an initial account of
the concepts, taxonomy, and system design that may eventually enter textbooks as
this area matures.

\subsection{Background: Two AI Fields, Two Different Outcomes}

Two of the largest areas of AI are computer vision (CV) and natural-language
processing (NLP). CV has approximately 1,400 tasks, while NLP has approximately
900 tasks.

With the emergence and large-scale commercialization of large language models,
a user can enter a few sentences and ask one model to perform essentially any
NLP task in a prompt--answer form. In this sense, \TutorialEmph{NLP tasks have been unified}
under a common model paradigm.

In general, if the input or output of a task contains a visual element, it can
be classified as a visual task. Unlike NLP, however, visual-task forms and data
formats are complex and cannot be reduced to a simple prompt--answer format.

Visual-task forms include classification, localization, detection,
segmentation, estimation, reconstruction, tracking, editing, rendering,
simulation, planning, and many others.

Visual data also have many different distributions. An image is only a way of
visualizing data, not a universal representation of data. Some visual data even
have special abstract constraints and specifications.

Text, by contrast, is still text. Even relatively structured information or
code in formats such as SQL, Word, and Excel remains text or a string.

As a result, \TutorialEmph{computer vision still lacks a general method or single model} that
can easily perform all of its tasks. Since the emergence of large language
models, this gap has become increasingly consequential for the direction of the
field and for the work of computer-vision researchers.

\subsection{Three Goals of Visual-Task Unification}

Our work therefore addresses \TutorialKey{three goals}, with three corresponding solutions:

\begin{enumerate}
    \item \textbf{Unifying visual-task forms.} This is addressed by the Unified
    Law, a conceptual solution that organizes visual tasks into four generative
    interfaces.
    \item \textbf{Unifying visual input and output formats.} This is addressed
    by World Snapshot Models and World Snapshot Graphics. WSM provides four
    training interfaces and their corresponding strategies, while WSG unifies
    the data format.
    \item \textbf{Unifying training objectives across visual tasks.} This is
    addressed by TunnelingNet. Every visual task takes a generative form and
    produces each type of visual or world information in the form appropriate
    to that task, while only slightly deflecting the original generation path.
\end{enumerate}

The three solutions form one connected sequence: the Unified Law defines the
shared task form; WSM and WSG define the interfaces and world-information
representation; and TunnelingNet provides the common learning mechanism.

\subsection{Why Large-Scale Visual Unification Remains Difficult}

The approximately 900 NLP tasks have relatively consistent task forms,
training objectives, and formats. Visual tasks take far more diverse forms;
their objectives can point in different or even opposite directions, and their
formats are inconsistent. Even so, vision researchers have attempted several
degrees of unification:

\begin{enumerate}
    \item Some methods train visual any-to-any systems from scratch or predict
    visual information as an LLM predicts text through end-to-end multimodal
    learning. The number of modalities grows slowly because the required data
    are enormous and can increase almost exponentially. Adding a new visual
    target may also interfere with existing capabilities.
    \item Training one model from scratch to predict all kinds of visual
    information usually fails to scale. Text tasks can reinforce one another
    because their format is uniform, whereas visual targets can be different or
    substantially conflicting.
    \item Under these conditions, existing systems usually dedicate one model
    to one visual task or unify only a comparatively small group of tasks under
    one method. \TutorialEmph{No existing visual model has unified thousands of visual tasks.}
\end{enumerate}

In other words, the visual world contains more task types, yet those tasks are
harder to unify. The textual world reached a common large-model paradigm first;
computer vision did not.

\subsection{Current Methods and a Sixth Paradigm}

In practice, stronger visual capabilities are usually built on an existing
backbone through one of \TutorialKey{five major extension paradigms}:

\begin{enumerate}
    \item \textbf{ControlNet:} add a new backbone-sized branch and train it to
    control an existing model;
    \item \textbf{LoRA:} modify a small number of low-rank parameters to change
    the behavior of the overall model;
    \item \textbf{MoE:} add knowledge or computation blocks and route inputs
    among them;
    \item \textbf{IP-Adapter:} train an adapter that coordinates conditioning
    components inside a model; and
    \item \textbf{IC-Control-style methods:} organize text and visual elements
    in one context so that different visual elements can interact.
\end{enumerate}

These five paradigms commonly extend or improve a particular visual capability
and can be applied to many kinds of visual generation backbones. They describe
where a capability is added and how it influences the existing generation
process.

\TutorialKey{Our approach constitutes a sixth paradigm.} All six can build on a generative
backbone, but they differ in where a visual capability is added, how it changes
generation, and what constitutes one unit of visual-task expansion.

\begin{figure*}[t]
    \centering
    \includegraphics[width=0.98\textwidth]{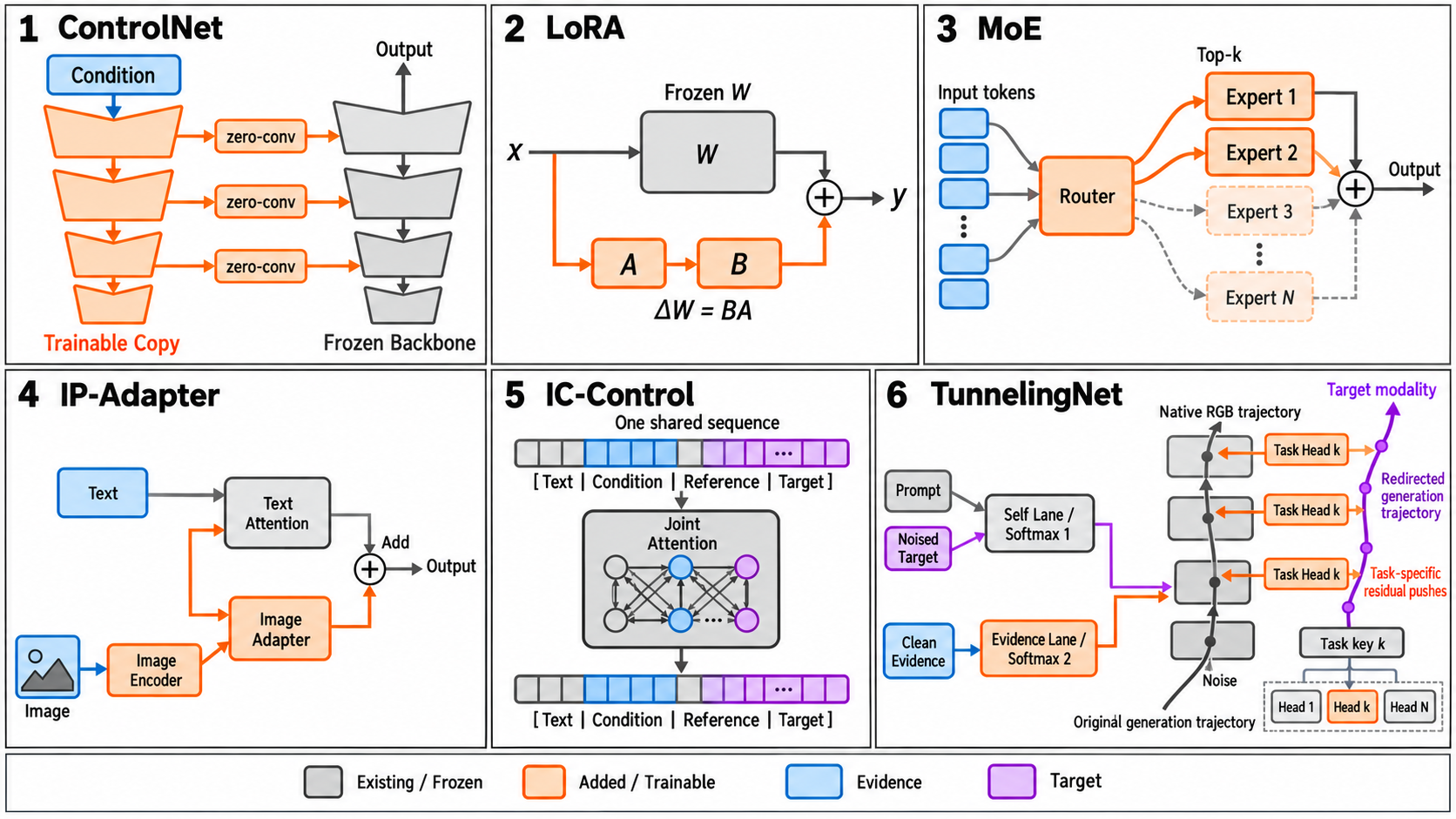}
    \caption{\textbf{A tutorial comparison of six paradigms for extending visual
    capabilities.} ControlNet adds a copied control body; LoRA adds low-rank
    weight updates; MoE selects expert computation; IP-Adapter adds an image
    conditioning branch; IC-Control-style methods organize conditions,
    references, and targets in a shared context; TunnelingNet applies
    task-specific residual pushes at multiple depths to redirect a frozen
    backbone's native RGB trajectory toward a selected target modality.}
    \label{fig:tutorial_six_paradigms}
\end{figure*}

Figure~\ref{fig:tutorial_six_paradigms} uses one visual notation to compare six
paradigms. Gray denotes existing or frozen computation, orange denotes added or
trainable computation, blue denotes condition evidence, and purple denotes the
generation target. The figure is deliberately simplified: it shows the defining
information route of each paradigm rather than every implementation detail.

\noindent\TutorialTerm{ControlNet.}
ControlNet~\cite{zhang2023adding} copies a large part of a pretrained generative
network and injects its features into the frozen backbone at multiple depths.
This produces strong and mature spatial control. Its defining unit of extension
is a substantial control body associated with a condition, usually while the
generation target remains RGB. Scaling to many independently maintained
capabilities therefore adds a correspondingly large collection of control
structures.

\noindent\TutorialTerm{LoRA.}
LoRA~\cite{hu2022lora} expresses a model update through low-rank matrices. It is
parameter efficient, general, easy to train, and easy to distribute. A LoRA can
learn a cross-modal conversion when its training setup provides the appropriate
inputs and targets. However, LoRA itself specifies how weights change; it does
not by itself define which world information is the condition, which modality
is the target, how clean evidence enters, or how one converted result becomes
evidence for the next task. If the host model retains native joint attention,
heterogeneous prompt, condition, reference, and target tokens continue to share
the same interaction route.

\noindent\TutorialTerm{Mixture of Experts.}
MoE~\cite{shazeer2017outrageouslylargeneuralnetworks,zhang2025mixtureexpertslargelanguage}
increases model capacity by routing tokens to selected experts. Its central
question is which computation or knowledge block should process an input.
Experts and routers can support large models, but adding many visual targets can
increase expert capacity, routing complexity, and coordinated-training cost.
Expert selection alone also does not define a directed graph among visual world
states.

\noindent\TutorialTerm{IP-Adapter.}
IP-Adapter~\cite{ye2023ipadapter} introduces an image encoder, projection, and
additional attention branch so that image prompts can complement text
conditioning. It provides a strong image-prompt interface while preserving the
pretrained text-to-image prior. Its principal role is to add an external image
condition to RGB generation, rather than to define a large graph of arbitrary
condition--target modality relationships.

\noindent\TutorialTerm{IC-Control style.}
In-context visual control methods
\cite{huang2024incontextlora,zhang2025icedit,dalva2025lorashop,guo2025longcontext}
place text, conditions, references, and targets into a common context. Their
strength is flexibility: contextual examples can describe a task without a
separate handcrafted interface for every interaction. As more modalities are
added, however, sequence length and joint-interaction cost grow, and new tokens
can change the shared attention field. Conditions and targets are jointly
represented rather than being organized as independently managed directed
routes.

\noindent\TutorialKey{TunnelingNet.}
TunnelingNet treats every target modality as a space that is genuinely
generated. Every condition--target relationship is a directed visual task with
a target trajectory, evidence route, task key, and lightweight target-side
tunnel. Small task-specific residual forces are applied repeatedly through the
depth of the frozen generative backbone, gradually deflecting its original RGB
generation trajectory toward the selected target modality. Clean conditions
act on the noised target through an evidence route. MTA normalizes target
self-generation and condition evidence separately, while NJA remains available
for tasks in which full native interaction is useful.

An intuitive comparison is that TunnelingNet internalizes the layer-wise
steering effect of ControlNet but uses ultralight residual paths inside the
backbone instead of attaching a copied backbone-sized branch outside it. The
result is not merely a smaller control module. The tunnel also has an explicit
semantic role in a task graph: it identifies a target, receives specified
evidence, follows a selected attention topology, and can be activated by a task
key. Under the parameter-matched comparison in this paper, this routing reaches
the selected structured outputs with \TutorialResult{$19.0$--$38.5\times$ fewer updates} and
\TutorialResult{recognizable structure $77$--$100\times$ earlier} than LoRA NJA.

\input{sec_preprint/X_tutorial_comparison}

\subsection{A Practical Mental Model of WSG, WSM, and TunnelingNet}

The three names describe different levels of the system.

\begin{itemize}
    \item \TutorialTerm{WSG is the shared representation.} RGB occupies channels
    0--2, while additional channel IDs store other world information. Readers
    and adapters expose dense maps, categories, keypoints, boxes, skeletons,
    body parameters, and audio-related targets through their native forms.

    \item \TutorialTerm{WSM is the shared generative system.} It contains a frozen
    visual prior and a library of directed capabilities. It can be understood
    as a computer with many specialized applications installed: all
    applications share the same operating system, and opening one application
    activates only the selected application instead of executing the entire
    library.

    \item \TutorialTerm{TunnelingNet opens and executes a route.} A task key selects
    the required target tunnel and evidence topology. The tunnel changes the
    generation path at multiple depths; it does not require another copy of the
    shared backbone.
\end{itemize}

This mental model explains why installing more capabilities need not increase
the active computation of an ordinary single-route request. A new task trains
only its target-side tunnel, leaving the backbone and existing tunnels
unchanged. Different tasks can be trained independently and in parallel. After
training, a capability can be installed, tested, repaired, replaced, upgraded,
or rolled back without retraining the other routes.

\noindent\TutorialKey{Adding one capability.}
Training a new capability has five steps: (1) prepare paired conditions and
targets for the desired conversion; (2) register the target's WSG channel
identity and native reader; (3) select the interface, evidence roles, and task
key; (4) freeze the shared backbone, text encoder, VAE, and existing routes while
training the target tunnel; and (5) validate, version, and install the resulting
capability. A later correction repeats only the affected route's training and
validation. This workflow is why the system can acquire capabilities on
different GPUs in parallel and why updating one route does not require a new
joint training cycle for the complete library.

The same organization also explains the role of visual chains. Switching,
merging, or stacking weight patches combines parameters, but it does not by
itself define an executable sequence among visual world states. In WSM, the
output of one directed task can become clean evidence for the next. A sequence
such as
\[
    \mathrm{RGB}\rightarrow\mathrm{Depth}\rightarrow
    \mathrm{Normal}\rightarrow\mathrm{Edge}
\]
is therefore a chain of explicit visual contracts. Each step has a declared
target and can be inspected, replaced, or reused independently.

\subsection{From a Route to a Visual Capability Operating System}

WSM can be viewed as a computer with many specialized applications installed. The
shared backbone plays the role of a common operating system; each registered
route supplies one application-level capability; and a task key opens only the
capability requested by the user. Installing another capability expands the
library but does not cause every installed route to execute on every request.

ControlWorld turns this model into an end-user workflow. A user selects the
condition and target in the task graph, supplies image, text, reference, or
world-information evidence, executes the corresponding tunnel, views the result
through the target's reader, and may pass that result into the next route. This
supports workflows such as

\begin{itemize}
    \item RGB $\rightarrow$ depth $\rightarrow$ normal $\rightarrow$ edge;
    \item RGB $\rightarrow$ roughness $\rightarrow$ base color;
    \item non-RGB conversions such as normal $\rightarrow$ edge;
    \item edge, depth, normal, or pose $\rightarrow$ RGB generation;
    \item first-frame appearance plus current structure $\rightarrow$ a new
    structural or appearance state; and
    \item structural evidence plus RGB $\rightarrow$ geometry, material,
    semantics, pose, or other world properties.
\end{itemize}

The engineering difference is direct. A conventional pipeline may maintain a
separate depth model, normal model, edge model, pose model, and RGB generator,
plus conversion code between their incompatible formats. WSG gives intermediate
world states stable identities, the shared backbone supplies a common prior, and
task tunnels provide replaceable conversions. The output of one step can be
reused immediately instead of being treated as an accidental by-product of an
opaque end-to-end model.

\subsection{Explicit Workflows and an Embedded Task Graph}

The task graph is a semantic organization, not a requirement that the system
must appear as a visible workflow or as thousands of unrelated model files. In
ControlWorld, the graph is exposed so that a user can select, inspect, and chain
routes. In a deployed WSM, the same task keys, condition--target roles, tunnel
parameters, and readers can reside inside one model package and be selected by
the request. Physical sharding is a storage decision; it does not define the
boundary of the method or the model.

This distinction also separates \emph{capability acquisition} from
\emph{capability execution}. Routes have different targets and can therefore be
learned independently, even on different GPUs, without placing all modalities,
losses, and targets in one joint training batch. At inference, only the route or
route chain named by the task key is activated. The library can consequently
grow through distributed acquisition while retaining one shared visual prior
and predictable per-request computation.

An additional research direction follows from this organization. A
self-contained WSM may use one of its own estimated world states as evidence for
a later route, or use an audited chain to construct training signals for a new
route. This is not required by the present evaluation; it is a path toward
recursive visual systems whose intermediate reasoning remains represented as
inspectable world information instead of hidden activations alone.

This organization points beyond a single benchmark. The same capability library
can support richer perception for robotics and digital twins; geometry, motion,
material, lighting, and rendering chains for film and games; pose, shape, and
segmentation routes for digital humans, health, and sports; physical and material
property estimation for manufacturing and logistics; editable product assets for
e-commerce; audiovisual localization and segmentation; and institution-specific
routes for medical, agricultural, satellite, or scientific imagery. These are
not separate claims that one frozen release already solves every domain. They
illustrate how a common task grammar and independently trainable routes turn
domain expansion into capability installation instead of complete-system
reconstruction. The measurable quantities are correspondingly concrete:
trainable parameters and updates per new route, time to recognizable structure,
resident library size, active parameters per request, number of independently
replaceable capabilities, and the length and success rate of executable chains.

\clearpage
\input{sec_preprint/X_application_landscape}

\FloatBarrier
\clearpage
\flushbottom

%% file: sec_preprint/X_tutorial_comparison.tex
\FloatBarrier
\clearpage
\onecolumn

\subsection{A Detailed Comparison of the Six Paradigms}

The compact drawing in Figure~\ref{fig:tutorial_six_paradigms} becomes more
informative when each paradigm is examined along the same system dimensions.
Table~\ref{tab:tutorial_six_paradigms_full} reconstructs that comparison. It is
not a ranking in which one method must replace all others: each established
paradigm solves an important problem. The comparison identifies the additional
organization required when the goal changes from adding one capability to
building a large, directed, maintainable capability system.

{\setlength{\tabcolsep}{2pt}
\renewcommand{\arraystretch}{1.14}
\scriptsize
\begin{longtable}{@{}p{0.12\textwidth}*{6}{p{0.135\textwidth}}@{}}
\caption{\textbf{Full tutorial comparison of six paradigms for extending visual capabilities.}
The rows distinguish an architectural primitive from a complete task-system organization.}
\label{tab:tutorial_six_paradigms_full}\\
\toprule
\textbf{Dimension} & \TutorialTerm{ControlNet} & \TutorialTerm{LoRA} & \TutorialTerm{MoE} &
\TutorialTerm{IP-Adapter} & \TutorialTerm{IC-Control style} & \TutorialKey{TunnelingNet / WSM} \\
\midrule
\endfirsthead
\multicolumn{7}{c}{\tablename\ \thetable\ continued}\\
\toprule
\textbf{Dimension} & \TutorialTerm{ControlNet} & \TutorialTerm{LoRA} & \TutorialTerm{MoE} &
\TutorialTerm{IP-Adapter} & \TutorialTerm{IC-Control style} & \TutorialKey{TunnelingNet / WSM} \\
\midrule
\endhead
\midrule
\multicolumn{7}{r}{Continued on the next page}\\
\endfoot
\bottomrule
\endlastfoot

\textbf{Method level} &
Large auxiliary control architecture &
Parameter-efficient weight update &
Conditional-computation and capacity architecture &
External image-conditioning adapter &
Joint visual-context organization &
Condition--target task architecture spanning representation, routing, trajectory deflection, training, and deployment \\

\textbf{Intuition} &
Copy a large control body beside the backbone &
Attach a small numerical patch to selected weights &
Add experts and route tokens among them &
Attach an image-prompt branch &
Place text, conditions, references, and target in one context &
Open a small target tunnel inside a shared backbone and steer its generation path repeatedly through depth \\

\textbf{Main modification site} &
Copied multi-layer network with injections into the backbone &
Weight-side branches of selected linear, attention, or feed-forward layers &
Expert layers and routers &
Image encoder, projector, and adapter attention &
Input sequence, joint attention, and associated fine-tuning layers &
Target-side residual paths plus a dedicated condition-to-target evidence route \\

\textbf{Added structure} &
Near-backbone-scale control body and injection links &
Low-rank matrices $A$ and $B$ &
Expert networks and router &
Image encoder, projector, and adapter attention &
Context organization, usually with fine-tuning or LoRA &
Task tunnel, task key, condition--target roles, interface type, and evidence routing \\

\textbf{Backbone treatment} &
Usually frozen &
Original weights usually frozen &
Implementation dependent; experts and routing are coordinated &
Usually frozen &
Shared model or LoRA is commonly fine-tuned &
Generative backbone, text encoder, and VAE are frozen \\

\textbf{How conditions enter} &
A copied control body extracts features and injects them layer by layer &
Uses condition interfaces already present in the host architecture &
Router assigns tokens to experts &
Independent image-conditioning branch enters attention &
Condition and target tokens interact in a common context &
Clean evidence acts on the noised target through a declared evidence route \\

\textbf{Meaning of a visual task} &
Usually a new spatial condition for an RGB target &
Defined by the training data; LoRA itself defines weight change, not an input--output task &
Defined jointly by the shared objective and expert roles &
Usually image prompt plus text to RGB &
A contextual relation learned among jointly represented tokens &
Every input--target mapping has its own generation trajectory, TunnelingNet, and task key \\

\textbf{Information route} &
External control route through copied layers and injections &
Network topology and token routes normally remain those of the host model &
Router selects computation modules &
Independent image-prompt route &
Shared context provides joint interaction &
Target self-generation is separated from condition evidence; the interface selects MTA or native joint attention \\

\textbf{Attention resources} &
Control features are extracted before backbone injection &
Under native joint attention, heterogeneous tokens share one Softmax budget &
Tokens compete for or receive routing capacity &
Text and image conditions commonly use separate attention before fusion &
All tokens interact in the shared attention field &
MTA separately normalizes target self-generation and condition evidence; NJA remains available when full interaction is useful \\

\textbf{How output changes} &
Strong layer-wise control of RGB generation &
Low-rank change to model function without explicit task routing &
Selected experts provide different computation or knowledge &
Image prompt augments text conditioning &
Joint context learns relationships among all elements &
Small task-specific residual forces progressively redirect the shared RGB trajectory into a selected target space \\

\textbf{Unit of expansion} &
Copied control body and injection links &
Set of low-rank weight differences &
Expert capacity and routing &
Adapter or conditioning interface &
Contextual task organization and its training &
Directed task edge jointly defined by conditions, target, interface, tunnel, and task key \\

\textbf{Typical range} &
Depth, edge, or pose to RGB &
Depends on data and objective &
Depends on expert definitions &
Image prompt and text to RGB &
Text and visual contexts to a visual target &
Directed generation among appearance, geometry, pose, motion, material, physical, semantic, sparse, and audio information \\

\textbf{Adding a capability} &
Adds another substantial control body &
Adds an independent patch, but native information routes remain &
May require coordination with experts and router &
Adds an independent conditioning adapter &
Changes the shared context and may require re-coordination &
Trains and installs one task tunnel without retraining the backbone or existing routes \\

\textbf{Capability composition} &
External workflow combines control modules &
Switch, merge, or stack weight patches &
Router selects or mixes experts &
Combines image prompt with text and other controls &
Combines evidence inside one context &
One route's decoded output becomes clean evidence for the next declared route, creating executable visual chains \\

\textbf{Scaling pressure} &
Large control structure per capability &
Separate training and management; native evidence competition may remain &
Expert count, routing, and coordinated-training cost &
Primarily expands image conditioning, not arbitrary modality graphs &
Sequence length, joint-interaction cost, and cross-modality interference &
Small independently trainable tunnels; shared prior, task graph, and chains can expand while one request activates only selected routes \\

\textbf{Principal strength} &
Mature spatial control &
Efficient, general, distributable adaptation &
Scalable conditional capacity &
Strong image-prompt conditioning &
Flexible contextual interaction &
Unified task grammar, world-information representation, target routing, lifecycle management, and composable directed generation \\
\end{longtable}
}

\clearpage
\twocolumn
\raggedbottom

\subsection{What the Comparison Means}

Put simply, natural-language processing now has large language models that
handle \TutorialEmph{approximately 900 language tasks} within a common paradigm. Vision has \TutorialEmph{no comparable model}; existing visual methods generally unify \TutorialEmph{no more than approximately 30 tasks}. The main reason is that visual tasks are harder to
train, use incompatible data formats, and optimize inconsistent objectives.

Training a model from scratch with dozens of input types and dozens of output
types is impractical. LLMs succeeded in part because even code is structured
language: linguistic data are inherently more consistent and easier to train on
than visual data.

Computer vision has therefore largely progressed by adding new capabilities to
existing image and video generation models, and the five paradigms above are
among the most common and effective ways to extend such models.

\TutorialKey{We take a different approach:} reaching hundreds or thousands of new targets by
deflecting the generation trajectory of a shared backbone. Because visual
information can be represented in image-like form, we call the resulting
representation World Snapshot Graphics, or a World Frame, and the model a World
Snapshot Model. Because the method reuses an existing generative backbone while
opening routes to different targets, the analogy of creating new tunnels
motivates the name TunnelingNet.

\subsection{What Constitutes the Unified Capability}

No single familiar ingredient is presented as the contribution in isolation.
The essential construction is their organization into a scalable visual-task
system:

\begin{itemize}
    \item \TutorialEmph{A clear mapping} from conditional world information to target world
    information;
    \item \TutorialEmph{Non-RGB world states treated as actual generation targets};
    \item \TutorialEmph{Independent addition, replacement, repair, and upgrade of task}
    capabilities;
    \item \TutorialEmph{Composable generation chains} among multiple visual tasks;
    \item separation between capability-library size and single-route active
    computation;
    \item \TutorialEmph{Target-side tunnels paired with condition-evidence routing};
    \item \TutorialEmph{Separate processing of target generation and condition evidence} by
    MTA when the interface requires it; and
    \item \TutorialEmph{A large, flexibly composable mapping structure over visual tasks.}
\end{itemize}

This organization supports uncommon and bidirectional conversions whenever the
modalities are sufficiently related, as well as many-to-one, one-to-many, and
many-to-many contracts. RGB can also serve as additional evidence when a
structural modality alone cannot determine texture, boundaries, material, or
scene details. The resulting interface covers dense continuous fields,
categorical maps, sparse structures, body representations, and audio tokens
without requiring all targets to share one decoder.

The scale numbers in the main paper can now be read in the correct way. WSM v1
evaluates 1,148 routes on one shared visual backbone; it does not execute all
routes simultaneously. Each dense route adds approximately 12.51M trainable
parameters, \TutorialResult{about $0.32\%$ of the 3.9B visual backbone}. Its complete resident
visual parameter bank is \TutorialResult{approximately 18.54B parameters}, including all dense
task tunnels and deployed categorical refiners. An ordinary dense request,
however, activates \TutorialResult{approximately 3.913B visual parameters}: the backbone and one
selected tunnel. Copying a 3.9B backbone for every evaluated task would require
\TutorialResult{approximately 4.48T visual parameters}. Sharing the backbone therefore reduces
this task-library parameterization by \TutorialResult{about $241\times$} while retaining a
dedicated route for every task. Capability-library size and per-request active
computation are separate quantities.

\subsection{How Parameter and Training Costs Scale with More Capabilities}

The qualitative differences above become decisive when capabilities are added
at scale. A dense TunnelingNet contains 12.51M trainable parameters. Under the
representative configurations used in our engineering comparison, the
per-capability parameter or weight-memory increment of established extension
paradigms is approximately the multiple shown in
Table~\ref{tab:tutorial_parameter_scaling}. These ratios are
configuration-dependent rather than universal constants, but they expose the
different scaling coefficients of the six paradigms.

\begin{table}[t]
\centering
\small
\setlength{\tabcolsep}{5pt}
\caption{\textbf{Illustrative per-capability parameter scale relative to one
12.51M-parameter TunnelingNet.} The comparison uses representative
configurations of each paradigm; implementations vary.}
\label{tab:tutorial_parameter_scaling}
\begin{tabular}{lc}
\toprule
\textbf{Extension paradigm} & \textbf{Relative scale} \\
\midrule
\TutorialResult{TunnelingNet} & \TutorialResult{$1.0\times$} \\
IP-Adapter & $\sim1.8\times$ \\
IC-Control style & $\sim4\times$ \\
LoRA & typically $1$--$8\times$ \\
ControlNet & $\sim29\times$ \\
MoE & $\sim300$--$10{,}000\times$ \\
Full fine-tuning & $\sim312.5\times$ \\
\bottomrule
\end{tabular}
\end{table}

\TutorialKey{A 10,000-capability library makes the scaling difference concrete.} Linearizing the representative increments in Table~\ref{tab:tutorial_parameter_scaling}, TunnelingNet adds \TutorialResult{approximately 125.1B resident parameters}, producing \TutorialResult{a roughly 129B-parameter complete model} while an ordinary request still activates only the backbone and one selected TunnelingNet. The corresponding increments are about 225B for IP-Adapter, 500B for IC-Control, 125B--1.00T for LoRA, 3.63T for ControlNet, 37.5T--1.25P for MoE, and 39.1T for separate full fine-tuning. At this scale, most alternatives become difficult to train, store, transfer, and repeatedly upgrade. The conclusion is supported by measured experiments, not parameter arithmetic alone. Under matched trainable parameters, TunnelingNet reduces pooled training time by \TutorialResult{$20.6\times$}; across five representative external tasks, WSM records \TutorialResult{23 first-place and three second-place results}. The scaling advantage is therefore achieved without trading away task quality. At the ultra-large scale of hundreds to 10,000 capabilities considered here, WSM is the only evaluated approach that simultaneously maintains a manageable resident library, near-backbone active computation per request, rapid capability acquisition, and competitive or leading task quality. These properties make it the most practical currently demonstrated route toward ultra-large, LLM-scale visual capability systems containing hundreds of billions, and eventually trillions, of resident parameters.

Parameter count alone does not explain the acquisition advantage. Our
parameter-matched LoRA NJA and TunnelingNet comparison holds the
trainable parameter count fixed. Across Tasks 003, 275, and 313, the measured
training time falls from 41.58 GPU-hours (1.73 GPU-days) to 2.02 GPU-hours
(0.084 GPU-days), a pooled $20.6\times$ acceleration. Applying the same
measured per-task averages to 1,148 capabilities gives an illustrative
capability-library acquisition cost of approximately 663 versus 32.2 GPU-days,
a reduction of about 631 GPU-days. This extrapolation describes repeated
independent capability acquisition; it is not substituted for the measured
three-task result.

The scale becomes larger under industrial training schedules. SenseNova-Vision
reports 50,000 training steps on at least 16 80GB GPUs
\cite{sensenova2026sensenovavision}, whereas the corresponding WSM task uses
1,740 single-GPU updates. Counting GPU-updates gives a conservative workload
proxy of at least
\[
    \frac{50{,}000\times16}{1{,}740}\approx460\times .
\]
If every one of 1,148 capabilities were acquired at that workload, the
illustrative total would be \TutorialResult{approximately 242,000 GPU-days}, compared with about
\TutorialResult{526 GPU-days} at the observed WSM task time, saving \TutorialResult{approximately 241,000 GPU-days}. This is a transparent scale scenario based on published steps and
GPU counts, not a measured SenseNova wall-clock or financial result.

Let \(c\) be GPU rental cost in dollars per GPU-hour and \(R\) the number of
full capability-library acquisition or upgrade cycles. The corresponding
GPU-rental-equivalent saving in this industrial scenario is
\[
    S(R,c)\approx241{,}000\times24\times c\times R .
\]
At \(c=\$2\)--\(\$4\), one complete cycle is approximately
\TutorialResult{\(\$11.6\)--\(\$23.2\) million}; ten cycles are
\TutorialResult{\(\$116\)--\(\$232\) million}; and one hundred hyperscale cycles are
\TutorialResult{\(\$1.16\)--\(\$2.32\) billion}. The arithmetic explains why the same method can
save only hundreds of thousands of dollars for a small deployment yet reach
tens of millions per full training cycle, and hundreds of millions or billions
under repeated hyperscale development. These estimates exclude data,
annotation, engineering labor, storage, networking, and inference.

The same acceleration also changes an individual research cycle. Using the
measured averages above, 50--100 training runs, including seeds, ablations, and
hyperparameter trials, decrease from \TutorialResult{approximately 29--58 to 1.4--2.8}
single-GPU days. This saves \TutorialResult{about 27--55 GPU-days}, or roughly
\TutorialResult{\(\$1{,}300\)--\(\$5{,}300\)} at \(\$2\)--\(\$4\) per GPU-hour, while allowing
results to be inspected and experiments redirected within the same day.

Deployment exhibits a separate saving. Replicating a 3.9B backbone for 1,148
tasks requires approximately \TutorialResult{4.48T parameters, or 8.96TB} of FP16/BF16 weights.
The complete 18.54B-parameter WSM occupies \TutorialResult{approximately 37.1GB of weights}, and
an ordinary dense request activates \TutorialResult{approximately 3.913B parameters, or 7.83GB}
of weights. These are weight-only footprints; runtime memory additionally
contains activations and framework state.

\subsection{Capabilities That Emerge at the System Level}

The table also reveals why the contribution cannot be reduced to another
adapter or a collection of LoRAs. A weight-update primitive answers how a
parameter is changed. A capability system must additionally specify what
information is the condition, what world state is generated, how evidence
reaches that target, how the route is selected, how routes compose, and how one
capability changes over its lifetime. Once these relations are explicit, several
system-level abilities become available together:

\begin{itemize}
    \item \TutorialTerm{Uncommon, bidirectional, and multi-input/output contracts.}
    Related modalities can be connected in either direction, and contracts can
    be one-to-many, many-to-one, or many-to-many instead of being restricted to
    an external condition that controls RGB.

    \item \TutorialTerm{Executable visual chains.} A decoded result becomes clean
    evidence for the next route. This is structurally different from merging
    weights or selecting experts because the intermediate world state and the
    next target remain explicit and inspectable.

    \item \TutorialTerm{Independent capability lifecycle.} A route can be trained in
    parallel, installed, tested, repaired, replaced, upgraded, or rolled back
    without changing the shared backbone or previously registered routes.

    \item \TutorialTerm{Target-specific information flow.} Only the target-side tunnel
    is optimized. MTA can give target self-generation and condition evidence
    separate normalization, while native joint attention remains selectable for
    interfaces that benefit from full interaction.

    \item \TutorialTerm{Heterogeneous native outputs.} Dense continuous fields,
    categorical maps, keypoints, boxes, skeletons, body parameters, and audio
    tokens coexist in one task system without pretending that every output has
    the same decoder or constraint.

    \item \TutorialTerm{RGB-assisted recovery.} RGB can accompany an incomplete
    structural condition when texture, boundaries, material, identity, or scene
    detail cannot be recovered from that condition alone, creating a distinct
    family of RGB-assisted inverse-rendering contracts.
\end{itemize}

This organization also creates visual tasks instead of merely collecting known
ones. Among 1,072 audited cross-variable contracts, \TutorialResult{at least 655} have no clear
historical input--output counterpart; including historically underspecified
cases gives an \TutorialResult{upper estimate of 1,056}. The result is a system that both covers
the established task inventory and exposes previously unnamed relationships
among world properties.

%% file: sec_preprint/X_application_landscape.tex
\subsection{World Snapshot Model Application Landscape}
\label{sec:tutorial_application_landscape}

The World Snapshot Model framework opens a broad application landscape across foundational vision, capability expansion, generation, world-state computing, scientific and industrial systems, and visual-AI infrastructure. The following forward-looking map organizes 882 prospective application scenarios into 19 families. It serves as a research and development roadmap for the task graph, capability library, and world-information interfaces introduced above.

\definecolor{wsmappmark}{HTML}{B54708}
\par\medskip
\noindent\textbf{1. Foundational Computer Vision, World Intelligence, and Reasoning (38)}\par

General-purpose computer vision; vision foundation models; multimodal vision foundation models; vision agents; multimodal vision agents; visual reasoning; visual chain-of-thought reasoning; compositional visual reasoning; large-scale visual reasoning; multimodal visual reasoning; cross-modal visual reasoning; visual physical reasoning; visual spatial reasoning; visual temporal reasoning; visual causal reasoning; visual counterfactual reasoning; visual commonsense reasoning; interactive visual-world reasoning; unified visual tasks; any-input, any-output vision systems; multi-input, single-output vision systems; multi-input, multi-output vision systems; open-world visual learning; automated visual task discovery; visual world-interaction models; interactive visual world models; persistent visual world models; dynamic visual world models; predictive visual world models; physical visual world models; in-cabin visual perception; visual search; video summarization; visual disaster damage assessment; cross-view verification; cross-temporal verification; cross-modal visual verification; \textcolor{wsmappmark}{*}pixel-level style transfer.
\par\medskip
\noindent\textbf{2. Model Capability Expansion, Composition, and Runtime Management (37)}\par

Multi-task visual learning; continual visual learning; incremental visual learning; modular visual learning; compositional visual learning; visual capability modules; hot-loading of model capabilities; hot updates of model capabilities; hot-swapping of model capabilities; incremental deployment of model capabilities; automated composition of visual capabilities; visual task routing; vision model routing; visual tool routing; visual expert routing; visual capability orchestration; world-capability orchestration; visual workflow optimization; visual workflow execution; visual workflow verification; vision model fusion; visual capability fusion; vision model distillation; visual capability distillation; vision model merging; vision model compression; visual capability versioning; visual capability rollback; visual capability diagnostics; visual capability verification; visual capability compatibility testing; automated expansion of visual capabilities; compositional expansion of visual capabilities; linear expansion of visual capabilities; nonlinear expansion of visual capabilities; homogeneous expansion of visual capabilities; heterogeneous expansion of visual capabilities.
\par\medskip
\noindent\textbf{3. Generation, Editing, Restoration, and Cross-Domain Translation (97)}\par

Generative visual perception; generative visual tasks; multi-condition visual generation; multi-target visual generation; automated visual task generation; visual workflow generation; visual capability repair; general-purpose image generation; general-purpose video generation; long-form video generation; real-time video generation; interactive video generation; multiview image generation; multiview video generation; 3D content generation; 4D content generation; spatial video generation; controllable image generation; controllable video generation; reference-guided image generation; scene-consistent generation; object-consistent generation; physically consistent generation; temporally consistent generation; illumination-consistent generation; motion-consistent generation; audio-visual consistent generation; image editing; video editing; 3D editing; 4D editing; scene editing; object editing; illumination editing; motion editing; camera editing; facial-expression editing; apparel editing; background editing; localized region editing; semantic editing; interactive editing; generative restoration; image restoration; video restoration; 3D restoration; old-photo restoration; digital restoration of cultural heritage; image denoising; video denoising; image deblurring; video deblurring; image super-resolution; video super-resolution; image completion; video completion; scene completion; object completion; missing-modality completion; image colorization; video colorization; image style transfer; video style transfer; domain translation; cross-domain generation; image-to-image translation; video-to-video translation; audio-to-image generation; audio-to-video generation; image-to-audio generation; video-to-audio generation; text-to-image generation; text-to-video generation; text-to-3D generation; image-to-3D generation; video-to-3D generation; 3D-to-image generation; 3D-to-video generation; World Snapshot generation; World Snapshot video generation; World Snapshot field generation; world-state generation; future-state generation; world-state restoration; world-state completion; generative world models; apparel generation; media content generation; general-purpose generative interfaces; \textcolor{wsmappmark}{*}unified 1D-to-4D generation; \textcolor{wsmappmark}{*}text-to-world generation; \textcolor{wsmappmark}{*}single-pass generation of tens to hundreds of world attributes; \textcolor{wsmappmark}{*}World Snapshot fine-tuning of pretrained generative models; \textcolor{wsmappmark}{*}auxiliary-attribute expansion for generated content; \textcolor{wsmappmark}{*}pixel-level motion transfer; \textcolor{wsmappmark}{*}multilayer human generation; \textcolor{wsmappmark}{*}joint generation of human surface and internal structures.
\par\medskip
\noindent\textbf{4. World Snapshots, World States, and World Data (38)}\par

World Snapshot Models; world-state prediction; world-state updating; world-state versioning; world-state branching; world-state rollback; world-state querying; world-state fusion; world-state compression; world-state transmission; world-state databases; world-state search engines; world-state knowledge bases; world-data platforms; world-state digital assets; world-attribute interfaces; World Snapshot APIs; world-state APIs; \textcolor{wsmappmark}{*}World Snapshot computer graphics; \textcolor{wsmappmark}{*}World Snapshot line representation; \textcolor{wsmappmark}{*}World Snapshot illumination representation; \textcolor{wsmappmark}{*}World Snapshot shape representation; \textcolor{wsmappmark}{*}latent-space entity representation; \textcolor{wsmappmark}{*}World Snapshot graphics data acquisition; \textcolor{wsmappmark}{*}World Snapshot video data acquisition; \textcolor{wsmappmark}{*}World Snapshot field data acquisition; \textcolor{wsmappmark}{*}world-information acquisition from machinery; \textcolor{wsmappmark}{*}World Snapshot data mining; \textcolor{wsmappmark}{*}World Snapshot exploration by agents; \textcolor{wsmappmark}{*}PNG-to-WSG conversion; \textcolor{wsmappmark}{*}WSG information extraction; \textcolor{wsmappmark}{*}WSG storage; \textcolor{wsmappmark}{*}WSG compression; \textcolor{wsmappmark}{*}World Snapshot database systems; \textcolor{wsmappmark}{*}secure World Snapshot computing; \textcolor{wsmappmark}{*}efficient World Snapshot computing architectures; \textcolor{wsmappmark}{*}cross-platform World Snapshot video players; \textcolor{wsmappmark}{*}World Snapshot software engineering.
\par\medskip
\noindent\textbf{5. Recognition, Localization, Detection, Segmentation, and Tracking (67)}\par

Generative visual understanding; visual completion of occluded regions; visual world-state retrieval; image classification; video classification; object classification; scene classification; visual action classification; visual behavior classification; visual event classification; audio-visual cross-modal classification; open-vocabulary image classification; fine-grained image classification; zero-shot image classification; few-shot image classification; image localization; object localization; keypoint localization; text-guided visual grounding; vision-language grounding; audio-visual sound-source localization; 3D visual localization; visual spatiotemporal localization; visual indoor localization; visual outdoor localization; object detection; open-vocabulary object detection; zero-shot object detection; few-shot object detection; 3D object detection; video object detection; small-object detection; dense object detection; occluded-object detection; visual anomaly detection; video event detection; video action detection; audio-visual sound-event detection; audio-guided visual instance detection; semantic segmentation; instance segmentation; panoptic segmentation; part segmentation; human segmentation; object segmentation; video segmentation; 3D segmentation; 4D segmentation; interactive segmentation; open-vocabulary segmentation; audio-guided visual instance segmentation; motion segmentation; scene segmentation; visual forest monitoring; visual vessel detection; video retrieval; video understanding; video question answering; visual question-answering assistance; image captioning; video captioning; visual object recognition; visual object tracking; visual terrain understanding; visual causal analysis; \textcolor{wsmappmark}{*}occlusion reasoning for overlapping objects; \textcolor{wsmappmark}{*}attribute inference for occluded regions.
\par\medskip
\noindent\textbf{6. Visual Geometry, Motion, Material, and Physical-Property Estimation (96)}\par

Visual Model Deployment at the Edge; Geometry-Consistent Generation; Material-Consistent Generation; Material Editing; Pose Editing; Visual Physical-Property Editing; Visual Scene-State Estimation; Visual Material Classification; Visual Material Segmentation; Visual Affordance Segmentation; Visual Contact-Region Segmentation; Visual Risk-Region Segmentation; Depth Estimation; Monocular Depth Estimation; Multi-View Depth Estimation; Video Depth Estimation; Metric Depth Estimation; Relative Depth Estimation; Disparity Estimation; Surface Normal Estimation; Curvature Estimation; Edge Estimation; Contour Estimation; Optical Flow Estimation; Scene Flow Estimation; Visual Motion Estimation; Camera Pose Estimation; Object Pose Estimation; Human Pose Estimation; Hand Pose Estimation; Facial Pose Estimation; Skeleton Estimation; Visual Human-Parameter Estimation; Visual Object-Structure Estimation; Visual Scene-Structure Estimation; Visual Geometry Estimation; Visual Scale Estimation; Visual Distance Estimation; Visual Velocity Estimation; Visual Acceleration Estimation; Visual Trajectory Estimation; Visual Deformation Estimation; Visual Articulation-State Estimation; Visual Occlusion Estimation; Visual Visibility Estimation; Visual Illumination Estimation; Visual Reflectance Estimation; Visual Material Estimation; Visual Texture Estimation; Visual Shadow Estimation; Visual Transparency Estimation; Visual Roughness Estimation; Visual Metallicity Estimation; Visual Refractive-Index Estimation; Visual Density Estimation; Visual Mass Estimation; Visual Volume Estimation; Visual Weight Estimation; Visual Support-Area Estimation; Visual Friction-Coefficient Estimation; Visual Friction-Force Estimation; Visual Elasticity Estimation; Visual Stiffness Estimation; Visual Hardness Estimation; Thermal-Image Temperature Estimation; Visual Pressure Estimation; Visual Stress Estimation; Visual Strain Estimation; Visual Flow-Rate Estimation; Visual Contact-State Estimation; Visual Collision-Risk Estimation; Visual Affordance Estimation; Visual Stability Estimation; Visual Load-Bearing-Capacity Estimation; Visual Physical-Property Estimation; Visual Material-Property Estimation; Visual Uncertainty Estimation; Visual Confidence Estimation; Visual Risk Estimation; Visual Future-State Estimation; Visual Pedestrian-Trajectory Prediction; Visual Collision Prediction; Visual Geometry-Consistency Verification; \textcolor{wsmappmark}{*}Wave-Function Visual Generative Models; \textcolor{wsmappmark}{*}Native Visual Computation of World Physical Parameters; \textcolor{wsmappmark}{*}Native Visual Modeling of Relative Distance; \textcolor{wsmappmark}{*}Native Visual Modeling of Absolute Distance; \textcolor{wsmappmark}{*}Visual Relative-Volume Computation; \textcolor{wsmappmark}{*}Visual Relative-Mass Prediction; \textcolor{wsmappmark}{*}Pixelwise Mass-Distribution Modeling; \textcolor{wsmappmark}{*}Multi-Wave-Function Distributional Representations of Visual Worlds; \textcolor{wsmappmark}{*}Attaching Visual Physical Properties to Generated Content; \textcolor{wsmappmark}{*}Pose- and Motion-Conditioned Visual World Classification; \textcolor{wsmappmark}{*}Depth- and Audio-Conditioned Visual World Classification; \textcolor{wsmappmark}{*}Multi-Depth-Layer Visual Occlusion Representation; \textcolor{wsmappmark}{*}Probabilistic Visual World-Line Prediction.
\par\medskip
\noindent\textbf{7. Reconstruction, Digital Twins, Simulation, and Scientific Computing (55)}\par

Past Visual-State Reconstruction; Visual Digital Twins; Generative Visual Digital Twins; Real-Time Visual Digital Twins; Urban Visual Digital Twins; Human Visual Digital Twins; Image Reconstruction; Video Reconstruction; 3D Reconstruction; 4D Reconstruction; Multi-View Reconstruction; Dynamic Scene Reconstruction; Human Reconstruction; Hand Reconstruction; Face Reconstruction; Object Reconstruction; Indoor Scene Reconstruction; Outdoor Scene Reconstruction; Urban Visual Reconstruction; Roadway Visual Reconstruction; Terrain Visual Reconstruction; Subsurface-Structure Visual Reconstruction; Microscopy Image Reconstruction; Magnetic Resonance Image Reconstruction; Computed Tomography Image Reconstruction; Ultrasound Image Reconstruction; Neural Radiance Field Reconstruction; Gaussian Splatting Reconstruction; Point-Cloud Reconstruction; Mesh Reconstruction; Voxel Reconstruction; Occupancy-Field Reconstruction; Material Reconstruction; Illumination Reconstruction; Acoustic-Field Visualization and Reconstruction; Physical-Field Visualization and Reconstruction; Generative Visual Reconstruction; Generative RGB-D Reconstruction; Generative Multi-Attribute Reconstruction; Visual Material Representation; Visual Physics Simulation; Visual Fluid Simulation; Visual Structural Simulation; Visual Thermodynamic Simulation; Visual Electromagnetic-Field Simulation; Visual Acoustic-Field Simulation; Microscopy Vision; Audiovisual Scene Understanding; Visual Sim-to-Real Transfer; Visual Real-to-Sim Reconstruction; Generative Reconstruction APIs; \textcolor{wsmappmark}{*}Visual Modeling of Dynamic Physical Fields; \textcolor{wsmappmark}{*}Billion-Scale Natural Simulation-Scene Generation; \textcolor{wsmappmark}{*}Recording World Information with Gaussian Representations; \textcolor{wsmappmark}{*}Fusion of Gaussian Fields and World Snapshots.
\par\medskip
\noindent\textbf{8. Robotics, Embodied AI, and Autonomous Systems (48)}\par

Vision-Based Embodied Intelligence; Robotic Visual Digital Twins; Visual Robot Localization; Robotic Visual World Models; Embodied Visual World Models; Vision-Language-Action Models; Robotic Visual Perception; Vision-Based Robot Planning; Vision-Based Robot Control; Vision-Guided Robotic Grasping; Vision-Guided Robotic Manipulation; Vision-Based Robot Navigation; Vision-Based Robot Obstacle Avoidance; Visual SLAM for Robotics; Robotic Visual Scene Understanding; Vision-Based Robot Task Planning; Visual Robot Skill Learning; Visual Robot Skill Composition; Visual Robot Imitation Learning; Visual Robot Reinforcement Learning; Vision-Based Autonomous Robot Exploration; Vision-Based Robot Fault Diagnosis; Robotic Visual-State Prediction; Visual Safety Control for Robots; Visual Multisensor Fusion for Robotics; Visual Human-Robot Collaboration; Home-Service Robot Vision; Industrial Robot Vision; Logistics Robot Vision; Warehouse Robot Vision; Aerial Robot Vision; Space Robot Vision; Biomimetic Robot Vision; General-Purpose Robot Vision; Vision-Based Drone Navigation; Vision-Based Drone Obstacle Avoidance; Visual Navigation-Map Generation; Robotic Vision for Space Stations; Vision-Based Autonomous Satellite Navigation; Vision-Guided Intraoperative Navigation; Marine Vessel Visual Navigation; Visual Navigation for Blind and Visually Impaired People; Military Robot Vision; Visual Search-and-Rescue Robotics; Home-Robot Visual Control; Vision-Based Multi-Robot Collaboration; Robotic Vision Operating-System Extensions; Embodied Visual Intelligence APIs.
\par\medskip
\noindent\textbf{9. Autonomous Driving, Transportation, Drones, and Aerospace (39)}\par

Vehicle Visual Digital Twins; Visual Vehicle Localization; Visual World Models for Autonomous Driving; Vision-Based Autonomous Driving; Vision-Based Driver Assistance; Vehicle Visual Perception; Vision-Based Vehicle Planning; Vision-Based Vehicle Control; Lane Detection; Road Segmentation; Drivable-Area Detection; Traffic-Participant Detection; Visual Traffic-Scene Prediction; Visual Vehicle-Trajectory Prediction; Visual Driving-Risk Assessment; Visual Driver-Behavior Understanding; Vision-Based Driving Decision-Making; Vision-Based Parking Assistance; Vision-Based Automated Parking; Visual Driver Monitoring; Cooperative Vehicle-Infrastructure Vision; Traffic Visual Digital Twins; Visual Traffic Simulation; Visual Data Generation for Autonomous Driving; Autonomous-Driving Scene Reconstruction; Visual Counterfactual Testing for Autonomous Driving; Vision-Based Drone Inspection; Vision-Based Drone Mapping; Vision-Based Drone Formation Control; Visual Object Tracking by Drones; Drone-Based Visual Environment Perception; Aerial Image Analysis; Satellite Image Analysis; Astronomical Image Analysis; Telescope Image Analysis; Planetary-Surface Image Analysis; Visual Space-Domain Awareness; Visual Search-and-Rescue Drones; Visual Traffic-Accident Analysis.
\par\medskip
\noindent\textbf{10. Remote Sensing, Earth, Cities, Architecture, and Infrastructure (46)}\par

Building Visual Digital Twins; Earth Visual Digital Twins; Environmental Visual Digital Twins; Visual Geolocation; Remote-Sensing Image Segmentation; Visual Building Reconstruction; Remote-Sensing Image Interpretation; Remote-Sensing Land-Cover Classification; Remote-Sensing Land-Use Analysis; Visual Disaster Monitoring; Visual Flood Monitoring; Visual Fire Monitoring; Visual Wildfire Prediction; Image-Based Earthquake-Damage Assessment; Visual Landslide Monitoring; Visual Glacier Monitoring; Visual Coastline Monitoring; Visual Water-Body Monitoring; Remote-Sensing Environmental Change Detection; Visual Weather Analysis; Visual Geospatial Reasoning; Map Generation; Visual Map Updating; 3D Mapping; High-Definition Map Generation; Indoor Visual Mapping; Visual Monitoring of Urban Infrastructure; Visual Road Inspection; Visual Bridge Inspection; Visual Tunnel Inspection; Visual Railway Inspection; Visual Airport Inspection; Visual Port Inspection; Visual Power-Grid Inspection; Visual Pipeline Inspection; Visual Water-Infrastructure Inspection; Visual Telecommunications-Infrastructure Inspection; Visual Building Information Modeling; Visual Construction-Site Monitoring; Visual Building-Quality Inspection; Visual Building-Energy Analysis; Visual Spatial-Layout Generation; Visual Furniture-Layout Generation; Visual Real-Estate Presentation; Virtual Property Tours; Architectural Visualization.
\par\medskip
\noindent\textbf{11. Industrial, Manufacturing, Logistics, Retail, and Commercial Applications (37)}\par

Industrial visual digital twins; Automated visual defect detection; Industrial defect segmentation; Machine vision; Automated visual inspection; Surface defect detection; Visual dimensional inspection; Visual assembly inspection; Visual weld inspection; Visual part recognition; Visual part pose estimation; Visual production-process monitoring; Vision-based equipment condition monitoring; Vision-based predictive maintenance; Visual fault localization; Visual root-cause analysis; Production-line visual digital twins; Supply-chain visualization; Vision-based warehouse automation; Vision-based logistics sorting; Visual package inspection; Vision-based inventory recognition; Visual shelf monitoring; Visual product tracking; Retail computer vision; Visual product recognition; Visual product search; Shelf-image analysis; Virtual try-on; Product image generation; Advertising image generation; E-commerce visual content generation; Brand visual asset generation; Marketing visual content generation; Industrial visual design; Industrial visual safety monitoring; Image-based insurance damage assessment.
\par\medskip
\noindent\textbf{12. Medicine, Life Sciences, Agriculture, Forestry, and Marine Applications (48)}\par

Medical visual digital twins; Medical image segmentation; Underwater visual reconstruction; Medical image reconstruction; Medical robot vision; Surgical robot vision; Rehabilitation robot vision; Agricultural robot vision; Underwater robot vision; Medical image analysis; Image-assisted clinical diagnosis; Lesion detection in medical images; Tumor segmentation in medical images; Organ segmentation in medical images; Anatomical reconstruction from medical images; Image-guided surgical planning; Image-guided radiotherapy planning; Disease-progression prediction from medical images; Longitudinal medical image analysis; Histopathology image analysis; Digital pathology; Cell image segmentation; Cell tracking in microscopy images; Tissue-structure analysis; Protein imaging analysis; Vision-based biomechanical analysis; Vision-based human motion analysis; Vision-based rehabilitation assessment; Vision-based vital-sign estimation; Healthcare visual digital twins; Agricultural computer vision; Vision-based crop monitoring; Visual pest and disease detection; Visual weed detection; Visual fruit counting; Visual fruit-ripeness estimation; Vision-based crop-yield prediction; Vision-based soil-condition estimation; Agricultural drone vision; Vision-based livestock monitoring; Vision-based aquaculture monitoring; Marine computer vision; Underwater scene understanding; Underwater object detection; Underwater 3D reconstruction; Vision-based marine ecosystem monitoring; Vision-based fisheries monitoring; \textcolor{wsmappmark}{*}World Snapshot-based medical information recording.
\par\medskip
\noindent\textbf{13. Film, Media, Games, XR, Design, and Culture (64)}\par

Immersive visual content generation; Identity-consistent character generation; Cross-shot visual character consistency; Visual character editing; Game visual world models; Film and television visual world models; AI-generated films; AI-generated television series; AI-generated animation; AI-generated short-form video; AI-powered visual live streaming; Virtual production; Visual previsualization; Visual shot planning; Visual storyboard generation; Visual shot-sequence generation; Visual character generation; Digital visual actors; Visual virtual characters; Visual digital humans; Visual virtual presenters; Vision-based motion capture; Vision-based facial-expression capture; Visual facial reenactment; Visual lip synchronization; Audio-driven visual animation; Music visualization; Visual effects; Visual post-production; Automated video editing; Visual media asset management; Visual game generation; Game-world visual generation; Visual game-level generation; Visual game-character generation; Visual game-animation generation; Visual game-material generation; Visual game-physics generation; Visual reconstruction of game scenes; Vision-based game agents; Visual intelligence for non-player characters; Visual interactive storytelling; Persistent visual game worlds; Open visual game worlds; Live visual-content updates for games; Visual game-mod ecosystem; Visual cloud gaming; Virtual-reality visual generation; Augmented-reality visual generation; Mixed-reality visual generation; Visual spatial computing; Virtual visual laboratories; Visual content for virtual museums; Digital visual exhibitions; Visual digitization of cultural heritage; Visual archaeological reconstruction; Artwork image analysis; AI visual art creation; AI-assisted visual design; AI-assisted visual product design; AI-assisted visual graphic design; \textcolor{wsmappmark}{*}World Snapshot film generation; \textcolor{wsmappmark}{*}Two-hour feature-length film generation; \textcolor{wsmappmark}{*}Infinite game-frame generation.
\par\medskip
\noindent\textbf{14. Education, Sports, Sign Language, and Accessibility (22)}\par

Vision-based teaching; Visual educational content generation; Sports computer vision; Vision-based athlete pose analysis; Sports video analysis; Tactical analysis of sports video; Vision-based movement scoring; Vision-assisted officiating; Vision-based fitness coaching; Golf video analysis; Soccer video analysis; Basketball video analysis; Tennis video analysis; Swimming video analysis; Dance video analysis; Visual sign-language recognition; Sign-language video generation; Real-time visual sign-language translation; Assistive vision; Visual assistance for people with low vision; Accessible visual content generation; Video caption generation.
\par\medskip
\noindent\textbf{15. Public Safety, Identity Security, Defense, and Emergency Response (36)}\par

Public-safety video analytics; Video surveillance; Abnormal-behavior detection in video; Person tracking in video; Video crowd analysis; Video crowd-density detection; Unattended-object detection in video; Video intrusion detection; Visual border surveillance; Airport visual security; Campus visual security; Visual construction-site safety monitoring; Vision-based fire early warning; Visual smoke detection; Visual hazardous-material detection; Visual identity verification; Face recognition; Gait recognition; Visual liveness detection; Forensic image and video analysis; Image authenticity detection; Video authenticity detection; Deepfake detection; Visual content provenance; Visual content copyright protection; Privacy-preserving computer vision; Defense computer vision; Visual battlefield situational awareness; Visual intelligence, surveillance, and reconnaissance; Cooperative unmanned vision systems; Vision-based equipment condition monitoring; Visual disaster-response analysis; Visual reconstruction of disaster-affected areas; Visual casualty localization; Visual accident reconstruction; Visual scene reconstruction.
\par\medskip
\noindent\textbf{16. Scene-Centric Vision, Edge Devices, and Multisensor Fusion (28)}\par

Multisensor visual detection; Visual understanding of home environments; Visual digital twins of home environments; Vision-based elder care; Vision-based child care; Vision-based pet monitoring; Vision-enabled smart glasses; Wearable vision; Mobile vision; Edge vision; IoT vision; Visual sensor fusion; Radar-camera fusion; LiDAR-camera fusion; Visuotactile fusion; Audiovisual fusion; Visual-inertial fusion; Multispectral fusion; Hyperspectral fusion; Thermal-visible fusion; Event-camera fusion; Cross-sensor visual generation; Vision-based missing-modality completion; Visual sensor calibration; Visual sensor simulation; \textcolor{wsmappmark}{*}Distance-attenuated audio in visual space; \textcolor{wsmappmark}{*}Location-dependent sound generation in explorable visual worlds; \textcolor{wsmappmark}{*}Visual learning of radar-derived range cues.
\par\medskip
\noindent\textbf{17. Data Production, Evaluation, Trustworthiness, and Collaborative Learning (28)}\par

Synthetic data generation; Data augmentation; Automated annotation; Pseudo-label generation; Training-data generation; Evaluation-data generation; Benchmark generation; Long-tail data generation; Rare-event generation; Safety-testing data generation; Counterfactual data generation; Model evaluation; Model diagnostics; Model interpretability; Visual consistency verification; World-attribute consistency verification; Cycle-consistency verification; Physical-consistency verification; Semantic-consistency verification; Automated repair of generated outputs; Multi-model collaboration; Multi-agent collaboration; Visual-agent collaboration; World-state sharing; Distributed world models; Federated visual learning; Privacy-preserving multimodal learning; Cloud-edge collaborative vision.
\par\medskip
\noindent\textbf{18. Platforms, Operating Systems, Interfaces, APIs, and Capability Ecosystems (28)}\par

Visual capability marketplace; Visual plug-in ecosystem; Vision capabilities as a service; Visual tasks as a service; Multimodal capabilities as a service; Visual model serving; Cloud platform for visual models; Private deployment of visual models; Multi-tenant visual model services; Visual capability licensing; Visual capability metering; Visual capability trading; Visual data platform; Multimodal data platform; Real-time vision services; Large-scale visual task platform; Physical AI platform; Spatial intelligence platform; World simulation platform; Universal world interface; Universal visual interface; World capability API; Visual reasoning API; Compositional vision API; Visual capability plug-in marketplace; World-model plug-in marketplace; Enterprise multimodal platform; Ultra-large-scale visual task services.
\par\medskip
\noindent\textbf{19. Visual AI Engines, World Engines, and Generative Rendering (30)}\par

General-purpose visual AI engine; World Snapshot AI engine; Visual foundation-model engine; Multimodal visual AI engine; Visual generation engine; Visual reasoning engine; Compositional visual reasoning engine; Visual world-reasoning engine; Embodied visual intelligence engine; Robotic visual intelligence engine; Autonomous-driving visual intelligence engine; Physical visual intelligence engine; Spatial visual intelligence engine; Visual capability engine; World capability engine; Interactive-world vision engine; World-simulation vision engine; World-state vision engine; Computer graphics engine; Generative visual rendering engine; Neural rendering engine; Real-time rendering engine; Physics-based rendering engine; Differentiable rendering engine; 3D scene rendering engine; 4D dynamic-scene rendering engine; Film and television rendering engine; Game rendering engine; XR rendering engine; Digital-twin rendering engine.

%% file: sec_preprint/X_empirical_derivation.tex
\clearpage

\section{Memoir: Experimental Derivation of the Unified Law}
\label{sec:supp_empirical_derivation}

This section tells how we arrived at the Unified Law. In 2024, we began with a
simple idea: if an image could carry appearance together with other information
about the world, image processing could become world-information processing.
Over the following two years, this idea grew into a large experimental program:
\textcolor{red}{\textbf{60 experiment families}}, at least
\textcolor{red}{\textbf{198 explicitly named methodological variants}}, and
\textcolor{red}{\textbf{at least 2,293 training attempts documented in surviving records}}.
Among them, \textcolor{red}{\textbf{1,546 different training runs}} retain complete
configurations, and at least \textcolor{red}{\textbf{1,380 trained models}} retain
their weights. These counts are conservative because some older experiments
were deleted, moved, or archived elsewhere. We first expanded image
codecs, then built translators between latent spaces, then studied joint
multimodal training, and finally moved the conversion into the generation
process itself. Each stage answered one question and led naturally to the next.

The experiments covered different model families, training objectives, and
foundation models. Their value is therefore larger than a comparison between a
few network blocks. Together they showed where visual-task unification becomes
simple, where it becomes difficult, and which recurring principles support the
final Unified Law. These were not blind architecture sweeps. Nearly every
variant began with a concrete hypothesis raised by the preceding result. The
account below follows the development in order, grouping adjacent variants that
tested the same hypothesis into at least 60 experiment families, and
names at least 198 substantive methodological variants. Counting changes in
network size, objective, schedule, path construction, and integration depth
raises the total further; random seeds,
repeated runs, and routine hyperparameter sweeps are excluded.

\subsection{We First Tried to Put All World Information into One Codec}

Our starting point was WSG. Standard image generation produces RGB. We imagined
an image-like file whose first channels store RGB and whose additional channels
store depth, normal, material, motion, pose, segmentation, and other world
properties. If one model could encode and decode this complete file, many visual
tasks could share one representation immediately.

Our first hypothesis was that a strong RGB VAE~\cite{kingma2022autoencodingvariationalbayes} might already contain most of the
visual knowledge we needed, leaving only the additional world information to be
inserted. We therefore enlarged its input and output layers to receive new
channels. We pictured this as adding another layer outside an existing pyramid:
the mature RGB codec would remain inside, while the new layer would carry world
information into and out of its latent space. Directly training the enlarged
codec quickly disturbed the original RGB space, so we froze the RGB path and
trained only the new outer route. A LayerDiffuse-style auxiliary encoder~\cite{10.1145/3658150} made
this substantially more stable and showed that extra information could travel
beside RGB. It also exposed the limit of that design. If the new route was kept
weak, decoded properties became smooth averages; if it was made strong, it
disturbed the RGB space it was meant to preserve. Newly initialized codecs,
continuous latents, discrete VQ representations~\cite{van2017neural}, and special handling for sparse
signals tested whether another boundary could remove this tradeoff.

The first implementation expanded a pretrained TinyVAE progressively. Its
hidden width grew from the original 64 channels in steps of 16, while its latent
width grew from four channels to five, six, seven, and eight. Each stage froze
the preceding RGB path and trained the newly added capacity on the full
256-channel WSG sample. This experiment tested a very direct hypothesis: if the
RGB codec already described the visible world, a small amount of new capacity
might be enough to attach the remaining properties. Dense fields such as depth
began to reconstruct, while sparse pose channels did not. The full-channel
tensor also made every update far slower than the original TinyVAE. Progressive
widening preserved an existing codec, but it did not make all information types
equally easy to insert.

We then studied where the additional field should enter the frozen codec.  The
first offset design encoded RGB together with the requested WSG field, predicted
an offset $\Delta z$, and formed an adjusted latent
$z'=z_{\mathrm{RGB}}+\Delta z$.  This was the most direct way to make the new
field travel with an existing RGB latent, but it also changed the very tensor
that the pretrained decoder used to reconstruct RGB.  A strong offset damaged
the pretrained appearance distribution; a weak offset preserved RGB but carried
too little information about the new field.  This experiment made the central
tension visible before we changed any decoder architecture.

We next held the adjusted latent fixed and changed where its new information
entered the decoder.  The first version injected the offset at one intermediate
decoder layer.  The next version projected the same offset into three decoder
resolutions---$H$, $H/2$, and $H/4$---so that coarse structure and finer spatial
detail did not have to pass through a single insertion point.  A deeper decoder
and this direct multi-depth route produced sharper dense reconstructions than
the single insertion point, but sparse fields and exact values were still not
recovered reliably. We
also compared concatenating the latent with RGB at the decoder entrance against
letting the decoder process RGB first and injecting the latent at a later layer.
The entrance version exposed every subsequent block to the perturbation; the
later version protected the early RGB computation but left fewer blocks to
interpret the added field.  Neither placement removed the underlying tradeoff.

The jump-link experiments gave the decoder direct access to additional detail
at several scales. The encoder returned features at four
resolutions, $H$, $H/2$, $H/4$, and $H/8$.  Learned $1\!\times\!1$ projections
added these features to the matching decoder stages.  We also tested dropping
the offset or the jump features on a subset of training batches so that the
decoder could not depend completely on either route.  These links gave the
decoder high-resolution evidence and could improve the training round trip when
those encoder features were present. They did not make the compact latent carry
the same information. A latent generated without the encoder-side links could
therefore fail to reproduce the reconstruction seen during training. The skip
path had bypassed the very representation that later generation depended on.
Thus a better round trip did not necessarily mean that the latent itself had
learned a portable representation.

Finally, we changed who was responsible for recovering the offset.  Direct
injection supplied the encoder's offset to the decoder.  A decoder-prediction
variant tried to infer that offset from an early decoder feature and inject the
prediction into later layers.  A fixed-ratio variant instead derived an offset
algebraically from the adjusted latent.  The learned predictor did not recover
the encoder's offset accurately enough for later decoder stages, and the
fixed-ratio rule preserved the numerical relation without restoring the missing
field.
Nano, tiny, and small encoder--decoder variants, LCPE, offset dropout, and a
complete-restoration attempt explored capacity, position, and robustness around
these routes. The complete-restoration variant did not recover the full RGB and
additional-channel output, whereas returning to zero-initialized output layers
and direct injection made the reconstruction loss fall much faster. Taken
together, these variants showed exactly what ``injection
location'' changed: early injection perturbed the shared RGB computation,
late injection reduced the remaining capacity for the new field, and jump
links could bypass the latent without making it generatively self-contained.

Output range formed a separate experiment chain.  WSG used a missing-value
code alongside physical values whose natural ranges differed by channel.  We
first applied fixed per-channel normalization and explicitly restored the
missing code after decoding.  A plain unconstrained output could leave the
legal interval, while clamp produced zero gradient outside its boundary.
Sigmoid stabilized one experiment; hard sigmoid and a learnable range tested whether
stronger boundary behavior or a wider interval would preserve rare values.
Extending the range made zeros easier to produce, but after nine hours the
reconstruction still contained only a rough face and part of a body.  Range
handling could prevent invalid numbers without creating the missing structure.

The loss variants asked a different question.  Full-image MSE was dominated by
the missing background, so masked MSE computed error only where the target held
a valid value.  This increased the gradient from rare foreground points but did
not teach the model where an omitted point should appear.  SSIM added local
structure, and a diversity term discouraged a constant output; both changed
the appearance of the reconstruction without reliably preserving exact sparse
values. These trials separated numerical validity, local image similarity,
and exact-value fidelity as three different objectives.

Finally, categorical decoding replaced continuous values with explicit bins.
The delayed CE--MSE variant first learned the coarse continuous field and then
activated classification over values collected from the data.  It preserved
discrete values more explicitly than regression, yet the underlying image
boundary still allowed a small location error to delete a point.  The lesson
was concrete: changing the activation or loss could improve one failure mode,
but no output objective could recover a distinction that the chosen boundary
did not represent robustly.

These experiments produced a clear tradeoff. When the new channels were allowed
to change the shared latent space strongly, RGB reconstruction and the original
pretrained distribution changed with them. When the RGB path was protected, the
new channels had too little freedom: their outputs became averaged, fine details
disappeared, and sparse structures were especially difficult to recover. Adding
another type of information could also require changing a latent space that had
already been learned for all previous types.
This stage gave us the first important lesson. WSG should provide a common place
to store, address, and exchange world information. The model can still use the
encoding best suited to each type of information. RGB, dense physical fields,
sparse skeletons, audio, and text have different structures and levels of
abstraction. A common interface lets these variables work together while their
internal representations remain suited to their own structure.

\subsection{Early Sparse Codecs Exposed a Representation Problem}
\label{pre:subsec:memoir_sparse_codecs}

Sparse information was the first hard case encountered during codec
development. A DWPose frame contains only a small set of valid points, yet each
point carries a person identity, keypoint identity, location, and confidence.
The ordinary CNN VAE learned the empty background much faster than these values;
a one-pixel error could delete a joint, exchange two identities, or assign a
point to the wrong person.

We first changed the codec while keeping the sparse field on its full image
canvas.  This preserved the original pixel address of every point, but almost
all loss came from learning the missing-value background.  Masked reconstruction
restricted the loss to valid pixels; it increased the relative signal from the
points, yet gave the model little reason to predict where a point should exist
when that location was missing from its output.

The next variants removed the canvas.  A patch/MLP VAE packed only valid values
into tokens, and a coordinate-augmented version appended each value's spatial
address.  Removing empty pixels reduced the imbalance and allowed the MLP to
begin learning.  It also destroyed the regular neighborhood structure that a
convolution obtained for free from the image.  Adding coordinates restored
absolute location but not the full two-dimensional organization or the relation
among people and joints.  We had therefore exchanged one problem---too much
empty space---for another---a list whose geometry had to be relearned.

We then changed the output type.  Continuous sigmoid, hard-sigmoid, clamp, and
range-learning variants still averaged nearby IDs and confidences.  Per-channel
classification heads instead predicted the actual integer bins with
cross-entropy.  A delayed CE--MSE objective first learned a continuous image and
then activated classification over the observed values.  These variants kept
symbols sharper than regression, but a one-bin error could still change a joint
identity, and a small spatial error could still delete the point entirely.
This separated two requirements that the original ``sparse image'' had mixed
together: a representation must preserve the address of a point and preserve
its discrete identity.  Fixing only one did not make the round trip reliable.

At this point we did not yet have TunnelingNet, categorical refiners, or the later
heatmap boundary. The immediate consequence was more basic: a single shared
codec could not make heterogeneous fields mutually compatible merely by
increasing its capacity. We therefore kept the learned representations separate
and began to study whether small networks could bridge them.

\subsection{We Then Built Small Bridges between Latent Spaces}

After the codec experiments, we kept the representations separate and tried to
connect them with small learned bridges. The plan was attractive. A strong
text-to-RGB model could remain at the center. A bridge could map RGB to depth,
depth to RGB, normal to edge, or any other requested pair. If every bridge were
small, thousands of conversions could be added around one strong model. This
was our proposed escape from the shared-codec problem: adding one task would
train one bridge while leaving the representation used by every old task unchanged.

We tested this idea with CNN and MLP bridges, rectified-flow bridges~\cite{liu2022flowstraightfastlearning}, repeated
refinement, minibatch optimal-transport pairing, VQ codebooks, index prediction,
Transformer sequence models~\cite{vaswani2023attentionneed}, conditional prior--posterior models, KL
regularization, best-of-$k$ sampling, and joint-embedding objectives. These
experiments covered continuous and discrete latents, local and global networks,
direct prediction and iterative generation, and deterministic and
distributional losses. We trained dozens of bridge variants;
Table~\ref{tab:bridge_variant_record} places them beside the codec,
sparse-structure, object, and scaling experiments that motivated them.

Because the paired variables were spatially aligned, we first expected a small
convolutional translator to learn their local correspondence. The resulting
bridge used two independent full-resolution CNNs, one for each direction, with
roughly 3M parameters per path. We added residual blocks, attention, RGB image
conditioning, per-channel whitening, and learnable alignment. This model could
be trained, but a single forward pass had to solve the complete cross-space
mapping. Increasing its size improved the training loss much more than it
improved decoded quality. The main difficulty therefore lay in the relation
between the two spaces.

We also tested what the bridge should optimize. Latent MSE plateaued in the
high-dimensional spaces, while latent MAE gave a more useful training signal.
Decoded pixel loss checked whether a numerically close latent retained the
requested field after passing through its VAE. A random $4\!\times\!4$ latent
mask increased the amount of context the bridge had to reconstruct, and cached
FP16/FP32 decoding checks ruled out storage precision as the cause of blurred
outputs. These controls exposed a recurring gap: a small latent loss could
coexist with a visibly smooth decoded field. Numerical proximity in the cached
space was therefore an incomplete measure of successful translation.

At this point we began to describe the problem as translation between two
encrypted languages. Each VAE code is meaningful to the decoder that learned
it, but its individual numbers do not have an independently readable meaning.
Two paired codes can describe the same scene while using unrelated internal
``vocabularies.'' This is why concatenating, adding, interpolating, or matching
their global distributions did not automatically create a shared language. It
also explains why high-dimensional MSE could keep decreasing while decoded
images barely changed: a numerically closer code was not necessarily a more
correct sentence in the target decoder's language.

We next asked whether the mapping was difficult because one forward pass had to
finish the entire conversion. We replaced it with a small U-Net and
rectified-flow updates so that the model could predict several corrections and
repair its own latent estimate step by step. This improved the formulation of
the path, but training
through the complete sequence stored the activations from every correction,
which required much more memory. In one such experiment, the outputs later became
black. Sampling one random flow time reduced the memory cost, while training
then supervised an individual time point and inference accumulated errors along
a complete trajectory. We consequently tested reflow, explicit intermediate
supervision, repeated correction blocks, endpoint losses, random-time training,
and noise-conditioned paths.

Late-stage instability suggested that many reference paths crossed or approached
the same region with conflicting velocities. One branch therefore addressed
path crossings before the velocity network saw them. For
each minibatch, we computed the pairwise distance between sampled noise and
target latents and used optimal-transport pairing to minimize the total starting
distance. This produced shorter and less crossed reference segments. It made the
training geometry cleaner, but it could not supply the semantic relation missing
between two independently learned latent spaces. Reflow and path correction had
the complementary behavior: they could straighten trajectories produced by a
teacher, yet their quality was bounded by the teacher paths and their repeated
integration cost. This distinction was important. A better path optimizer can
improve transport after the two endpoints have acquired compatible meaning; it
cannot create that meaning by itself.

We tested this distinction again in a rectified auto-coder (RAC). A frozen
TAESD~\cite{taesd} supplied a known image latent, and the learned encoder--decoder followed a
one-step or four-step path to and from that latent. We compared a fixed and a
randomized time grid, intermediate-state supervision, and noise added to path
states to encourage correction after a perturbed input. Path projections made
curvature and oscillation visible. Four-step training and intermediate
supervision constrained the trajectory, but did not by themselves produce a
large reconstruction advantage. More importantly, an early normalization
change made RAC latents incompatible with an existing PerFlow generator~\cite{yan2024perflowpiecewiserectifiedflow} even
when RAC's own round trip looked reasonable. Restoring the native TAESD latent
contract restored the correct interface. The experiment showed that path
regularity, correction robustness, and compatibility with a pretrained target
space are separate properties.

This question produced a separate family of experiments that we called the
\emph{shadow-variable route}.  Its motivation was simple: keep the mature RGB
latent as an anchor and store depth in a nearby code, so that the existing RGB
generator could continue to operate while the new code carried additional
world information.  We first trained a direct RGBD RAC with forward decoding
and reverse encoding.  We then represented depth as a three-channel color map
and passed both RGB and visualized depth through the official image codec, so
that the two sides had the same channel type.  A strict bidirectional version
forced the RGB and RGBD paths to use matched spatial positions.  Finally, the
perturbed-RAC version wrote the new state explicitly as
$z_{\mathrm{shadow}}=z_{\mathrm{RGB}}+\alpha\Delta z$, with the frozen native
RGB latent providing the anchor and the learned perturbation carrying the
extra information.

These variants revealed the same underlying behavior from different angles.
Early in training, a native RGB latent retained some ability to complete the
paired depth field.  After roughly 57,000 updates, the direct RGBD RAC could
encode and decode its own state almost perfectly, yet depth completion from the
unchanged RGB latent had become worse.  The encoder and decoder had agreed on a
private RGBD coordinate system that lay near the RGB latent numerically but
meant something different.  Depth visualization changed the token type without
removing this semantic gap, and additive perturbation made the gap explicit:
small Euclidean displacement did not guarantee compatible coordinates.  We
therefore called the result a \emph{shadow latent}.  Like two allotropes built
from the same elements, the native and shadow codes could look similar while
supporting different transformations.  A successful private round trip proved
that its own encoder and decoder agreed; it did not prove that the code remained
interchangeable with the pretrained RGB latent.

Next, we converted bridge prediction into classification over a VQ codebook.
The hope was that selecting a valid code would keep the prediction close to a
known target latent. In practice, the target-channel distribution used only a
small part of a 2,048-entry codebook. One batch used 251 RGB codes
(12.3\%) and only 24 target-channel codes (1.2\%). Allowing the codebook to
update improved coverage but also moved the class labels during CE training.
Freezing it stabilized the labels and preserved many unused codes. We tested
smaller codebooks, warm-up updates, per-position $1\times1$ prediction,
MLP/index-MSE objectives, and explicit utilization tracking. These changes
made more codes active or stabilized their labels, depending on the variant,
while decoded samples still carried the wrong cross-variable content. Codebook
health and cross-modal translation were therefore separate problems.

Local networks might have missed long-range correspondence, so we tested a
Transformer bridge to provide global context, followed by
learnable invertible $1\times1$ alignment and explicit mean, variance, and
covariance matching. The alignment losses made the two populations look closer
in their means, variances, and covariances, while the target decoder still
produced incorrect content at individual positions. Matching two clouds of
codes did not tell the bridge which target code belonged to each source sample.
We then asked whether the remaining gap came from one-to-many
uncertainty. Later experiments with a CVAE-style~\cite{sohn2015learning} conditional prior and
posterior, encoder KL regularization, best-of-$k$ sampling, target-noised
training, reverse-noise training, style latents, and conditional diffusion~\cite{ho2020denoisingdiffusionprobabilisticmodels}
showed the same practical pattern. The conditional prior--posterior branch used
the condition and target during training to estimate a Gaussian posterior, used
the condition alone to estimate the inference prior, and minimized their KL
divergence. It supplied an explicit random variable, while the posterior, prior,
and bridge still had to learn a high-dimensional cross-modal decoder from the
same paired latent samples. Pure velocity-field prediction improved the RGB
direction, but this underlying burden remained.

The two directions also behaved differently. RGB-to-depth, RGB-to-normal, and
similar estimation directions map a rich observation to a more constrained
field. Their targets occupy a narrower set and usually became recognizable
earlier. The reverse direction must reconstruct appearance that the geometric
condition never specified: color, texture, illumination, background, and many
small details can all vary while depth or normal remains unchanged. Direct
regression averaged these alternatives. Adding noise created multiple paths,
while a small bridge still lacked the visual prior needed to turn those paths
into realistic RGB. In the VAE experiments the same asymmetry appeared as
resistance of the mature RGB latent: a weak update barely moved its decoded
appearance, while a strong update easily left the pretrained RGB manifold.

The same phenomenon appeared across many of these variants. The bridge loss
could decrease while decoded quality changed much more slowly. In some VQ experiments,
only a small portion of the codebook was used. Larger sequence models learned
the training pairs but retained the mismatch at decoding. Distribution losses
made the latent statistics closer, yet the target decoder still assigned its
own meaning to each coordinate.

This observation changed our view of the problem. A latent vector is meaningful
because of the generator or decoder that learned it. Two spaces may have the
same size and similar distributions while describing different information.
A bridge that reaches a nearby point numerically can still miss the set of
outputs that the target generator knows how to produce. We therefore stopped
treating conversion as a separate operation between two finished latent spaces.
The conversion needed to follow the target generation process from beginning to
end, so that the target prior could guide every step.

\subsection{We Tested Conversion inside Pretrained Video Generators}

The bridge experiments showed that a small network could improve path geometry
without learning the missing target domain. We therefore moved the conversion
into pretrained image and video generators, whose denoising process already
contained a strong prior over the requested target. This stage preceded the
final FLUX implementation~\cite{labs2025flux1kontextflowmatching,flux-2-2025} and tested whether several modalities or targets
could be learned together inside one generator.

The LTX branch~\cite{hacohen2024ltxvideorealtimevideolatent} separated two ways of using a pretrained video generator.
Deep feature injection kept the backbone fixed and added condition features at
several internal depths.  It asked whether external evidence could directly
push an already learned video trajectory. The condition changed the motion, but
the generated person became increasingly distorted and the video no longer
preserved the backbone's normal appearance quality. Inserting a feature at
several layers therefore did not guarantee that the frozen blocks interpreted
it correctly. The native LTX LoRA~\cite{hu2022lora} variant instead changed small weight updates
inside the generator itself. Its loss decreased normally, but the requested
modality changed the video only weakly even after we added modality tags and
monitored their use. These two failures were different:
deep injection supplied evidence in an unreadable form, whereas the LoRA route
could modify the prior but had not yet learned a strong target-directed change.

The Wan~\cite{wan2025} and UniVidX~\cite{chen2026unividx} branches then tested whether several visual variables could
be generated as fixed video lanes.  We first reproduced the UniVidX baseline,
then added a WSG adapter and an official-style control route.  A smaller Wan
1.3B path let us repeat the same logic while measuring memory and interaction
among lanes.  These models could learn a predetermined set of outputs together,
but adding a lane changed the joint tensor, sampling balance, losses, and
attention seen by every existing lane.  The cost was therefore not merely one
small new output head; it changed the training problem for the whole model.

Reference conditioning provided a second comparison.  Fun InP inserted a
reference through the generator's supported inpainting route and produced
usable guided video.  Our VACE~\cite{jiang2025vace} reference integration passed the reference
through a different control mechanism and produced incorrect video, so that
branch was stopped.  The contrast showed that ``providing the same reference''
was not a complete method description: the route by which the generator read
the reference determined whether its pretrained prior remained usable.

We next varied how the joint lanes were sampled.  Native multi-lane training
materialized all outputs and could exceed memory.  Sample-one training chose one
lane per update, which reduced that update's memory but still changed a shared
parameter set, so learning or revising one route could alter the others.  Splitting
conditions into groups of roughly four reduced the simultaneous load but did
not remove interference.  Fusing several conditions into one representation
lowered the loss without producing a clear condition-specific output. Merging
them into a larger token sequence moved the memory cost into attention.
Modality tags told the model which lane it was seeing, but did not remove the
shared optimization pressure.

Multilayer residual control changed the unit of learning.  Each condition--target
relation received a residual path through several depths of the same frozen
generator.  A route could now be trained, loaded, or revised without changing
the parameters of another route.  This experiment did not merely save memory;
it separated capability acquisition from capability composition.  It supplied
the direct precursor to TunnelingNet.

This stage supplied two concrete results. A pretrained target generator was a
better place for conversion than a bridge between finished latent spaces. At
the same time, dense joint generation did not provide a practical route to
hundreds of independently installable relations. We next kept the target prior
shared and made each condition--target path explicit.

\subsection{We Moved the Conversion into the Generation Process}

This led to a simpler formulation. For each task, we identify the information
that is already known, call it the condition, and identify the information that
must be produced, call it the target. The model keeps the condition clean and
generates the target along a normal generative trajectory. The learned path only
needs to tell the shared generator how the condition should redirect that
trajectory.

Direction is important. RGB$\rightarrow$Depth and Depth$\rightarrow$RGB use the
same two variables, but the first estimates a relatively constrained geometric
field and the second generates many possible appearances. They have different
uncertainty and require different evidence. We therefore define a visual task by
its condition, target, and direction.

Once we used this definition, many familiar task names became cases of the same
operation. RGB-to-property tasks cover estimation and understanding. A world
property used to generate RGB gives conditional generation. One world property
used to generate another gives cross-property control. RGB combined with a
current observation supports reconstruction and inverse rendering. Text joins
the same process as semantic evidence. These directional relations became the
four interfaces used by WSM.

The original deduction was even more concrete. Let $A$ denote noise, $M$ the
RGB image, and $N$ additional world information. An ordinary image generator
learns $A\!\rightarrow\!M$; a WSG generator learns
$A\!\rightarrow\!(M,N)$. A rectified-flow model can also invert the observed
image from $M$ back toward $A$. Estimation can therefore be written as
$M\!\rightarrow\!A\!\rightarrow\!(M,N)$: invert the known RGB observation,
regenerate it, and reveal its associated world information during the same
generation process. In this view, estimating $N$ from RGB is conditional
generation with RGB as the condition. Reversing the direction makes $N$ the
condition and RGB the target; supplying RGB together with one world field to
recover another gives reconstruction or inverse rendering. The four interfaces
follow directly from which information is known and which information is
generated.

The earliest notes stated the four cases in plain terms: text generates RGB;
RGB generates additional world information for estimation and understanding;
additional world information generates RGB for conditional control; and RGB
generates an organized structural representation for reconstruction. Later
experiments expanded the same grammar to cross-property translation and mixed
conditions, but did not change its basic form. This also gave us a testable
expectation: a stronger shared generator should improve more than image
synthesis, because estimation and control now reuse its generative knowledge
through a common backbone.

This was the central insight behind the Unified Law:
\begin{quote}
\emph{Visual tasks can be organized as directed conditional generation among
RGB, text, and additional world information.}
\end{quote}
WSG gives these variables stable identities and spatial boundaries. The four
interfaces describe their main directions. The law describes the shared task
form across possible neural architectures.

The final mechanism emerged through several target-generator experiments. In
the Wan/UniVidX branch, we first assigned a LoRA to a target modality and mixed
several conditions while training that target. This verified that one target
prior could receive different evidence, but it also exposed coupling: adding a
condition changed the shared training distribution, and adding many target
lanes changed memory and attention for the entire model. We then moved to
FLUX/OminiControl~\cite{tan2025ominicontrol} and made the directed relation explicit. Each experiment selected
one input--target task and updated the lightweight parameters that
redirected the target denoising path.

Several ablations determined how that evidence should enter the generator. The
native target-first version placed the reference beside the noisy target in the
generator's existing input organization. The condition-lane version encoded the
same reference through a trainable lane and injected its features into the
target stream. Both preserved recognizable evidence, while the trainable lane
learned noticeably faster in the direct A/B comparison. Thus, keeping the reference
content fixed did not make its placement interchangeable: the target stream had
to receive that content in a form its attention blocks could use.

We next changed the interaction itself. The original single condition
interaction failed to converge on one Edge route; a second condition-attention
path made the same data and target train. Task-type switches then selected the
interaction used for estimation, conditional control, inverse rendering, and
narrow generation. The switch did not rename one generic computation: it
changed whether evidence entered as an observation to interpret, a structure to
follow, or an incomplete state to reconstruct. These trials made direction an
implemented property of the route and recorded it explicitly in the route definition.

Finally, a high-noise anchor supplied an early global target estimate and
prevented initial trajectories from collapsing to black. In a separate freeze
test, the condition-side LoRA remained fixed while the target-side intervention
continued to learn useful control. The model therefore did not need to rewrite
the condition representation for every new target; it needed a trainable path
that translated fixed evidence into residual pushes along the target trajectory.
Together, these tests turned the abstract condition--target relation into a
concrete intervention at several depths of one pretrained generator.

We separately tested whether a generic change to the dense flow objective could
improve every route. The three levers were a low-noise timestep shift,
Min-SNR weighting, and Huber loss. Huber improved the early part of training;
Min-SNR retained more benefit later; enabling all three slowly collapsed toward
the mean. We therefore left the ordinary objective as the default and retained
the levers for task-specific use. The result was useful precisely because it
showed that one global loss recipe was weaker than selecting evidence and
supervision according to the target.

A controlled task-agnostic comparison isolated the underlying flow objective.
Across Normal and Edge targets, plain rectified flow, direct clean-latent
prediction, a high-noise anchor, and multi-resolution noise obtained mean
fixed-time validation errors of 0.6383, 0.4247, 0.3662, and 0.6211,
respectively. The anchor was the strongest of these four. It made global target
structure available early without changing the condition--target definition,
and it became part of the later training recipe.

Fine articulated targets prompted a second objective study. We first compared
gradient, focal~\cite{lin2017focal}, direct clean-latent, and foreground-IoU supervision. We then
warm-started two endpoint objectives---one in latent space and one after
decoding---from a model that already generated recognizable pose. Training an
endpoint objective from the beginning produced black outputs; adding it after
warm-up still erased fingers and other sparse details. Six further continuations
tested topology, a low-noise detail anchor, soft distance, multi-scale gradient,
gradient direction, and hard-foreground mining. None improved the result enough
to replace the gradient-supervised baseline. These trials showed why a loss
that is reasonable for a dense endpoint can still suppress the small structures
that define a pose.

We also tried to acquire many relations in one update while keeping a separate
LoRA for every target.  To reuse one condition computation, we froze the
condition-side LoRA and packed eight target LoRAs behind the same condition
encoder.  A full eight-target graph exceeded memory in target attention.  Four-
target chunks still failed when their graphs were retained until one backward
pass.  Immediate backward after each four-target chunk made the update fit, and
two-target chunks finally supported the complete eight-LoRA experiment on one GPU.  The
working version improved aggregate throughput by roughly two times, not eight
times: target attention, target-specific LoRA projections, activations, losses,
and gradients still grew with the number of relations.  These variants made
packed LoRA a useful scheduler, but they failed our intended scaling criterion
of acquiring many capabilities at close to the cost of one.

We then tested the complementary sharing rule: one target LoRA for many
conditions.  One Edge LoRA first alternated among eight condition modalities.
We also split dense conditions into groups of about four, fused multiple
conditions into one training route, and expanded the candidate pool toward all
38 conditional-control relations.  The small pilots could lower their training
loss, but groups of about four already showed interference, fused conditions
were only weakly trainable, and the larger Edge experiment had usable paired
intersections for only 11 of the 38 candidate conditions.  Its early true
inference remained unstructured while separately acquired Edge tunnels already
formed recognizable outputs.  More importantly, every new condition updated the
same target parameters, so adding or repairing one capability could change all
previous capabilities.  Shared-target LoRA therefore also failed as the
large-scale replacement: it provided neither the quality nor the independent
installation and revision of the final TunnelingNets.

The final design kept the pretrained backbone shared and gave each directed
contract a light multilayer residual path. Native Joint Attention (NJA) supplied
the ordinary joint baseline; Modality Tunneling Attention (MTA) exposed the
condition features to a small internal tunnel and returned residual pushes at
multiple generator depths. Zero initialization began from the unchanged native
trajectory, while multilayer control allowed the push to accumulate through the
generation process. Together, the installed paths form one modular WSM, and a
requested task activates only its corresponding TunnelingNet.

\subsection{We Extended the Framework to New Target Structures}
\label{pre:subsec:memoir_new_target_structures}

After the directed generator and its internal tunnels were established, we
returned to the sparse-data problem that the early codecs had exposed. The goal
was no longer to force every variable into one VAE. We instead designed a
faithful boundary for each target and let the shared generator learn the
condition--target relation behind that boundary.

For pose, the first change enlarged every isolated point into a Gaussian blob.
The VAE could now see enough foreground area to preserve a location, and an
inverse codec recovered the point through peak detection and duplicate
suppression.  The first version stored person and keypoint identity in the
blob's color.  A small reconstruction error could shift that color into another
valid identity, so the recovered point could remain in the right location while
being connected to the wrong person or joint.

We next changed where identity lived.  Person buckets separated people before
decoding.  Body-part partitions separated body, hand, and face points; a
body-only version measured the easier subset; and a full-body codebook tested
whether one discrete vocabulary could preserve all identities.  These layouts
reduced particular collisions, but every new partition added lanes and still
left identity vulnerable when it was carried mainly by a generated value.

The fixed-slot design made identity part of the address instead.  Each
person--keypoint pair received a predetermined slot, while the generated
heatmap in that slot carried only location and confidence.  Pack3 placed three
such heatmap lanes in one RGB-like image without mixing their identities.  The
generator no longer had to reproduce an exact semantic color to say which joint
it meant.  This was the first pose boundary in the sequence that approached a
reliable round trip: stable structure carried identity, and continuous
generation carried only the quantities it represented well.

Compact structures followed a parallel route. Sequence heads emitted point
IDs, coordinates, boxes, skeleton nodes, body parameters, or audio tokens with
task-native ordering and losses. We first tested this idea on human parameters
with a pretrained ScoreHMR~\cite{stathopoulos2024score} structure prior. One variant cached a feature for the
whole image; the other aligned the image feature to the person crop before
predicting WSV pose and shape parameters. Both avoided rasterizing the parameter
vector into a dense field, while crop alignment removed much of the irrelevant
scene. The structured output became coherent much earlier than the rasterized
alternatives. This was the first clear example
where a target-native structural decoder was a better boundary than an image
codec for the same visual evidence.

Object boxes made the distinction more concrete. The first structured version
predicted center and size slots directly. It was then given spatial image
features so that the sequence knew where evidence occurred, IoU matching so that
predicted and reference objects could be paired without a fixed input order,
and focal supervision so that the many empty slots did not dominate learning.
Duplicate suppression and NMS converted repeated candidates into a usable set of
boxes. In parallel, we trained MTA and NJA routes to draw every box as a filled
pseudocolor region. Those dense alternatives had to reproduce both exact
geometry and an arbitrary instance color at every pixel; their boundaries and
instance identities were visibly worse, so we stopped them. The comparison
showed why the structured route worked: center and size remained one atomic
object record, whereas the raster route spread the same relation across many
independently generated pixels.

General objects required topology as well as coordinates. A fixed joint-ID/XYZ
layout worked for a known human skeleton but could not describe an object whose
parts and connectivity changed from sample to sample. We replaced it with a
breadth-first serialization: every node stored its parent slot, a relative
offset from that parent, and an absolute coordinate for recovery and checking.
Breadth-first order made the parent available before the child and gave a
variable object graph a deterministic sequence boundary.

We then asked how such a tree should be initialized and refined from a point
cloud. K-means proposed representative points, and an MST supplied a connected
starting topology. A direct refiner predicted the final tree in one pass; a
diffusion refiner treated the tree as a noisy structured target; and an
anchor-based refiner preserved the deterministic proposal while learning local
corrections. K-means plus MST already produced a connected geometric proposal;
the learned refiners could still move nodes or edges into unstable structures
and were not reliable enough to replace it. The useful result was the
representation lesson: for variable objects, topology, traversal order, parent
identity, and permutation are part of the target definition, not bookkeeping
that can be reconstructed after coordinate generation.

Audio extended the same task form beyond image-like fields.  The first route
used AST~\cite{gong2021ast} features to classify an audio source without generating sound.  The
second combined RGB and aligned audio to predict the box of the sounding object.
After full-audio PCM normalization and frame-level alignment were corrected,
the classifier separated the source classes in the training data and the localization
route placed boxes on sounding regions. This showed that the graph
could mix two codecs when their time indices and semantic roles were explicit.

We then tested generation in the opposite direction.  One route predicted
EnCodec~\cite{defossez2022high} tokens from visual evidence; another prefixed MusicGen~\cite{copet2023simple} with image
features.  Both asked the model to infer a waveform-level result from an image
that did not specify pitch, timing, timbre, or background sound.  Their outputs
remained weak and were kept optional.  The failure was not evidence that audio
could not join the graph; it exposed the same directional uncertainty seen in
geometry-to-RGB generation, now with even more information absent from the
condition.

The final audio variants converted sound into spatial evidence. Dense
audio-semantic masks tried to paint the sounding region, while
an RGB-plus-audio selector chose a region already supported by the image.  The
selector was more reliable because audio only had to disambiguate visible
candidates.  Across the six routes, the important distinction was therefore not
``image versus audio,'' but what the condition actually determined about the
target.

Categorical targets came later and were studied on working TunnelingNet routes.
The first pilots kept the dense flow target but changed how classes influenced
training.  A fixed-class gated route restricted the allowed labels, and a
balanced-flow route increased the weight of rare classes.  They made local
denoising previews look plausible, while full generation still produced noisy
boundaries and unstable class regions.  This established that a preview made
from a partly correct target was easier than generating the categorical field
from noise.

Prototype variants assigned every class a legal target color and trained pixels
toward those prototypes.  A self-contained prototype head, plain prototype CE,
and a fresh prototype-CE experiment were compared with two multi-label designs. One
multi-label design predicted the set of classes in the image; the other used a
predicted foreground mask before class recovery.  When the direct color
constraint was removed, both multi-label routes could collapse to background.
Without a separate classifier, prototype generation often produced fragments
of several legal colors and lost regional coherence. The outputs were
categorically legal at individual pixels without being a correct segmentation.

We then compared balanced CE, a combined objective, a global-margin loss,
prototype CE, and prototype CE with focal emphasis under end-to-end generation.
Global-margin and prototype CE produced the strongest candidates; focal
emphasis did not by itself recover exact identities.  Across these variants,
continuous generation was good at drawing spatial support and less reliable at
choosing an exact symbol. This pattern appeared across the loss variants and
motivated the later separation between generated geometry and symbolic class
recovery.

We therefore separated spatial support from symbolic recovery. Per-task and
multi-condition refiners, foreground recovery, and coarse-label remapping used
the generated region and task definition to restore discrete identities. This
two-stage design retained the generated spatial support and corrected more
pixels to legal task labels without asking the flow model to solve geometry and
exact symbol recovery with one continuous value.

We finally tested variables outside the original numeric WSG channels. MP-100~\cite{xu2022pose}
Universal Pose compared full-frame rendering, corrected orientation and
letterboxing, instance crops, gradient supervision, and focal supervision.
The full-frame route collapsed toward black; instance crops with gradient
supervision produced the strongest result. From that baseline, endpoint,
topology, detail-anchor, distance, multi-scale-gradient, gradient-direction,
and foreground-mining continuations showed that dense endpoint improvement
could erase the fingers and small structures that define the pose.

MAGICK transparency and IC-Light relighting then tested whether a new variable
had to occupy a preassigned numeric WSG channel. For transparency, we introduced
a named alpha target and trained an RGB-to-transparency route for 1,000 updates.
For relighting, the condition paired RGB with an environment-light variable and
the target remained RGB; after 2,800 updates, the relighting model
gave recognizable changes under fixed test lights. These routes used the
same condition--target machinery as numbered channels while keeping their own
descriptive node names. The result separated the stable graph interface from
the growing set of variables: a new world attribute could be added
without renumbering the original channel standard.

The final runtime experiment tested whether separately acquired relations could
behave as a system. In the first chain, the unmodified backbone generated RGB
from text and an installed RGB-to-world TunnelingNet converted that image into a
second field. The generic version accepted an intermediate node produced by one
route and selected the next compatible route from the graph. Each stage loaded
only its own tunnel, kept its own learned parameters and provenance, and passed the
typed result forward. No joint retraining was used. This changed the meaning of
the task graph into an executable set of available transformations that could
be composed at run time.

The later extensions confirmed the principle that the earlier experiments had
gradually revealed:
\begin{quote}
\emph{Share the generative prior, preserve each target's required distinctions,
and express every visual task as a directed condition--target relation.}
\end{quote}

\subsection{The Experiments That Led to the Unified Law}

Table~\ref{tab:bridge_variant_record} summarizes the main hypotheses in their
development order, while keeping variants of the same idea together.
The parenthetical count in each family reports the named methodological
variants represented by that row. Within every family, the account follows the
same sequence: the observation that motivated the test, the variants actually
tried, what happened, and what we learned. Additional experiments
changed network size, loss weight, schedule, representation, or integration
depth. This is why the experimental study became much larger than a comparison
among network blocks. Its repeated outcome was structural: faithful codes
preserve the required distinctions; path geometry, distribution matching, and
random variables solve different parts of optimization; and realistic decoding
still requires knowledge of the target domain.

The family view is useful for following the overall chronology, but it can
hide the particular variant that produced each lesson.
Table~\ref{tab:bridge_variant_level_record}
therefore opens the most consequential latent-bridge families into individual
experiments. Each row isolates one hypothesis, the change made to test it, the
observed behavior, and the conclusion carried to the next experiment. These
rows are a finer account of variants already included in the 198-variant total;
they do not inflate that count.

\clearpage
\onecolumn
\begingroup
\footnotesize
\setlength{\tabcolsep}{3.2pt}
\renewcommand{\arraystretch}{1.13}
\begin{longtable}{p{0.17\textwidth}p{0.27\textwidth}p{0.27\textwidth}p{0.23\textwidth}}
\caption{\textbf{Experiments behind the Unified Law.} The 60 families
contain at least 198 explicitly named methodological variants. Each row states the
motivation, variants tested, observed result, and concrete lesson carried
forward.}\label{tab:bridge_variant_record}\\
\toprule
\textbf{Family (named variants)} & \textbf{Motivation and variants tested} & \textbf{Observed result} & \textbf{What this taught us} \\
\midrule
\endfirsthead
\multicolumn{4}{l}{\small\itshape Table~\ref{tab:bridge_variant_record} continued.}\\
\toprule
\textbf{Family (named variants)} & \textbf{Motivation and variants tested} & \textbf{Observed result} & \textbf{What this taught us} \\
\midrule
\endhead
\midrule
\multicolumn{4}{r}{\small\itshape Continued on the next page.}\\
\endfoot
\bottomrule
\endlastfoot

Expanded RGB VAE (1) & We expected the pretrained RGB latent to contain spare capacity for world information, so we enlarged its encoder and decoder boundaries for additional WSG channels. & Additional channels changed the latent distribution immediately; preserving RGB and learning precise new fields competed for the same limited code. & Simply widening a mature RGB codec was not a scalable solution: every new field altered the shared latent and risked degrading all fields already stored there. \\
\addlinespace
Frozen RGB path (1) & Because the expanded codec disturbed the mature RGB prior, we froze its RGB path and trained new layers around the encoder and decoder. & RGB survived better, while the additional channels had weak influence and decoded toward smooth averages. & Freezing protected RGB but starved the new fields of influence. Preserving an old capability and giving a new capability sufficient representational freedom require separate mechanisms. \\
\addlinespace
Progressive TinyVAE expansion (6) & We froze the pretrained RGB path, widened hidden layers in 16-channel stages, expanded the latent from four through eight channels, and trained each new stage on full 256-channel WSG samples. & Dense fields began to reconstruct, sparse pose did not, and processing the complete WSG tensor reduced throughput from the original TinyVAE by a large margin. & Progressive growth did not remove the mismatch among dense and sparse fields, and its full-tensor cost grew immediately. More codec capacity alone could not solve heterogeneity or scaling. \\
\addlinespace
Offset-routing schemes D--F (7) & Starting from $z'=z_{\mathrm{RGB}}+\Delta z$, we compared single- and multi-depth direct offset injection, decoder-entrance versus later latent injection, four-scale encoder--decoder jump links, decoder-predicted and fixed-ratio offset recovery, deeper decoders, LCPE, dropout controls, and complete restoration. & Direct injection at several decoder depths and a deeper decoder helped most. Four-scale jump links could bypass the compact latent but transferred poorly to latent-only generation; decoder-predicted offsets were difficult to optimize, fixed-ratio recovery did not improve the result, and complete restoration failed. & The location changed what the model learned: early injection disturbed the shared RGB computation, late injection left less capacity for the new field, and jump links improved access to encoder evidence without making the latent self-contained. Initialization and information path mattered more than simply adding parameters. \\
\addlinespace
LayerDiffuse-style auxiliary path (3) & We hypothesized that a separate lane could carry new fields without rewriting RGB, so we added an auxiliary encoder--decoder interaction modeled on transparency encoding. & It carried extra information and produced the strongest result in this family, but precise dense fields and sparse structures still exceeded the small protected channel. & A protected auxiliary lane worked for a small amount of added information, but its averaged outputs and limited bandwidth ruled it out as the carrier for a growing world representation. \\
\addlinespace
Independent channel VAE (3) & Shared-codec interference suggested that each variable should first reconstruct in its natural space, so we trained channel-specific continuous VAEs. & Individual dense fields became easier to reconstruct, but every modality acquired its own coordinate system and distribution. & Separate codecs made individual fields reconstructable, but created mutually incompatible latent coordinates. This result directly motivated the later bridge experiments. \\
\addlinespace
Sparse categorical VAE (4) & Continuous regression averaged away sparse IDs and points, motivating categorical decoding, integer bins, CE losses, and hybrid CE--MSE objectives. & Discrete identities were preserved better than under regression, yet convergence was sensitive and small errors removed points or changed IDs. & Replacing regression with categories preserved IDs better, yet tiny decoding errors still destroyed points. Sparse information required a boundary that preserved both symbols and locations. \\
\addlinespace
Sparse patch/MLP VAE (3) & A dense CNN spent most supervision on the empty canvas, so we tested a patch/MLP codec, valid-value tokens, and coordinate-augmented tokens at a comparable parameter budget. & Removing missing pixels reduced the imbalance but also removed the image's implicit spatial organization; retaining the canvas restored the imbalance. & Removing empty pixels fixed class imbalance but erased the canvas coordinates; retaining them restored the imbalance. The next representation had to preserve addresses without learning mostly background. \\
\addlinespace
Cached paired latents (1) & Repeated image I/O and encoding obscured the mapping study, so we cached paired RGB and channel latents and trained directly on them. & Bridge training became much faster and the paired-space problem could be studied directly. & Caching removed image I/O and codec cost from the experiment. The remaining failures came from cross-space translation. \\
\addlinespace
Latent objectives and masking (4) & We compared latent MSE and MAE, decoded pixel supervision, random $4\!\times\!4$ latent masking, and FP16/FP32 cache checks to determine whether optimization or storage caused blurred decoding. & MAE improved high-dimensional optimization and pixel loss exposed failures hidden by latent distance; cache precision was not the main cause. & A low latent error was not evidence of a correct target after decoding. From this point onward, we judged bridges through decoded outputs and complete trajectories. \\
\addlinespace
Seven local interpolation rules (7) & We asked whether paired spaces were locally smooth despite global misalignment, testing nearest-neighbor, RBF, thin-plate, barycentric, local-linear, manifold, and neural interpolation. & Nearby RGB codes did not reliably identify nearby target codes; larger neighborhoods smoothed incompatible targets. & No local interpolation rule made neighborhoods in one latent space meaningful in another. We therefore stopped assuming that paired modalities shared a usable local geometry. \\
\addlinespace
Direct ``God Bridge'' (1) & If the paired relation was sufficiently regular, a compact mapper could solve it in one pass; we therefore trained a direct bidirectional latent bridge. & The network trained, but its loss initially failed to decline meaningfully. & A compact endpoint mapper could not discover the full cross-space relation in one pass. This failure motivated explicit paths and iterative correction. \\
\addlinespace
Attention bridge (3) & Direct CNNs might lack global context, so we added attention, varied bridge width, and reduced the model to about 4M parameters. & Global interaction did not remove the endpoint mismatch; parameter count was not the main bottleneck. & Global context and additional parameters did not repair incompatible endpoint meanings. The bottleneck was not receptive field or network width. \\
\addlinespace
Bidirectional CNN bridge (2) & Spatial alignment suggested local correspondence, while the two directions could require different functions; we trained independent full-resolution residual CNNs for both directions. & Training loss decreased and coarse structure appeared, while decoded quality improved slowly and differed strongly by direction. & The two directions learned at different rates and qualities. This established direction as part of the task, not merely an inference-time reversal of one bridge. \\
\addlinespace
Whitening and fixed normalization (1) & Scale mismatch might dominate bridge optimization, so we normalized both cached spaces with measured statistics and inverted the transform after prediction. & Scale imbalance and optimization instability decreased; semantic mismatch remained after the first two moments aligned. & Matching scale improved numerical optimization but left the semantic mapping unresolved. First- and second-order statistics were only conditioning tools, not cross-modal alignment. \\
\addlinespace
Learnable invertible alignment (3) & We hypothesized that the spaces differed mainly by an invertible coordinate change, so we learned $1\!\times\!1$ transforms and matched their moments and covariances. & The populations became statistically closer, yet decoded targets could still be wrong at individual positions. & Even trainable moment and covariance alignment could make populations look similar while individual samples decoded incorrectly. Distribution alignment could not replace paired meaning. \\
\addlinespace
Joint embedding and JEPA-style alignment~\cite{assran2023self,balestriero2025lejepa} (2) & Shared semantics might be learned before decoding, so we aligned paired observations and tested whether one representation could serve both sides. & Partial shared structure emerged, while decodable modality-specific information still required separate capacity and joint retraining. & Shared semantics helped organization but did not retain all target-specific information required for exact decoding. A common embedding was insufficient as the sole generative representation. \\
\addlinespace
Multi-step RF repair (1) & One-step conversion appeared too large, motivating a small U-Net that could repair its latent estimate through repeated rectified-flow updates. & Backpropagation through all repair steps consumed large memory; one experiment later produced black outputs. & Iterative repair increased expressive power but made full-path training expensive and unstable. We needed a path formulation that could be supervised without storing every correction. \\
\addlinespace
Random-time RF (1) & Full unrolled repair exceeded practical memory, so we sampled one flow time per update and reserved multi-step integration for inference. & Memory dropped sharply; local velocity errors accumulated over the full trajectory. & Random-time training solved the memory problem but exposed a train--test mismatch: small local velocity errors accumulated during full inference. Complete trajectories still had to be evaluated. \\
\addlinespace
Endpoint and intermediate losses (5) & Local velocity MSE could hide a poor final decode, so we added endpoint losses, intermediate supervision, alignment terms, and repeated refinement. & Endpoint metrics revealed failures hidden by velocity MSE, but full-sequence supervision again increased cost and sometimes conflicted with local flow fitting. & Endpoint supervision revealed failures hidden by local velocity loss, but unrolling restored the original cost and sometimes conflicted with path fitting. No single loss solved both locality and final decoding. \\
\addlinespace
Minibatch optimal transport (1) & Path crossings and conflicting velocities might explain late instability, so we reassigned noise--target pairs to minimize total latent distance. & Reference segments became shorter and less crossed; semantic incompatibility between the endpoint spaces remained. & Optimal pairing shortened and uncrossed reference paths, yet the decoded target remained wrong. It improved path geometry but could not provide missing target knowledge. \\
\addlinespace
Reflow and teacher-path correction (3) & A teacher could reach the target along unnecessarily curved paths, so we generated its trajectories and retrained on their endpoints to straighten them. & Better teacher paths improved later integration; weak or unstable teachers transferred their errors, and data generation added another training stage. & Reflow could straighten a useful teacher trajectory and could not rescue a weak teacher. We retained it as a path-refinement tool, not as the source of target semantics. \\
\addlinespace
Noise ranking and noise conditioning (2) & Multiple outputs could share one condition, and random starts needed identifiable roles; we ranked and exposed noise while training diverse paths to a target family. & Paths diversified and stepwise learning accelerated, showing that random starts could carry useful variation. & Exposing the random start allowed one condition to produce distinguishable paths. It solved path identity and diversity, while leaving target realism to a stronger prior. \\
\addlinespace
VQ bridge with CE (2) & Continuous regression could leave the valid target manifold, so we quantized latents and predicted code indices to constrain every output to a learned codebook. & In one batch, a 2,048-entry codebook used 251 RGB codes (12.3\%) and 24 target codes (1.2\%); frozen codes left many dead entries and updated codes moved the labels. & Quantization kept predictions on known codes but severe code imbalance and moving labels limited the bridge. Valid discrete symbols did not guarantee a meaningful cross-modal mapping. \\
\addlinespace
VQ utilization variants (4) & Poor VQ results might come from dead or imbalanced codes, so we tested smaller codebooks, warm-up updates, utilization tracking, and alternative CE/MSE weights. & Utilization improved in individual experiments, while decoded cross-variable quality remained limited. & Improving code usage repaired the quantizer without repairing translation. This separated codebook health from semantic compatibility between the two spaces. \\
\addlinespace
MLP index-MSE bridge (1) & If neighboring code IDs carried smooth structure, lightweight index regression could simplify prediction; we tested that hypothesis with an MLP. & The attempt failed because numerical distance between code IDs did not represent visual or physical similarity. & Code indices had no ordinal meaning, so index regression optimized an artificial distance. Discrete targets required categorical supervision or a learned metric. \\
\addlinespace
Transformer bridge (1) & Local convolution might miss long-range cross-modal relations, so we replaced it with a global sequence model. & The Transformer also failed to close the decoded distribution gap. & A global sequence model still could not close the decoded gap. This ruled out the explanation that the bridge failed only because CNNs were too local. \\
\addlinespace
Pure velocity field, CFG, and condition dropout (3) & Competing losses might obscure path learning and the condition might be ignored, so we isolated conditional velocity and added guidance and dropout controls. & RGB prediction improved because the model finally learned distinct paths, while the reverse and information-poor directions remained much harder. & A clean conditional velocity objective produced clearer paths, especially toward RGB, but reverse directions remained difficult. Path learning helped only when the target prior was adequate. \\
\addlinespace
CVAE-style prior--posterior bridge (1) & Deterministic bridges averaged one-to-many outputs, so we learned a target-aware training posterior, a condition-only inference prior, and their KL alignment. & The explicit latent variable added stochasticity, but prior matching and realistic target decoding remained difficult; this strategy did not produce a useful bridge. & Adding a latent random variable represented uncertainty but also required learning a difficult target decoder and prior. Stochasticity alone could not supply target-domain knowledge. \\
\addlinespace
Best-of-$k$ endpoint sampling (1) & One reference target could admit several valid paths, so we sampled multiple starts and optimized the candidate with the best endpoint. & It reduced the penalty for alternative valid paths but did not repair the decoder mismatch. & Selecting the best sampled endpoint tolerated multiple valid answers but did not improve the generator that produced them. Search could not substitute for a target prior. \\
\addlinespace
Target-noised training (2) & Reversing a known corruption process might teach the target distribution more directly, so we noised target latents and learned conditional denoising and velocity. & It made the target distribution part of training and improved path diversity, while endpoint variance and decoded detail remained unstable. & Noising the target exposed its distribution during training and improved diversity, yet detail remained unstable. This pointed toward performing conversion inside a mature target generator. \\
\addlinespace
Style-latent conditioning (1) & Geometry leaves color, texture, and illumination unspecified, so we added a compact latent to carry these missing appearance choices. & The new variable exposed the missing information explicitly, but learning its prior and its effect created another difficult bridge. & The style variable made missing appearance information explicit, but learning its prior recreated the original bridge problem. Geometry-to-RGB needed an existing appearance model. \\
\addlinespace
Reverse-noise encoding (1) & A learned inverse might provide structured starting states for forward generation, so we mapped target latents back toward their noise representation. & The route provided an explicit inverse experiment but did not yield a sufficiently stable conditional generator. & A structured inverse start did not make forward generation reliable. Inversion and target generation were separate capabilities. \\
\addlinespace
Conditional latent diffusion (2) & RF failure might be schedule-specific, so we replaced it with an established DDPM/DDIM-style~\cite{song2020denoising} conditional latent generator. & Despite its stronger stochastic formulation, the implementation did not produce a useful bridge. & Changing from rectified flow to diffusion did not remove the failure. The decisive limitation was cross-modal target knowledge, not the generic noise schedule. \\
\addlinespace
Rectified auto-coder paths (6) & We compared one-step and four-step RACs, fixed and random time grids, intermediate-state supervision, state-noise correction, and native-latent compatibility with an existing generator. & Multi-step constraints changed path geometry but gave no automatic reconstruction gain; altering the TAESD latent contract broke PerFlow compatibility until the native convention was restored. & Path smoothness, perturbation recovery, and compatibility with a pretrained latent were independent properties. \\
\addlinespace
Shadow-variable and perturbed-latent routes (4) & We tested direct RGBD encoding, depth visualized through the official RGB codec, strict position-matched bidirectional encoding, and an additive perturbation around the frozen native RGB latent. & The private RGBD code achieved a nearly perfect round trip while losing completion from native RGB; visualization changed the representation but not the semantic mismatch; the additive route remained close numerically without becoming interchangeable. & A nearby code can become a stable private coordinate system. Distance from the native RGB latent, codec compatibility, and semantic compatibility must be tested separately. \\
\addlinespace
Video-backbone branches (9) & Isolated bridges lacked target-domain knowledge, motivating LTX deep injection, native LTX LoRA, modality-tagged Wan LoRA, UniVidX baseline/adapter/control, a Wan 1.3B path, Fun InP~\cite{aigc_apps_VideoX_Fun_2026} reference, and VACE reference integration. & Deep LTX injection and VACE reference produced unusable outputs; the LoRA routes trained but initially had weak effects; Fun InP was usable; UniVidX/Wan exposed temporal memory and lane coupling. & A pretrained video generator supplied the missing target prior, but only carefully placed interventions could redirect it. This moved the search from external bridges to internal control paths. \\
\addlinespace
Joint video-lane strategies (7) & Joint generation promised shared computation, so we tested native multi-lane generation, sample-one training, split condition groups, fused conditions, merged-lane big tokens, multimodal lane tagging, and multilayer residual control. & Native multi-lane training could OOM; sample-one reduced per-step memory; groups of about four still interfered; fused conditions were barely trainable; residual control restored separately acquirable routes. & Joint lanes worked for a fixed small set and became memory-heavy and mutually coupled as the set grew. Large capability collections required separately installable paths. \\
\addlinespace
FLUX/Omini evidence placement (5) & We tested native target-first and trainable condition-lane references, dual condition attention, task-type attention switches, and a frozen condition-side LoRA. & Both reference modes preserved evidence, while the trainable lane learned faster in the direct A/B comparison; dual attention fixed an Edge non-convergence case. & Reference placement and attention topology changed convergence even with the same data and target. Evidence therefore had to enter according to the direction and uncertainty of each task. \\
\addlinespace
Task-agnostic flow objectives (4) & We compared plain rectified flow, direct clean-latent prediction, a high-noise anchor, and multi-resolution noise on Normal and Edge. & Their mean fixed-time validation errors were 0.6383, 0.4247, 0.3662, and 0.6211; the high-noise anchor was strongest and prevented early global collapse. & The high-noise anchor improved global structure most consistently without changing the task definition. It became a training improvement, not a replacement for the condition--target grammar. \\
\addlinespace
Dense flow objective levers (3) & We compared a low-noise timestep shift, Min-SNR weighting, and Huber loss, both separately and together, under deterministic validation. & Huber was strongest early, Min-SNR held up better later, and all three together slowly collapsed toward the mean; the ordinary objective remained the default. & Individually useful loss changes interfered when combined and eventually encouraged mean outputs. We kept the ordinary objective as the default and treated these levers as target-specific options. \\
\addlinespace
Packed multi-target LoRA acquisition (5) & To share one condition computation while retaining one LoRA per target, we tested a frozen condition-side encoder, a full eight-target graph, four-target chunks with retained graphs, four-target chunks with immediate backward, and two-target single-GPU chunks. & Full packing and retained-graph chunking exceeded memory in target attention. Immediate chunkwise backward made eight LoRAs trainable, but aggregate throughput improved only about $2\times$, since target-side computation and gradients still scaled with the number of relations. & Packing was a useful scheduler, not a near-constant-cost capability expansion method. Sharing the backbone execution did not remove the target-specific activation, optimization, or parameter cost. \\
\addlinespace
Shared-target multi-condition LoRA (4) & To compress routes with the same output, we alternated eight conditions through one Edge LoRA, split dense conditions into groups of about four, fused conditions into one route, and expanded toward the 38 conditional-control candidates. & Small pilots lowered loss, but four-condition groups interfered, fused conditions were weak, and only 11 candidate conditions had usable paired intersections in the larger experiment. Early generated outputs remained behind separately acquired Edge tunnels, while every condition modified the same weights. & A common target did not make its condition-to-target relations interchangeable. Large-scale capability growth required routes that could be acquired, installed, repaired, and removed without changing earlier capabilities. \\
\addlinespace
NJA/MTA TunnelingNet routes (4) & We compared native joint attention, modality tunneling attention, zero-initialized residual control, and multilayer residual control inside the shared FLUX generator. & The internal tunnel redirected the denoising trajectory with a small active parameter set and could be trained separately for each input--output task. & A small internal residual path could redirect a strong shared generator without retraining its target prior. This was the first mechanism that combined low marginal cost with independent task installation. \\
\addlinespace
Semantic heatmap and inverse codec (2) & Isolated pixels gave the VAE too little signal, so we expanded points into Gaussian heatmaps and inverted them with peak detection, NMS, and ID quantization. & Locations survived round-trip encoding more reliably, while semantic colors could still change and reconnect the wrong joints. & Enlarging points preserved their locations, but color errors still changed identity. A usable sparse boundary had to move identity out of fragile pixel values. \\
\addlinespace
Person--keypoint slot heatmaps (2) & Color was carrying too many meanings, so we assigned every person--keypoint pair a fixed slot and packed three heatmap lanes into each RGB image. & The slot index stabilized identity and left the VAE to preserve only a smooth blob and its position; the representation approached a usable round trip. & Fixed slots made identity structural and left the generator to predict only location and confidence. This produced the first nearly reversible boundary for sparse pose. \\
\addlinespace
Body-part and full-body heatmaps (3) & Collisions among people, hands, faces, and body points motivated body partitions, body-only maps, and a full-body codebook. & Partitioning reduced semantic collisions, although each extra partition increased the number of lanes and the caching cost. & Partitioning reduced collisions at the cost of more lanes. The experiment showed exactly when extra representation capacity was worthwhile: when it preserved a distinction needed for decoding. \\
\addlinespace
Direct sparse sequence heads (5) & Image decoding remained inefficient for compact outputs, so we predicted ordered slots for points, poses, boxes, skeleton parameters, and audio tokens. & Training became compact and task-native, while each structure required explicit ordering, masks, matching, decoding, and losses. & Typed sequences were efficient for compact outputs but required explicit ordering and matching. The unified task form could therefore support different codecs without forcing one tensor topology. \\
\addlinespace
ScoreHMR WSV structure route (2) & We tested full-frame and crop-aligned cached image features in a pretrained human-regression model while predicting WSV-derived SMPL~\cite{loper2023smpl} pose and shape parameters. & The image-feature route learned quickly and provided the first successful direct structured adapter in this sequence. & A pretrained structural prior learned body parameters quickly from image evidence. Target-specific decoders could play the same role for sequences that image generators played for dense fields. \\
\addlinespace
Atomic object-box slots (3) & We compared a structured center/size predictor with a Gaussian spatial prior against MTA and NJA routes that generated filled pseudocolor box masks. & The structured route became usable with spatial features, IoU matching, focal supervision, duplicate suppression, and NMS; both dense-mask routes were stopped after inferior geometry and instance colors. & The structured box route outperformed dense pseudocolor masks because it kept each object's coordinates together. Object membership had to be encoded as an atomic relation. \\
\addlinespace
Object BFS tree (2) & Fixed joint-ID slots assumed a shared skeleton that general objects do not possess, so we serialized parent slots, relative offsets, and absolute coordinates in breadth-first order. & The target represented variable object topology more faithfully, although the resulting skeleton quality still required improvement. & Fixed human-style slots could not represent arbitrary object topology. Parent order and traversal had to become explicit parts of the target code. \\
\addlinespace
Point-cloud skeleton initialization (4) & Pure tree prediction lacked a geometric anchor, so we derived k-means/MST-style initial skeletons from object point clouds and tested refinement, diffusion, and anchor corrections. & Initialization produced a promising geometric baseline; downstream refinement remained experimental. & Deterministic geometry supplied a useful skeleton before learning, whereas unconstrained refinement remained unstable. A strong analytical initialization reduced what the learned route had to discover. \\
\addlinespace
Sparse and mixed audio routes (6) & We tested AST source classification, RGB-plus-audio source boxes, EnCodec-token regression, image-prefix MusicGen, dense audio-semantic masks, and an RGB-plus-audio region selector. & Source classification, localization, and the mixed region selector became usable; direct visual-to-audio generation remained weak and optional. & Aligned audio could classify or select a visual region, while direct waveform/token generation remained weak. Cross-codec tasks were feasible when timing and semantic roles were explicit. \\
\addlinespace
Dense categorical-mask pilots (8) & Before introducing the refiner, we tested a gated fixed-class route, balanced flow, a self-contained prototype head, set-level and predicted-mask multi-label heads, pixelwise semantic prototype CE, and prototype CE with and without a multi-label classifier. & True generation overturned optimistic partial-denoising previews: balanced flow retained noisy geometry, multi-label variants collapsed to background when direct color constraints were removed, and classifier-free prototype CE fragmented masks across legal colors. & True generation exposed failures hidden by partially denoised previews. Spatial support and exact class identity were separate problems and had to be evaluated separately. \\
\addlinespace
Categorical objective modes (5) & We compared class-balanced CE, a combined categorical objective, global-margin supervision, prototype CE, and prototype CE with focal regression on a second target. & Four matched categorical-objective experiments completed 1,000 updates; global-margin and prototype CE gave the strongest visual candidates, while focal emphasis did not resolve discrete identity by itself. & Global-margin and prototype objectives improved candidates without making continuous flow exactly categorical. Better losses narrowed the gap but did not remove the need for symbolic recovery. \\
\addlinespace
Categorical region refiners (4) & Pseudo-RGB generation could draw the right support while crossing a class boundary, so we tested per-task refinement, multi-condition refinement, foreground recovery, and coarse-label remapping. & Discrete accuracy improved for several tasks, with difficulty remaining for fine-grained classes. & A second stage could preserve generated geometry and recover exact labels. This became the practical solution when one continuous generator could not reliably satisfy both requirements. \\
\addlinespace
Universal Pose custom TunnelingNet (5) & To add a category-independent articulated target, we compared full-frame rendering, letterboxing, instance crops, gradient supervision, and focal supervision on MP-100. & Full-frame training collapsed toward black; corrected geometry and instance crops helped, and the gradient-supervised RGB-to-pose pilot gave the strongest result. & Correct geometry, cropping, and gradient supervision mattered more than simply enlarging the model. This successful custom target also showed that new variables need not consume fixed WSG channel IDs. \\
\addlinespace
Pose-detail objective pilots (8) & Starting from a recognizable gradient-supervised pose route, we tested latent and decoded endpoint losses, topology, a low-noise detail anchor, soft distance, multi-scale gradient, gradient direction, and hard-foreground mining. & Endpoint losses produced black outputs from scratch and eroded fingers after warm-start; the other six continuations were also weaker than the original gradient-supervised baseline. & Endpoint-oriented losses repeatedly removed fingers and other sparse details. We kept the gradient-supervised baseline and learned that dense endpoint improvement could oppose articulated fidelity. \\
\addlinespace
Named custom-variable routes (2) & We added an RGB-to-transparency route from MAGICK data and an RGB-plus-environment-light to relighted-RGB route from IC-Light data. & The transparency route completed 1,000 updates; after 2,800 updates, the relighting model produced recognizable results under fixed test lights. & Transparency and relighting confirmed that named graph variables generalized beyond pose. The task graph could grow without redesigning the original numeric channel allocation. \\
\addlinespace
Runtime route composition (2) & We tested text-to-RGB followed by an RGB-to-world route and a generic intermediate-node chain assembled from installed tunnels. & The stages executed without joint retraining and preserved separate provenance for every edge. & Installed routes could be chained without joint retraining while preserving their provenance. Independently acquired capabilities therefore formed an executable system. \\
\end{longtable}

\subsubsection{Latent-Bridge Variants in Development Order}

For this table, a \emph{source latent} is the compressed tensor produced by
the frozen source encoder, a \emph{target latent} is the tensor expected by
the frozen target decoder, and a \emph{bridge} is the trainable mapping tested
between them.  Each row states exactly what changed relative to the preceding
design, why we made that change, what we observed, and what that experiment
allowed us to conclude.

\begin{longtable}{p{0.09\textwidth}p{0.18\textwidth}p{0.22\textwidth}p{0.20\textwidth}p{0.23\textwidth}}
\caption{\textbf{Variant-level chronology of the latent-bridge search.} This
table unpacks the bridge families in Table~\ref{tab:bridge_variant_record}.
Rows follow the approximate development sequence, including points where work moved between
families; the first column preserves each recurring research line without
rearranging that chronology.}
\label{tab:bridge_variant_level_record}\\
\toprule
\textbf{Family} & \textbf{Variant and key change} & \textbf{Why we tried it} & \textbf{What happened} & \textbf{What we learned} \\
\midrule
\endfirsthead
\multicolumn{5}{l}{\small\itshape Table~\ref{tab:bridge_variant_level_record} continued.}\\
\toprule
\textbf{Family} & \textbf{Variant and key change} & \textbf{Why we tried it} & \textbf{What happened} & \textbf{What we learned} \\
\midrule
\endhead
\midrule
\multicolumn{5}{r}{\small\itshape Continued on the next page.}\\
\endfoot
\bottomrule
\endlastfoot

Direct map & \textbf{Direct ``God Bridge''} Feed the source latent to one compact CNN and regress the paired target latent in a single pass. & The most optimistic hypothesis was that a compact endpoint mapper could solve the paired relation in one pass. & The bridge was trainable, but its loss initially showed little meaningful decline and its outputs lacked fidelity. & A small one-pass network could reduce latent error, but it could not reproduce the target images faithfully. We therefore stopped treating the entire cross-space conversion as ordinary pointwise regression. \\
\addlinespace
Direct map & \textbf{Attention-augmented CNN} Insert attention blocks into the compact CNN so every latent location can use distant locations. & Convolution might miss global relationships, so attention blocks were added while the bridge was kept lightweight. & Global interaction did not remove the endpoint mismatch. & Because attention did not solve the mismatch, the failure was not caused mainly by a limited receptive field. Global communication alone could not translate coordinates whose meanings differed across the two latent spaces. \\
\addlinespace
Measurement & \textbf{Cached paired latents} Encode every paired RGB/target sample once, save both latent tensors, and train the bridge only from this cache. & Image loading and repeated VAE encoding obscured the mapping experiment, so paired RGB and target latents were precomputed once and streamed during bridge training. & Training reached roughly thousands of updates per hour and became much easier to inspect, while decoded failures remained. & After removing repeated encoding from the loop, training became fast while the same decoded errors remained. This ruled out the data pipeline as the main bottleneck and isolated latent-space translation as the problem. \\
\addlinespace
Alignment & \textbf{Whitening} Measure each space's mean and scale, standardize both before the bridge, and undo target standardization before decoding. & Mean and scale mismatch might dominate optimization, so each latent population was standardized with measured statistics and inverted after prediction. & Numerical conditioning improved, while semantic errors remained after first- and second-order scale alignment. & Standardizing both latent populations made the optimizer behave better, but the same samples still decoded to the wrong content. Similar means and scales therefore did not imply corresponding latent meanings. \\
\addlinespace
Direct map & \textbf{Wider residual CNN} Enlarge each directional residual CNN to roughly 3M parameters while keeping the paired data and objective fixed. & Insufficient bridge capacity was a simple explanation for the plateau, so each directional CNN was enlarged to roughly three million parameters. & Training loss improved more than decoded quality. & The larger mapper fit the paired coordinates better without producing comparably better images. The plateau was therefore not simply caused by too few bridge parameters. \\
\addlinespace
Measurement & \textbf{FP16 versus FP32 cache} Store and decode the same cached latent pairs in FP16 and FP32, then compare the resulting images. & Blurred decoding might have come from quantization in the cached latent pairs, so the same pairs were checked and decoded under both cache precisions. & The precision change did not explain the large visual gap. & Both cache precisions produced the same large visual gap. Quantization during latent storage was therefore ruled out as the source of the bridge failure. \\
\addlinespace
Measurement & \textbf{Latent MSE} Predict the target latent directly and minimize mean squared error at every latent coordinate. & The first direct bridge treated paired coordinates as ordinary regression targets and minimized squared latent error. & The loss decreased and then plateaued while decoded details remained visibly wrong. & MSE could decrease even when the target decoder produced visibly incorrect details. A small coordinate error in latent space was therefore not a reliable measure of correct target content. \\
\addlinespace
Measurement & \textbf{Latent MAE} Keep the same direct mapper and paired latents, but replace MSE with mean absolute error. & High-dimensional MSE concentrated gradient on a subset of large residuals, so the bridge objective was switched to MAE. & Optimization became steadier and late residuals were less dominated by outliers, without closing the decoded gap. & MAE made optimization less sensitive to large residuals but did not improve the decoded endpoint enough. Better numerical conditioning alone did not solve the semantic mismatch between the spaces. \\
\addlinespace
Measurement & \textbf{Decoded pixel supervision} Decode the predicted latent with the frozen target VAE and add an image-space error against the target image. & To test whether a close latent remained correct after decoding, the predicted target latent was passed through its frozen decoder and compared in image space. & Pixel supervision supplied a stronger visual signal, yet fine structure improved slowly and the bridge still produced smooth averages. & Comparing decoded images exposed errors hidden by latent loss and gave a more useful training signal. Its slow, averaged outputs also showed that supervision alone could not supply the target-domain prior missing from a small bridge. \\
\addlinespace
Measurement & \textbf{Random latent masking} Hide random $4\!\times\!4$ source-latent regions during training and ask the mapper to predict the full target. & The direct mapper might have copied local coordinates without learning context, so random $4\!\times\!4$ latent regions were hidden during training. & The intervention did not produce a clear improvement in decoded quality. & Forcing the mapper to infer hidden neighborhoods did not improve decoded quality. The dominant error was therefore not simple local copying or insufficient spatial context. \\
\addlinespace
Direct map & \textbf{Image-conditioned bridge} Give the mapper both the source latent and features from the original RGB image. & The latent pair alone might omit information visible in the source image, so RGB image evidence was added as an explicit bridge condition. & Extra evidence helped individual reconstructions but did not make the two completed latent spaces mutually decodable. & Adding the source image helped some individual predictions, but the target decoder still interpreted many outputs incorrectly. More source evidence reduced ambiguity without defining the target latent language. \\
\addlinespace
RF path & \textbf{U-Net repair bridge} Replace the one-pass mapper with a multiscale U-Net that repeatedly updates its current target-latent estimate. & A multiscale U-Net was introduced so the bridge could repeatedly correct its latent estimate. & Repeated repair increased memory sharply, and one experiment later produced black outputs. & Repeated U-Net correction added capacity, memory use, and instability, yet it did not reliably recover the target. Iterative repair was therefore not a practical substitute for a correct target-space model. \\
\addlinespace
RF path & \textbf{Fully unrolled RF} Run every rectified-flow integration step during training and backpropagate the final endpoint loss through all steps. & All rectified-flow corrections were differentiated through so the final endpoint could supervise the complete path. & Activation storage grew with the number of integration steps and exceeded the intended lightweight budget. & Supervising the final endpoint through every integration step was possible, but memory grew with the full trajectory. This established global path supervision as useful but too expensive to serve as the default training procedure. \\
\addlinespace
RF path & \textbf{Random-time RF} Sample one time $t$ per update and train the velocity at that point, which avoids storing the full integration path. & One random time point on the reference path was supervised per update. & Memory fell substantially, while small local velocity errors accumulated during complete inference. & Training one randomly sampled time step greatly reduced memory, but small velocity errors accumulated into a poor final image. Good local flow predictions therefore did not guarantee a correct full trajectory. \\
\addlinespace
Direct map & \textbf{Independent directional CNNs} Use one CNN for RGB-to-channel and a different CNN for channel-to-RGB; do not share their weights. & One shared bidirectional rule assumed that reversal preserved difficulty, so RGB$\rightarrow$channel and channel$\rightarrow$RGB received separate networks. & The estimation direction improved faster; the reverse direction remained substantially harder. & RGB-to-structure learned much faster than structure-to-RGB under otherwise similar settings. Reversing condition and target therefore created a different task with different uncertainty and could not share one symmetric rule. \\
\addlinespace
Alignment & \textbf{Learned $1\!\times\!1$ alignment} Place a learned invertible $1\!\times\!1$ channel projection between the two latent spaces before translation. & If the two spaces differed mainly by a channel-wise coordinate transform, a learned invertible projection could align them before translation. & Population statistics moved closer, but individual predictions still decoded incorrectly. & The projection aligned population statistics, yet paired samples still decoded to the wrong targets. A channel-wise invertible transform was therefore insufficient to recover sample-level correspondence. \\
\addlinespace
Alignment & \textbf{Moment and covariance matching} Add losses that match the predicted and target latent means, variances, and covariance matrices. & To go beyond whitening, mean, variance, and covariance penalties were added between paired populations. & Distribution plots looked more similar while output details remained wrong. & The two latent clouds looked more alike after matching their statistics, while output details remained incorrect. Distribution-level similarity therefore did not guarantee that corresponding samples carried the same meaning. \\
\addlinespace
Alignment & \textbf{Joint-embedding alignment} Encode the paired RGB and target observations into one shared embedding, then decode the required modality from it. & Paired RGB and target observations were pulled into a common representation before decoding. & Shared structure emerged, but exact modality-specific information required separate capacity and joint retraining. & The shared embedding captured common structure but discarded details needed only by one modality. A common space could organize modalities, yet faithful generation still required target-specific information and capacity. \\
\addlinespace
Probabilistic & \textbf{Encoder KL regularization} Add a KL penalty that pushes the bridge bottleneck toward a smooth reference distribution. & The bridge bottleneck was regularized toward a smoother common distribution. & The latent became easier to sample, while modality-specific detail weakened when regularization dominated. & KL regularization made the bottleneck smoother and easier to sample, but strong regularization erased modality-specific details. Sampling convenience therefore had to be balanced against the information required for exact conversion. \\
\addlinespace
RF path & \textbf{Noise ranking} Assign a persistent rank/identity to each sampled noise start and condition the path on that identity. & Random starts were assigned stable ranks so different paths for one condition could remain distinguishable. & The bridge produced more differentiated trajectories and stepwise training became faster. & Giving each random start a stable identity reduced averaging among different possible paths. This showed that part of the collapse came from indistinguishable trajectories, although path identity alone did not solve target generation. \\
\addlinespace
RF path & \textbf{One-step endpoint loss} Take one numerical integration step during training and penalize its predicted endpoint against the target latent. & A cheap endpoint term was added to correct trajectory drift, initially using a single integration step. & The endpoint number could become small while full-step inference and decoded output remained poor. & A low one-step endpoint error could coexist with poor eight-step inference and poor decoded images. Endpoint loss was therefore trustworthy only when it used the same trajectory and representation as actual inference. \\
\addlinespace
RF path & \textbf{Matched multi-step endpoint loss} Replace the one-step approximation with the same eight-step integrator used at inference. & The endpoint computation was changed to use the same eight-step integration schedule as inference. & Training and visualization became more comparable, at greater computational cost. & Using the inference-time integration schedule made the training endpoint and evaluated endpoint refer to the same object. The additional cost was necessary to interpret endpoint improvements correctly. \\
\addlinespace
RF path & \textbf{Checkpointed endpoint integration} Keep eight-step endpoint training, but discard intermediate activations and recompute them during backpropagation. & Multi-step endpoint supervision was retained while activations were recomputed during backward propagation. & Peak memory grew much less, although training slowed. & Recomputing intermediate states kept multi-step endpoint supervision within memory, although each update became slower. This provided a workable memory--time trade-off without changing the underlying translation problem. \\
\addlinespace
Measurement & \textbf{Full-step decoded metrics} Run the complete inference integrator, decode its final latent with the target VAE, and score that decoded output separately. & Complete inference was evaluated separately from the differentiable endpoint term and then decoded through the target VAE. & Low local or aligned-space error could coexist with much larger full-trajectory and visual error. & Several models looked good under local training loss but failed after full integration and decoding. We therefore retained complete decoded inference as the decisive bridge evaluation. \\
\addlinespace
RF path & \textbf{Minibatch OT pairing} Within each minibatch, use optimal transport to reassign sampled noise starts to target latents before flow training. & Intersecting noise-to-target segments could assign conflicting velocities to nearby states, so noise samples were reassigned to targets by minibatch optimal transport. & Reference segments became shorter and less crossed, while semantic incompatibility between endpoint spaces remained. & OT shortened and untangled the reference segments, but the decoded targets remained wrong. It improved transport geometry while leaving the missing target-domain semantics unchanged. \\
\addlinespace
RF path & \textbf{Noise conditioning} Expose the sampled starting noise to the velocity network at every prediction. & The sampled start itself was exposed to the velocity network as a condition. & Path diversity improved, especially toward RGB, while difficult reverse mappings remained. & Exposing the sampled start helped the network distinguish multiple trajectories and increased diversity. Reverse mappings still failed, showing that trajectory identity could not replace knowledge of how to generate the target modality. \\
\addlinespace
RF path & \textbf{Reflow} Use a trained bridge to generate start--endpoint pairs, then train a second bridge on those generated paths. & A trained bridge was used to generate teacher endpoints, and a second model learned straighter paths between the observed starts and endpoints. & Useful teacher paths could be straightened; weak or unstable teacher outputs were inherited. & Reflow successfully straightened paths only when the first bridge already supplied useful endpoints. Errors in the teacher endpoint persisted in the second model. \\
\addlinespace
RF path & \textbf{Intermediate-state supervision} Integrate to selected interior times and supervise both the intermediate states and final endpoint. & Errors could accumulate between randomly supervised time points, so intermediate integrated states were explicitly constrained. & Path behavior became more regular, without an automatic gain in final decoded reconstruction. & The additional constraints made intermediate states smoother and more regular, but the final decoded target did not improve automatically. Controlling the path interior therefore did not supply the missing endpoint semantics. \\
\addlinespace
RF path & \textbf{Repeated correction blocks} Apply the same lightweight correction block several times, feeding each predicted state into the next update. & The same lightweight correction rule was applied several times to let later steps repair earlier errors. & Error correction improved locally but multiplied integration cost and could amplify bias across steps. & Repeated updates repaired some local errors but also repeated systematic bias and increased inference cost. Iteration helped only when the correction rule had been trained on the off-path states encountered during inference. \\
\addlinespace
VQ bridge & \textbf{VQ bridge with CE} Quantize target latents into codebook IDs and train the bridge as a per-location classifier with cross-entropy. & Continuous predictions could leave the valid target manifold, so target latents were quantized and the bridge predicted code indices. & Predictions remained on codebook entries, but target usage was extremely imbalanced and many codes were inactive. & Quantization kept predictions on valid codebook entries, but most target codes were rarely or never used. Discreteness prevented invalid coordinates without making the codes balanced or semantically predictable from the source. \\
\addlinespace
VQ bridge & \textbf{Frozen VQ codebook} Pretrain the VQ codebook, freeze its entries, and train only the bridge that predicts their IDs. & The codebook was frozen to make CE labels stable throughout bridge training. & Labels stopped moving, but dead entries and poor target coverage remained. & Freezing kept class labels stable, but it also preserved dead codes and poor coverage of target patterns. Stable labels alone could not make an unsuitable codebook represent the target distribution. \\
\addlinespace
VQ bridge & \textbf{Updated VQ codebook} Allow the VQ entries to update jointly while the bridge learns to predict their current IDs. & The codebook was allowed to adapt so more target patterns could receive codes. & Utilization improved while the class meanings moved during classifier training. & Updating activated more codes, but their meanings moved while the bridge classifier was learning them. Better coverage therefore introduced a moving-target problem for cross-modal prediction. \\
\addlinespace
VQ bridge & \textbf{Smaller VQ dictionaries} Reduce the original 2,048-entry dictionary and repeat the same VQ bridge training. & The 2,048-entry codebook was reduced to test whether excessive capacity caused dead codes. & Utilization rose in some experiments, while decoded translation remained limited. & Reducing the dictionary increased code usage but did not recover accurate decoded translation. Dead-code reduction was therefore a codebook diagnostic, not evidence that the bridge had learned the target mapping. \\
\addlinespace
VQ bridge & \textbf{Index-MSE MLP} Output one real number per location and regress the numerical VQ index with MSE before rounding to an ID. & Code IDs were regressed as numbers under the hypothesis that nearby indices represented nearby content. & The attempt failed because code indices had no ordinal geometry. & Regression failed because code IDs were unordered labels. A VQ bridge had to predict them categorically or learn a meaningful code metric. \\
\addlinespace
Direct map & \textbf{Transformer bridge} Flatten latent locations into a sequence and replace the convolutional mapper with a Transformer. & A sequence model replaced local convolution to test whether long-range interaction was the missing ingredient. & The Transformer also retained the decoded distribution gap. & The Transformer added global context and more flexible interactions, yet the same decoded gap remained. The failure was therefore not specific to convolution or short-range processing. \\
\addlinespace
RAC & \textbf{One-step RAC} Map the input directly to a TAESD-compatible latent with one learned rectified-auto-coder update. & A rectified auto-coder tested whether a direct learned path could reproduce a known TAESD latent contract. & It provided a simple baseline but little opportunity to correct perturbations along the path. & One-step reconstruction measured whether the endpoint could be reached directly, but it revealed nothing about recovery from errors along a trajectory. We therefore separated endpoint fidelity from path robustness. \\
\addlinespace
RAC & \textbf{Four-step RAC} Replace the single update with four sequential learned updates and inspect every intermediate state. & The same auto-coder was expanded to four learned stages. & Path geometry changed and intermediate behavior became visible, without an automatic large reconstruction gain. & Four stages exposed and reshaped the intermediate path, but additional stages did not automatically improve the final reconstruction. Step count was a control variable, not a substitute for a compatible latent contract. \\
\addlinespace
RAC & \textbf{Fixed versus random time grids} Train otherwise matched RAC models using either fixed integration times or newly sampled times per update. & RAC compared a prescribed integration grid with randomly sampled training times. & The two schedules emphasized different parts of the trajectory and neither alone solved compatibility. & The schedules changed which trajectory regions were learned best, while both retained compatibility errors. Time sampling affected optimization coverage but could not repair a mismatched latent interface. \\
\addlinespace
RAC & \textbf{State-noise correction} Add noise to intermediate RAC states during training and require later updates to recover the clean trajectory. & Noise was added to intermediate RAC states so later updates learned to recover from off-path inputs. & Perturbation recovery improved while clean reconstruction changed less. & Training on perturbed intermediate states improved recovery when inference drifted away from the ideal path. Clean reconstruction changed little, confirming that robustness to accumulated errors had to be trained separately. \\
\addlinespace
Shadow variable & \textbf{Direct RGBD shadow code} Train forward RGB-latent-to-RGBD decoding together with reverse RGBD encoding, then test both the learned code and the unchanged native RGB latent. & We expected reverse encoding to place RGBD around the mature RGB manifold while preserving the original RGB coordinates. & After roughly 57,000 updates, the learned RGBD code reconstructed almost perfectly, while depth completion from unchanged RGB became worse than early in training. & Encoder and decoder had formed a private coordinate system. Their agreement did not preserve the meaning of the native RGB latent. \\
\addlinespace
Shadow variable & \textbf{Visualized-depth shadow code} Convert depth into a three-channel color image and encode it with the official image codec before learning the RGB--depth interaction. & Matching the channel type and codec was intended to remove the possibility that raw scalar depth was simply incompatible with an RGB encoder. & The representation became easier for the image codec to process, yet the cross-modal semantic gap remained. & Making a modality look like RGB solves a representation-format problem, but it does not make its latent coordinates mean the same thing as native RGB coordinates. \\
\addlinespace
Shadow variable & \textbf{Strict bidirectional alignment} Replace the loose expand--pool projection with a subpixel-seeded mapping that gives forward decoding and reverse encoding matched spatial positions. & We tested whether the shadow code drifted because the two directions were allowed to use different spatial correspondences. & Position matching improved bookkeeping and exposed intermediate alignment error, but private-code reconstruction still did not imply native-code completion. & Spatial correspondence is necessary for a reversible interface, yet coordinate alignment alone cannot impose semantic identity between modalities. \\
\addlinespace
Shadow variable & \textbf{Native-RGB additive perturbation} Parameterize the new code as $z_{\mathrm{RGB}}+\alpha\Delta z$, freeze the native RGB anchor, and restrict the learned contribution to a scaled perturbation. & This was the strongest test of the shadow hypothesis: extra world information had to remain close to a valid pretrained RGB code. & The route kept the new code close in magnitude while exposing conflicting objectives between RGB preservation and RGBD recovery. & Euclidean proximity can be enforced explicitly, but semantic compatibility cannot. A perturbation becomes useful only when the downstream generator has learned how to interpret the information carried by that perturbation. \\
\addlinespace
RAC & \textbf{Native-latent compatibility} Remove the added latent normalization and use the original TAESD scaling expected by PerFlow. & An extra latent normalization was removed after it broke decoding by an existing PerFlow generator. & Restoring the original TAESD convention restored the interface. & The normalized codec could reconstruct its own samples while failing when connected to the existing PerFlow generator. Restoring native TAESD conventions showed that a useful latent contract must remain compatible with the pretrained model that consumes it. \\
\addlinespace
Probabilistic & \textbf{Pure conditional velocity} Remove auxiliary alignment and endpoint objectives, leaving only conditional velocity prediction. & Auxiliary objectives were removed so the bridge learned only the conditional vector field. & RGB-directed prediction improved, while information-poor reverse directions stayed difficult. & Removing auxiliary losses clarified the flow and improved prediction toward RGB, but reverse directions remained difficult. The remaining gap followed the information and generative knowledge required by the target, not objective complexity. \\
\addlinespace
Probabilistic & \textbf{Classifier-free guidance} Randomly drop the condition during training, then combine conditional and unconditional predictions at inference. & Condition dropout and guided inference were added to strengthen use of the source modality. & Conditional influence became easier to control; it did not repair the target decoder mismatch. & Guidance strengthened or weakened how much the model used an available condition. It could not create appearance or target-domain knowledge that the bridge had never learned. \\
\addlinespace
Probabilistic & \textbf{CVAE prior--posterior} Train a target-aware posterior and a condition-only prior, and align them with a KL loss. & A target-aware posterior and condition-only prior were trained with KL alignment to represent one-to-many outputs. & The random variable represented uncertainty, but prior matching and realistic decoding remained difficult. & The latent variable represented multiple possible outputs, but the condition-only prior still could not decode them realistically. Modeling uncertainty helped only after the model had learned a capable target generator. \\
\addlinespace
Probabilistic & \textbf{Best-of-$k$ sampling} Generate $k$ candidate starts or paths and backpropagate through the candidate whose endpoint is closest to the reference. & Several candidate starts were generated and the endpoint closest to the reference was selected for optimization. & Alternative valid paths were penalized less, but the generated candidates did not become intrinsically better. & Choosing the closest of several samples stopped penalizing every alternative as wrong, but it did not improve the samples being proposed. Selection handled evaluation ambiguity without strengthening the generator. \\
\addlinespace
Probabilistic & \textbf{Target-noised training} Add noise directly to target latents and train the model to recover them while conditioning on the source. & Target latents were corrupted and conditionally denoised so training directly visited the target distribution. & Diversity improved, while endpoint detail and stability remained insufficient. & Denoising corrupted target latents exposed the model directly to the target distribution and improved diversity. Fine detail and stability still lagged, so exposure alone did not replace a mature target generative prior. \\
\addlinespace
Probabilistic & \textbf{Style-latent conditioning} Add a compact latent carrying color, texture, and illumination alongside the geometric condition. & Geometry leaves appearance unspecified, so a compact latent was added for color, texture, illumination, and other residual choices. & The missing variables became explicit, but their prior and decoder formed another difficult bridge. & The style latent made the missing appearance variables explicit, but those variables still needed a learned prior and decoder. Merely naming color, texture, and illumination did not make them inferable from geometry. \\
\addlinespace
Probabilistic & \textbf{Reverse-noise encoding} Train an inverse model from target latent to structured noise, then use that noise to initialize forward conditional generation. & Target latents were mapped backward toward structured noise to provide a better starting state for forward prediction. & The inverse route was measurable but did not yield a stable conditional generator. & The inverse model could move a known target latent toward structured noise, but the forward model still could not regenerate a stable target from that noise and the condition. Thus, learning an inverse route did not provide the target knowledge required for conditional generation. \\
\addlinespace
Probabilistic & \textbf{Conditional latent diffusion} Replace rectified flow's velocity prediction along a continuous interpolation path with conditional DDPM/DDIM denoising across multiple noise levels. & Flow matching and diffusion define different learning problems. We tested whether iterative stochastic denoising could model the target distribution that the learned flow path could not reach. & The diffusion objective trained under these different dynamics, but its decoded targets were still not practically useful. & The same failure survived a change from velocity-field learning to stochastic denoising. This ruled out a flow-specific path or objective as the main cause and showed that the small bridge still lacked the target-domain knowledge needed to generate correct outputs. \\
\end{longtable}
\endgroup

\clearpage
\twocolumn

\subsection{How the Experiments Led to the Unified Law}
\label{pre:subsec:memoir_experimental_logic}

The law did not begin as a slogan. It was the common answer left by a long
sequence of failed and successful experiments. A single codec could reconstruct
some world fields, but every newly added field changed the shared latent space.
Small bridges could align statistics, shorten paths, or reconstruct their own
private codes, yet they could not make independently trained modal spaces mean
the same thing. Sparse points disappeared until we gave every point a visible
address. RGB-to-depth learned more easily than depth-to-RGB because the second
direction must invent appearance that depth never contains. Finally, conversion
became reliable when the requested target was produced through a generator that
already knew how that target should look. These observations gave us one simple
description: known RGB, text, and world information are conditions; the
requested information is the generated target; reversing the direction defines
a different task.

WSM v1 was built after this experimental history and tests the resulting rule at
scale. Its 1,148 evaluated tasks span 39 output groups and all four interfaces.
Task-native metrics, qualitative matrices, external comparisons, and executable
chains test many different condition--target directions under the same
definition. The earlier experiments explain where the rule came from; the WSM study
tests how broadly it works.

\subsection{The Deeper Insight: a Model Must First Preserve What It Is Asked to Learn}
\label{pre:subsec:memoir_preservation}

The experiments eventually changed our question. We first asked which
architecture could learn every channel. We later found that an earlier question
decides whether learning is possible at all: \emph{does the representation keep
the difference that the task asks the model to predict?} If a keypoint encoder
turns two different joints into the same nearly empty image, or a semantic-color
encoding gives two objects an ambiguous color, the later network no longer has
the information needed to tell them apart. Formally, if an encoder maps two
different world states $x_a$ and $x_b$ to the same code,
\begin{equation}
E(x_a)=E(x_b),
\end{equation}
then every later network that sees only that code must treat the two states in
the same way, even when their correct answers differ. More layers, more data, a
stronger loss, or a larger attention module cannot recover information that was
already erased. Our sparse-pixel, semantic-color, shared-VAE, and latent-
alignment experiments reached this same limit through different routes.

The useful side of this result is simple. Once every important distinction has a
stable address and can be encoded and decoded accurately, it becomes learnable.
A model does not need to sense that variable in the same way as a person. A
red--green color-blind observer cannot recover color after a sensor has already
collapsed red and green into one value. If an auxiliary code records which
wavelength class was observed, however, the observer can compare and communicate
the two classes without acquiring ordinary color vision. Our heatmap experiments
did the same thing for sparse structures. A VAE often erased tiny points on a
large empty canvas. Giving each point identity its own heatmap lane turned that
identity into a stable address, so the model only had to generate its position
and confidence.

This also changed how we understood modality. RGB, depth, pose, material,
object structure, audio, and language can describe the same event, but their
usual encodings place that information in very different coordinate systems.
The difference may lie in value range, spatial layout, density, abstraction, or
the meaning assigned to each coordinate. From the viewpoint of a world
snapshot, these are different fields of one state. They do not need identical
latent coordinates. They need faithful codecs, stable field identities, and a
learned relation that says how a known field constrains a requested field. In
short:
\begin{quote}
\emph{Modality is often distance in representation, not distance in the world.}
\end{quote}

This idea can be written compactly. Let a world snapshot contain typed fields
$W=\{X_1,\ldots,X_m\}$, and let every field have an encoder and decoder
$(E_i,D_i)$. A task chooses the fields already known, $C\subseteq W$, and the
fields to be produced, $Y\subseteq W$, then learns
\begin{equation}
p\!\left(E_Y(Y)\mid E_C(C),\,s\right),
\end{equation}
where $s$ may contain text or other context. In ordinary language, the model
generates the requested fields from the known fields. Generation, estimation,
control, reconstruction, classification, and structured prediction differ in
which fields are known, which are requested, and which direction connects them.
Adding a field expands what the model can read and write. Connecting existing
fields creates new tasks without requiring a new learning principle.

Large language models provide a useful analogy. A question, a document, a class
label, a program, and an answer can all be written as tokens. Once they share
that writable form, many language tasks become ``given these tokens, generate
the next tokens.'' Computer vision has strong generators, but depth, skeletons,
materials, object structures, and sound have usually been treated as outputs of
different systems. WSG makes these answers writable as named world fields. The
Unified Law supplies the prompt--answer grammar: known world fields form the
prompt, and requested world fields form the generated answer.

Vision adds a difficulty that text often hides. Two visual fields may be aligned
pixel by pixel and still use incompatible latent coordinates. A codec designed
for natural RGB images may also erase sparse points or distort physical values.
Visual unification therefore needs two ingredients. First, every field must
cross the input--output boundary without losing the information required by the
task. Second, the generator must know how to produce a valid sample in the
requested target domain. The codec and sparse experiments revealed the first
requirement; the failed external bridges revealed the second.

The resulting scaling strategy has two practical steps. First, enlarge the set
of world information that the system can receive and emit faithfully. Second,
learn lightweight directed paths among those fields while reusing a shared
generative prior. The first step makes the information available; the second
makes it usable. This suggests a general rule for foundation-model development:
\begin{quote}
\emph{Before scaling what a model knows, scale what the model can faithfully
express.}
\end{quote}

The same reasoning can extend beyond vision. If a foundation model can receive
and emit a new variable accurately, the new variable can become a condition, a
target, or both. A tokenizer adds something the model can read or write; a
control branch learns how one variable guides another; a router selects the
relation required by the current request. Together they give a direct sequence:
preserve the information, learn the relation, and then scale the capability.

\subsection{What the Experiments Taught Us about Modalities}
\label{pre:subsec:memoir_modalities}

\noindent\textbf{Visual information is relatively low in abstraction, while
textual information is highly abstract; the useful learning form changes with
this progression.}
During development, we gradually realized that moving from visual information
to textual information is a process from low abstraction to high abstraction.
An image retains concrete location, color, geometry, and texture; text discards
most of those sample-specific values and keeps a compact semantic description.
The learning objective changes in the same direction, from MSE-like
reconstruction of continuous values toward CE-like selection of semantic
symbols. A compact, low-channel SD-style VAE latent~\cite{rombach2021highresolution} has already removed much of
the spatial detail, so conversion involving it behaves more like prediction. A
higher-channel FLUX latent or representation-autoencoder feature retains more
visual structure and is consequently better suited to conditional image
generation, masking, and spatial translation.

This view places visual fields and text on one representational continuum.  If
the target exposes more low-level variables, part of the work normally performed
by the VAE moves into the generation process; a strong generator can often carry
that work.  If the same information is converted into reliable semantic tokens,
the task moves toward categorical prediction, and greater abstraction can help
when sufficient data are available. In this sense, a VAE is the visual
counterpart of a tokenizer: it decides which distinctions become the model's
symbols. It also divides the work between codec and generator. A compact codec
performs more abstraction before learning begins; a less compressed target
leaves more low-level structure for the generator to model. Neither choice is
universally better. The suitable choice depends on whether the task requires
exact continuous reconstruction, spatial translation, or semantic prediction.

Our experiments also found no automatic connection between representations at
different abstraction levels.  Equal tensor shapes, similar distributions, or
paired observations did not give their coordinates the same meaning.  Whitening,
interpolation, invertible alignment, VQ codes, joint embeddings, flow matching,
and diffusion all preserved this mismatch in different forms.  We call this
the \emph{principle of modal incompatibility}: independently learned modal
spaces do not contain a ready-made path between corresponding samples. They do
not become mutually readable until a sample-level relation and a usable target
decoder connect them. This explains why two encoders can emit arrays of the same
shape and scale while a small bridge still cannot translate between them.

\noindent\textbf{A condition constrains only the information it contains, so
reversing the direction of a task can change its difficulty completely.}
Text can precisely narrow the region of a target manifold that should be
generated.  This helps denoising because the pretrained generator supplies
appearance while text selects the intended semantic region.  A dense condition
such as depth, pose, edges, or material carries strong spatial information, but
it does not necessarily specify the appearance variables missing from that
condition. A weak semantic condition can be easy to use because it selects
within knowledge already stored by the generator. A strong modal condition
carries many exact values, yet those values help only when the architecture
preserves their spatial identity and gives them a direct path into generation.
Strong information is therefore not automatically strong supervision.

The direction of conversion therefore changes the problem.  RGB usually
constrains depth strongly, so RGB-to-depth is relatively close to prediction.
Depth-to-RGB remains one-to-many because depth alone does not determine texture,
color, lighting, identity, or background.  Without text, noise, style, or other
information that selects among these possibilities, velocity-field supervision
receives many incompatible valid RGB targets and, in our bridge experiments,
often averaged them or failed to reach a useful RGB sample. A
condition can guide only the information it contains; direction and missing
information must therefore remain part of the task definition.

\noindent\textbf{Dense and sparse fields need different boundaries because they
lose different information when encoded.}
Our experiments gave a simple working picture.  Dense visual data can usually
condition another dense field, and a dense observation can be used to estimate
a sparse structure.  The reverse direction is harder: a sparse structure does
not contain all the values needed to produce a dense appearance.  Converting
the sparse structure into heatmaps or fixed visual lanes makes it easier for an
image generator to read, but it does not create texture, illumination, identity,
or other information that was never present.

The same distinction explains why one universal image codec performed poorly.
Dense fields can be compressed by a VAE, discretized by a VQ codec, or sometimes
used directly in pixel space.  A sparse field placed on a mostly empty image is
dominated by its background, and a tiny reconstruction error can erase a point
or change its identity.  We therefore tested four practical boundaries:
visualize sparse values as heatmaps, predict an ordered sequence, predict the
original parameters directly, or use a one-shot structured decoder.  Fixed
heatmap slots worked because they turned identity into an address and left the
generator to predict location and confidence.  This did not create an exception
to the generative formulation; it showed that different information structures
may need different codecs before they enter the same formulation.

Density is not the only distinction. Two fields may cover only partly
overlapping facts about the scene. Removing empty regions or constructing a
path through the informative regions can reduce wasted computation, but it
cannot create values that the condition never observes. A useful practical test
is therefore: can the field be converted into an RGB-like representation and
decoded back with the distinctions required by the task still intact? If yes,
it can often reuse an image generator directly. If no, a sequence, parameter,
token, waveform, or other target-native boundary is needed. Our heatmap, object,
body-parameter, categorical, and audio experiments are instances of this rule.

\noindent\textbf{The four WSM interfaces are the four directional roles of RGB,
text, and additional world information, not arbitrary task labels.}
Vision, language, and audio are treated as large domains because their usual
spaces, abstraction levels, codecs, and pretrained priors are different.  Their
primary tasks can nevertheless be written in a generative form and can
eventually participate in one broader model once each domain can be encoded and
decoded faithfully.

The basic capability of an image generator is narrow RGB generation.
Generalized visual generation allows the target to be depth, pose, material,
motion, topology, or other world information.  It does not require one dense
joint generator to emit every variable simultaneously: visual chains can
compose separately acquired relations and create a requested combination only
when it is needed. This mattered in our experiments: joint generation coupled
memory and optimization across all outputs, while a chain could reuse
independently learned relations. A genuinely different domain such as audio
first needs its own faithful codec and target prior, after which it can join the
same conditional-generation framework.

Audio illustrates this point. A spectrogram can be handled as a dense visual
field, learned audio tokens can be handled as a sequence, and a waveform can be
generated by a waveform-native model. These forms carry different inductive
biases and should not be mixed into image channels simply because they can be
stored in a tensor. Once timing, codec, and target semantics are explicit,
however, audio can participate in the same condition--target grammar.

The familiar visual task names then become simple directional cases.  Producing
additional world information from RGB is estimation.  Producing RGB from text
or additional world information is generation or conditional control.
Reconstruction is a special case of estimation; rerendering is a special case
of conditional control in which RGB and another condition are both provided.
Spatially aligned world fields can be translated into one another, and using
RGB together with property $A$ to recover property $B$ is inverse rendering.
These relations directly produce the four WSM interfaces.

\noindent\textbf{Accurate encoding and decoding establish representational
access; cross-field use establishes mastery.}
\label{pre:par:memoir_accessibility_mastery}
A codec can read an image and reproduce the same pixels without learning what
those pixels imply for depth, motion, material, sound, language, or any other
field. This is analogous to memorizing and repeating a statement without
understanding its content. An accurate private round trip measures whether a
representation can be stored and recovered. Mastery is demonstrated when the
backbone uses that representation to infer, generate, or control another one.

What we need from a foundation-model backbone is stronger. A mastered field
should be usable as a condition, a target, or an intermediate representation:
one field should be able to produce several requested fields, several known
fields should be able to determine one requested field, and different fields
should participate in composed or arbitrary directed conversions. In this
operational sense, mastery is demonstrated by relations, not by identity
reconstruction alone. A modality becomes part of the model's usable knowledge
only when the model can do something with it beyond returning what it was
given.

This distinction also clarifies what a foundation model should seek. It should
first acquire basic representations through which vision, language, audio, and
other descriptions of the world can be read and written faithfully. It should
then learn relations among those representations so that they can serve as
inputs and outputs of one-to-one, one-to-many, many-to-one, and composed tasks.
Encoding many modalities into tensors is therefore only the representational
boundary of a general model; making them mutually usable is the capability.

Training one model that jointly acquires all such relations at once is often
computationally impractical. The Unified Law instead describes incremental
expansion from a backbone that already masters part of this representational
space: preserve a new field, connect it to existing fields, and install the new
directed capability without relearning the whole model. The formal distinction
between representational accessibility and operational mastery is summarized
in Sec.~\ref{pre:sec:formal_unified_law}.

\noindent\textbf{Which experiments failed, and why.}
Read against the requirements above, the failures in
Table~\ref{tab:bridge_variant_record} fall into three groups, and each
group was repaired only when its missing condition was restored, never by more
parameters or longer training.

\emph{The representation erased a distinction.} A shared VAE erased tiny points
on a large empty canvas. Continuous output heads averaged neighboring IDs and
confidences, and sparse categorical codecs kept identities better yet still
removed points or changed IDs after small decoding errors. A patch/MLP codec
removed the empty canvas and with it the spatial organization that a
convolution had obtained for free. Semantic-color heatmaps preserved locations
while color errors reconnected the wrong joints, and dense pseudocolor object
masks separated what a box slot kept together. The same failure occurred on the
condition side: deep feature injection into a frozen video generator delivered
evidence in a form its blocks could not read, so motion changed while
appearance degraded. Heatmaps with fixed person--keypoint slots, typed
sequences, object atoms, and named custom fields repaired this group by giving
each required distinction a stable address.

\emph{Each field lived in its own space.} Independent channel VAEs
reconstructed individual fields while giving every modality its own latent
space. Spaces trained separately in this way are heterogeneous, because their
structures differ, and heteronymous, because even two points at corresponding
positions of similar-looking structures carry unrelated meanings. Whitening,
learnable invertible alignment, and joint-embedding objectives matched shapes
and global statistics while individual samples still decoded incorrectly. The
shadow-variable route made the failure especially clear: a private RGBD code
achieved a nearly perfect round trip yet lost the ability to complete depth
from the native RGB latent, and a small normalization change made RAC latents
unreadable by an existing PerFlow generator although their own round trip
looked correct. External latent bridges---the direct ``God Bridge,'' attention
and Transformer bridges, VQ bridges, and conditional latent diffusion---had no
shared reference from which to start and could not close the decoded gap.
Improving the path did not help either: optimal transport, reflow,
intermediate supervision, and noise conditioning made trajectories shorter,
straighter, or more diverse, but did not supply the target semantics missing
from the bridge. This group was repaired only when all fields were written into
one space that the generator already reads and writes: pseudo-RGB serialization
through the RGB codec, which may lose some precision but keeps every field in a
common space. An encoder that writes every modality into one space and decodes
it accurately would serve the same role; the shared representation itself
becomes the bridge, even when the codec does not understand what it encodes.
Conversion then moved inside a pretrained generator whose prior covered that
space, first in video backbones and then in the FLUX-based TunnelingNet routes.

\emph{Capabilities shared parameters.} Widening a mature RGB codec let every new
field alter the shared latent. Joint video lanes interfered once groups reached
about four targets, fused conditions were barely trainable, and packed LoRAs
raised aggregate throughput by only about two times for eight relations because
target-side computation and gradients still grew with each relation.
Shared-target LoRAs let every new condition update the same parameters, so
adding or repairing one capability could change all previous ones. Separately
acquired tunnels repaired this group by storing each condition--target relation
in its own small path inside the shared generator.

Across all three groups, apparent success was judged only by complete
inference. Low latent losses, partial denoising previews, private round trips,
and aligned global statistics repeatedly looked successful while full
generation produced black images, fragmented categories, missing structures,
or codes unreadable by the native decoder. Section~\ref{pre:sec:formal_unified_law}
states these groups as the preservation, coverage, and installation conditions
of the Unified Law and lists the predictions they imply.

\subsection{Formal Lessons from the Experiments}
\label{pre:subsec:memoir_parameterized_knowledge}

The experiments above produced a set of measurable lessons. The following
definitions state those lessons in a form that can be tested on new models,
representations, and tasks.

\noindent\textbf{$\mathsf{K}_1$. Information preservation is decided at the model boundary.}
Let $T$ map known fields $C$ to requested fields $Y$, and let
$s=(s_C,s_Y)$ be the representation presented to the model. Write
$y_a\sim_T y_b$ when two targets are equivalent under the task's native answer
and evaluation. A task-sufficient representation must satisfy
\begin{equation}
\begin{aligned}
s_C(c_a)=s_C(c_b)&\Rightarrow
p_T(Y\mid c_a)=p_T(Y\mid c_b),\\
s_Y(y_a)=s_Y(y_b)&\Rightarrow y_a\sim_T y_b.
\end{aligned}
\label{pre:eq:memoir_preservation}
\end{equation}
The first line says that collapsing two conditions is harmless only when they
have the same correct answer distribution. The second says that collapsing two
targets is harmless only when the task itself does not need to distinguish
them. Once either implication fails, every downstream model receives the same
code for two cases that require different behavior. More parameters, data, or
training cannot reconstruct the erased distinction; the remedy is a better
boundary. Fixed heatmap lanes, typed sequences, object atoms, and named fields
worked because they restored such addresses.

\noindent\textbf{$\mathsf{K}_2$. Private reconstruction does not establish mutual
readability.}
For field $X_i$ with encoder--decoder pair $(E_i,D_i)$, define its private
round-trip error and its directed cross-field error by
\begin{equation}
\begin{aligned}
r_i &= \mathbb{E}\,\ell_i\!\left(D_i(E_i(X_i)),X_i\right),\\
e_{i\rightarrow j}(h) &=
\mathbb{E}\,\ell_j\!\left(D_j(h(E_i(X_i))),X_j\right).
\end{aligned}
\label{pre:eq:memoir_modal_compatibility}
\end{equation}
Low $r_i$ and $r_j$, equal tensor shapes, and matched latent statistics do not
imply low $e_{i\rightarrow j}$ for a small bridge $h$. Two private codecs may
reconstruct perfectly while assigning unrelated meanings to corresponding
coordinates. Whitening, invertible alignment, joint embeddings, and the
shadow-variable route repeatedly produced this pattern. Mutual readability
requires sample-level correspondence, stable field identities, and a target
decoder or generator that already gives those coordinates a usable meaning.

\noindent\textbf{$\mathsf{K}_3$. Target abstraction determines how much new output
behavior the backbone must learn.}
For representation $Z$ of underlying field $X$, define
\begin{equation}
\alpha(Z;X)=H(X\mid Z).
\label{pre:eq:memoir_abstraction}
\end{equation}
Larger $\alpha$ means that more sample-specific information has been discarded
before task learning begins. A detailed continuous representation retains
location, geometry, color, and texture and therefore favors reconstruction,
masking, or spatial translation. A highly abstract representation keeps a
smaller semantic choice and behaves more like a categorical prediction target.
This is why an image and a sentence describing it occupy different positions
on one abstraction continuum, and why MSE-like reconstruction gradually gives
way to CE-like prediction. A VAE acts as a visual tokenizer: whatever it
removes becomes uncertainty that the generator must supply.

Consider a backbone that is already accustomed to producing dense RGB images.
It can continue producing dense outputs from many kinds of conditions. The
condition may be text, depth, pose, or another sparse or abstract input; once
encoded, it only needs to steer a generation process the backbone already
knows. Learning can therefore remain fast even when the condition leaves some
image details unresolved. Asking the same backbone to produce text, keypoints,
categories, or another sparse and abstract target is a larger change. The model
must learn what the new output elements mean, where they are placed, and how
they are decoded. The target therefore affects acquisition more strongly than
the condition: a familiar target permits many kinds of conditions, while an
unfamiliar target requires new output behavior.

This separates two sources of difficulty. A geometry-to-RGB task may leave
color, texture, and illumination unresolved yet learn quickly because the
backbone already knows how to generate RGB. An RGB-to-structure task may be more
fully determined by its condition and still require a new sparse or symbolic
output boundary. Directional uncertainty in $\mathsf{K}_4$ measures what the
condition leaves unknown; target knowledge in $\mathsf{K}_6$ measures how much
of the requested output behavior the backbone already possesses. Both affect
learning difficulty.

\noindent\textbf{$\mathsf{K}_4$. Directional difficulty is the information missing from the
condition.}
Define the unresolved information of a directed task by
\begin{equation}
\mu(C\!\rightarrow\!Y\mid t)=H(Y\mid C,t),
\label{pre:eq:memoir_directional_uncertainty}
\end{equation}
where $t$ is optional text or other context. In general,
$H(Y\mid C,t)\ne H(C\mid Y,t)$, so reversing two fields creates a different
task. RGB often constrains depth, while depth leaves RGB color, texture,
illumination, identity, and background unresolved. Text helps only by reducing
the missing target information,
\begin{equation}
H(Y\mid C)-H(Y\mid C,t)=I(Y;t\mid C).
\label{pre:eq:memoir_text_gain}
\end{equation}
A condition can therefore guide only the information it contains and the route
preserves. High dimensionality is not automatically strong supervision, and a
one-to-many reverse direction needs text, noise, another field, or a target
prior to supply its missing choices.

\noindent\textbf{$\mathsf{K}_5$. Field density determines whether an image-like boundary is
appropriate.}
Let $\rho(Y)$ be the fraction of spatial positions occupied by a sparse target,
and let $d_T$ be its task-native error after decoding. Under an unweighted
pixel loss, an empty prediction can obtain
\begin{equation}
\ell_{\mathrm{pix}}(0,Y)=\mathcal{O}(\rho(Y))
\label{pre:eq:memoir_sparse_background}
\end{equation}
while $d_T$ remains maximal because every point or identity is missing. This is
why a dense-image VAE can appear numerically successful while erasing sparse
structures. Heatmap lanes, fixed slots, sequences, direct parameters, and
one-shot structured decoders change which errors receive weight and which
identities have stable addresses. They preserve sparse information; they do
not invent appearance or physical values absent from the condition.

\noindent\textbf{$\mathsf{K}_6$. A better path cannot replace missing target knowledge.}
Let $\ell_{\mathrm{path}}(P)$ measure the length, curvature, or transport cost
of a path $P$. When the coverage condition $\kappa(T,G,s)=0$, the bridge
experiments showed
\begin{equation}
\ell_{\mathrm{path}}(P_1)<\ell_{\mathrm{path}}(P_2)
\quad\nRightarrow\quad Q_T(P_1)>Q_T(P_2)
\label{pre:eq:memoir_path_knowledge}
\end{equation}
when target-domain knowledge was absent. Optimal transport, reflow, intermediate supervision,
noise conditioning, and repeated correction could make a path shorter,
straighter, or more robust without teaching valid target texture or structure.
Path design controls how generation moves; the target prior determines what
valid destinations look like. Once that prior exists, a lightweight route can
redirect it efficiently.

\noindent\textbf{$\mathsf{K}_7$. Independent capability installation is measured by
interference.}
For an installed task $i$ and newly added route $j$, define
\begin{equation}
\Delta Q_{i\leftarrow j}=Q_i(G,\Phi)-Q_i(G,\Phi\cup\{\phi_j\}).
\label{pre:eq:memoir_interference}
\end{equation}
Independent installation requires $\Delta Q_{i\leftarrow j}=0$ up to evaluation
tolerance, even though every route remains part of one shared model and
runtime. Joint lanes, fused conditions, packed LoRAs, and shared-target LoRAs
coupled gradients, memory, schedules, or later revisions. Separately optimized
tunnels restrict each update to $\phi_j$. For $N$ installed routes,
\begin{equation}
P_{\mathrm{resident}}(N)=P_G+\sum_{i=1}^{N}|\phi_i|,
\qquad P_{\mathrm{active}}(j)=P_G+|\phi_j|.
\label{pre:eq:memoir_resident_active}
\end{equation}
The resident capability collection grows with the number of installed
relations, while task-specific active parameters and marginal acquisition cost
remain bounded by one selected route. This distinction, not a claim of constant
total storage, explains why thousand-task expansion becomes practical.

\noindent\textbf{$\mathsf{K}_8$. Success is a property of the decoded terminal object.}
Let $\mathcal{L}_{\mathrm{sur}}$ denote a training loss, latent alignment, or
intermediate preview, and let $Q_T$ evaluate the complete decoded output in the
task's native space. For two candidate designs $a$ and $b$, the experiments
repeatedly gave
\begin{equation}
\mathcal{L}_{\mathrm{sur}}(a)<\mathcal{L}_{\mathrm{sur}}(b)
\quad\nRightarrow\quad Q_T(a)>Q_T(b).
\label{pre:eq:memoir_terminal_success}
\end{equation}
Low surrogate loss can accompany black images, missing points, fragmented
categories, or codes unreadable by the native decoder. Intermediate quantities
diagnose a route; they do not establish output quality. The complete trajectory,
decoder, and target-native evaluator determine $Q_T$.

The remaining families supplied six more specific diagnostics. They refine the
eight core principles without changing the law's statement.

\noindent\textbf{$\mathsf{K}_9$. Local path accuracy and rollout accuracy are
different.}
For an $n$-step update with local errors $\varepsilon_k$ and state-transition
Lipschitz factors $\lambda_k$, terminal deviation can accumulate as
\begin{equation}
\varepsilon_{\mathrm{roll}}^{(n)}\ \lesssim\ \sum_{k=1}^{n}
\left(\prod_{q=k+1}^{n}\lambda_q\right)\varepsilon_k.
\label{pre:eq:memoir_rollout_error}
\end{equation}
Random-time training, one-step endpoint losses, and intermediate supervision
change different terms in this expression. A small local velocity error does
not ensure a small full-trajectory error; matched integration and full decoded
evaluation are required.

\noindent\textbf{$\mathsf{K}_{10}$. A discrete code needs stable coverage and a
meaningful metric.}
For codebook size $K$, let $u=|\mathcal{K}_{\mathrm{used}}|/K$ be utilization
and let $\delta_{\mathrm{code}}$ measure how much code meanings move during
training. High utilization with large $\delta_{\mathrm{code}}$ gives a moving
classification target, while frozen codes can give stable labels with poor
coverage. Moreover, nominal IDs have no default ordinal geometry:
\begin{equation}
d_{\mathrm{semantic}}(k_a,k_b)\ \not\propto\ |k_a-k_b|.
\label{pre:eq:memoir_discrete_contract}
\end{equation}
This separates codebook health, label stability, and cross-field predictability.

\noindent\textbf{$\mathsf{K}_{11}$. Diversity and fidelity are independent.}
Generated diversity is measured directly by $H(\hat Y\mid C,t)$. Noise
identities, CVAE variables, best-of-$k$ selection, target noising, and guidance
can increase or expose it, but
\begin{equation}
H(\hat Y\mid C,t)\uparrow\quad\nRightarrow\quad Q_T\uparrow.
\label{pre:eq:memoir_diversity_fidelity}
\end{equation}
Stochasticity represents unresolved choices; it does not teach the model which
samples are valid in the target domain.

\noindent\textbf{$\mathsf{K}_{12}$. Engineering speedups and semantic progress
must be accounted for separately.}
Write the measured acquisition cost as
\begin{equation}
\mathcal{C}_{\mathrm{total}}=\mathcal{C}_{\mathrm{data}}+
\mathcal{C}_{\mathrm{codec}}+\mathcal{C}_{\mathrm{route}}+
\mathcal{C}_{\mathrm{target}}.
\label{pre:eq:memoir_cost_decomposition}
\end{equation}
Caching reduces the first two terms; packing can share part of
$\mathcal{C}_{\mathrm{route}}$; neither operation necessarily reduces
translation error or the target-specific gradients in
$\mathcal{C}_{\mathrm{target}}$. This distinction
prevented throughput improvements from being mistaken for a solved mapping.

\noindent\textbf{$\mathsf{K}_{13}$. The best routing and objective depend on the
declared task.}
For permitted route family $\mathcal{M}$ and budget $B$, define
\begin{equation}
\phi_T^{*}=\arg\max_{\phi\in\mathcal{M},\,\mathcal{C}(\phi)\le B} Q_T(G,\phi).
\label{pre:eq:memoir_task_dependent_route}
\end{equation}
The experiments give $\phi_T^{*}\ne\phi_{T'}^{*}$ across interfaces: structured
targets favored separately normalized evidence routing, RGB retained an
advantage from native joint attention, and categorical outputs benefited from
explicit symbolic recovery. A unified task grammar therefore does not require
one identical internal route or loss for every information structure.

\noindent\textbf{$\mathsf{K}_{14}$. Installed relations form executable
compositions.}
For installed directed kernels $p_{\phi_1},\ldots,p_{\phi_L}$, a visual chain
is the explicit composition
\begin{equation}
p(Y_L\mid C_0)=\int\prod_{\ell=1}^{L}
p_{\phi_\ell}(Y_\ell\mid Y_{\ell-1})\,dY_{1:L-1}.
\label{pre:eq:memoir_route_composition}
\end{equation}
Separate route provenance is retained at every edge, so a chain can be assembled
without joint retraining. Its quality remains an empirical property of the
component routes and their intermediate representations, not an automatic
consequence of combinatorial path count.

These fourteen parameterized principles retain the plain conclusions of the
development record while making their terms observable. Together they give the
sequence that ultimately produced the Unified Law: preserve what the task must
distinguish; place fields in mutually usable representations; measure the
missing information in the declared direction; choose a boundary suited to the
field's abstraction and density; reuse a prior that knows valid targets;
install each relation without disturbing previous ones; and judge success only
after complete inference. The last six additionally formalize rollout error,
discrete codes, diversity, resource accounting, task-dependent routing, and
runtime composition. Section~\ref{pre:sec:formal_unified_law} extracts the
smallest set needed to state and refute the law.

The following traceability table closes the loop between the experimental
record and the formal statements. Each of the 60 rows covers the named
variants counted inside that family; the 52 most consequential variants are
also unpacked individually in
Table~\ref{tab:bridge_variant_level_record}. Thus the 198 currently indexed named variants
enter the formal account without pretending that every engineering variant
requires a new symbol or a separate law.

\clearpage
\onecolumn
\begingroup
\footnotesize
\setlength{\tabcolsep}{5pt}
\renewcommand{\arraystretch}{1.12}
\begin{longtable}{p{0.72\textwidth}p{0.19\textwidth}}
\caption{\textbf{Formal traceability of all 60 experimental families.}
The right column identifies the parameterized conclusions supported or
qualified by every family. Table~\ref{pre:tab:formal_variant_traceability}
then opens the family counts into 198 individual variants.}
\label{pre:tab:formal_family_traceability}\\
\toprule
\textbf{Experimental family} & \textbf{Formal knowledge} \\
\midrule
\endfirsthead
\multicolumn{2}{l}{\small\itshape Table~\ref{pre:tab:formal_family_traceability} continued.}\\
\toprule
\textbf{Experimental family} & \textbf{Formal knowledge} \\
\midrule
\endhead
\midrule
\multicolumn{2}{r}{\small\itshape Continued on the next page.}\\
\endfoot
\bottomrule
\endlastfoot
Expanded RGB VAE & $\mathsf{K}_1,\mathsf{K}_2,\mathsf{K}_7$ \\
Frozen RGB path & $\mathsf{K}_1,\mathsf{K}_7$ \\
Progressive TinyVAE expansion & $\mathsf{K}_1,\mathsf{K}_3,\mathsf{K}_5,\mathsf{K}_{12}$ \\
Offset-routing schemes D--F & $\mathsf{K}_1,\mathsf{K}_2,\mathsf{K}_6,\mathsf{K}_{13}$ \\
LayerDiffuse-style auxiliary path & $\mathsf{K}_1,\mathsf{K}_7,\mathsf{K}_{13}$ \\
Independent channel VAE & $\mathsf{K}_1,\mathsf{K}_2$ \\
Sparse categorical VAE & $\mathsf{K}_1,\mathsf{K}_5,\mathsf{K}_{10}$ \\
Sparse patch/MLP VAE & $\mathsf{K}_1,\mathsf{K}_5,\mathsf{K}_{13}$ \\
Cached paired latents & $\mathsf{K}_{12}$ \\
Latent objectives and masking & $\mathsf{K}_8,\mathsf{K}_{13}$ \\
Seven local interpolation rules & $\mathsf{K}_2,\mathsf{K}_6$ \\
Direct ``God Bridge'' & $\mathsf{K}_2,\mathsf{K}_6$ \\
Attention bridge & $\mathsf{K}_2,\mathsf{K}_6,\mathsf{K}_{13}$ \\
Bidirectional CNN bridge & $\mathsf{K}_2,\mathsf{K}_4$ \\
Whitening and fixed normalization & $\mathsf{K}_2,\mathsf{K}_6$ \\
Learnable invertible alignment & $\mathsf{K}_2,\mathsf{K}_6$ \\
Joint embedding and JEPA-style alignment & $\mathsf{K}_1,\mathsf{K}_2,\mathsf{K}_3$ \\
Multi-step RF repair & $\mathsf{K}_6,\mathsf{K}_9,\mathsf{K}_{12}$ \\
Random-time RF & $\mathsf{K}_8,\mathsf{K}_9$ \\
Endpoint and intermediate losses & $\mathsf{K}_8,\mathsf{K}_9,\mathsf{K}_{13}$ \\
Minibatch optimal transport & $\mathsf{K}_6,\mathsf{K}_9$ \\
Reflow and teacher-path correction & $\mathsf{K}_6,\mathsf{K}_9$ \\
Noise ranking and noise conditioning & $\mathsf{K}_4,\mathsf{K}_{11}$ \\
VQ bridge with CE & $\mathsf{K}_1,\mathsf{K}_2,\mathsf{K}_{10}$ \\
VQ utilization variants & $\mathsf{K}_2,\mathsf{K}_{10}$ \\
MLP index-MSE bridge & $\mathsf{K}_1,\mathsf{K}_{10},\mathsf{K}_{13}$ \\
Transformer bridge & $\mathsf{K}_2,\mathsf{K}_6,\mathsf{K}_{13}$ \\
Pure velocity field, CFG, and condition dropout & $\mathsf{K}_4,\mathsf{K}_6,\mathsf{K}_{11}$ \\
CVAE-style prior--posterior bridge & $\mathsf{K}_4,\mathsf{K}_6,\mathsf{K}_{11}$ \\
Best-of-$k$ endpoint sampling & $\mathsf{K}_6,\mathsf{K}_8,\mathsf{K}_{11}$ \\
Target-noised training & $\mathsf{K}_6,\mathsf{K}_9,\mathsf{K}_{11}$ \\
Style-latent conditioning & $\mathsf{K}_4,\mathsf{K}_6,\mathsf{K}_{11}$ \\
Reverse-noise encoding & $\mathsf{K}_6,\mathsf{K}_9$ \\
Conditional latent diffusion & $\mathsf{K}_6,\mathsf{K}_9,\mathsf{K}_{13}$ \\
Rectified auto-coder paths & $\mathsf{K}_2,\mathsf{K}_6,\mathsf{K}_9$ \\
Shadow-variable and perturbed-latent routes & $\mathsf{K}_1,\mathsf{K}_2,\mathsf{K}_6$ \\
Video-backbone branches & $\mathsf{K}_6,\mathsf{K}_7,\mathsf{K}_{13}$ \\
Joint video-lane strategies & $\mathsf{K}_7,\mathsf{K}_{12},\mathsf{K}_{13}$ \\
FLUX/Omini evidence placement & $\mathsf{K}_4,\mathsf{K}_{13}$ \\
Task-agnostic flow objectives & $\mathsf{K}_8,\mathsf{K}_9,\mathsf{K}_{13}$ \\
Dense flow objective levers & $\mathsf{K}_8,\mathsf{K}_{13}$ \\
Packed multi-target LoRA acquisition & $\mathsf{K}_7,\mathsf{K}_{12}$ \\
Shared-target multi-condition LoRA & $\mathsf{K}_4,\mathsf{K}_7,\mathsf{K}_{12}$ \\
NJA/MTA TunnelingNet routes & $\mathsf{K}_6,\mathsf{K}_7,\mathsf{K}_{13}$ \\
Semantic heatmap and inverse codec & $\mathsf{K}_1,\mathsf{K}_5$ \\
Person--keypoint slot heatmaps & $\mathsf{K}_1,\mathsf{K}_5$ \\
Body-part and full-body heatmaps & $\mathsf{K}_1,\mathsf{K}_5,\mathsf{K}_{12}$ \\
Direct sparse sequence heads & $\mathsf{K}_1,\mathsf{K}_5,\mathsf{K}_{13}$ \\
ScoreHMR WSV structure route & $\mathsf{K}_6,\mathsf{K}_{13}$ \\
Atomic object-box slots & $\mathsf{K}_1,\mathsf{K}_5,\mathsf{K}_{13}$ \\
Object BFS tree & $\mathsf{K}_1,\mathsf{K}_5$ \\
Point-cloud skeleton initialization & $\mathsf{K}_5,\mathsf{K}_6,\mathsf{K}_{13}$ \\
Sparse and mixed audio routes & $\mathsf{K}_1,\mathsf{K}_3,\mathsf{K}_4,\mathsf{K}_{13}$ \\
Dense categorical-mask pilots & $\mathsf{K}_5,\mathsf{K}_8,\mathsf{K}_{10}$ \\
Categorical objective modes & $\mathsf{K}_8,\mathsf{K}_{10},\mathsf{K}_{13}$ \\
Categorical region refiners & $\mathsf{K}_8,\mathsf{K}_{10},\mathsf{K}_{13}$ \\
Universal Pose custom TunnelingNet & $\mathsf{K}_1,\mathsf{K}_5,\mathsf{K}_{13}$ \\
Pose-detail objective pilots & $\mathsf{K}_5,\mathsf{K}_8,\mathsf{K}_{13}$ \\
Named custom-variable routes & $\mathsf{K}_1,\mathsf{K}_7,\mathsf{K}_{13}$ \\
Runtime route composition & $\mathsf{K}_7,\mathsf{K}_{14}$ \\
\end{longtable}
\clearpage
\begin{longtable}{p{0.035\textwidth}p{0.155\textwidth}p{0.225\textwidth}p{0.43\textwidth}p{0.095\textwidth}}
\caption{\textbf{Variant-level evidence and formal traceability.}
The 198 currently indexed methodological variants are listed independently. Each row
records the intervention actually distinguished in the development record,
the direct observation retained from it, and the formal principle(s) it
supports. Engineering-only repairs, repeated seeds, and ordinary
hyperparameter sweeps remain excluded.}
\label{pre:tab:formal_variant_traceability}\\
\toprule
\textbf{ID} & \textbf{Family} & \textbf{Variant} &
\textbf{Direct observation and knowledge retained} & \textbf{Formal link} \\
\midrule
\endfirsthead
\multicolumn{5}{l}{\small\itshape Table~\ref{pre:tab:formal_variant_traceability} continued.}\\
\toprule
\textbf{ID} & \textbf{Family} & \textbf{Variant} &
\textbf{Direct observation and knowledge retained} & \textbf{Formal link} \\
\midrule
\endhead
\midrule
\multicolumn{5}{r}{\small\itshape Continued on the next page.}\\
\endfoot
\bottomrule
\endlastfoot
\input{tables/empirical_variant_traceability_rows.tex}

\end{longtable}
\endgroup
\clearpage
\twocolumn

\subsection{From Four Initial Tunnels to 1,148 Trained Baselines}
\label{pre:subsec:memoir_prospective_confirmation}

The order of these experiments is important. We first studied four
representative TunnelingNets 003, 270, 275, and 313, covering the four
interfaces. Together with the earlier bridge, codec, sparse-representation, and
routing experiments, these four Tunnels revealed one common rule: keep the
shared generator fixed, describe the known information as the condition,
describe the requested information as the target, and train the corresponding
TunnelingNet to redirect generation toward that target.

The development record also compares harder and easier designs for the same
task. The rule requires the chosen representation to preserve the distinctions
needed by the task and to present them in a form the generator can use. Early
sparse-codec experiments showed that empty background could overwhelm isolated
points. A one-shot dense rendering of keypoints or boxes therefore risked
losing locations, identities, or connections. The later single-canvas,
semantic-color, and pseudocolor-box experiments were harder to optimize than
the revised designs and lost exactly these distinctions. Multilane heatmaps,
fixed person--keypoint slots, and native structured outputs preserved the
missing distinctions and made the same tasks markedly easier to learn.

The categorical experiments showed a second case. Continuous generation could
draw the spatial support of a region, but an approximate color could not
reliably identify an exact class. Eight dense categorical pilots produced
background collapse, mixed legal colors, or fragmented regions. Separating
generated geometry from a lightweight categorical refiner then recovered the
exact labels. Universal Pose supplied a third case: in a full-frame one-shot
pose image, the sparse body signal is dominated by background and small joints
can disappear. The full-frame route collapsed toward black, whereas corrected
geometry, instance crops, and gradient supervision preserved the body
structure. The tasks themselves were possible; their first designs had
discarded information that the tasks required. Once that information was
preserved, the same tasks became markedly easier to learn. The earlier versions
might also learn with more data, parameters, or optimization; the comparison
concerns their relative difficulty. The underlying sparse-codec experiments
are described in Sec.~\ref{pre:subsec:memoir_sparse_codecs}; the later heatmap,
categorical, structured-output, and Universal-Pose experiments appear in
Sec.~\ref{pre:subsec:memoir_new_target_structures}. Their family-level record
is given in Table~\ref{tab:bridge_variant_record}, and the individual
interventions are indexed in
Table~\ref{pre:tab:formal_variant_traceability}: variants 136--142 record the
heatmap representations, 165--181 record the categorical pilots and refiners,
and 182--194 record the Universal-Pose and pose-detail experiments. Each row
states the implemented change, direct observation, and formal lesson.

We then used this rule to fix the interface and representation of the remaining
routes, without changing the shared backbone or previously learned tasks. We
subsequently trained and evaluated the other 1,144 tasks. Together with the
original four, this produced 1,148 trained and evaluated visual tasks across 39
output groups, which serve as baselines for directed visual tasks that mostly
had none. The development and training records show this order: the rule came
first, and the 1,144 routes were trained afterward.

We also added two capabilities introduced by other papers as ordinary
TunnelingNets: LayerDiffuse-style transparency~\cite{10.1145/3658150} and
IC-Light relighting~\cite{zhang2025scaling}. Transparency used 1,000 updates,
and relighting used 2,800.
Section~\ref{subsec:custom_capability_showpiece} shows the complete experiments
and outputs.

\subsection{Summary of the Unified Law}
\label{pre:subsec:memoir_summary}

The two-year experimental study spans \textcolor{red}{\textbf{60 architectural and
methodological families}} and at least \textcolor{red}{\textbf{198 explicitly named model and
training variants}}. Table~\ref{tab:bridge_variant_record} attaches a concrete
experimental lesson to every family, while
Table~\ref{tab:bridge_variant_level_record} describes \textcolor{red}{\textbf{52 representative variants}}
into their motivation, exact intervention, observed behavior, and causal
conclusion. Altogether, we accumulated
\textcolor{red}{\textbf{at least 2,293 training attempts documented in surviving records}}.
Among them, \textcolor{red}{\textbf{1,546 different training runs}} retain complete
configurations, and at least \textcolor{red}{\textbf{1,380 trained models}} retain
their weights. These conservative counts exclude older experiments that were
deleted, moved, or archived elsewhere. From these experiments,
we obtained the following \textcolor{red}{\textbf{eight principles}}.

\noindent\textcolor{blue}{\textbf{1. If the representation erases a distinction,
the model cannot learn that distinction later.}}
The sparse-pixel, semantic-color, shared-VAE, and object-structure experiments
looked different, yet each representation had collapsed two states that the
requested output needed to distinguish. No architecture or loss can recover a
joint identity, object membership, or physical value after the boundary has
erased it. Heatmaps, fixed slots, typed sequences, object atoms, and named
custom fields became learnable when each required distinction received a stable
address and a faithful encoding. The first test for a new capability is thus
whether its information survives both entry into and exit from the model.

\noindent\textcolor{blue}{\textbf{2. Two latent spaces can have the same shape
and distribution while speaking different internal languages.}}
Dense fields, sparse structures, audio, language, and RGB can describe one
world state while assigning different meanings to their coordinates. Whitening,
alignment losses, invertible transforms, and the shadow-variable experiments
could match dimensions or global statistics while decoded samples remained
wrong. Mutual readability requires three things: a faithful codec for each
side, stable identities for the information being exchanged, and paired
sample-level relations that teach what one coordinate system means to the
other.

\noindent\textcolor{blue}{\textbf{3. As a representation moves from detailed
visual information toward abstract symbols, the useful learning objective moves
from reconstruction toward prediction.}}
An image keeps concrete location, color, geometry, and texture; a sentence keeps
a much more abstract semantic account of the same scene. Moving from the first
toward the second is therefore also a move from MSE-like reconstruction toward
CE-like prediction of symbols. We observed the same difference among visual
latents. A compact, low-channel SD-style VAE latent has already discarded much
of the spatial detail and often behaves more like a prediction target. A
high-channel FLUX or RAE~\cite{zheng2026diffusion} representation retains more visual structure and is
better suited to masking, aligned translation, and conditional generation. The
VAE acts as a visual tokenizer: details removed by the codec must be supplied by
the generator, while details preserved by the codec can be reconstructed or
transformed directly. Channel count alone is not the rule; what information the
representation keeps is the rule.

\noindent\textcolor{blue}{\textbf{4. Reversing condition and target changes what
information is missing and can turn an easy estimate into a hard generation
problem.}}
RGB often constrains depth strongly, while depth does not determine color,
texture, illumination, identity, or background. RGB-to-depth can therefore
behave like estimation; depth-to-RGB must choose among many valid appearances.
The missing choices must come from text, another condition, sampled noise, or a
pretrained target-domain prior. Reversing the same pair of fields creates a new
learning problem, so a visual task is defined by its condition, target, and
direction.

\noindent\textcolor{blue}{\textbf{5. A shorter or smoother path cannot teach a
model what a valid target looks like.}}
Optimal transport shortened and uncrossed reference segments; reflow,
intermediate supervision, repeated correction, and noise conditioning improved
trajectory geometry or robustness. None of them supplied textures, shapes, or
other target semantics absent from the bridge. Reliable conversion emerged
when a pretrained target generator already knew the target distribution and a
small learned path redirected that generator's native trajectory. Path design
controls how the model travels; the target prior determines what valid
destinations look like.

\noindent\textcolor{blue}{\textbf{6. Each new condition--target relation should
be learned separately and then installed inside the shared model.}} Joint video lanes, fused conditions,
packed LoRAs, and shared-target LoRAs coupled the memory, gradients, sampling
schedule, and future revision of otherwise different relations. Adding or
repairing one relation could then disturb the others. Separately acquired
tunnels kept the generative backbone shared while storing each condition--target
relation in its own small path; inference activates only the path requested by
the current task. Capability addition becomes a local installation, while the
installed collection still executes as one model.

\noindent\textcolor{blue}{\textbf{7. Thousands of tasks become practical only
when each new relation is cheap and leaves earlier relations untouched.}}
Full-codec retraining changed a shared representation; external bridges had to
learn a target generator; joint multimodal generators coupled every relation;
packed LoRAs still repeated target-side computation and gradients. Residual
interventions inside a mature generator changed only a small task-relevant path
and left installed paths untouched. This bounded marginal parameter, training,
and revision cost is what made 1,148 routes practical; a method whose new task
reopens all old tasks cannot grow at the same rate.

\noindent\textcolor{blue}{\textbf{8. A training loss or intermediate preview is
not success; the complete inference process must produce a valid final target.}}
Partial denoising previews, low latent losses, private round trips, and aligned
global statistics sometimes looked successful while complete inference produced
black images, fragmented categories, missing structures, or codes unreadable by
the native decoder. We therefore run the full trajectory, decode it through the
target-native interface, and evaluate the resulting object with task-native
readers and metrics. Intermediate diagnostics explain failure; they do not by
themselves establish a usable capability.

These eight principles supplied the experimental foundation of the Unified
Law. Their common structure is direct: known RGB, text, and world information
act as conditions, and requested information is generated as the target.
Estimation, control, reconstruction, inverse rendering, structured prediction,
and cross-property translation differ in the placement and direction of those
variables.

\phantomsection\label{pre:par:memoir_development_validation}
Beginning with the initial WSG idea in 2024, the next two years of development
covered \textcolor{red}{\textbf{60 experimental families}} and
\textcolor{red}{\textbf{198 named variants}}. Across this process, we accumulated
\textcolor{red}{\textbf{at least 2,293 training attempts documented in surviving records}}.
Complete configurations remain for 1,546 different training runs, and weights
remain for at least \textcolor{red}{\textbf{1,380 trained models}}.
We repeatedly changed representations,
objectives, architectures, training paths, and condition--target directions,
then compared which behaviors survived those changes. The recurring results
gradually became the eight principles above. Once this evidence converged, we
formally stated the Unified Law and used it to develop the complete framework:
WSG organizes RGB and additional world information, WSM installs the resulting
visual capabilities around a shared generative backbone, and TunnelingNet
redirects that backbone toward each requested target at low marginal cost.
Using this final framework, we then trained and evaluated
\textcolor{red}{\textbf{1,148 visual tasks}} across 39 output groups and all four interfaces. The two-year experimental process
and the final evaluation are therefore consecutive stages of one study: the
first explains how the law and mechanism were discovered, while the second shows
how broadly the resulting method works. These numbers describe different
stages: the documented training attempts show how we developed the
method, and the 1,148 tasks measure how broadly the completed method was
evaluated.

%% file: tables/empirical_variant_traceability_rows.tex
% Auto-maintained variant-level evidence rows. Public text contains no private paths, commit hashes, or chat identifiers.
1 & Expanded RGB VAE & Direct expanded RGB input/output boundary & Directly widening the mature RGB boundary disturbed its latent; added information and preserved RGB competed. & $\mathsf{K_1},\mathsf{K_2},\mathsf{K_7}$ \\
2 & Frozen RGB path & Frozen native RGB path with trainable outer layers & Freezing preserved RGB but left the added field too little influence. & $\mathsf{K_1},\mathsf{K_7}$ \\
3 & Progressive TinyVAE expansion & Hidden-width expansion in 16-channel stages & Extra codec capacity helped dense reconstruction but did not solve sparse structure or full-tensor scaling. & $\mathsf{K_1},\mathsf{K_3},\mathsf{K_5},\mathsf{K_12}$ \\
4 & Progressive TinyVAE expansion & Five-channel latent stage & Extra codec capacity helped dense reconstruction but did not solve sparse structure or full-tensor scaling. & $\mathsf{K_1},\mathsf{K_3},\mathsf{K_5},\mathsf{K_12}$ \\
5 & Progressive TinyVAE expansion & Six-channel latent stage & Extra codec capacity helped dense reconstruction but did not solve sparse structure or full-tensor scaling. & $\mathsf{K_1},\mathsf{K_3},\mathsf{K_5},\mathsf{K_12}$ \\
6 & Progressive TinyVAE expansion & Seven-channel latent stage & Extra codec capacity helped dense reconstruction but did not solve sparse structure or full-tensor scaling. & $\mathsf{K_1},\mathsf{K_3},\mathsf{K_5},\mathsf{K_12}$ \\
7 & Progressive TinyVAE expansion & Eight-channel latent stage & Extra codec capacity helped dense reconstruction but did not solve sparse structure or full-tensor scaling. & $\mathsf{K_1},\mathsf{K_3},\mathsf{K_5},\mathsf{K_12}$ \\
8 & Progressive TinyVAE expansion & Full 256-channel WSG training & Extra codec capacity helped dense reconstruction but did not solve sparse structure or full-tensor scaling. & $\mathsf{K_1},\mathsf{K_3},\mathsf{K_5},\mathsf{K_12}$ \\
9 & Offset-routing schemes D{-}{-}F & Single-depth direct offset injection & Injection position changed the information available to the decoder; better round trips did not ensure a self-contained latent. & $\mathsf{K_1},\mathsf{K_2},\mathsf{K_6},\mathsf{K_13}$ \\
10 & Offset-routing schemes D{-}{-}F & Multi-depth direct offset injection & Injection position changed the information available to the decoder; better round trips did not ensure a self-contained latent. & $\mathsf{K_1},\mathsf{K_2},\mathsf{K_6},\mathsf{K_13}$ \\
11 & Offset-routing schemes D{-}{-}F & Decoder-entrance injection & Injection position changed the information available to the decoder; better round trips did not ensure a self-contained latent. & $\mathsf{K_1},\mathsf{K_2},\mathsf{K_6},\mathsf{K_13}$ \\
12 & Offset-routing schemes D{-}{-}F & Late-decoder injection & Injection position changed the information available to the decoder; better round trips did not ensure a self-contained latent. & $\mathsf{K_1},\mathsf{K_2},\mathsf{K_6},\mathsf{K_13}$ \\
13 & Offset-routing schemes D{-}{-}F & Four-scale encoder{-}{-}decoder jump links & Injection position changed the information available to the decoder; better round trips did not ensure a self-contained latent. & $\mathsf{K_1},\mathsf{K_2},\mathsf{K_6},\mathsf{K_13}$ \\
14 & Offset-routing schemes D{-}{-}F & Decoder-predicted offset recovery & Injection position changed the information available to the decoder; better round trips did not ensure a self-contained latent. & $\mathsf{K_1},\mathsf{K_2},\mathsf{K_6},\mathsf{K_13}$ \\
15 & Offset-routing schemes D{-}{-}F & Fixed-ratio offset recovery & Injection position changed the information available to the decoder; better round trips did not ensure a self-contained latent. & $\mathsf{K_1},\mathsf{K_2},\mathsf{K_6},\mathsf{K_13}$ \\
16 & LayerDiffuse-style auxiliary path & Auxiliary encoder lane & A protected auxiliary path carried limited extra information but lacked bandwidth for a growing world representation. & $\mathsf{K_1},\mathsf{K_7},\mathsf{K_13}$ \\
17 & LayerDiffuse-style auxiliary path & Auxiliary decoder lane & A protected auxiliary path carried limited extra information but lacked bandwidth for a growing world representation. & $\mathsf{K_1},\mathsf{K_7},\mathsf{K_13}$ \\
18 & LayerDiffuse-style auxiliary path & Transparency-style protected auxiliary path & A protected auxiliary path carried limited extra information but lacked bandwidth for a growing world representation. & $\mathsf{K_1},\mathsf{K_7},\mathsf{K_13}$ \\
19 & Independent channel VAE & Per-channel continuous VAE & Private codecs reconstructed fields but created mutually unreadable coordinate systems. & $\mathsf{K_1},\mathsf{K_2}$ \\
20 & Independent channel VAE & Field-specific normalized VAE & Private codecs reconstructed fields but created mutually unreadable coordinate systems. & $\mathsf{K_1},\mathsf{K_2}$ \\
21 & Independent channel VAE & Independent field latent space & Private codecs reconstructed fields but created mutually unreadable coordinate systems. & $\mathsf{K_1},\mathsf{K_2}$ \\
22 & Sparse categorical VAE & Categorical sparse decoder & Discrete supervision preserved symbols better, while small spatial or class errors still destroyed sparse identities. & $\mathsf{K_1},\mathsf{K_5},\mathsf{K_10}$ \\
23 & Sparse categorical VAE & Integer-bin output representation & Discrete supervision preserved symbols better, while small spatial or class errors still destroyed sparse identities. & $\mathsf{K_1},\mathsf{K_5},\mathsf{K_10}$ \\
24 & Sparse categorical VAE & Cross-entropy sparse objective & Discrete supervision preserved symbols better, while small spatial or class errors still destroyed sparse identities. & $\mathsf{K_1},\mathsf{K_5},\mathsf{K_10}$ \\
25 & Sparse categorical VAE & Hybrid CE{-}{-}MSE sparse objective & Discrete supervision preserved symbols better, while small spatial or class errors still destroyed sparse identities. & $\mathsf{K_1},\mathsf{K_5},\mathsf{K_10}$ \\
26 & Sparse patch/MLP VAE & Patch/MLP sparse codec & Removing empty pixels reduced imbalance but removed spatial organization; coordinates restored only part of it. & $\mathsf{K_1},\mathsf{K_5},\mathsf{K_13}$ \\
27 & Sparse patch/MLP VAE & Valid-value tokenization & Removing empty pixels reduced imbalance but removed spatial organization; coordinates restored only part of it. & $\mathsf{K_1},\mathsf{K_5},\mathsf{K_13}$ \\
28 & Sparse patch/MLP VAE & Coordinate-augmented valid-value tokens & Removing empty pixels reduced imbalance but removed spatial organization; coordinates restored only part of it. & $\mathsf{K_1},\mathsf{K_5},\mathsf{K_13}$ \\
29 & Cached paired latents & Cached paired source/target latents & Caching separated image I/O and codec cost from the scientific cross-space mapping problem. & $\mathsf{K_12}$ \\
30 & Latent objectives and masking & Latent MSE & Numerically better latent optimization did not guarantee correct decoded endpoints. & $\mathsf{K_8},\mathsf{K_13}$ \\
31 & Latent objectives and masking & Latent MAE & Numerically better latent optimization did not guarantee correct decoded endpoints. & $\mathsf{K_8},\mathsf{K_13}$ \\
32 & Latent objectives and masking & Decoded pixel supervision & Numerically better latent optimization did not guarantee correct decoded endpoints. & $\mathsf{K_8},\mathsf{K_13}$ \\
33 & Latent objectives and masking & Random 4-by-4 latent masking & Numerically better latent optimization did not guarantee correct decoded endpoints. & $\mathsf{K_8},\mathsf{K_13}$ \\
34 & Seven local interpolation rules & Nearest-neighbor interpolation & No tested local rule made neighborhoods in independently trained latent spaces semantically corresponding. & $\mathsf{K_2},\mathsf{K_6}$ \\
35 & Seven local interpolation rules & RBF interpolation & No tested local rule made neighborhoods in independently trained latent spaces semantically corresponding. & $\mathsf{K_2},\mathsf{K_6}$ \\
36 & Seven local interpolation rules & Thin-plate interpolation & No tested local rule made neighborhoods in independently trained latent spaces semantically corresponding. & $\mathsf{K_2},\mathsf{K_6}$ \\
37 & Seven local interpolation rules & Barycentric interpolation & No tested local rule made neighborhoods in independently trained latent spaces semantically corresponding. & $\mathsf{K_2},\mathsf{K_6}$ \\
38 & Seven local interpolation rules & Local-linear interpolation & No tested local rule made neighborhoods in independently trained latent spaces semantically corresponding. & $\mathsf{K_2},\mathsf{K_6}$ \\
39 & Seven local interpolation rules & Manifold interpolation & No tested local rule made neighborhoods in independently trained latent spaces semantically corresponding. & $\mathsf{K_2},\mathsf{K_6}$ \\
40 & Seven local interpolation rules & Neural interpolation & No tested local rule made neighborhoods in independently trained latent spaces semantically corresponding. & $\mathsf{K_2},\mathsf{K_6}$ \\
41 & Direct God Bridge & Direct bidirectional God Bridge & A compact one-pass mapper could not discover the complete cross-space relation. & $\mathsf{K_2},\mathsf{K_6}$ \\
42 & Attention bridge & Self-attention bridge & Global interaction and added width did not repair incompatible endpoint meanings. & $\mathsf{K_2},\mathsf{K_6},\mathsf{K_13}$ \\
43 & Attention bridge & Cross-attention bridge & Global interaction and added width did not repair incompatible endpoint meanings. & $\mathsf{K_2},\mathsf{K_6},\mathsf{K_13}$ \\
44 & Attention bridge & Compact 4M attention bridge & Global interaction and added width did not repair incompatible endpoint meanings. & $\mathsf{K_2},\mathsf{K_6},\mathsf{K_13}$ \\
45 & Bidirectional CNN bridge & RGB-to-field residual CNN & Opposite directions learned at different rates and qualities, making direction part of the task. & $\mathsf{K_2},\mathsf{K_4}$ \\
46 & Bidirectional CNN bridge & Field-to-RGB residual CNN & Opposite directions learned at different rates and qualities, making direction part of the task. & $\mathsf{K_2},\mathsf{K_4}$ \\
47 & Whitening and fixed normalization & Fixed mean/variance whitening & Matching scale improved conditioning without aligning sample-level semantics. & $\mathsf{K_2},\mathsf{K_6}$ \\
48 & Learnable invertible alignment & Learned 1-by-1 invertible projection & Population statistics could align while corresponding samples still decoded incorrectly. & $\mathsf{K_2},\mathsf{K_6}$ \\
49 & Learnable invertible alignment & Moment-matching alignment & Population statistics could align while corresponding samples still decoded incorrectly. & $\mathsf{K_2},\mathsf{K_6}$ \\
50 & Learnable invertible alignment & Covariance-matching alignment & Population statistics could align while corresponding samples still decoded incorrectly. & $\mathsf{K_2},\mathsf{K_6}$ \\
51 & Joint embedding and JEPA-style alignment & Joint-embedding alignment & Shared semantics emerged, but exact modality-specific information still required separate capacity. & $\mathsf{K_1},\mathsf{K_2},\mathsf{K_3}$ \\
52 & Joint embedding and JEPA-style alignment & JEPA-style predictive alignment & Shared semantics emerged, but exact modality-specific information still required separate capacity. & $\mathsf{K_1},\mathsf{K_2},\mathsf{K_3}$ \\
53 & Multi-step RF repair & Fully unrolled multi-step RF repair & Repeated repair increased memory and instability without supplying missing target knowledge. & $\mathsf{K_6},\mathsf{K_9},\mathsf{K_12}$ \\
54 & Random-time RF & Random-time single-point RF training & Random-time supervision saved memory, while local velocity errors accumulated during full rollout. & $\mathsf{K_8},\mathsf{K_9}$ \\
55 & Endpoint and intermediate losses & One-step endpoint loss & Endpoint-aware losses exposed hidden failures, but matched full paths restored substantial cost and could conflict with local fitting. & $\mathsf{K_8},\mathsf{K_9},\mathsf{K_13}$ \\
56 & Endpoint and intermediate losses & Matched multi-step endpoint loss & Endpoint-aware losses exposed hidden failures, but matched full paths restored substantial cost and could conflict with local fitting. & $\mathsf{K_8},\mathsf{K_9},\mathsf{K_13}$ \\
57 & Endpoint and intermediate losses & Checkpointed endpoint integration & Endpoint-aware losses exposed hidden failures, but matched full paths restored substantial cost and could conflict with local fitting. & $\mathsf{K_8},\mathsf{K_9},\mathsf{K_13}$ \\
58 & Endpoint and intermediate losses & Intermediate-state supervision & Endpoint-aware losses exposed hidden failures, but matched full paths restored substantial cost and could conflict with local fitting. & $\mathsf{K_8},\mathsf{K_9},\mathsf{K_13}$ \\
59 & Endpoint and intermediate losses & Aligned/outside-space endpoint supervision & Endpoint-aware losses exposed hidden failures, but matched full paths restored substantial cost and could conflict with local fitting. & $\mathsf{K_8},\mathsf{K_9},\mathsf{K_13}$ \\
60 & Minibatch optimal transport & Minibatch OT pairing & Optimal pairing shortened and uncrossed paths without supplying target semantics. & $\mathsf{K_6},\mathsf{K_9}$ \\
61 & Reflow and teacher-path correction & Reflow retraining & Path refinement helped only when the teacher already reached useful endpoints. & $\mathsf{K_6},\mathsf{K_9}$ \\
62 & Reflow and teacher-path correction & Teacher-endpoint path correction & Path refinement helped only when the teacher already reached useful endpoints. & $\mathsf{K_6},\mathsf{K_9}$ \\
63 & Reflow and teacher-path correction & Repeated correction blocks & Path refinement helped only when the teacher already reached useful endpoints. & $\mathsf{K_6},\mathsf{K_9}$ \\
64 & Noise ranking and noise conditioning & Persistent noise ranking & Identifiable random starts increased path diversity but did not create target realism. & $\mathsf{K_4},\mathsf{K_11}$ \\
65 & Noise ranking and noise conditioning & Explicit noise conditioning & Identifiable random starts increased path diversity but did not create target realism. & $\mathsf{K_4},\mathsf{K_11}$ \\
66 & VQ bridge with CE & Frozen-codebook VQ bridge with CE & Legal code predictions still suffered from low utilization, moving labels, and cross-space semantic mismatch. & $\mathsf{K_1},\mathsf{K_2},\mathsf{K_10}$ \\
67 & VQ bridge with CE & Updated-codebook VQ bridge with CE & Legal code predictions still suffered from low utilization, moving labels, and cross-space semantic mismatch. & $\mathsf{K_1},\mathsf{K_2},\mathsf{K_10}$ \\
68 & VQ utilization variants & Smaller codebook & Healthier code usage did not by itself improve cross-field translation. & $\mathsf{K_2},\mathsf{K_10}$ \\
69 & VQ utilization variants & Codebook warm-up updates & Healthier code usage did not by itself improve cross-field translation. & $\mathsf{K_2},\mathsf{K_10}$ \\
70 & VQ utilization variants & Utilization-aware monitoring & Healthier code usage did not by itself improve cross-field translation. & $\mathsf{K_2},\mathsf{K_10}$ \\
71 & VQ utilization variants & Alternative CE{-}{-}MSE weighting & Healthier code usage did not by itself improve cross-field translation. & $\mathsf{K_2},\mathsf{K_10}$ \\
72 & MLP index-MSE bridge & MLP index-MSE prediction & Numerical distance between nominal code IDs was not a semantic metric. & $\mathsf{K_1},\mathsf{K_10},\mathsf{K_13}$ \\
73 & Transformer bridge & Transformer latent bridge & A global sequence model also failed to close the decoded distribution gap. & $\mathsf{K_2},\mathsf{K_6},\mathsf{K_13}$ \\
74 & Pure velocity field, CFG, and condition dropout & Pure conditional velocity objective & A cleaner conditional path helped toward RGB, while information-poor reverse directions remained difficult. & $\mathsf{K_4},\mathsf{K_6},\mathsf{K_11}$ \\
75 & Pure velocity field, CFG, and condition dropout & Classifier-free guidance & A cleaner conditional path helped toward RGB, while information-poor reverse directions remained difficult. & $\mathsf{K_4},\mathsf{K_6},\mathsf{K_11}$ \\
76 & Pure velocity field, CFG, and condition dropout & Condition dropout & A cleaner conditional path helped toward RGB, while information-poor reverse directions remained difficult. & $\mathsf{K_4},\mathsf{K_6},\mathsf{K_11}$ \\
77 & CVAE-style prior{-}{-}posterior bridge & CVAE prior{-}{-}posterior bridge & A stochastic latent represented uncertainty but introduced another difficult prior and target decoder. & $\mathsf{K_4},\mathsf{K_6},\mathsf{K_11}$ \\
78 & Best-of-k endpoint sampling & Best-of-k endpoint selection & Search tolerated several valid answers but could not repair the generator producing them. & $\mathsf{K_6},\mathsf{K_8},\mathsf{K_11}$ \\
79 & Target-noised training & Target-noised conditional denoising & Making the target distribution explicit improved diversity but left decoded detail unstable. & $\mathsf{K_6},\mathsf{K_9},\mathsf{K_11}$ \\
80 & Target-noised training & Target-noised conditional velocity & Making the target distribution explicit improved diversity but left decoded detail unstable. & $\mathsf{K_6},\mathsf{K_9},\mathsf{K_11}$ \\
81 & Style-latent conditioning & Style-latent conditioning & The style variable exposed missing appearance choices but recreated the bridge/prior problem. & $\mathsf{K_4},\mathsf{K_6},\mathsf{K_11}$ \\
82 & Reverse-noise encoding & Reverse-noise encoder & A structured inverse start did not make forward target generation reliable. & $\mathsf{K_6},\mathsf{K_9}$ \\
83 & Conditional latent diffusion & Conditional DDPM bridge & Changing the noise schedule from RF to diffusion did not remove the missing-target-prior failure. & $\mathsf{K_6},\mathsf{K_9},\mathsf{K_13}$ \\
84 & Conditional latent diffusion & Conditional DDIM bridge & Changing the noise schedule from RF to diffusion did not remove the missing-target-prior failure. & $\mathsf{K_6},\mathsf{K_9},\mathsf{K_13}$ \\
85 & Rectified auto-coder paths & One-step RAC & Path regularity, perturbation recovery, and compatibility with a pretrained latent contract were separate properties. & $\mathsf{K_2},\mathsf{K_6},\mathsf{K_9}$ \\
86 & Rectified auto-coder paths & Four-step RAC & Path regularity, perturbation recovery, and compatibility with a pretrained latent contract were separate properties. & $\mathsf{K_2},\mathsf{K_6},\mathsf{K_9}$ \\
87 & Rectified auto-coder paths & Fixed-time-grid RAC & Path regularity, perturbation recovery, and compatibility with a pretrained latent contract were separate properties. & $\mathsf{K_2},\mathsf{K_6},\mathsf{K_9}$ \\
88 & Rectified auto-coder paths & Random-time-grid RAC & Path regularity, perturbation recovery, and compatibility with a pretrained latent contract were separate properties. & $\mathsf{K_2},\mathsf{K_6},\mathsf{K_9}$ \\
89 & Rectified auto-coder paths & Intermediate-state RAC supervision & Path regularity, perturbation recovery, and compatibility with a pretrained latent contract were separate properties. & $\mathsf{K_2},\mathsf{K_6},\mathsf{K_9}$ \\
90 & Rectified auto-coder paths & State-noise correction RAC & Path regularity, perturbation recovery, and compatibility with a pretrained latent contract were separate properties. & $\mathsf{K_2},\mathsf{K_6},\mathsf{K_9}$ \\
91 & Shadow-variable and perturbed-latent routes & Direct RGBD shadow code & A nearby or privately reconstructable code was not necessarily interchangeable with the native RGB latent. & $\mathsf{K_1},\mathsf{K_2},\mathsf{K_6}$ \\
92 & Shadow-variable and perturbed-latent routes & Depth visualized through the official RGB codec & A nearby or privately reconstructable code was not necessarily interchangeable with the native RGB latent. & $\mathsf{K_1},\mathsf{K_2},\mathsf{K_6}$ \\
93 & Shadow-variable and perturbed-latent routes & Position-matched bidirectional shadow encoding & A nearby or privately reconstructable code was not necessarily interchangeable with the native RGB latent. & $\mathsf{K_1},\mathsf{K_2},\mathsf{K_6}$ \\
94 & Shadow-variable and perturbed-latent routes & Additive perturbation around native RGB latent & A nearby or privately reconstructable code was not necessarily interchangeable with the native RGB latent. & $\mathsf{K_1},\mathsf{K_2},\mathsf{K_6}$ \\
95 & Video-backbone branches & LTX deep injection & A pretrained video generator supplied a target prior, but only appropriately placed interventions redirected it reliably. & $\mathsf{K_6},\mathsf{K_7},\mathsf{K_13}$ \\
96 & Video-backbone branches & Native LTX LoRA & A pretrained video generator supplied a target prior, but only appropriately placed interventions redirected it reliably. & $\mathsf{K_6},\mathsf{K_7},\mathsf{K_13}$ \\
97 & Video-backbone branches & Modality-tagged Wan LoRA & A pretrained video generator supplied a target prior, but only appropriately placed interventions redirected it reliably. & $\mathsf{K_6},\mathsf{K_7},\mathsf{K_13}$ \\
98 & Video-backbone branches & UniVidX baseline & A pretrained video generator supplied a target prior, but only appropriately placed interventions redirected it reliably. & $\mathsf{K_6},\mathsf{K_7},\mathsf{K_13}$ \\
99 & Video-backbone branches & WSG UniVid adapter & A pretrained video generator supplied a target prior, but only appropriately placed interventions redirected it reliably. & $\mathsf{K_6},\mathsf{K_7},\mathsf{K_13}$ \\
100 & Video-backbone branches & Official-style video control & A pretrained video generator supplied a target prior, but only appropriately placed interventions redirected it reliably. & $\mathsf{K_6},\mathsf{K_7},\mathsf{K_13}$ \\
101 & Video-backbone branches & Wan 1.3B route & A pretrained video generator supplied a target prior, but only appropriately placed interventions redirected it reliably. & $\mathsf{K_6},\mathsf{K_7},\mathsf{K_13}$ \\
102 & Video-backbone branches & Fun InP reference route & A pretrained video generator supplied a target prior, but only appropriately placed interventions redirected it reliably. & $\mathsf{K_6},\mathsf{K_7},\mathsf{K_13}$ \\
103 & Video-backbone branches & VACE reference route & A pretrained video generator supplied a target prior, but only appropriately placed interventions redirected it reliably. & $\mathsf{K_6},\mathsf{K_7},\mathsf{K_13}$ \\
104 & Joint video-lane strategies & Native multi-lane generation & Fixed small lane sets were feasible, while larger collections became memory-heavy and mutually coupled. & $\mathsf{K_7},\mathsf{K_12},\mathsf{K_13}$ \\
105 & Joint video-lane strategies & Sample-one lane training & Fixed small lane sets were feasible, while larger collections became memory-heavy and mutually coupled. & $\mathsf{K_7},\mathsf{K_12},\mathsf{K_13}$ \\
106 & Joint video-lane strategies & Split condition groups & Fixed small lane sets were feasible, while larger collections became memory-heavy and mutually coupled. & $\mathsf{K_7},\mathsf{K_12},\mathsf{K_13}$ \\
107 & Joint video-lane strategies & Fused conditions & Fixed small lane sets were feasible, while larger collections became memory-heavy and mutually coupled. & $\mathsf{K_7},\mathsf{K_12},\mathsf{K_13}$ \\
108 & Joint video-lane strategies & Merged-lane big tokens & Fixed small lane sets were feasible, while larger collections became memory-heavy and mutually coupled. & $\mathsf{K_7},\mathsf{K_12},\mathsf{K_13}$ \\
109 & Joint video-lane strategies & Multimodal lane tags & Fixed small lane sets were feasible, while larger collections became memory-heavy and mutually coupled. & $\mathsf{K_7},\mathsf{K_12},\mathsf{K_13}$ \\
110 & Joint video-lane strategies & Multilayer residual control & Fixed small lane sets were feasible, while larger collections became memory-heavy and mutually coupled. & $\mathsf{K_7},\mathsf{K_12},\mathsf{K_13}$ \\
111 & FLUX/Omini evidence placement & Target-first native reference & Evidence placement and attention topology changed convergence even with identical data and targets. & $\mathsf{K_4},\mathsf{K_13}$ \\
112 & FLUX/Omini evidence placement & Trainable condition-lane reference & Evidence placement and attention topology changed convergence even with identical data and targets. & $\mathsf{K_4},\mathsf{K_13}$ \\
113 & FLUX/Omini evidence placement & Dual condition attention & Evidence placement and attention topology changed convergence even with identical data and targets. & $\mathsf{K_4},\mathsf{K_13}$ \\
114 & FLUX/Omini evidence placement & Task-type attention switch & Evidence placement and attention topology changed convergence even with identical data and targets. & $\mathsf{K_4},\mathsf{K_13}$ \\
115 & FLUX/Omini evidence placement & Frozen condition-side LoRA & Evidence placement and attention topology changed convergence even with identical data and targets. & $\mathsf{K_4},\mathsf{K_13}$ \\
116 & Task-agnostic flow objectives & Plain rectified flow & The high-noise anchor best preserved global structure under fixed validation, while objectives remained task-independent tools. & $\mathsf{K_8},\mathsf{K_9},\mathsf{K_13}$ \\
117 & Task-agnostic flow objectives & Direct clean-latent prediction & The high-noise anchor best preserved global structure under fixed validation, while objectives remained task-independent tools. & $\mathsf{K_8},\mathsf{K_9},\mathsf{K_13}$ \\
118 & Task-agnostic flow objectives & High-noise anchor & The high-noise anchor best preserved global structure under fixed validation, while objectives remained task-independent tools. & $\mathsf{K_8},\mathsf{K_9},\mathsf{K_13}$ \\
119 & Task-agnostic flow objectives & Multi-resolution noise & The high-noise anchor best preserved global structure under fixed validation, while objectives remained task-independent tools. & $\mathsf{K_8},\mathsf{K_9},\mathsf{K_13}$ \\
120 & Dense flow objective levers & Low-noise timestep shift & Individually useful loss changes interfered when combined and could encourage mean outputs. & $\mathsf{K_8},\mathsf{K_13}$ \\
121 & Dense flow objective levers & Min-SNR weighting & Individually useful loss changes interfered when combined and could encourage mean outputs. & $\mathsf{K_8},\mathsf{K_13}$ \\
122 & Dense flow objective levers & Huber loss & Individually useful loss changes interfered when combined and could encourage mean outputs. & $\mathsf{K_8},\mathsf{K_13}$ \\
123 & Packed multi-target LoRA acquisition & Full eight-target packed graph & Packing shared some execution but target-specific activations, gradients, and cost still scaled with relation count. & $\mathsf{K_7},\mathsf{K_12}$ \\
124 & Packed multi-target LoRA acquisition & Four-target retained-graph chunks & Packing shared some execution but target-specific activations, gradients, and cost still scaled with relation count. & $\mathsf{K_7},\mathsf{K_12}$ \\
125 & Packed multi-target LoRA acquisition & Four-target immediate-backward chunks & Packing shared some execution but target-specific activations, gradients, and cost still scaled with relation count. & $\mathsf{K_7},\mathsf{K_12}$ \\
126 & Packed multi-target LoRA acquisition & Two-target single-GPU chunks & Packing shared some execution but target-specific activations, gradients, and cost still scaled with relation count. & $\mathsf{K_7},\mathsf{K_12}$ \\
127 & Packed multi-target LoRA acquisition & Frozen condition-side packed encoder & Packing shared some execution but target-specific activations, gradients, and cost still scaled with relation count. & $\mathsf{K_7},\mathsf{K_12}$ \\
128 & Shared-target multi-condition LoRA & Eight-condition alternating Edge LoRA & A common target did not make distinct condition-to-target relations independently maintainable. & $\mathsf{K_4},\mathsf{K_7},\mathsf{K_12}$ \\
129 & Shared-target multi-condition LoRA & Four-condition dense groups & A common target did not make distinct condition-to-target relations independently maintainable. & $\mathsf{K_4},\mathsf{K_7},\mathsf{K_12}$ \\
130 & Shared-target multi-condition LoRA & Fused-condition shared route & A common target did not make distinct condition-to-target relations independently maintainable. & $\mathsf{K_4},\mathsf{K_7},\mathsf{K_12}$ \\
131 & Shared-target multi-condition LoRA & Thirty-eight-condition expansion & A common target did not make distinct condition-to-target relations independently maintainable. & $\mathsf{K_4},\mathsf{K_7},\mathsf{K_12}$ \\
132 & NJA/MTA TunnelingNet routes & Native joint attention route & Small internal residual paths could redirect a shared prior; the preferred attention topology depended on the interface. & $\mathsf{K_6},\mathsf{K_7},\mathsf{K_13}$ \\
133 & NJA/MTA TunnelingNet routes & Modality tunneling attention route & Small internal residual paths could redirect a shared prior; the preferred attention topology depended on the interface. & $\mathsf{K_6},\mathsf{K_7},\mathsf{K_13}$ \\
134 & NJA/MTA TunnelingNet routes & Zero-initialized residual control & Small internal residual paths could redirect a shared prior; the preferred attention topology depended on the interface. & $\mathsf{K_6},\mathsf{K_7},\mathsf{K_13}$ \\
135 & NJA/MTA TunnelingNet routes & Multilayer residual tunnel & Small internal residual paths could redirect a shared prior; the preferred attention topology depended on the interface. & $\mathsf{K_6},\mathsf{K_7},\mathsf{K_13}$ \\
136 & Semantic heatmap and inverse codec & Gaussian semantic heatmap & Heatmaps preserved locations better, while semantic color still entangled identity. & $\mathsf{K_1},\mathsf{K_5}$ \\
137 & Semantic heatmap and inverse codec & Peak/NMS inverse heatmap codec & Heatmaps preserved locations better, while semantic color still entangled identity. & $\mathsf{K_1},\mathsf{K_5}$ \\
138 & Person{-}{-}keypoint slot heatmaps & Fixed person{-}{-}keypoint slots & Fixed slots made identity structural and left only location/confidence to generation. & $\mathsf{K_1},\mathsf{K_5}$ \\
139 & Person{-}{-}keypoint slot heatmaps & Three-lane packed slot heatmaps & Fixed slots made identity structural and left only location/confidence to generation. & $\mathsf{K_1},\mathsf{K_5}$ \\
140 & Body-part and full-body heatmaps & Body-part partitioned heatmaps & Partitioning reduced identity collisions at the cost of additional lanes and caching. & $\mathsf{K_1},\mathsf{K_5},\mathsf{K_12}$ \\
141 & Body-part and full-body heatmaps & Body-only heatmaps & Partitioning reduced identity collisions at the cost of additional lanes and caching. & $\mathsf{K_1},\mathsf{K_5},\mathsf{K_12}$ \\
142 & Body-part and full-body heatmaps & Full-body heatmap codebook & Partitioning reduced identity collisions at the cost of additional lanes and caching. & $\mathsf{K_1},\mathsf{K_5},\mathsf{K_12}$ \\
143 & Direct sparse sequence heads & Keypoint sequence head & Typed sequences were compact but required explicit ordering, masks, matching, decoding, and native losses. & $\mathsf{K_1},\mathsf{K_5},\mathsf{K_13}$ \\
144 & Direct sparse sequence heads & Pose sequence head & Typed sequences were compact but required explicit ordering, masks, matching, decoding, and native losses. & $\mathsf{K_1},\mathsf{K_5},\mathsf{K_13}$ \\
145 & Direct sparse sequence heads & Box sequence head & Typed sequences were compact but required explicit ordering, masks, matching, decoding, and native losses. & $\mathsf{K_1},\mathsf{K_5},\mathsf{K_13}$ \\
146 & Direct sparse sequence heads & SMPL parameter head & Typed sequences were compact but required explicit ordering, masks, matching, decoding, and native losses. & $\mathsf{K_1},\mathsf{K_5},\mathsf{K_13}$ \\
147 & Direct sparse sequence heads & Audio-token sequence head & Typed sequences were compact but required explicit ordering, masks, matching, decoding, and native losses. & $\mathsf{K_1},\mathsf{K_5},\mathsf{K_13}$ \\
148 & ScoreHMR WSV structure route & Full-frame cached ScoreHMR features & A pretrained structural prior learned body parameters quickly from aligned image evidence. & $\mathsf{K_6},\mathsf{K_13}$ \\
149 & ScoreHMR WSV structure route & Crop-aligned cached ScoreHMR features & A pretrained structural prior learned body parameters quickly from aligned image evidence. & $\mathsf{K_6},\mathsf{K_13}$ \\
150 & Atomic object-box slots & Structured center/size box slots & Atomic center/size slots preserved object membership better than dense pseudocolor masks. & $\mathsf{K_1},\mathsf{K_5},\mathsf{K_13}$ \\
151 & Atomic object-box slots & MTA pseudocolor box masks & Atomic center/size slots preserved object membership better than dense pseudocolor masks. & $\mathsf{K_1},\mathsf{K_5},\mathsf{K_13}$ \\
152 & Atomic object-box slots & NJA pseudocolor box masks & Atomic center/size slots preserved object membership better than dense pseudocolor masks. & $\mathsf{K_1},\mathsf{K_5},\mathsf{K_13}$ \\
153 & Object BFS tree & Parent-relative BFS serialization & Explicit parent order and traversal represented variable object topology better than fixed human joint IDs. & $\mathsf{K_1},\mathsf{K_5}$ \\
154 & Object BFS tree & Absolute-coordinate BFS serialization & Explicit parent order and traversal represented variable object topology better than fixed human joint IDs. & $\mathsf{K_1},\mathsf{K_5}$ \\
155 & Point-cloud skeleton initialization & K-means point-cloud anchors & Analytical geometry supplied a useful starting skeleton; unconstrained learned refinement remained unstable. & $\mathsf{K_5},\mathsf{K_6},\mathsf{K_13}$ \\
156 & Point-cloud skeleton initialization & MST skeleton initialization & Analytical geometry supplied a useful starting skeleton; unconstrained learned refinement remained unstable. & $\mathsf{K_5},\mathsf{K_6},\mathsf{K_13}$ \\
157 & Point-cloud skeleton initialization & Diffusion skeleton refinement & Analytical geometry supplied a useful starting skeleton; unconstrained learned refinement remained unstable. & $\mathsf{K_5},\mathsf{K_6},\mathsf{K_13}$ \\
158 & Point-cloud skeleton initialization & Anchor-correction refinement & Analytical geometry supplied a useful starting skeleton; unconstrained learned refinement remained unstable. & $\mathsf{K_5},\mathsf{K_6},\mathsf{K_13}$ \\
159 & Sparse and mixed audio routes & AST audio-only classification & Aligned audio supported classification and spatial selection; direct token/waveform generation remained weak. & $\mathsf{K_1},\mathsf{K_3},\mathsf{K_4},\mathsf{K_13}$ \\
160 & Sparse and mixed audio routes & RGB-plus-audio source box & Aligned audio supported classification and spatial selection; direct token/waveform generation remained weak. & $\mathsf{K_1},\mathsf{K_3},\mathsf{K_4},\mathsf{K_13}$ \\
161 & Sparse and mixed audio routes & EnCodec visual-to-audio baseline & Aligned audio supported classification and spatial selection; direct token/waveform generation remained weak. & $\mathsf{K_1},\mathsf{K_3},\mathsf{K_4},\mathsf{K_13}$ \\
162 & Sparse and mixed audio routes & Image-prefix MusicGen & Aligned audio supported classification and spatial selection; direct token/waveform generation remained weak. & $\mathsf{K_1},\mathsf{K_3},\mathsf{K_4},\mathsf{K_13}$ \\
163 & Sparse and mixed audio routes & Dense audio-semantic mask & Aligned audio supported classification and spatial selection; direct token/waveform generation remained weak. & $\mathsf{K_1},\mathsf{K_3},\mathsf{K_4},\mathsf{K_13}$ \\
164 & Sparse and mixed audio routes & RGB-plus-audio region selector & Aligned audio supported classification and spatial selection; direct token/waveform generation remained weak. & $\mathsf{K_1},\mathsf{K_3},\mathsf{K_4},\mathsf{K_13}$ \\
165 & Dense categorical-mask pilots & Gated fixed-class route & Full generation exposed background collapse and fragmented legal colors hidden by partial previews. & $\mathsf{K_5},\mathsf{K_8},\mathsf{K_10}$ \\
166 & Dense categorical-mask pilots & Balanced-flow route & Full generation exposed background collapse and fragmented legal colors hidden by partial previews. & $\mathsf{K_5},\mathsf{K_8},\mathsf{K_10}$ \\
167 & Dense categorical-mask pilots & Self-contained prototype head & Full generation exposed background collapse and fragmented legal colors hidden by partial previews. & $\mathsf{K_5},\mathsf{K_8},\mathsf{K_10}$ \\
168 & Dense categorical-mask pilots & Set-level multi-label head & Full generation exposed background collapse and fragmented legal colors hidden by partial previews. & $\mathsf{K_5},\mathsf{K_8},\mathsf{K_10}$ \\
169 & Dense categorical-mask pilots & Predicted-mask multi-label head & Full generation exposed background collapse and fragmented legal colors hidden by partial previews. & $\mathsf{K_5},\mathsf{K_8},\mathsf{K_10}$ \\
170 & Dense categorical-mask pilots & Pixelwise semantic prototype CE & Full generation exposed background collapse and fragmented legal colors hidden by partial previews. & $\mathsf{K_5},\mathsf{K_8},\mathsf{K_10}$ \\
171 & Dense categorical-mask pilots & Prototype CE with multi-label classifier & Full generation exposed background collapse and fragmented legal colors hidden by partial previews. & $\mathsf{K_5},\mathsf{K_8},\mathsf{K_10}$ \\
172 & Dense categorical-mask pilots & Fresh prototype CE without classifier & Full generation exposed background collapse and fragmented legal colors hidden by partial previews. & $\mathsf{K_5},\mathsf{K_8},\mathsf{K_10}$ \\
173 & Categorical objective modes & Class-balanced CE & Better categorical losses improved candidates but continuous generation still needed symbolic recovery. & $\mathsf{K_8},\mathsf{K_10},\mathsf{K_13}$ \\
174 & Categorical objective modes & Combined categorical objective & Better categorical losses improved candidates but continuous generation still needed symbolic recovery. & $\mathsf{K_8},\mathsf{K_10},\mathsf{K_13}$ \\
175 & Categorical objective modes & Global-margin supervision & Better categorical losses improved candidates but continuous generation still needed symbolic recovery. & $\mathsf{K_8},\mathsf{K_10},\mathsf{K_13}$ \\
176 & Categorical objective modes & Prototype CE & Better categorical losses improved candidates but continuous generation still needed symbolic recovery. & $\mathsf{K_8},\mathsf{K_10},\mathsf{K_13}$ \\
177 & Categorical objective modes & Prototype CE with focal regression & Better categorical losses improved candidates but continuous generation still needed symbolic recovery. & $\mathsf{K_8},\mathsf{K_10},\mathsf{K_13}$ \\
178 & Categorical region refiners & Per-Head categorical refiner & A second stage preserved spatial support while recovering exact labels, with fine-grained limits. & $\mathsf{K_8},\mathsf{K_10},\mathsf{K_13}$ \\
179 & Categorical region refiners & Multi-condition categorical refiner & A second stage preserved spatial support while recovering exact labels, with fine-grained limits. & $\mathsf{K_8},\mathsf{K_10},\mathsf{K_13}$ \\
180 & Categorical region refiners & Foreground recovery & A second stage preserved spatial support while recovering exact labels, with fine-grained limits. & $\mathsf{K_8},\mathsf{K_10},\mathsf{K_13}$ \\
181 & Categorical region refiners & Coarse-label remapping & A second stage preserved spatial support while recovering exact labels, with fine-grained limits. & $\mathsf{K_8},\mathsf{K_10},\mathsf{K_13}$ \\
182 & Universal Pose custom Head & Full-frame Universal Pose & Correct geometry, cropping, and gradient supervision mattered more than simply enlarging the model. & $\mathsf{K_1},\mathsf{K_5},\mathsf{K_13}$ \\
183 & Universal Pose custom Head & Orientation-corrected letterbox & Correct geometry, cropping, and gradient supervision mattered more than simply enlarging the model. & $\mathsf{K_1},\mathsf{K_5},\mathsf{K_13}$ \\
184 & Universal Pose custom Head & Instance-crop Universal Pose & Correct geometry, cropping, and gradient supervision mattered more than simply enlarging the model. & $\mathsf{K_1},\mathsf{K_5},\mathsf{K_13}$ \\
185 & Universal Pose custom Head & Gradient-supervised Universal Pose & Correct geometry, cropping, and gradient supervision mattered more than simply enlarging the model. & $\mathsf{K_1},\mathsf{K_5},\mathsf{K_13}$ \\
186 & Universal Pose custom Head & Focal-supervised Universal Pose & Correct geometry, cropping, and gradient supervision mattered more than simply enlarging the model. & $\mathsf{K_1},\mathsf{K_5},\mathsf{K_13}$ \\
187 & Pose-detail objective pilots & Latent endpoint continuation & Endpoint-oriented and detail losses often removed fingers; the gradient baseline remained strongest. & $\mathsf{K_5},\mathsf{K_8},\mathsf{K_13}$ \\
188 & Pose-detail objective pilots & Decoded endpoint continuation & Endpoint-oriented and detail losses often removed fingers; the gradient baseline remained strongest. & $\mathsf{K_5},\mathsf{K_8},\mathsf{K_13}$ \\
189 & Pose-detail objective pilots & Topology continuation & Endpoint-oriented and detail losses often removed fingers; the gradient baseline remained strongest. & $\mathsf{K_5},\mathsf{K_8},\mathsf{K_13}$ \\
190 & Pose-detail objective pilots & Low-noise detail anchor & Endpoint-oriented and detail losses often removed fingers; the gradient baseline remained strongest. & $\mathsf{K_5},\mathsf{K_8},\mathsf{K_13}$ \\
191 & Pose-detail objective pilots & Soft-distance continuation & Endpoint-oriented and detail losses often removed fingers; the gradient baseline remained strongest. & $\mathsf{K_5},\mathsf{K_8},\mathsf{K_13}$ \\
192 & Pose-detail objective pilots & Multi-scale-gradient continuation & Endpoint-oriented and detail losses often removed fingers; the gradient baseline remained strongest. & $\mathsf{K_5},\mathsf{K_8},\mathsf{K_13}$ \\
193 & Pose-detail objective pilots & Gradient-direction continuation & Endpoint-oriented and detail losses often removed fingers; the gradient baseline remained strongest. & $\mathsf{K_5},\mathsf{K_8},\mathsf{K_13}$ \\
194 & Pose-detail objective pilots & Hard-foreground mining & Endpoint-oriented and detail losses often removed fingers; the gradient baseline remained strongest. & $\mathsf{K_5},\mathsf{K_8},\mathsf{K_13}$ \\
195 & Named custom-variable routes & MAGICK transparency route & Named graph variables expanded the task graph without consuming fixed WSG channel IDs. & $\mathsf{K_1},\mathsf{K_7},\mathsf{K_13}$ \\
196 & Named custom-variable routes & IC-Light relighting route & Named graph variables expanded the task graph without consuming fixed WSG channel IDs. & $\mathsf{K_1},\mathsf{K_7},\mathsf{K_13}$ \\
197 & Runtime route composition & Text-to-RGB then RGB-to-world chain & Installed routes executed as chains without joint retraining while retaining edge provenance. & $\mathsf{K_7},\mathsf{K_14}$ \\
198 & Runtime route composition & Generic intermediate-node route composition & Installed routes executed as chains without joint retraining while retaining edge provenance. & $\mathsf{K_7},\mathsf{K_14}$ \\